\documentclass[sw]{iosart2x_mod}

\usepackage{array}
\usepackage[none]{hyphenat}
\usepackage{caption}
\usepackage[font=footnotesize]{caption}
\usepackage{makecell}
\usepackage{multirow}
\usepackage{pifont}
\usepackage{xspace}
\usepackage[table]{xcolor}
\usepackage{colortbl}
\usepackage{anyfontsize}
\usepackage{pdflscape}
\usepackage{diagbox}
\usepackage[draft,authormarkup=none]{changes}
\usepackage{svg}
\usepackage{longtable}

\newcommand{\cmark}{\ding{51}}
\newcommand{\omark}{\ding{61}}
\newcommand{\xmark}{\ding{55}\xspace}
\newdimen\NetTableWidth
\definecolor{lgrey}{HTML}{D7D7D7}
\newcommand{\greyline}{\arrayrulecolor{lgrey}\hline\arrayrulecolor{black}}
\newcommand{\greyclineone}{\arrayrulecolor{lgrey}\cline{2-5}\arrayrulecolor{black}}
\newcommand{\greyclinetwo}{\arrayrulecolor{lgrey}\cline{2-3}\arrayrulecolor{black}}
\newcommand{\greyclinethree}{\arrayrulecolor{lgrey}\cline{5-5}\arrayrulecolor{black}}
\newcolumntype{P}[1]{>{\centering\arraybackslash}p{#1}}
\makeatletter
\newcommand*{\@rowstyle}{}
\newcommand*{\rowstyle}[1]{
  \gdef\@rowstyle{#1}%
  \@rowstyle\ignorespaces%
}
\newcolumntype{=}{
  >{\gdef\@rowstyle{}}%
}
\newcolumntype{+}{
  >{\@rowstyle}%
}
\makeatother

\pubyear{0000}
\volume{0}
\firstpage{1}
\lastpage{1}
\begin{document}

\makeatletter
\let\put@numberlines@box\relax
\makeatother

\NetTableWidth=\dimexpr
\linewidth
- 8\tabcolsep
\relax

\begin{frontmatter}

\title{Semantic Web technologies in sensor-based personal health monitoring systems: A systematic mapping study}
\runtitle{Semantic Web technologies in sensor-based personal health monitoring systems}

\begin{aug}
\author[A,B]{\inits{M.}\fnms{Mbithe} \snm{Nzomo}\ead[label=e1]{mnzomo@cs.uct.ac.za}%
\thanks{Corresponding author. \printead{e1}.}}
\author[A,B]{\inits{D.}\fnms{Deshendran} \snm{Moodley}\ead[label=e2]{deshen.moodley@uct.ac.za}}
\address[A]{Department of Computer Science, \orgname{University of Cape Town},
\cny{South Africa}\printead[presep={\\}]{e1,e2}}
\address[B]{\orgname{Centre for Artificial Intelligence Research (CAIR)},
\cny{South Africa}}
\end{aug}

\begin{abstract}
In recent years, there has been an increased focus on early detection, prevention, and prediction of diseases. 
This, together with advances in sensor technology and the Internet of Things, has led to accelerated efforts in the development of personal health monitoring systems. 
This study analyses the state of the art in the use of Semantic Web technologies in sensor-based personal health monitoring systems.
Using a systematic approach, a total of 48 systems are selected as representative of the current state of the art. 
We critically analyse the extent to which the selected systems address seven key challenges: interoperability, situation detection, situation prediction, decision support, context awareness, explainability, and uncertainty handling.
We discuss the role and limitations of Semantic Web technologies in managing each challenge.
We then conduct a quality assessment of the selected systems based on the data and devices used, system and components development, rigour of evaluation, and accessibility of research outputs.
Finally, we propose a reference architecture to provide guidance for the design and development of new systems.
This study provides a comprehensive mapping of the field, identifies inadequacies in the state of the art, and provides recommendations for future research.
\end{abstract}

\begin{keyword}
\kwd{Semantic Web technologies}
\kwd{ontologies}
\kwd{knowledge graphs}
\kwd{linked data}
\kwd{sensors}
\kwd{Internet of Things}
\kwd{health monitoring}
\end{keyword}

\end{frontmatter}


\section{Introduction}
\label{introduction}

Non-communicable diseases are on the rise globally, resulting not only in decreased quality of life but also increasing healthcare costs \cite{murphy_household_2020}. For this reason, there have been accelerated efforts to develop personal health monitoring systems for early detection, prediction, and prevention of diseases. 
The emerging paradigm of precision health goes beyond treating existing diseases and rather focuses on preventing disease before it strikes.
Eschewing the one-size-fits-all approach in favour of assessing individual circumstances, precision health encourages people to actively monitor and work towards improving their health so as to lower the risk of disease \cite{gambhir_toward_2018}.
Personal health monitoring is part of this vision, allowing people to not only increase understanding of their health but also to receive recommendations for any necessary interventions. 
Significant advances in the Internet of Things (IoT) over the last decade has led to the rapid rise of wearable sensors, which are increasingly being used for health monitoring outside traditional clinical settings. 
Wearable sensors can collect and measure physiological data such as vital signs, which can be combined with health records and questionnaires to determine lifestyle habits and medical history.
Beyond physiological data, ambient sensors can monitor environmental factors such as air quality and weather, which have a significant impact on health.
Additionally, the widespread adoption of artificial intelligence (AI) in the health domain has led to personal health monitoring systems becoming increasingly AI-driven. Such systems use techniques such as knowledge representation and reasoning and machine learning to analyse health data and provide actionable insights.

There are several crucial issues affecting sensor-based personal health monitoring systems, which can be distilled into seven key challenges.
The first of these is \textbf{interoperability}. 
Heterogeneous sensor observations, differing data transmission technologies, and disparate standards for describing health data all contribute to interoperability issues in personal health monitoring systems.
Additionally, the representation of health domain knowledge and its integration with sensor data remains a challenging task \cite{de_brouwer_context-aware_2023}.
By its nature, sensor data is dynamic and complex, necessitating interpretation into higher-level concepts or situations \cite{ye_situation_2011}.
Situation analysis involves the use of sensor data to detect the current state of a given environment (\textbf{situation detection}), while anticipating possible future states (\textbf{situation prediction}) \cite{adeleke_integrating_2017}.
The representation of domain knowledge is essential for facilitating situation analysis from sensor data and supporting subsequent decision-making. 
The \textbf{decision support} process augments human judgement, assisting clinicians in navigating complex medical decisions \cite{sutton_overview_2020} and supporting patients in making informed health decisions outside clinical settings \cite{dullabh_challenges_2022}. 
Both the situation analysis and decision support processes must incorporate \textbf{context awareness}.
Dey and Abowd \cite{dey_towards_2000} define context as any information that can be used to characterize the situation of an entity, including location, identity, activity, and time.
Such information is essential for accurate situation analysis and targeted decision support.
Moreover, since the health domain is a high-stakes one, \textbf{explainability} is gaining traction as a pivotal aspect of AI-driven health monitoring systems \cite{haque_semantic_2022}.
Finally, given the probabilistic nature of health outcomes and the limitations of sensor and other health data, there is inherent uncertainty in the situation analysis and decision-making processes \cite{kaplan_decision_2005}. Thus, effective \textbf{uncertainty handling} is critical in sensor-based personal health monitoring systems.

Semantic Web technologies have been used in numerous health applications ranging from health data integration \cite{peng_literature_2020,hammad_semantic-based_2020} to clinical decision support \cite{jing_ontologies_2023,cui_review_2025}. While they have shown promise in alleviating the seven key challenges,
the degree to which they address each challenge differs.
Therefore, the goal of this study is to systematically map the state of the art in the use of Semantic Web technologies in sensor-based personal health monitoring systems.
We analyse the effectiveness of the systems in addressing the seven key challenges, identify the role and limitations of Semantic Web technologies, and assess the overall quality of the systems.
Accordingly, a systematic mapping study was selected as the most appropriate approach.
Systematic mapping studies have become increasingly popular in software engineering \cite{khan_landscaping_2019}.
While systematic reviews aim at synthesizing evidence for specific research questions, mapping studies go further and provide a high-level view of the research landscape \cite{khan_landscaping_2019}. 
By structuring a research area through classification and categorisation, mapping studies seek to discover emerging research trends and identify potential gaps for further lines of inquiry \cite{petersen_guidelines_2015,budgen_using_2008}.

The contributions of this study are as follows:
\begin{enumerate}
    \item We present a \textbf{systematic mapping} of the field based on 48 systems that are systematically selected as representative of the current state of the art. Our selection criteria includes systems across all development stages, from research prototypes to production-level systems.
    \item We \textbf{critically evaluate} the extent to which the systems address the seven key challenges, i.e. interoperability, situation detection, situation prediction, decision support, context awareness, explainability, and uncertainty handling.
    \item We \textbf{assess the role and limitations of Semantic Web technologies} in managing each challenge.
    \item We undertake a \textbf{quality assessment} of the selected systems based on the data and devices used, system and components development, rigour of evaluation, and accessibility of research outputs.
    \item Following an analysis of the current architectures, components, functionalities, and development tools, we propose a \textbf{reference architecture} to provide guidance for the design and development of new systems.
    \item We highlight inadequacies in existing systems and outstanding issues in the field, thereby identifying potential \textbf{directions for future research}. 
\end{enumerate}        

The remainder of this paper is structured as follows. 
Section~\ref{background} provides an overview of personal health monitoring using sensors and highlights how Semantic Web technologies can enhance sensor-based health monitoring systems.
Section~\ref{related-reviews} discusses related reviews and surveys, motivating the novelty and importance of this study.
Section~\ref{methodology} details the methodology used to conduct the study, including the search strategy and the inclusion and exclusion criteria, culminating in a summary of the selected systems.
Section~\ref{key-challenges} discusses the seven key challenges that such systems must address, and critically analyses the capacity of the systems to deal with these challenges, while Section~\ref{system-quality} analyses the quality of each system.
The architectures of the selected systems are discussed in Section~\ref{architectures} and a reference architecture is proposed.
Section~\ref{discussion} summarises the main findings of the study, discusses its limitations, and makes recommendations for future research directions. 
Finally, Section~\ref{conclusion} concludes the study.

\section{Background}
\label{background}

\subsection{Sensor-based personal health monitoring}
Sensors used for health monitoring are typically worn, implanted, or placed in close proximity to the human body.
When several such sensors are used at the same time, they form a wireless body sensor network (BSN), also known as a body area network (BAN) \cite{gravina_wearable_2021}.
This is part of the IoT paradigm, in which sensor-based ``things'' connect and exchange data over a shared network such as the Internet.
Two categories of physiological data can be collected from health monitoring sensors: vital signs and biological signals (biosignals).
The primary vital signs are heart rate, blood pressure, respiratory rate, temperature, and blood oxygen saturation \cite{dias_wearable_2018}.
Biosignals are space- or time-based records produced from electrical, chemical, or mechanical activity within the body during a biological event such as a beating heart \cite{escabi_biosignal_2012}. 
They include records of electrical activity in the body such as electrocardiograms (ECG) for the heart, electromyograms (EMG) for the skeletal muscles, and electroencephalograms (EEG) for the brain, as well as data from photoplethysmography (PPG), an optical sensing technology consisting of an LED and a photodetector to detect blood volume changes \cite{ferlini_in-ear_2022}.
In addition to physiological data, physical activity data such as daily step count can also be captured by sensors.
This data provides important contextual information about an individual's lifestyle, which can enhance health monitoring. 

Health monitoring sensors are usually either wearable or implantable. 
Wearable sensors are worn on the body or are otherwise integrated with clothes and shoes. 
Such sensors include electrodes for measuring electrical signals, thermal sensors for measuring temperature, and PPG sensors. 
Smart watches and bands are the most commonly used wearable sensors, but earables (devices placed in the ear) have recently emerged as a promising alternative \cite{choudhury_earable_2021}.
In contrast, implantable sensors operate from within the human body. Although they are much less commonly used than wearable sensors, they are particularly useful for monitoring chronic illness as well as post-surgery monitoring to minimise complications and avoid readmission \cite{andreu-perez_wearable_2015}.
Health monitoring sensors also include portable devices that can measure physiological and activity data but cannot be practically worn or used for prolonged periods of time. 
Examples of these include blood pressure monitors and pulse oximeters, as well as smartphones which contain sensors such accelerometers, which measure acceleration, and gyroscopes, which measure orientation and angular velocity \cite{straczkiewicz_systematic_2021}. 

Besides wearable and implantable sensors, there exist contact-free sensors that can monitor health-related factors. 
For example, the commercially available Emfit QS device\footnote{\url{https://emfit.com/heart-rate-variability-hrv-during-sleep}} sensor uses ballistocardiography (BGC), a measure of ballistic forces generated by the heart, to measure heart rate variability during sleep when placed underneath one's mattress.
Additionally, device-free human sensing is also gaining traction in many domains, including health \cite{xiao_survey_2022}.
Examples include the use of radio frequency signal reflections and WiFi channel state information to passively monitor vital signs \cite{kumar_cnn-based_2022} and detect falls \cite{tian_rf-based_2018} without requiring physical devices.
Additionally, ambient sensors are increasingly being incorporated in health monitoring to monitor the state of the external environment, such as temperature, humidity, and air quality, as these factors have a significant impact on human health \cite{dias_wearable_2018,cusack_reviewsmart_2024}. 

Sensors, while essential for personal health monitoring, also contribute significantly to the identified challenges.
The heterogeneity of sensor devices, observation data, and measurement procedures can hinder interoperability in personal health monitoring systems \cite{compton_ssn_2012}.
The dynamicity and complexity of sensor data requires expert knowledge to interpret and analyse it. 
This affects both situation analysis and decision support.
Furthermore, sensors can contribute to uncertainty.
Data is uncertain when the degree of confidence about what is stated by the data is less than 100\% \cite{khaleghi_multisensor_2011}. 
This can arise when there is missing data or when all the relevant attributes cannot be measured by the available sensors \cite{gravina_multi-sensor_2017}.
Some of these challenges can be addressed by the incorporation of Semantic Web technologies.

\subsection{Semantic Web technologies}

\label{semantic-web-technologies}
The Semantic Web provides foundational mechanisms for knowledge representation and reasoning, thereby playing an important role in the design of AI-driven health monitoring systems.
Three overlapping Semantic Web technologies have emerged as the most prominent over the years: ontologies, knowledge graphs, and linked data \cite{hitzler_semantic_2021}.
We will begin with an overview of key Semantic Web languages and standards, after which we will discuss each of the three technologies in turn, highlighting how they contribute to health monitoring.

\subsubsection{Languages and standards}

The development of Semantic Web technologies is facilitated using different languages and standards.
Resource Description Framework (RDF)\footnote{\url{https://www.w3.org/RDF}}, a standard for the description and exchange of interconnected data in the form of subject-predicate-object triples, can be considered one of the core building blocks of the Semantic Web.
As RDF is an abstract data model, it can be serialised in different formats, including N-Triples, Terse RDF Triple Language (Turtle), eXtensible Markup Language (XML), and JavaScript Object Notation for Linked Data (JSON-LD) \cite{schreiber_rdf_2014}.
Several extensions to RDF have been proposed. 
These include RDF Schema (RDFS)\footnote{\url{https://www.w3.org/wiki/RDFS}}, which provides a vocabulary to enrich RDF data;
RDF-star\footnote{\url{https://w3c.github.io/rdf-star/cg-spec}}, which allows an RDF triple to be embedded as the subject or object of another triple, without necessarily asserting the embedded triple, thereby enabling richer metadata annotation;
and the Notation3\footnote{\url{https://w3c.github.io/N3/spec}} specification, which extends the representational abilities of RDF by supportive declarative programming and allowing the access of online knowledge.
Other important standards in the Semantic Web community are:
Web Ontology Language (OWL)\footnote{\url{https://www.w3.org/OWL}}, a language for constructing ontologies;
Semantic Web Rule Language (SWRL)\footnote{\url{https://www.w3.org/Submission/SWRL}}, a language for expressing rules and logic; 
Shapes Constraint Language (SHACL)\footnote{\url{https://www.w3.org/TR/shacl}}, a language for describing RDF graphs, which also includes a rules language;
and SPARQL Protocol and RDF Query Language (SPARQL)\footnote{\url{https://www.w3.org/TR/rdf-sparql-query}}, a language for retrieving and manipulating RDF data.
SPARQL-star extends SPARQL to allow querying and updating of RDF-star data, while SPARQL Inferencing Notation (SPIN)\footnote{\url{https://spinrdf.org}} is a rules language based on SPARQL.

\subsubsection{Ontologies}
Arguably, the key technology underpinning the Semantic Web is ontologies, which have been widely used for reasoning and representation in sensor-based systems \cite{ye_situation_2011}. 
Their ability to represent knowledge formally and unambiguously not only enhances interoperability but is also useful in capturing the domain knowledge necessary for situation analysis and subsequent decision support.
Several ontologies have been developed to support the description of sensors and their observations, which is critical in any sensor-based system.
Two particularly prominent sensor ontologies are the semantic sensor network (SSN) ontology \cite{compton_ssn_2012} and the Smart Appliances REFerence (SAREF)\footnote{\url{https://saref.etsi.org}}  ontology \cite{garcia_etsi_2023}.
Both are standardised ontologies developed by the World Wide Web Consortium (W3C) and the European Telecommunication Standardization Institute (ETSI) respectively, with the aim of enabling semantic interoperability.
However, while SSN was developed for sensors and sensor-based systems in general, SAREF focuses on smart appliances and IoT devices.
The latest version of SSN is based on the Sensor, Observation, Sample, and Actuator (SOSA) ontology \cite{janowicz_sosa_2019}, which provides it with a lightweight, user-friendly, and extendable core.
SAREF has mappings to SSN, from which it borrows modelling patterns for several classes \cite{garcia_etsi_2023,moreira_saref4health_2020}.

As domain-agnostic ontologies, both SSN and SAREF require augmentation to meet application-specific requirements \cite{poveda-villalon_ontological_2020}.
SAREF provides a suite of ontologies that extend the core ontology for different domains, including two that are relevant for personal health monitoring: SAREF4EHAW\footnote{\url{https://saref.etsi.org/saref4ehaw}} for eHealth and ageing well and SAREF4WEAR\footnote{\url{https://saref.etsi.org/saref4wear}} for wearable devices.
SAREF4EHAW provides support for modelling concepts such as health system actors (including patients and caregivers) and health devices (including wearables), with the wearable concept linked to the SAREF4WEAR ontology.
An additional extension, SAREF4Health, was developed to address the limitations of SSN and SAREF in representing real-time ECG time series data exchanged between mobile devices and cloud gateways \cite{moreira_saref4health_2020}.
In contrast to SSN, SAREF is targeted at industry developers rather than ontology experts \cite{moreira_saref4health_2020}, making it more readily adoptable by those without extensive ontology development experience.
Furthermore, its extensions for the health domain provide a solid foundation for building semantic personal health monitoring systems.
Additional representational support can be obtained by integrating resources such as standardized clinical terminologies and medical knowledge bases.

\subsubsection{Knowledge graphs}

\label{knowledge-graphs}

A knowledge graph can generally be understood as a knowledge base of real-world data represented in a graph-based data model.
Ontologies are a vital building block in many knowledge graphs, and are used to define their data schema (such as properties, restrictions, and relationships) as well as enable semantic reasoning and entailment \cite{hogan_knowledge_2022}.
Knowledge graphs have seen increasingly widespread use in the health domain.
Their graph structure enables the conceptualisation, representation, and integration of data \cite{hogan_knowledge_2022}.
This is advantageous in health monitoring systems, where the integration of various sources of health data is critical.
An example of this is the Precision Medicine Knowledge Graph (PrimeKG), which integrates diverse biomedical data from multiple sources with the goal of enabling precision medicine analyses \cite{chandak_building_2023}.
Previous research has also explored the automatic construction of knowledge graphs from electronic health records \cite{rotmensch_learning_2017,chen_robustly_2019}, which can then be used for clinical decision support. 
This is related to personal health knowledge graphs, which are used to represent and reason over individual health data, including data from sensors \cite{gyrard_interdisciplinary_2022} and electronic health records \cite{jiang_graphcare_2024,tao_mining_2020}.
Additionally, knowledge graphs have been proposed for drug discovery \cite{zeng_toward_2022} and as a tool for explainability in AI-driven health monitoring systems \cite{rajabi_knowledge_2022,lecue_role_2020}.
Knowledge graphs have also proven useful in sensor-based systems, for example by providing graph-based visualisations of the data generated by IoT devices, which can then be queried in real time \cite{le-phuoc_graph_2016}.

\subsubsection{Linked data}

Both knowledge graphs and ontologies can be published using a linked data approach \cite{hitzler_semantic_2021}, whereby uniform resource identifiers (URIs) are used to identify distinct resources \cite{bizer_linked_2011}.
When the emphasis is on free use, modification, and sharing, it is referred to as Linked Open Data \cite{hitzler_semantic_2021}.
Linked data has been proposed for augmenting and representing sensor data in order to improve its accessibility and interoperability \cite{yu_using_2015}.
In the health domain, it has been explored in applications ranging from drug discovery \cite{gray_applying_2014} to the representation of electronic health records \cite{pathak_using_2013}.
Linked data can contribute to interoperability by ensuring heterogeneous health data is stored in a consistent format and structure.
However, its use in health monitoring is not well explored in the literature. 

\section{Related reviews}
\label{related-reviews}
\begin{figure}[ht]
\includegraphics[width=7cm]{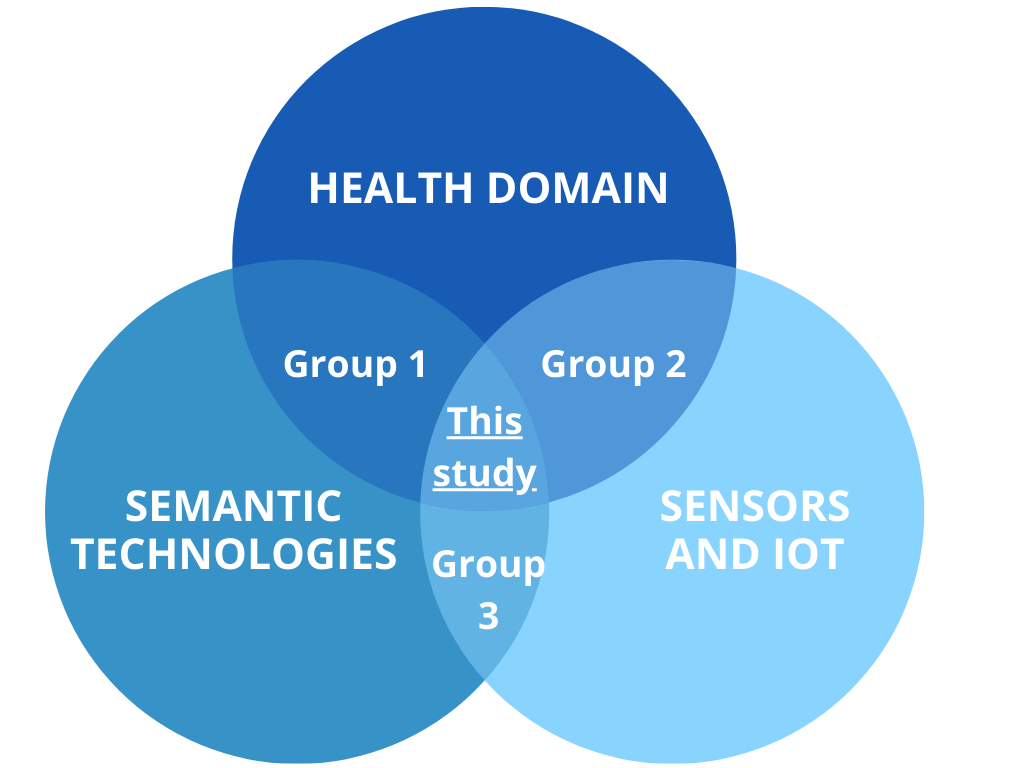}
\caption{Venn diagram illustrating the three focus areas of this study as well as the different groups of related reviews.\label{venn_diagram}}
\end{figure}

Several reviews related to sensors, Semantic Web technologies, and the health domain have been published.
These reviews can generally be categorised into three overlapping groups, which are illustrated as a Venn diagram in Figure~\ref{venn_diagram}.
The reviews in Group 1 focus on the use of Semantic Web technologies in the health domain; those in Group 2 review the use of sensors and IoT in the health domain; those in Group 3 review the use of Semantic Web technologies with sensor and IoT data; and finally,
Group 4 consists of other related reviews that do not fit neatly into any of the first three groups.
The related reviews are discussed in detail in the remainder of this section and summarised in Table~\ref{related-reviews-table}. 

\subsection{Group 1: Semantic Web technologies in the health domain}
This group of reviews explores the use of Semantic Web technologies in healthcare.
Zenuni et al. \cite{zenuni_state_2015} review ontologies and semantic data repositories used in different aspects of the health domain, including hospital systems and health datasets.
A similar review is conducted by Haque et al. \cite{haque_semantic_2022}, who explore themes such as e-healthcare, disease diagnosis, and information management.
Peng et al. \cite{peng_literature_2020} and Hammad et al. \cite{hammad_semantic-based_2020} focus on semantic approaches for health data integration and management, including data from wearable devices.
Dimitrieski et al. \cite{dimitrieski_survey_2016} review ontologies and ontology alignment approaches in healthcare, while
Jing et al. \cite{jing_ontologies_2023} focus on ontologies for rule management in clinical decision support systems.
More recently, Amar et al. \cite{amar_electronic_2024} examine semantic interoperability issues in electronic health records based on the Fast Healthcare Interoperability Resources (FHIR) standard, highlighting how RDF and OWL can improve interoperability. 
The review by Miranda et al. \cite{miranda_semantic_2024} explores how Semantic Web technologies can be used to enhance the interoperability and management of electronic health records in healthcare systems.
Similarly, Wu et al. \cite{wu_semantics-driven_2025} analyse recent work at the intersection of Semantic Web technologies and electronic health records, with a particular focus on how such technologies can improve data quality.
Finally, Cui et al. \cite{cui_review_2025} provide a comprehensive review of healthcare knowledge graphs, discussing their construction and use in a wide range of health applications, including clinical decision support and pharmaceutical research.
Although the reviews in this group provide a good overview of the ways in which Semantic Web technologies have been used in the health domain, five of them do not mention sensors at all, while the remaining five reviews do not include sensor data as a major focus.

\subsection{Group 2: Sensors and IoT in the health domain}
This group considers the use of sensors and IoT in the health domain.
Islam et al. \cite{islam_internet_2015} and Yin et al. \cite{yin_internet_2016} conduct general surveys on IoT for healthcare, covering a broad range of considerations on the topic including networks, communication standards and protocols, and cybersecurity.
The review by Qi et al. \cite{qi_advanced_2017} focuses on the use of IoT in personalised healthcare systems, including sensor devices and data processing techniques.
Philip et al. \cite{philip_internet_2021} explore advances in the field such as cloud computing, while  
Albahri et al. \cite{albahri_real-time_2018} focus on health monitoring systems for telemedicine applications, highlighting techniques that support the connection of hospital services to remote patients.
There have also been reviews specifically focusing on the state of the art in wearable sensors for health monitoring, such as those by Babu et al. \cite{babu_wearable_2024}, Cusack et al. \cite{cusack_reviewsmart_2024}, Dias and Cunha \cite{dias_wearable_2018}, and
Majumder et al. \cite{majumder_wearable_2017}.
Kim et al. \cite{kim_wearable_2019} hone in on biosensors that detect biofluids, such as sweat and tears, while
Baig et al. \cite{baig_systematic_2017} highlight the potential of remote monitoring systems for clinical adoption.
Punj and Kumar \cite{punj_technological_2019}, Banaee et al. \cite{banaee_data_2013}, and Andreu-Perez et al. \cite{andreu-perez_wearable_2015} explore advances in wearable sensor data collection, mining, and processing, and
Dang et al. \cite{dang_human-centred_2023} focus on statistical analysis and machine learning (ML) as modelling tools.
Bollineni et al. \cite{bollineni_iot_2025} adopt a forward-looking perspective in their review, highlighting both emerging technologies and future prospects for health-IoT architectures.
While these reviews provide useful analyses on the role of sensors and IoT in health monitoring, ten of them do not mention Semantic Web technologies, while the remaining six do so briefly without an in-depth analysis of their role in health monitoring.

\subsection{Group 3: Semantic Web technologies for sensors and IoT}
This group reviews the intersection between Semantic Web technologies and sensors without being limited to a particular domain.
Honti and Abonyi \cite{honti_review_2019} and Rhayem et al. \cite{rhayem_semantic_2020} explore the use of ontologies in IoT-based systems in different domains.
Bajaj et al. \cite{bajaj_study_2017} adopt a similar focus on ontologies, reviewing both general sensor ontologies as well as domain-specific ones for IoT.
Compton et al. \cite{compton_survey_2009} present a review of the semantic specification of sensors using ontologies, analysing the range and expressive power of sensor ontologies. 
The review by Harlamova et al. \cite{harlamova_survey_2017} explores the challenges in the use of Semantic Web technologies in IoT, while
Ye et al. \cite{ye_semantic_2015} review the application of Semantic Web technologies in pervasive and sensor-driven systems.
Although these reviews highlight the use of Semantic Web technologies with sensors and IoT, they are not specific to the health domain.

\subsection{Group 4: Other reviews related to AI and technology in the health domain}
A small number of reviews take a broader lens and consider different aspects of AI and technology in the health domain.
This includes the concept of Healthcare 4.0, a term referring to the increasing digitisation of the healthcare industry.
The reviews by Tortorella et al. \cite{tortorella_healthcare_2020} and Jayaraman et al. \cite{jayaraman_healthcare_2020} broadly cover Healthcare 4.0, and highlight health monitoring systems that use IoT and sensors.
However, only the review by Jayaram et al. \cite{jayaraman_healthcare_2020} mentions ontologies and other knowledge representation techniques.
More recent reviews, such as those by Rahman et al. \cite{rahman_ai_2025} and Rashid and Nemati \cite{rashid_human-centered_2024}, have begun addressing the transition toward Healthcare 5.0, which emphasizes explainability and human-centricity in health systems.
While both of these reviews mention the role of sensors, IoT, and AI in healthcare, neither mention the Semantic Web.
Another review in this group is by Kumar et al. \cite{kumar_artificial_2023}, who discuss AI in healthcare. Although they mention IoT and knowledge graphs, neither of these are the focus of the review.
Lastly, the review by Behera et al. \cite{behera_emerging_2019} focuses on techniques used to create healthcare systems modelled on human cognitive processes such as perception and thought. 
They highlight cognitive IoT as a future research direction through wearable sensors, while also mentioning Semantic Web technologies for knowledge representation.
However, neither the Semantic Web technologies nor sensors are discussed in detail.

\subsection{Summary}
Table~\ref{related-reviews-table} summarises the related reviews. The current study differs from existing work by focusing on the use of sensors and Semantic Web technologies for personal health monitoring, with both sensor data and Semantic Web technologies being primary points of focus.
Additionally, the majority of the related reviews and surveys do not take a systems perspective, whereas this study highlights how the different system components are integrated and discusses the development methodologies and tools, evaluation approaches, and architectures of the included systems.

\footnotesize
\setlength{\extrarowheight}{2pt}
\begin{longtable}{lllccc}
\caption{Summary of related reviews and their focus areas.} \label{related-reviews-table}\\
\hline
\textbf{Group} & \textbf{Review} & \textbf{Year} & \bfseries{\makecell[l]{Semantic Web \\technologies}} & \bfseries{\makecell[l]{Healthcare/\\health monitoring}} & \bfseries{\makecell[l]{Sensors/IoT}}\tabularnewline
\hline
\endfirsthead

\multicolumn{6}{c}{\tablename\ \thetable\ -- continued from previous page} \\
\hline
\textbf{Group} & \textbf{Review} & \textbf{Year} & \bfseries{\makecell[l]{Semantic Web \\technologies}} & \bfseries{\makecell[l]{Healthcare/\\health monitoring}} & \bfseries{\makecell[l]{Sensors/IoT}}\tabularnewline
\hline
\endhead

\hline
\multicolumn{4}{>{\raggedright\arraybackslash}p{0.715\NetTableWidth}}{\makecell[l]{\textbf{\cmark} - the area is a main focus area of the review \\\textbf{\omark} - the area is partially addressed, but is neither discussed in depth nor a focus area of the review \\\textbf{\xmark} - the area is not addressed at all in the review}} & \multicolumn{2}{>{\raggedleft\arraybackslash}p{0.285\NetTableWidth}}{\textbf{Table continued on next page.}}\\
\endfoot

\hline
\multicolumn{6}{>{\raggedright\arraybackslash}p{\NetTableWidth}}{\makecell[l]{\textbf{\cmark} - the area is a main focus area of the review \\\textbf{\omark} - the area is partially addressed, but is neither discussed in depth nor a focus area of the review \\\textbf{\xmark} - the area is not addressed at all in the review}}
\endlastfoot

\multirow{2}{*}{\makecell[l]{1. Semantic Web technologies \\in the health domain}} & Amar et al. \cite{amar_electronic_2024} & 2024 & \cmark & \cmark & \omark \tabularnewline
& \cellcolor{gray!4}Cui et al. \cite{cui_review_2025} & \cellcolor{gray!4}2025 & \cellcolor{gray!4}\cmark & \cellcolor{gray!4}\cmark & \cellcolor{gray!4}\xmark \tabularnewline
\multirow{8}{*}{\makecell[l]{1. Semantic Web technologies \\in the health domain}} & Dimitrieski et al. \cite{dimitrieski_survey_2016} & 2016 & \cmark & \cmark & \xmark \tabularnewline
& \cellcolor{gray!4}Hammad et al. \cite{hammad_semantic-based_2020} & \cellcolor{gray!4}2020 & \cellcolor{gray!4}\cmark & \cellcolor{gray!4}\cmark & \cellcolor{gray!4}\omark \tabularnewline
& Haque et al. \cite{haque_semantic_2022} & 2022 & \cmark & \cmark & \omark \tabularnewline
& \cellcolor{gray!4}Jing et al. \cite{jing_ontologies_2023} & \cellcolor{gray!4}2023 & \cellcolor{gray!4}\cmark & \cellcolor{gray!4}\cmark & \cellcolor{gray!4}\xmark \tabularnewline
& Miranda et al. \cite{miranda_semantic_2024} & 2024 & \cmark & \cmark & \xmark \tabularnewline
& \cellcolor{gray!4}Peng et al. \cite{peng_literature_2020} & \cellcolor{gray!4}2020 & \cellcolor{gray!4}\cmark & \cellcolor{gray!4}\cmark & \cellcolor{gray!4}\omark \tabularnewline
& Wu et al. \cite{wu_semantics-driven_2025} & 2025 & \cmark & \cmark & \omark \tabularnewline
& \cellcolor{gray!4}Zenuni et al. \cite{zenuni_state_2015} & \cellcolor{gray!4}2015 & \cellcolor{gray!4}\cmark & \cellcolor{gray!4}\cmark & \cellcolor{gray!4}\xmark \tabularnewline
\greyline
\multirow{16}{*}{\makecell[l]{2. Sensors and IoT in \\the health domain}} & Albahri et al. \cite{albahri_real-time_2018} & 2018 & \xmark & \cmark & \cmark \tabularnewline 
& \cellcolor{gray!4}Andreu-Perez et al. \cite{andreu-perez_wearable_2015} & \cellcolor{gray!4}2015 & \cellcolor{gray!4}\omark & \cellcolor{gray!4}\cmark & \cellcolor{gray!4}\cmark \tabularnewline
& Babu et al. \cite{babu_wearable_2024} & 2024 & \xmark & \cmark & \cmark \tabularnewline
& Baig et al. \cite{baig_systematic_2017}  & 2017 & \xmark & \cmark & \cmark \tabularnewline  
& \cellcolor{gray!4}Banaee et al. \cite{banaee_data_2013} & \cellcolor{gray!4}2013 & \cellcolor{gray!4}\omark & \cellcolor{gray!4}\cmark & \cellcolor{gray!4}\cmark \tabularnewline
& Bollineni et al. \cite{bollineni_iot_2025} & 2025 & \xmark & \cmark & \cmark \tabularnewline
& \cellcolor{gray!4}Cusack et al. \cite{cusack_reviewsmart_2024} & \cellcolor{gray!4}2024 & \cellcolor{gray!4}\xmark & \cellcolor{gray!4}\cmark & \cellcolor{gray!4}\cmark \tabularnewline
& Dang et al. \cite{dang_human-centred_2023} & 2023 & \xmark & \cmark & \cmark \tabularnewline
& \cellcolor{gray!4}Dias and Cunha \cite{dias_wearable_2018} & \cellcolor{gray!4}2018 & \cellcolor{gray!4}\xmark & \cellcolor{gray!4}\cmark & \cellcolor{gray!4}\cmark \tabularnewline
& Islam et al. \cite{islam_internet_2015} & 2015 & \omark & \cmark & \cmark \tabularnewline
& \cellcolor{gray!4}Kim et al. \cite{kim_wearable_2019} & \cellcolor{gray!4}2019 & \cellcolor{gray!4}\xmark & \cellcolor{gray!4}\cmark & \cellcolor{gray!4}\cmark \tabularnewline
& Majumder et al. \cite{majumder_wearable_2017} & 2017 & \xmark & \cmark & \cmark \tabularnewline
& \cellcolor{gray!4}Philip et al. \cite{philip_internet_2021} & \cellcolor{gray!4}2021 & \cellcolor{gray!4}\omark & \cellcolor{gray!4}\cmark & \cellcolor{gray!4}\cmark \tabularnewline
& Punj and Kumar \cite{punj_technological_2019} & 2019 & \xmark & \cmark & \cmark \tabularnewline
& \cellcolor{gray!4}Qi et al. \cite{qi_advanced_2017} & \cellcolor{gray!4}2017 & \cellcolor{gray!4}\omark & \cellcolor{gray!4}\cmark & \cellcolor{gray!4}\cmark \tabularnewline
& Yin et al. \cite{yin_internet_2016} & 2016 & \omark & \cmark & \cmark \tabularnewline
\greyline
\multirow{6}{*}{\makecell[l]{3. Semantic Web technologies \\for sensors and IoT}} & Bajaj et al. \cite{bajaj_study_2017} & 2017 & \cmark & \omark & \cmark \tabularnewline    
& \cellcolor{gray!4}Compton et al. \cite{compton_survey_2009} & \cellcolor{gray!4}2009 & \cellcolor{gray!4}\cmark & \cellcolor{gray!4}\xmark & \cellcolor{gray!4}\cmark \tabularnewline
& Harlamova et al. \cite{harlamova_survey_2017} & 2017 & \cmark & \omark & \cmark \tabularnewline
& \cellcolor{gray!4}Honti and Abonyi \cite{honti_review_2019} & \cellcolor{gray!4}2019 & \cellcolor{gray!4}\cmark & \cellcolor{gray!4}\omark & \cellcolor{gray!4}\cmark \tabularnewline
& Rhayem et al. \cite{rhayem_semantic-enabled_2021} & 2020 & \cmark & \xmark & \cmark \tabularnewline
& \cellcolor{gray!4}Ye et al. \cite{ye_semantic_2015} & \cellcolor{gray!4}2015 & \cellcolor{gray!4}\cmark & \cellcolor{gray!4}\xmark & \cellcolor{gray!4}\cmark \tabularnewline
\greyline
\multirow{6}{*}{\makecell[l]{4. Other related reviews}} & Behera et al. \cite{behera_emerging_2019} & 2019 & \omark & \cmark & \omark \tabularnewline   
& \cellcolor{gray!4}Jayaraman et al. \cite{jayaraman_healthcare_2020} & \cellcolor{gray!4}2020 & \cellcolor{gray!4}\omark & \cellcolor{gray!4}\cmark & \cellcolor{gray!4}\omark \tabularnewline
& Kumar et al. \cite{kumar_artificial_2023} & 2023 & \omark & \cmark & \omark \tabularnewline
& \cellcolor{gray!4}Rahman et al. \cite{rahman_ai_2025} & \cellcolor{gray!4}2025 & \cellcolor{gray!4}\xmark & \cellcolor{gray!4}\cmark & \cellcolor{gray!4}\cmark \tabularnewline
& Rashid and Nemati \cite{rashid_human-centered_2024} & 2024 & \xmark & \cmark & \cmark \tabularnewline
& \cellcolor{gray!4}Tortorella et al. \cite{tortorella_healthcare_2020} & \cellcolor{gray!4}2020 & \cellcolor{gray!4}\xmark & \cellcolor{gray!4}\cmark & \cellcolor{gray!4}\omark \tabularnewline
\hline
\multicolumn{2}{c}{\textbf{This study}} & \textbf{2025} & \cmark & \cmark & \cmark \tabularnewline
\end{longtable}
\normalsize

\section{Methodology}
\label{methodology}
\subsection{Objectives and reporting strategy}
In order to achieve our goal of mapping the state of the art in the use of Semantic Web technologies in sensor-based personal health monitoring systems, the following are the objectives of this study:
\begin{enumerate}
    \item To systematically select systems that represent the state of the art in the use of Semantic Web technologies in sensor-based personal health monitoring systems.
    \item To determine the extent to which the seven key challenges are addressed by the selected systems.
    \item To assess the role and limitations of Semantic Web technologies in addressing these challenges. 
    
    \item To conduct a comprehensive quality assessment of selected systems based on data and devices used, system and component development, evaluation rigour, and accessibility of research outputs.
    \item{To propose a reference architecture that provides guidance for the design and development of new systems.}
    
    \item To highlight inadequacies in existing systems and provide recommendations for future research.
\end{enumerate}
The study was conducted and is reported using the Preferred Reporting Items for Systematic Reviews and Meta-Analyses (PRISMA) \cite{moher_preferred_2009} framework.
To further ensure the quality of the study, we adhered to the following quality assessment criteria as described by Kitchenham et al. \cite{kitchenham_systematic_2010}:

\begin{enumerate}
    \item ``The inclusion criteria are explicitly defined in the paper'': The inclusion and exclusion criteria are specified in Section~\ref{inclusion-criteria}.
    \item ``The authors have either searched four or more digital libraries and included additional search strategies or identified and referenced all journals addressing the topic of interest'': Six digital libraries were searched, and additional records were identified by using the preliminary search results and related reviews to search for similar studies. More details on the search strategy are given in Section~\ref{strategy}. 
    \item ``The authors have explicitly defined quality criteria and extracted them from each primary study'': The systems are analysed in Section~\ref{key-challenges} based on the seven identified challenges and the criteria is outlined in Table~\ref{challenges-aspects}. 
    Additionally, the quality of each system is assessed and discussed in Section~\ref{system-quality} based on criteria outlined in Table~\ref{quality-critera}.
    \item ``Information is presented about each paper so that the data summaries can clearly be traced to relevant papers'': A summary of all the included systems is shown in Table~\ref{selected-systems}, with all systems fully cited. 
    A GitHub repository\footnote{\url{https://github.com/mbithenzomo/semantic_phms_mapping_study}} has been created for this study, which includes copies and links of the selected papers and other supplementary material.
\end{enumerate}

\subsection{Search strategy}
\label{strategy}
Six digital libraries were searched: ACM Digital Library\footnote{\url{https://dl.acm.org}}, IEEE Xplore\footnote{\url{https://ieeexplore.ieee.org/Xplore}}, PubMed\footnote{\url{https://pubmed.ncbi.nlm.nih.gov}}, ScienceDirect\footnote{\url{https://www.sciencedirect.com}}, Scopus\footnote{\url{https://www.scopus.com}}, and Web of Science\footnote{\url{https://www.webofscience.com}}. 
An initial search was conducted between 9th and 12th February 2024, and a second search was conducted between 14th and 15th August 2025 to ensure inclusion of the most recent literature.
Abstracts, titles, and/or keywords were searched using terms related to the topic of the study, at the intersection of five areas: Semantic Web technologies, sensors, the health domain, monitoring, and systems.
The search strings used are shown in Table~\ref{library-search}.
Boolean operators were used for a more specific search, although the ScienceDirect library had a limit on the number of Boolean operators that could be used per search. This library also did not allow the use of wildcard characters.
To ensure a state of the art study, all results were filtered to only include literature published in or after 2012 during the first search, and between 2024 and 2025 during the second search, thus covering the period between 2012 and 2025.
Additionally, where possible, the results were filtered to only include conference papers and journal articles published in English.
This filtered out other types of literature such as surveys and reviews, books and book chapters, research abstracts, posters and conference proceedings, as well as articles written in languages other than English.
The first search yielded 960 records, while the second search yielded a further 217 records, resulting in a total of 1,177 records from the digital library search. 

\begin{table*}[ht]
\caption{Search strings used in digital library search.} \label{library-search}
\begin{tabular}{ll}
\hline
\textbf{Area} & \textbf{Search strings}\\
\hline
Semantic Web technologies & semantic*, ontolog*, knowledge graph, linked data\\
\rowcolor{gray!4}Sensors & sensor*, iot, internet of things, wearable*, device*, body area network\\
Health domain & health*, medic*\\
\rowcolor{gray!4}Monitoring & monitor*, track*, remote, tele*, distributed, continuous, daily\\
Systems & system, framework, application, architecture\\
\hline
\end{tabular}
\end{table*}

The selected results from the digital library search, together with the related review articles discussed in Section~\ref{related-reviews}, were then used to identify further potentially relevant studies 
through a related paper search. 
This was done using two online tools, Connected Papers\footnote{\url{https://www.connectedpapers.com}} and Semantic Scholar\footnote{\url{https://www.semanticscholar.org}}. 
The first related paper search in February 2024 identified 62 records while the second one in August 2025 identified 22 records, resulting in 84 additional records.
Thus, the entire search process yielded a total of 1,261 records, which were then assessed for eligibility.
Rayyan{\footnote{\url{https://www.rayyan.ai}}}\cite{ouzzani_rayyanweb_2016}, an online tool for the management of systematic reviews, was used to facilitate the screening and assessment process.
The search results from each phase of the search process are summarized in Table~\ref{search-results}.

\begin{table*}[ht]
\caption{Summary of search results from both the first (February 2024) and the second (August 2025) searches.} \label{search-results}

\begin{tabular}{lllll}
\hline
\textbf{Search type} & \textbf{Digital library/Online tool} & \textbf{First search} & \textbf{Second search} & \textbf{Total}\\
\hline
\multirow{7}{*}{Digital library search} & ACM Digital Library & 171 & 41 & 212\\
& \cellcolor{gray!4}IEEE Xplore & \cellcolor{gray!4}202 & \cellcolor{gray!4}29 & \cellcolor{gray!4}231\\
& PubMed & 90 & 19 & 109\\
& \cellcolor{gray!4}ScienceDirect & \cellcolor{gray!4}74 & \cellcolor{gray!4}36 & \cellcolor{gray!4}110\\
& Scopus & 202 & 35 & 237\\
& \cellcolor{gray!4}Web of Science & \cellcolor{gray!4}221 & \cellcolor{gray!4}57 & \cellcolor{gray!4}278\\
& \textbf{Total} & \textbf{960} & \textbf{217} & \textbf{1,177}\\
\greyline
\multirow{3}{*}{Related papers search} & Connected Papers & 43 & 16 & 59 \\
& \cellcolor{gray!4}Semantic Scholar & \cellcolor{gray!4}19 & \cellcolor{gray!4}6 & \cellcolor{gray!4}25\\
& \textbf{Total} & \textbf{62} & \textbf{22} & \textbf{84} \\
\greyline
\multicolumn{2}{l}{\textbf{Total number of identified records}} & \textbf{1,022} & \textbf{239} & \textbf{1,261} \\
\hline
\end{tabular}
\end{table*}

\subsection{Inclusion and exclusion criteria}
\label{inclusion-criteria}
This study includes only peer-reviewed journal articles and conferences papers written in English. 
Further, we only include systems that incorporate one of the three Semantic Web technologies (i.e. an ontology, knowledge graph, or the explicit use of linked data) as an integral system component, with a clear description of the technical implementation. Systems lacking sufficient implementation details were excluded, as this would compromise the quality assessment.
Additionally, because a system consists of several integrated components, studies reporting the development of only one component (for example, an ontology) were excluded.
Of particular interest to this study are sensors that measure physiological data (that is, biosignals and vital signs) and/or physical activity data (for example, daily step count).
Applications of sensors outside health monitoring , such as activity recognition, fitness, or nutrition, were excluded.
Furthermore, systems that do not have an analysis, inferencing, or reasoning component were also excluded. 
These inclusion and exclusion criteria are summarised in Table~\ref{criteria}.

\footnotesize
\begin{longtable}{m{.035\NetTableWidth}>{\raggedright}m{.18\NetTableWidth}>{\raggedright}m{.38\NetTableWidth}>{\raggedright}m{.38\NetTableWidth}}
\caption{Inclusion and exclusion criteria.} \label{criteria}\\
\hline
\textbf{\#} & \textbf{Criteria} & \textbf{Inclusion Criteria} & \textbf{Exclusion Criteria}\tabularnewline
\hline
\endfirsthead

\multicolumn{4}{c}{\tablename\ \thetable\ -- continued from previous page} \\
\hline
\textbf{\#} & \textbf{Criteria} & \textbf{Inclusion Criteria} & \textbf{Exclusion Criteria}\tabularnewline
\hline
\endhead

\hline
\multicolumn{4}{r}{\textbf{Table continued on next page.}}\\
\endfoot

\hline
\endlastfoot

C1 & Publication year & The year of publication is 2012 or later. & The year of publication is earlier than 2012.\tabularnewline
\rowcolor{gray!4}
C2 & Language & The publication is written in English. & The publication is written in a language other than English.\tabularnewline
C3 & Publication type & The publication is a peer-reviewed journal article or conference paper reporting original research. & The publication is either not peer-reviewed (e.g. research abstracts, posters, books, and keynotes), is a collection of works (e.g. conference proceedings), or does not report original research (e.g. reviews, surveys, and position papers).\tabularnewline
\rowcolor{gray!4}
C4 & Accessibility & The publication is open access or can otherwise be accessed by the authors, e.g. through institutional access. & The publication cannot be accessed without additional payment.\tabularnewline
C5 & Multiple integrated components & The publication must report on a system, framework, application, or architecture consisting of several integrated components. & Studies reporting the development of only one component (e.g. an ontology).\tabularnewline
\rowcolor{gray!4}
C6 & Semantic Web technologies & The system incorporates at least one of the three Semantic Web technologies as an integral component. & Semantic Web technologies either do not form an integral component of the system or else their technical implementation is insufficiently described.\tabularnewline
C7 & Health monitoring & The system focuses on {personal} health monitoring. & The system has a focus outside the domain of human health, or is related to health but does not focus on health monitoring (e.g. systems focusing solely on other areas such as activity recognition, sports, fitness, and nutrition).\tabularnewline
\rowcolor{gray!4}
C8 & Sensors for physiological and/or physical activity data & The system incorporates sensors that measure physiological data (i.e. biosignals and vital signs) and/or physical activity data (e.g. daily steps). & The system does not incorporate sensors or the sensors incorporated do not measure physiological or physical activity data.\tabularnewline
C9 & Analysis \& Reasoning & The system has an analysis, inferencing, or reasoning component. & The system does not analyse or reason over the sensor data.\tabularnewline
\rowcolor{gray!4}
C10 & Extended work & If the system has been extended in later work, the more recent version is included in the review. & The system is extended in later work.\tabularnewline
\end{longtable}
\normalsize

\subsection{Selection results}

From the 1,261 identified records, 366 duplicates were removed resulting in 895 unique records.
Next, preliminary screening was done by reviewing the title and abstract of each record.
At this stage, records were excluded for reasons such as not being focused on the health domain or not involving health monitoring.
We also found that a number of records had bypassed some of the filters that were applied in the initial identification stage, such as publication year and language.
664 records were excluded based on the title and abstract screening.
The remaining 231 papers were read in full to determine if they still met the inclusion criteria.
One reason for exclusion at this stage was if the system had been extended in later work and the extension was one of the systems being assessed.
In such cases, the extension was included in the study while the previous work was excluded.
Additionally, a small number of publications were excluded for reasons such as the full text being inaccessible without additional payment or the article being retracted.
Ultimately, 48 systems were selected for inclusion in this study.
Figure~\ref{flow_diagram} shows a PRISMA flow diagram illustrating the identification, screening, eligibility, and inclusion process.
The diagram also provides details on the specific reasons for exclusion and the corresponding number of publications excluded at each stage.

\begin{figure}[ht]
\includesvg[width=15cm]{figures/flow_diagram.svg}
\caption{PRISMA flow diagram outlining the selection process, combining the results from both the .\label{flow_diagram}}
\end{figure}

\subsection{Summary of selected systems}
\begin{figure}[ht]
\includesvg[width=9cm]{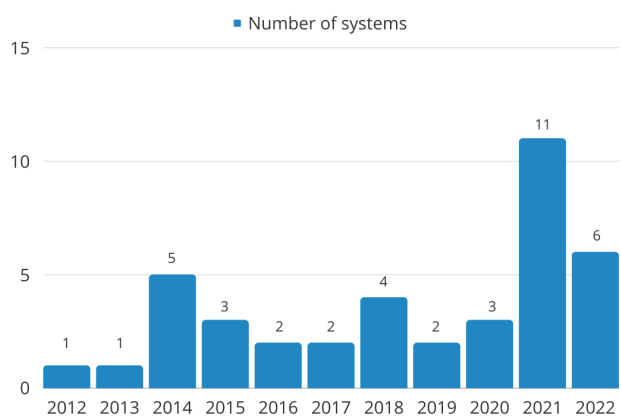}
\caption{Bar graph showing the distribution of the systems by year of publication.\label{publication-year}}
\end{figure}

A summary of the 48 selected systems is shown in Table~\ref{selected-systems}, while Figure~\ref{publication-year} shows the distribution of the systems according to the publication year. 
The year of publication ranges from 2012 to 2025, with 2021 being the most common.
In terms of the application area, 29 focus on a particular disease or diseases, while the remaining 19 provide a solution for general health monitoring.
Additionally, 17 of the systems mention elderly people as a target user \cite{bampi_ontology-driven_2025,chiang_context-aware_2015,elhadj_do-care_2021,esposito_smart_2018,garcia-moreno_systematic_2023,garcia-valverde_heart_2014,henaien_combined_2020,hooda_semantic_2020,ivascu_activity-aware_2021,ivascu_multi-agent_2015,kilintzis_supporting_2019,spoladore_ontology-based_2021,stavropoulos_detection_2021,titi_ontology-based_2019,vadillo_enhancement_2013,villarreal_mobile_2014,zhou_design_2022}.
Personal health monitoring systems can be classified into three development stages: research prototypes (functional systems validated on synthetic or existing datasets but not tested with real users), user validated (systems that have undergone user studies or real-world testing), and deployed (production-level systems that are currently operational). 
Of the 48 systems, a majority are research prototypes (36 systems), with 12 user validated systems and no deployed system.

Regarding the types of Semantic Web technologies used in the systems, nearly all make use of ontologies. 
The three exceptions are Yu et al. \cite{yu_improving_2022} and Zhou et al. \cite{zhou_design_2022}, who use only knowledge graphs, and Xu et al. \cite{xu_design_2017}, who use linked data and knowledge graphs.
Six systems use multiple Semantic Web technologies: 
Kilintzis et al. \cite{kilintzis_supporting_2019} and Reda et al. \cite{reda_heterogeneous_2022} combine linked data with ontologies; 
Stavropoulos et al. \cite{stavropoulos_detection_2021} and Zafeiropoulos et al. \cite{zafeiropoulos_evaluating_2024} combine knowledge graphs with ontologies; 
Xu et al. \cite{xu_design_2017} combine linked data with knowledge graphs; and 
Ammar et al. \cite{ammar_using_2021} incorporate all three technologies.
Table~\ref{selected-systems} also provides an overview of other complimentary technologies and techniques used, as well as the architecture type of each system.
These aspects are discussed in more detail in Sections~\ref{key-challenges} and ~\ref{architectures} respectively.

\scriptsize
\setlength{\extrarowheight}{2pt}
\begin{longtable}{>{\raggedright\arraybackslash}m{0.015\NetTableWidth}>{\raggedright\arraybackslash}m{0.2\NetTableWidth}>{\raggedright\arraybackslash}m{0.03\NetTableWidth}>{\raggedright\arraybackslash}m{0.1\NetTableWidth}>{\raggedright\arraybackslash}m{0.1\NetTableWidth}>{\raggedright\arraybackslash}m{0.13\NetTableWidth}>{\raggedright\arraybackslash}m{0.15\NetTableWidth}>{\raggedright\arraybackslash}m{0.135\NetTableWidth}}
\caption{Summary of systems selected for this study.} \label{selected-systems}\\
\hline
\textbf{\#} & \textbf{System} & \textbf{Year} & \textbf{Application} & \bfseries\makecell[l]{Development \\Stage} & \bfseries\makecell[l]{Semantic Web \\Technologies} & \bfseries\makecell[l]{Other Technologies \\\& Techniques} & \bfseries\makecell[l]{Architecture \\Type}\tabularnewline
\hline
\endfirsthead

\multicolumn{8}{c}{\tablename\ \thetable\ -- continued from previous page} \\
\hline
\textbf{\#} & \textbf{System} & \textbf{Year} & \textbf{Application} & \bfseries\makecell[l]{Development \\Stage} & \bfseries\makecell[l]{Semantic Web \\Technologies} & \bfseries\makecell[l]{Other Technologies \\\& Techniques} & \bfseries\makecell[l]{Architecture \\Type}\tabularnewline
\hline
\endhead

\hline
\multicolumn{6}{>{\raggedright\arraybackslash}p{0.715\NetTableWidth}}{ASP - Answer set programming; BN - Bayesian network; CBR - Case-based reasoning; CDL - Contextual defeasible logic; COPD - Chronic obstructive pulmonary disease; FL - Fuzzy logic; KG - knowledge graph; ML - machine learning; NLP - Natural language processing} & \multicolumn{2}{>{\raggedleft\arraybackslash}p{0.285\NetTableWidth}}{\textbf{Table continued on next page.}}\\
\endfoot

\hline
\multicolumn{8}{>{\raggedright\arraybackslash}p{\NetTableWidth}}{ASP - Answer set programming; BN - Bayesian network; CBR - Case-based reasoning; CDL - Contextual defeasible logic; COPD - Chronic obstructive pulmonary disease; FL - Fuzzy logic; KG - knowledge graph; ML - machine learning; NLP - Natural language processing}
\endlastfoot

1 & Akhtar et al. \cite{akhtar_multi-agent_2022} & 2022 & Parkinson's disease & Prototype & Ontology & Agents; CDL; rules & Layered; multi-agent \tabularnewline
\rowcolor{gray!4}
2 & Ali et al. \cite{ali_intelligent_2021} & 2021 & Diabetes; abnormal blood pressure & Prototype & Ontology & ML; NLP; rules & Layered \tabularnewline
3 & Ali et al. \cite{ali_smart_2020} & 2020 & Heart disease & Prototype & Ontology & ML; rules & Layered \tabularnewline
\rowcolor{gray!4}
4 & Ali et al. \cite{ali_type-2_2018} & 2018 & Diabetes & User validated & Ontology & FL; queries; rules & Layered \tabularnewline
5 & Alti et al. \cite{alti_agent-based_2021} & 2021 & Diabetes & Prototype & Ontology & Agents; queries; rules & \makecell[l]{Layered; multi-agent; \\service-oriented} \tabularnewline
\rowcolor{gray!4}
6 & Ammar et al. \cite{ammar_using_2021} & 2021 & Diabetes & Prototype & Linked data; KG; ontology & Agents; queries; rules & Layered; multi-agent \tabularnewline
7 & Bampi et al. \cite{bampi_ontology-driven_2025} & 2025 & General monitoring & Prototype & Ontology & Rules; queries & Modular \tabularnewline
\rowcolor{gray!4}
8 & Chatterjee et al. \cite{chatterjee_automatic_2021} & 2021 & Obesity & Prototype & Ontology & Queries; rules & Modular \tabularnewline
9 & Chiang and Liang \cite{chiang_context-aware_2015} & 2015 & General monitoring & Prototype & Ontology & FL; rules & Modular \tabularnewline
\rowcolor{gray!4}
10 & De Brouwer et al. \cite{de_brouwer_mbrain_2022} & 2022 & Headache & User validated & Ontology & ML; queries & Modular \tabularnewline
11 & El-Sappagh et al. \cite{el-sappagh_mobile_2019} & 2019 & Diabetes & Prototype & Ontology & Queries; rules & Modular \tabularnewline
\rowcolor{gray!4}
12 & Elhadj et al. \cite{elhadj_do-care_2021} & 2021 & General monitoring & Prototype & Ontology & Rules & Layered \tabularnewline
13 & Esposito et al. \cite{esposito_smart_2018} & 2018 & Arrhythmia & User validated & Ontology & FL; rules & Layered \tabularnewline
\rowcolor{gray!4}
14 & Fenza et al. \cite{fenza_hybrid_2012} & 2012 & General monitoring & Prototype & Ontology & Agents; FL; rules & \makecell[l]{Layered; multi-agent; \\service-oriented} \tabularnewline
15 & Garcia-Moreno et al. \cite{garcia-moreno_systematic_2023} & 2023 & Frailty; dependence & User validated & Ontology & ML & Layered; service-oriented \tabularnewline
\rowcolor{gray!4}
16 & Garcia-Valverde et al. \cite{garcia-valverde_heart_2014} & 2014 & General monitoring & Prototype & Ontology & ML; rules & Unclear \tabularnewline
17 & Hadjadj and Halimi \cite{hadjadj_integration_2021} & 2021 & General monitoring & User validated & Ontology & Queries; rules & Layered \tabularnewline
\rowcolor{gray!4}
18 & Henaien et al. \cite{henaien_combined_2020} & 2020 & General monitoring & Prototype & Ontology & ML; queries; rules & Layered \tabularnewline
19 & Hooda and Rani \cite{hooda_semantic_2020} & 2020 & Diabetes; heart disease & Prototype & Ontology & Queries; rules & Modular \tabularnewline
\rowcolor{gray!4}
20 & Hristoskova et al. \cite{hristoskova_ontology-driven_2014} & 2014 & Heart failure & User validated & Ontology & Rules & Service-oriented \tabularnewline
21 & Hussain and Park \cite{hussain_big-ecg_2021} & 2021 & Stroke & User validated & Ontology & ML; queries; rules & Modular \tabularnewline
\rowcolor{gray!4}
22 & Ivașcu and Negru \cite{ivascu_activity-aware_2021} & 2021 & General monitoring & Prototype & Ontology & Agents; ML; queries; rules & Modular; multi-agent \tabularnewline
23 & Ivașcu et al. \cite{ivascu_multi-agent_2015} & 2015 & Mental \& degenerative disorders & Prototype & Ontology & Agents; rules & Modular; multi-agent \tabularnewline
\rowcolor{gray!4}
24 & Khozouie et al. \cite{khozouie_ontological_2018} & 2018 & General monitoring & Prototype & Ontology & Rules & Modular \tabularnewline
25 & Kilintzis et al. \cite{kilintzis_supporting_2019} & 2019 & COPD & User validated & Linked data; ontology & Rules; queries & Layered; service-oriented \tabularnewline
\rowcolor{gray!4}
26 & Kim et al. \cite{kim_ontology_2014} & 2014 & General monitoring & Prototype & Ontology & Queries; rules & Layered \tabularnewline
27 & Kordestani et al. \cite{kordestani_extended_2021} & 2021 & Kidney disease; skin disease & Prototype & Ontology & ASP; rules; BN & Layered \tabularnewline
\rowcolor{gray!4}
28 & Lopes de Souza et al. \cite{lopes_ontology-driven_2023} & 2023 & Hypertension & Prototype & Ontology & Queries; rules & Layered; modular \tabularnewline
29 & Martella et al. \cite{martella_semantically_2025} & 2025 & General monitoring & Prototype & Ontology & Agents; rules; queries & \makecell[l]{Layered; modular; \\multi-agent; service-\\oriented} \tabularnewline
\rowcolor{gray!4}
30 & Mavropoulos et al. \cite{mavropoulos_smart_2021} & 2021 & General monitoring & User validated & Ontology & Agents; ML; NLP; rules & Layered; modular; single-agent \tabularnewline
31 & Mcheick et al. \cite{mcheick_stroke_2016} & 2016 & Stroke & Prototype & Ontology & BN & Layered \tabularnewline
\rowcolor{gray!4}
32 & Mezghani et al. \cite{mezghani_semantic_2015} & 2015 & Diabetes & Prototype & Ontology & ML; queries; rules & Layered; service-oriented \tabularnewline
33 & Minutolo et al. \cite{minutolo_hybrid_2016} & 2016 & Arrhythmia & Prototype & Ontology & FL; rules & Modular \tabularnewline
\rowcolor{gray!4}
34 & Peral et al. \cite{peral_ontology-oriented_2018} & 2018 & Diabetes & Prototype & Ontology & ML; NLP; rules & Unclear \tabularnewline
35 & Reda et al. \cite{reda_heterogeneous_2022} & 2022 & General monitoring & Prototype & Linked data; ontology & Queries; rules & Layered \tabularnewline
\rowcolor{gray!4}
36 & Rhayem et al. \cite{rhayem_semantic-enabled_2021} & 2021 & Gestational diabetes & Prototype & Ontology & Queries; rules & Modular \tabularnewline
37 & Spoladore et al. \cite{spoladore_ontology-based_2021} & 2021 & Diabetes; COPD & Prototype & Ontology & Queries; rules & Layered \tabularnewline
\rowcolor{gray!4}
38 & Stavropoulos et al. \cite{stavropoulos_detection_2021} & 2021 & Multiple sclerosis & User validated & KG; ontology & Rules & Modular \tabularnewline
39 & Titi et al. \cite{titi_ontology-based_2019} & 2019 & General monitoring & Prototype & Ontology & Queries; rules & Layered \tabularnewline
\rowcolor{gray!4}
40 & Vadillo et al. \cite{vadillo_enhancement_2013} & 2013 & General monitoring & User validated & Ontology & Agents & Layered; multi-agent \tabularnewline
41 & Villarreal et al. \cite{villarreal_mobile_2014} & 2014 & Diabetes & User validated & Ontology & None specified & Layered \tabularnewline
\rowcolor{gray!4}
42 & Xu et al. \cite{xu_design_2017} & 2017 & General monitoring & Prototype & Linked data; KG & CBR; queries & Layered; service-oriented \tabularnewline
43 & Yu et al. \cite{yu_improving_2022} & 2022 & Paediatric asthma & Prototype & KG & ML; NLP; rules & Modular \tabularnewline
\rowcolor{gray!4}
44 & Yu et al. \cite{yu_semantic_2017} & 2017 & General monitoring & Prototype & Ontology & Queries; rules & Layered \tabularnewline
45 & Zafeiropoulos et al. \cite{zafeiropoulos_evaluating_2024} & 2024 & Parkinson's disease & Prototype & KG; ontology & ML; queries; rules & Modular \tabularnewline
\rowcolor{gray!4}
46 & Zeshan et al. \cite{zeshan_iot-enabled_2023} & 2023 & General monitoring & User validated & Ontology & Queries; rules & Modular \tabularnewline
47 & Zhang et al. \cite{zhang_knowledge-based_2014} & 2014 & General monitoring & Prototype & Ontology & Rules & Layered; modular \tabularnewline
\rowcolor{gray!4}
48 & Zhou et al. \cite{zhou_design_2022} & 2022 & General monitoring & Prototype & KG & ML & Modular \tabularnewline
\end{longtable}
\normalsize

\section{Challenges in health monitoring systems}
\label{key-challenges}

This section examines the role of Semantic Web technologies in addressing the seven key challenges identified in Section~\ref{introduction}, as well as the contribution of other complimentary technologies and techniques that are incorporated into the systems.
Additionally, a critical evaluation is provided assessing the extent to which each system succeeds in addressing these key challenges.
Although there is a broader range of challenges facing sensor-based health monitoring systems, we have necessarily had to delimit the scope of this article. 
By focusing our analysis on these seven salient challenges, we aim to provide an in-depth assessment of how effectively they have been addressed in the current state of the field.
However, we briefly discuss some other considerations, including privacy and security as well as usability, at the end of the section.

\subsection{Interoperability} 
\label{interoperability}
Interoperability can be defined as the ability of different components or systems not only to exchange information but also to make use of it \cite{benson_principles_2021}. 
There are three types of interoperability identified in the health domain: technical, semantic, and process interoperability \cite{gibbons_coming_2007,benson_principles_2021}.
Technical interoperability refers to the way data or information moves from one system or component to another.
Related to this is syntactic interoperability, which provides a structure and syntax for the transmitted data \cite{hosseini_chapter_2016}.
Semantic interoperability refers to the ability of the recipient to understand and make use of the received data \cite{benson_principles_2021}, whereas
process interoperability concerns the seamless coordination of workflows in care delivery \cite{kuziemsky_framework_2016}.
A subset of this is clinical interoperability, through which patients can be seamlessly transferred between different care teams \cite{benson_principles_2021}.
This review focuses on technical, syntactic, and semantic interoperability.
We also discuss the role of interoperability in enabling data fusion, a crucial functionality in health monitoring systems.

\subsubsection{Technical interoperability}
Differing data transmission technologies can contribute to a lack of technical interoperability in health monitoring systems, particularly those that use a range of different sensors. 
Data transmission protocols used in sensors include Bluetooth, Bluetooth Low Energy, ANT+, and Zigbee, with the first three being the most common among wearable devices today \cite{gravina_wearable_2021}.
Interoperability among these different protocols can be achieved using gateway devices, which receive data from different sensors and transmit it to cloud services \cite{rahmani_exploiting_2018}.
This is done by Ali et al. \cite{ali_type-2_2018}, who use a router as a gateway to receive sensor data and transmit it to the internet.
11 of the systems \cite{ammar_using_2021}\cite{ali_smart_2020,alti_agent-based_2021,el-sappagh_mobile_2019,elhadj_do-care_2021,hussain_big-ecg_2021,khozouie_ontological_2018,lopes_ontology-driven_2023,peral_ontology-oriented_2018,villarreal_mobile_2014,zhang_knowledge-based_2014} use a mobile phone as a gateway device or base station, typically receiving sensor data via Bluetooth or Bluetooth Low Energy and transmitting it to the cloud via WiFi or mobile data.

\subsubsection{Syntactic interoperability}
While technical interoperability is associated with hardware components and infrastructure, syntactic interoperability is usually associated with data formats \cite{veer_achieving_2008}. 
There are several standards that are widely used to promote syntactic interoperability among systems. 
Among them is the ISO/IEEE 11073 standard, which provides a common format for communication involving medical devices and patient health data, with an emphasis on vital signs. 
This is used by El-Sappagh et al. \cite{el-sappagh_mobile_2019} for formatting data for transmission to the base unit or gateway.
Other important standards for health data are provided by Health Level 7 (HL7).
One of these is FHIR, which describes data formats, resources, and an application programming interface (API) through which health information can be exchanged \cite{benson_principles_2021}. 
Two of the 48 systems make use of FHIR.
El-Sappagh et al. \cite{el-sappagh_mobile_2019} convert sensor data from the ISO/IEEE 11073 standard to FHIR resource formats, while also receiving data in FHIR format from hospital information systems.
In this way, both sensor data and data from hospital systems are in the same format.
Similarly, Kilintzis et al. \cite{kilintzis_supporting_2019} propose a semantic model based on FHIR, using its data types and defining classes as FHIR resource categories.
FHIR resources can be defined using different data formats\footnote{\url{https://build.fhir.org/resource-formats.html}}, including XML, RDF serialised in Turtle, and JSON.
Both systems use the JSON format.

\subsubsection{Semantic interoperability}
The next type of interoperability is semantic interoperability, which is concerned with the meaning of the exchanged information.
Semantic interoperability can be achieved through the use of unambiguous codes and identifiers, which can be provided by existing standard classifications and terminologies \cite{benson_principles_2021}.
Ontologies are, of course, a well-established way to embed semantic interoperability in a system \cite{sheth_semantic_2008}.
Within the medical domain, many existing medical terminologies are available as ontologies, including SNOMED CT\footnote{\url{https://bioportal.bioontology.org/ontologies/SNOMEDCT}}, the International Classification of Diseases (ICD)\footnote{\url{https://bioportal.bioontology.org/ontologies/ICDO}}, and the International Classification for Nursing Practice (ICNP)\footnote{\url{https://bioportal.bioontology.org/ontologies/ICNP}}.
Among the systems, SNOMED CT is the most commonly used \cite{bampi_ontology-driven_2025,kilintzis_supporting_2019} \cite{chatterjee_automatic_2021,el-sappagh_mobile_2019,kordestani_extended_2021,lopes_ontology-driven_2023,reda_heterogeneous_2022,rhayem_semantic-enabled_2021,titi_ontology-based_2019,zhou_design_2022}.
ICNP is used by Elhadj et al. \cite{elhadj_do-care_2021} and Henaien et al. \cite{henaien_combined_2020} , while ICD is used by Spoladore et al. \cite{spoladore_ontology-based_2021} and Yu \cite{yu_improving_2022} (ICD-11, the latest version) as well as Titi et al. \cite{titi_ontology-based_2019} (ICD-10). 
The Unified Medical Language System (UMLS) \cite{bodenreider_unified_2014} is a large thesaurus that integrates multiple terminologies of medical knowledge. It is used by Peral et al. \cite{peral_ontology-oriented_2018}, and Zhou et al. \cite{zhou_design_2022}.
Another thesaurus is Medical Subject Headings (MeSH), which is used for indexing, cataloguing, and searching health information, and is integrated in the system proposed by Reda et al. \cite{reda_heterogeneous_2022}.
Garcia Moreno et al. \cite{garcia-moreno_systematic_2023} and Spoladore et al. \cite{spoladore_ontology-based_2021} incorporate the International Classification of Functioning, Disability and Health (ICF)\footnote{\url{https://icd.who.int/dev11/l-icf/en}}.

Terminologies for specific diseases and conditions also exist. 
For example, Ali et al. \cite{ali_intelligent_2021} and El-Sappagh et al. \cite{el-sappagh_mobile_2019} reuse ontologies specific to diabetes.
Similarly, De Brouwer et al. \cite{de_brouwer_mbrain_2022} use the third edition of the International Classification of Headache Disorders (ICHD-3)\footnote{\url{https://ichd-3.org}}, while
Hristoskova et al. \cite{hristoskova_ontology-driven_2014} and Zafeiropoulos et al. \cite{zafeiropoulos_evaluating_2024} reuse the Heart Failure Ontology and Parkinson and Movement Disorder Ontology respectively.
The Vital Sign Ontology is extended by El-Sappagh \cite{el-sappagh_mobile_2019} and Ivașcu and Negru \cite{ivascu_activity-aware_2021}.
Xu et al. \cite{xu_design_2017} posit that it is difficult to build scalable ontology-based systems suitable for large amounts of healthcare data and instead opt for a linked data approach to add semantic information to the data. 
Their proposed system uses linked open data medical knowledge graphs, namely Diseasome, DBpedia, and DrugBank.
Using these resources, they create a knowledge graph showing the relationships between symptoms and diseases.
Domain-independent concepts can also be referenced from Semantic Web technologies.
For instance, Peral et al. \cite{peral_ontology-oriented_2018} and Reda et al. \cite{reda_heterogeneous_2022} both use WordNet, a lexical English language database of semantic relations between words, linking them into semantic relations.

Semantic Web technologies also provide a means to represent sensors and the data they capture.
Sensors can be represented with varying degrees of expressiveness. Concepts that can be captured about sensors include unique identifier, manufacturer, location of deployment, dimensions, operating conditions, type of data captured, and hierarchy with regard to related sensors \cite{compton_survey_2009}. 
Similarly, various sensor data concepts can be represented, such as the property being observed, units of measurement, and measurement timestamps.
38 systems represent sensor and sensor data concepts in ontologies.
The reuse of existing sensor ontologies, particularly established ones such as SAREF, can contribute to a higher degree of expressiveness for sensor and sensor data concepts. 
This is because these validated ontologies provide rich modelling of such concepts, facilitating more effective querying of and reasoning on sensor data, which is essential for situation analysis.
Comprehensive sensor ontologies also support sensor management, allowing sensors to be catalogued based on their attributes as captured in ontologies \cite{compton_survey_2009}.
Despite these benefits, only 16 systems reuse existing sensor or device ontologies, namely SSN/SOSA \cite{bampi_ontology-driven_2025,garcia-moreno_systematic_2023,martella_semantically_2025} \cite{chatterjee_automatic_2021,el-sappagh_mobile_2019,elhadj_do-care_2021,ivascu_activity-aware_2021,rhayem_semantic-enabled_2021,stavropoulos_detection_2021,titi_ontology-based_2019}, SAREF and its extensions \cite{de_brouwer_mbrain_2022,hadjadj_integration_2021,lopes_ontology-driven_2023,zafeiropoulos_evaluating_2024}, the Amigo device ontology \cite{hristoskova_ontology-driven_2014}, and the Moving Objects ontology \cite{rhayem_semantic-enabled_2021}.
This could be attributed to the existing ontologies providing more complexity than the systems require, although this can be mitigated by selectively importing only the relevant classes.

Foundational ontologies can contribute to semantic interoperability by providing unambiguous and domain-independent concept definitions \cite{amaral_foundational_2021}.
Three of the selected systems directly incorporate a foundational ontology. 
El-Sappagh et al. \cite{el-sappagh_mobile_2019} use the Basic Formal Ontology, while De Brouwer et al. \cite{de_brouwer_mbrain_2022} and Stavropoulos et al. \cite{stavropoulos_detection_2021} use the DOLCE+DnS (Description and Situation) Ultra Lite (DUL) ontology.
Other systems indirectly integrate foundational ontologies by reusing other ontologies that have already incorporated them.
For example, the SSN ontology uses DUL as its upper ontology \cite{compton_ssn_2012}, and the SAREF ontology also has an indirect reference to DUL through its mappings to the SSN ontology \cite{daniele_created_2015}.
Consequently, any system that reuses the SSN or SAREF ontologies inherits an indirect connection to DUL.

\subsubsection{Supporting data fusion}
For real-time health monitoring, streaming sensor data must be retrieved and dynamically fused with other heterogeneous, multimodal, and distributed sources of data. 
This data fusion is pivotal for downstream situation detection, situation prediction, and decision support. 
Interoperability serves as an enabler for effective data fusion. 
Technical interoperability provides the protocols and hardware necessary to collect data from diverse sources; syntactic interoperability ensures data formats and structures are compatible across systems; and semantic interoperability establishes shared meaning for data elements, allowing accurate interpretation.
Additionally, data fusion supports process interoperability by creating a unified and comprehensive patient view from diverse data sources, which enables different healthcare providers to coordinate their workflows effectively \cite{kuziemsky_framework_2016}.

As seen in Table~\ref{supported-data}, the selected systems support a wide range of heterogeneous and multimodal data.
All 48 systems collect physiological data from body sensors, while 
15 systems additionally incorporate weather data from ambient sensors.
Health and medical records are the most frequently incorporated data source other than sensor data, with 21 systems supporting the incorporation of such data from external hospital systems.
These  records provide additional information that is useful for health monitoring, such as an individual's disease history, laboratory test results, medications taken, allergies, and previous hospital admissions.
The systems proposed by Ali et al. \cite{ali_smart_2020,ali_type-2_2018}, El-Sappagh \cite{el-sappagh_mobile_2019}, and Rhayem et al. \cite{rhayem_semantic-enabled_2021} have the most comprehensive records, capturing laboratory tests, prior disease diagnoses, and lifestyle information such as exercise, nutrition, alcohol consumption, and smoking status.
Some systems use medical records to extract diagnosis status \cite{ali_intelligent_2021}, while others use them to extract an individual's risk factors for disease \cite{ali_smart_2020}.
These records can also be used to overcome limitations of sensor data such as missing values, as was done by Ali et al. \cite{ali_intelligent_2021}.
Besides health and medical records, data from social networks can also be used to complement sensor data. 
This is done in two systems: Ali et al. \cite{ali_intelligent_2021}, who use social networking data for monitor individuals' mental health through sentiment analysis; and Ammar et al. \cite{ammar_using_2021}, who use social networks, blogs, and news articles as sources of public knowledge.

A detailed analysis of the data sources, including sensor devices and existing datasets, is provided in Section~\ref{data-devices}.


\scriptsize
\setlength{\extrarowheight}{2pt}
\begin{longtable}{ll>{\raggedright\arraybackslash}p{0.81\NetTableWidth}}
\caption{Sensor data and other types of data used in the systems.} \label{supported-data}\\
\hline
\textbf{\#} & \textbf{System} & \textbf{Supported sensor data and other types of data}\tabularnewline
\hline
\endfirsthead

\multicolumn{3}{c}{\tablename\ \thetable\ -- continued from previous page} \\
\hline
\textbf{\#} & \textbf{System} & \textbf{Supported sensor data and other types of data}\tabularnewline
\hline
\endhead

\hline
\multicolumn{3}{>{\raggedright\arraybackslash}p{\NetTableWidth}}{BG - blood glucose; BP - blood pressure; BT - body temperature; CO\textsubscript{2} - carbon dioxide; CO - carbon monoxide; ECG - electrocardiogram; EEG - electroencephalogram; EIT - electroimpedance tomography; EMG - electromyogram; GPS - global positioning system; HR - heart rate; O\textsubscript{2} - oxygen; RR - respiratory rate; SpO\textsubscript{2} - blood oxygen saturation; TVOC - total volatile organic compounds \hfill \textbf{Table continued on next page.}}\\
\endfoot
\hline
\multicolumn{3}{>{\raggedright\arraybackslash}p{\NetTableWidth}}{BG - blood glucose; BP - blood pressure; BT - body temperature; CO\textsubscript{2} - carbon dioxide; CO - carbon monoxide; ECG - electrocardiogram; EEG - electroencephalogram; EIT - electroimpedance tomography; EMG - electromyogram; GPS - global positioning system; HR - heart rate; O\textsubscript{2} - oxygen; RR - respiratory rate; SpO\textsubscript{2} - blood oxygen saturation; TVOC - total volatile organic compounds}
\endlastfoot

1 & Akhtar et al. \cite{akhtar_multi-agent_2022} & \textbf{Body} (BP, HR, BT, ECG, EEG, EMG); \textbf{Ambient} (temperature, CO \& CO\textsubscript{2} levels, motion)\tabularnewline
\rowcolor{gray!4}
2 & Ali et al. \cite{ali_intelligent_2021} & \textbf{Body} (BP, SpO\textsubscript{2}, BT, HR, ECG, EEG, BG); \textbf{Other} (health/medical records, social networks, smartphone)\tabularnewline
3 & Ali et al. \cite{ali_smart_2020} & \textbf{Body} (RR, SpO\textsubscript{2}, BP, BT, HR, EMG, EEG, ECG, BG, cholesterol, position, activity); \textbf{Other} (health/medical records)\tabularnewline
\rowcolor{gray!4}
4 & Ali et al. \cite{ali_type-2_2018} & \textbf{Body} (ECG, EEG, EMG, HR, BP, BG, cholesterol, range of motion)\tabularnewline
5 & Alti et al. \cite{alti_agent-based_2021} & \textbf{Body} (HR, BG, motion); \textbf{Other} (GPS)\tabularnewline
\rowcolor{gray!4}
6 & Ammar et al. \cite{ammar_using_2021} & \textbf{Body} (BP, activity); \textbf{Other} (health/medical records, social determinants of health, social networks, blogs)\tabularnewline
7 & Bampi et al. \cite{bampi_ontology-driven_2025} & \textbf{Body} (HR, BR); \textbf{Ambient} (Humidity, other unnamed)\tabularnewline
\rowcolor{gray!4}
8 & Chatterjee et al. \cite{chatterjee_automatic_2021} & \textbf{Body} (BP, BG, activity); \textbf{Ambient} (temperature, humidity); \textbf{Other} (interviews, questionnaires, weather forecast, health/medical records)\tabularnewline
9 & Chiang and Liang \cite{chiang_context-aware_2015} & \textbf{Body} (BP, HR, BG, cholesterol);  \textbf{Ambient} (motion, indoor \& outdoor temperature, humidity)\tabularnewline
\rowcolor{gray!4}
10 & De Brouwer et al. \cite{de_brouwer_mbrain_2022} &  \textbf{Body} (Acceleration, HR, blood volume pulse, galvanic skin response, skin temperature);  \textbf{Other} (daily headache diary, questionnaires)\tabularnewline
11 & El-Sappagh et al. \cite{el-sappagh_mobile_2019} & \textbf{Body} (BP, HR, BG); \textbf{Other} (health/medical records)\tabularnewline
\rowcolor{gray!4}
12 & Elhadj et al. \cite{elhadj_do-care_2021} & \textbf{Body} (BT, HR, BP, RR, SpO\textsubscript{2}); \textbf{Ambient} (temperature, humidity, location, motion); \textbf{Other} (health/medical records)\tabularnewline
13 & Esposito et al. \cite{esposito_smart_2018} & \textbf{Body} (BT, HR, SpO\textsubscript{2}, acceleration)\tabularnewline
\rowcolor{gray!4}
14 & Fenza et al. \cite{fenza_hybrid_2012} & \textbf{Body} (HR, BP, BT, SpO\textsubscript{2}, BG);  \textbf{Ambient} (temperature)\tabularnewline
15 & Garcia-Moreno et al. \cite{garcia-moreno_systematic_2023} & \textbf{Body} (BT, HR, accelerometer, electrodermal activity); \textbf{Ambient} (unnamed)\tabularnewline
\rowcolor{gray!4}
16 & Garcia-Valverde et al. \cite{garcia-valverde_heart_2014} & \textbf{Body} (HR, acceleration, orientation/angular velocity, magnetoresistance)\tabularnewline
17 & Hadjadj and Halimi \cite{hadjadj_integration_2021} & \textbf{Body} (BP, HR, BT, BG); \textbf{Other} (vehicle sensor data)\tabularnewline
\rowcolor{gray!4}
18 & Henaien et al. \cite{henaien_combined_2020} & \textbf{Body} (SpO\textsubscript{2}, BP, HR, RR, BT); \textbf{Ambient} (temperature, light, motion); \textbf{Other} (health/medical records)\tabularnewline
19 & Hooda and Rani \cite{hooda_semantic_2020} & \textbf{Body} (BP, HR, BG, ECG); \textbf{Other} (health/medical records)\tabularnewline
\rowcolor{gray!4}
20 & Hristoskova et al. \cite{hristoskova_ontology-driven_2014} & \textbf{Body} (BP, HR, SpO\textsubscript{2}, ECG); \textbf{Other} (health/medical records, WiFi location tag)\tabularnewline
21 & Hussain and Park \cite{hussain_big-ecg_2021} & \textbf{Body} (ECG); \textbf{Other} (health/medical records)\tabularnewline
\rowcolor{gray!4}
22 & Ivașcu and Negru \cite{ivascu_activity-aware_2021} & \textbf{Body} (HR, RR, ECG, acceleration)\tabularnewline
23 & Ivașcu et al. \cite{ivascu_multi-agent_2015} & \textbf{Body} (EEG, acceleration); \textbf{Ambient} (video, audio, motion, bed sensor data)\tabularnewline
\rowcolor{gray!4}
24 & Khozouie et al. \cite{khozouie_ontological_2018} & \textbf{Body} (BP, BT, SpO\textsubscript{2}, ECG, EMG, acceleration, orientation/angular velocity); \textbf{Ambient} (temperature, humidity, CO \& O\textsubscript{2} levels); \textbf{Other} (GPS)\tabularnewline
25 & Kilintzis et al. \cite{kilintzis_supporting_2019} & \textbf{Body} (BP, ECG, EIT, SpO\textsubscript{2}); \textbf{Other} (health/medical records)\tabularnewline
\rowcolor{gray!4}
26 & Kim et al. \cite{kim_ontology_2014} & \textbf{Body} (BP, other unnamed vital signs); \textbf{Ambient} (temperature, illumination, humidity, wind); \textbf{Other} (weather forecast, news, weather indices)\tabularnewline
27 & Kordestani et al. \cite{kordestani_extended_2021} & \textbf{Body} (BT, other unnamed vital signs); \textbf{Ambient} (temperature); \textbf{Other} (health/medical records)\tabularnewline
\rowcolor{gray!4}
28 & Lopes de Souza et al. \cite{lopes_ontology-driven_2023} & \textbf{Body} (HR, BP, BT, acceleration, orientation/angular velocity); \textbf{Other} (user-submitted data)\tabularnewline
29 & Martella et al. \cite{martella_semantically_2025} & \textbf{Body} (HR, temperature); \textbf{Ambient} (CO\textsubscript{2}, TVOC, particulate matter, temperature, humidity, ambient light, atmospheric pressure, sound pressure, electromagnetic emission); \textbf{Other} (GPS)\tabularnewline
\rowcolor{gray!4}
30 & Mavropoulos et al. \cite{mavropoulos_smart_2021} & \textbf{Body} (BP, BG, sleep); \textbf{Ambient} (video); \textbf{Other} (health/medical records)\tabularnewline
31 & Mcheick et al. \cite{mcheick_stroke_2016} & \textbf{Body} (BP, blood flow velocity)\tabularnewline
\rowcolor{gray!4}
32 & Mezghani et al. \cite{mezghani_semantic_2015} & \textbf{Body} (BP, HR, BG); \textbf{Other} (health/medical records)\tabularnewline
33 & Minutolo et al. \cite{minutolo_hybrid_2016} & \textbf{Body} (BT, HR, SpO\textsubscript{2}, acceleration)\tabularnewline
\rowcolor{gray!4}
34 & Peral et al. \cite{peral_ontology-oriented_2018} & \textbf{Body} (BG); \textbf{Other} (the web, existing databases, health/medical records)\tabularnewline
35 & Reda et al. \cite{reda_heterogeneous_2022} & \textbf{Body} (HR, BT, BP, weight, calories burned, step count); \textbf{Other} (self-reported data)\tabularnewline
\rowcolor{gray!4}
36 & Rhayem et al. \cite{rhayem_semantic-enabled_2021} & \textbf{Body} (BT, BP, HR, BG, cholesterol, activity); \textbf{Ambient} (temperature, humidity); \textbf{Other} (health/medical records)\tabularnewline
37 & Spoladore et al. \cite{spoladore_ontology-based_2021} & \textbf{Body} (HR, SpO\textsubscript{2})\tabularnewline
\rowcolor{gray!4}
38 & Stavropoulos et al. \cite{stavropoulos_detection_2021} & \textbf{Body} (HR, step count, sleep)\tabularnewline
39 & Titi et al. \cite{titi_ontology-based_2019} & \textbf{Body} (BT, BP, HR, BG); \textbf{Ambient} (temperature, humidity)\tabularnewline
\rowcolor{gray!4}
40 & Vadillo et al. \cite{vadillo_enhancement_2013} & \textbf{Body} (HR, BT, BP, SpO\textsubscript{2}, BG); \textbf{Ambient} (motion, temperature, occupancy of bed / chair, CO levels)\tabularnewline
41 & Villarreal et al. \cite{villarreal_mobile_2014} & \textbf{Body} (BP, BT, BG)\tabularnewline
\rowcolor{gray!4}
42 & Xu et al. \cite{xu_design_2017} & \textbf{Body} (BP, ECG, BG); \textbf{Other} (health/medical records)\tabularnewline
43 & Yu et al. \cite{yu_improving_2022} & \textbf{Body} (BP, HR, sleep, exercise, weight); \textbf{Other} (health/medical records, self-reported data)\tabularnewline
\rowcolor{gray!4}
44 & Yu et al. \cite{yu_semantic_2017} & \textbf{Body} (HR, BP, body fat); \textbf{Other} (mobile applications)\tabularnewline
45 & Zafeiropoulos et al. \cite{zafeiropoulos_evaluating_2024} & \textbf{Body} (HR, movement data, sleep); \textbf{Other} (health/medical records)\tabularnewline
\rowcolor{gray!4}
46 & Zeshan et al. \cite{zeshan_iot-enabled_2023} & \textbf{Body} (BT, BP, HR); \textbf{Other} (GPS)\tabularnewline
47 & Zhang et al. \cite{zhang_knowledge-based_2014} & \textbf{Body} (BP, BT, HR, SpO\textsubscript{2}); \textbf{Other} (health/medical records)\tabularnewline
\rowcolor{gray!4}
48 & Zhou et al. \cite{zhou_design_2022} & \textbf{Body} (BP, HR, RR, BT, SpO\textsubscript{2}, BG, uric acid, cholestrol, lipoproteins, triglycerides, sleep); \textbf{Ambient} (particulate matter, CO\textsubscript{2}, temperature, formaldehyde, TVOC); \textbf{Other} (health/medical records)\tabularnewline
\end{longtable}
\normalsize

\subsubsection{The role of Semantic Web technologies}
Semantic Web technologies do not inherently provide support for technical interoperability, since they operate at a higher, more abstract level to formally represent and derive meaning from the data. 
Therefore, in order to achieve technical interoperability, health monitoring systems must leverage data transmission protocols and devices.
However, Semantic Web technologies are critical in the achievement of semantic interoperability among the selected systems.
20 of the systems make use of terminologies such as SNOMED CT and ICD through ontologies, 
and access knowledge graphs such as DrugBank and Diseasome which are published as linked data.
This allows the systems to access and reason with standardised health domain knowledge.

Semantic Web technologies can also contribute to syntactic interoperability.
For instance, El-Sappagh et al. \cite{el-sappagh_mobile_2019} and Kilintzis et al. \cite{kilintzis_supporting_2019} map FHIR resources to an ontology, allowing for interoperability between their proposed system and hospital information systems that use FHIR. 
 These are the only two systems that use Semantic Web technologies to achieve interoperability with established health standards. 
This can be attributed to the historical gap in user-friendly tools for this purpose.
For example, although FHIR has an RDF representation serialised in Turtle format, early adopters noted several issues precluding its ease of use, such as literal values and FHIR references being nested under blank nodes, and unnecessarily long predicate names \cite{sharma_implementing_2022}.
However, more recent versions of FHIR RDF have largely addressed these issues \cite{sharma_implementing_2022}.

Finally, with regard to the representation of heterogeneous and multimodal data, 42 systems use ontologies for this purpose.
The exceptions are three systems which represent the data in knowledge graphs \cite{ammar_using_2021,xu_design_2017,yu_improving_2022}, and three systems which only use Semantic Web technologies (i.e. domain ontologies, external knowledge graphs, and linked data repositories) as a source of domain knowledge \cite{ali_intelligent_2021,hussain_big-ecg_2021,zhou_design_2022}.

\subsection{Situation detection}
\label{situation-detection}

A situation can be understood as a higher-level interpretation of sensor data that is relevant and of interest in an application domain \cite{ye_situation_2011}. 
Personal health monitoring systems should be capable of situation analysis, which entails both the detection and the prediction of health situations.
We discuss situation detection below, and situation prediction in Section~\ref{situation-prediction}.

\subsubsection{Forms of situation detection}
In health monitoring systems, situation detection can take a variety of forms.
One of these is the categorisation of individual sensor observations based on whether they are within or outside a given range as determined by domain knowledge.
For example, in the system proposed by Akhtar et al. \cite{akhtar_multi-agent_2022}, when vital signs such as temperature and heart rate are outside the normal range, the situation is classified as an emergency.
Likewise, Elhadj et al. \cite{elhadj_do-care_2021} classify expected observations as normal, while observations outside the normal ranges are classified as abnormal. They also include a third classification, wrong, for faulty observations from malfunctioning sensors.
Similar threshold-based situation categories are used in 19 of the systems \cite{ammar_using_2021,bampi_ontology-driven_2025,kilintzis_supporting_2019,martella_semantically_2025} \cite{alti_agent-based_2021,de_brouwer_mbrain_2022,garcia-valverde_heart_2014,hadjadj_integration_2021,hristoskova_ontology-driven_2014,ivascu_activity-aware_2021,khozouie_ontological_2018,lopes_ontology-driven_2023,peral_ontology-oriented_2018,rhayem_semantic-enabled_2021,titi_ontology-based_2019,villarreal_mobile_2014,zafeiropoulos_evaluating_2024,zeshan_iot-enabled_2023,zhang_knowledge-based_2014}.
Thresholds have also been used to classify physical activity based on level of intensity \cite{chatterjee_automatic_2021,el-sappagh_mobile_2019,esposito_smart_2018,garcia-valverde_heart_2014,ivascu_activity-aware_2021}. 
A better approach than using individual sensor observations is to consider different observations and personal attributes to classify individuals.
This is done by Ali et al. \cite{ali_type-2_2018}, who classify the patient health condition as either healthy, moderate, or serious based on multiple sensor outputs and properties such as sex, weight, and height.
Similarly, Chiang and Liang \cite{chiang_context-aware_2015} classify situations as either healthy, moderate, or severe based on age, blood pressure, blood glucose, heart rate, and cholesterol.

Another form of situation detection in health monitoring is the detection of medical conditions and diseases. 
Some conditions such as hypertension and hyperglycemia can be diagnosed based on individual sensor observation thresholds.
This is done by Kim et al. \cite{kim_ontology_2014}, who detect prehypertension and step 1 and 2 hypertension based on defined blood pressure thresholds.
Similarly, hyperglycemia is detected by Rhayem et al. \cite{rhayem_semantic-enabled_2021} based on blood glucose levels.
Other diseases require the analysis of signs and symptoms based on a combination of different sensor observations and other sources of data.
For example, Ivașcu et al. \cite{ivascu_multi-agent_2015} detect mental disorders (Parkinson’s, Alzheimer’s, psychosis, and depression) using signs and symptoms related to behaviour, motor skills, cognitive skills, facial appearance, mood, sleep, weight, and speech.
Other systems are able to detect types of headaches \cite{de_brouwer_mbrain_2022}, heart disease \cite{ali_smart_2020}, diabetes \cite{ali_intelligent_2021,ali_type-2_2018,hooda_semantic_2020}, frailty \cite{garcia-moreno_systematic_2023}, stroke \cite{hussain_big-ecg_2021}, and skin and kidney diseases \cite{kordestani_extended_2021}.
Beyond the detection of diseases, Zafeiropoulos et al. \cite{zafeiropoulos_evaluating_2024} detect medication adherence in Parkinson’s disease patients by recognising missed doses based on two features: tremors and bradykinisia.

\subsubsection{Techniques for situation detection}
39 of the systems implement some form of rule-based reasoning for situation detection.
Of these, 28 systems specify using Semantic Web rules in particular \cite{akhtar_multi-agent_2022,ali_type-2_2018,alti_agent-based_2021,bampi_ontology-driven_2025,chatterjee_automatic_2021,chiang_context-aware_2015,el-sappagh_mobile_2019,elhadj_do-care_2021,esposito_smart_2018,garcia-valverde_heart_2014,hadjadj_integration_2021,henaien_combined_2020,hooda_semantic_2020,hristoskova_ontology-driven_2014,kilintzis_supporting_2019,kim_ontology_2014,lopes_ontology-driven_2023,mezghani_semantic_2015,minutolo_hybrid_2016,reda_heterogeneous_2022,rhayem_semantic-enabled_2021,spoladore_ontology-based_2021,stavropoulos_detection_2021,titi_ontology-based_2019,yu_semantic_2017,zafeiropoulos_evaluating_2024,zeshan_iot-enabled_2023,zhang_knowledge-based_2014}.
A discussion of the specific rule languages used in the systems can be found in Section~\ref{development-tools}.
Rules provide a way to implement expert knowledge in an if-then form, whereby if certain conditions are met, then a consequent conclusion is made or action taken.
Despite their widespread use, rules have several limitations.
Firstly, crisp rules are unable to handle uncertainty and ambiguity in sensor observations and the determination of health situations.
To mitigate this, five systems incorporate fuzzy logic \cite{ali_type-2_2018,chiang_context-aware_2015,esposito_smart_2018,fenza_hybrid_2012,minutolo_hybrid_2016} while one incorporates defeasible logic \cite{akhtar_multi-agent_2022} within the rules.
These techniques are discussed in greater detail in Section~\ref{uncertainty}, which focuses on techniques for handling uncertainty in health monitoring.
Secondly, rules are typically based on existing expert knowledge, and therefore cannot incorporate new knowledge that experts may be unaware of.
Additionally, manually updating rules is time-consuming, making them difficult to scale.
This challenge can be overcome using learned rules based on ML algorithms, which can acquire new, high quality knowledge automatically \cite{hitzler_neural-symbolic_2020} and contribute to dynamic and adaptable rule-based systems.
The systems proposed by Hussain et al. \cite{hussain_big-ecg_2021}, Henaien et al. \cite{henaien_combined_2020} and Peral et al. \cite{peral_ontology-oriented_2018} extract rules from decision trees.
However, caution should be exercised when using ML-derived rules, as they may still need verification and validation from domain experts.
As an alternative to rule-based reasoning, Xu et al. \cite{xu_design_2017} implement case-based reasoning, arguing that it is easier to capture human experiences using cases rather than rules. 
By searching for historical cases that are similar to the current case, their proposed system is able to obtain treatment plans that have been successful in the past.

In addition to the development of rules as discussed above, ML is also used in a number of the systems for the classification of diseases based not only on sensor data but also other data sources.
Ali et al. \cite{ali_intelligent_2021} use a bidirectional long short-term memory (BiLSTM) model to detect diabetes and blood pressure, to classify sentiments from social networking data for mental health monitoring, and to classify drug side effects.
Their proposed system uses domain ontologies to extract important features that can enhance the ML classification.
Zhou et al. \cite{zhou_design_2022} also use a BiLSTM model for disease prediction, while Garcia-Moreno et al. \cite{garcia-moreno_systematic_2023} use k-nearest neighbours to classify elderly individuals based on frailty and dependence.
Other ML algorithms used include multi-layer perceptron for heart disease detection \cite{ali_smart_2020} and random forest for stroke detection \cite{hussain_big-ecg_2021}.
ML is also used for physical activity classification, for example using the k-nearest neighbours \cite{garcia-valverde_heart_2014,mavropoulos_smart_2021}, decision trees \cite{mavropoulos_smart_2021}, and random forest \cite{ivascu_activity-aware_2021,mavropoulos_smart_2021} algorithms.
Finally, ML can also be used to classify situation severity for reporting purposes.
This is done by Zafeiropoulos et al. \cite{zafeiropoulos_evaluating_2024}, who use a graph neural network to distinguish between medium and high alerts.
A full review of ML techniques for situation analysis in the health domain is outside the scope of this study.
Readers are referred to the reviews by Rav\`i et al. \cite{ravi_deep_2017} and Li et al. \cite{li_comprehensive_2021}.

\subsubsection{The role of Semantic Web technologies}
Semantic Web technologies can support situation detection in two main ways: firstly, they formally represent important concepts and the relationships between them, i.e. sensor data, domain knowledge, contextual information, and even the situations themselves; 
and secondly, they support reasoning through which new knowledge can be derived from existing knowledge \cite{ye_situation_2011}.
Although several situation-focused ontologies have been developed, including the Situation Theory Ontology \cite{kokar_ontology-based_2009} and the Scenes and Situations ontology \cite{almeida_towards_2018}, none of the selected systems reuse any such ontologies. 
Despite this, Semantic Web technologies remain vital for situation detection among the selected systems.
Rule-based reasoning is the most common mechanism for situation detection among the systems. 
These rules rely heavily on concepts that are formally defined in ontologies, and they are more often than not expressed in standard Semantic Web languages such as SWRL.

\subsection{Situation prediction} 
\label{situation-prediction}
All 48 selected systems detect current situations. In contrast, only 12 of the systems go beyond this to predict some future outcome \cite{ali_smart_2020,alti_agent-based_2021,chiang_context-aware_2015,de_brouwer_mbrain_2022,fenza_hybrid_2012,hristoskova_ontology-driven_2014,mcheick_stroke_2016,peral_ontology-oriented_2018,reda_heterogeneous_2022,rhayem_semantic-enabled_2021,yu_semantic_2017,zhou_design_2022}.

\subsubsection{Forms of situation prediction}
All 12 of these systems explore the concept of risk as a situation prediction feature, since determining an individual's risk profile for a certain condition can be used to predict future adverse health situations.
This includes the risks of heart disease \cite{ali_intelligent_2021,hristoskova_ontology-driven_2014}, arthritis recurrence \cite{chiang_context-aware_2015}, stroke \cite{mcheick_stroke_2016}, and fetal loss in gestational diabetes patients \cite{rhayem_semantic-enabled_2021}.
Zhou et al. \cite{zhou_design_2022} use a multi-label classification approach to simultaneously assess the risk of multiple chronic diseases, including hypertension, diabetes, and arthritis.
De Brouwer et al. \cite{de_brouwer_mbrain_2022} detect triggers as a means to anticipate potential headache attacks in the future, which can be considered a form of risk prediction.
To support the identification of potential risks, future physiological readings can also be predicted using historical sensor observations, as is done by Peral et al. \cite{peral_ontology-oriented_2018}. 
Their proposed system predicts blood glucose levels over three-day and five-day windows. These predictions of sensor measurements can then be analysed to determine future health risks.
Hristoskova et al. \cite{hristoskova_ontology-driven_2014} similarly adopt a temporal analysis, determining the risk of congestive heart failure over a four-year time horizon.

Another useful aspect of situation prediction is the determination of the prognosis, i.e. expected progression, of a detected disease, although this is poorly explored in the systems. 
Hussain and Park \cite{hussain_big-ecg_2021} mention the intention to extend their system in future work to include automated stroke prognosis; however, this is not implemented in the current version of the system.
In contrast, Yu et al. \cite{yu_semantic_2017} include a disease progression class in their proposed ontology, representing past diagnoses or potential health risks and their associated times. 
However, the system does not include any methods to predict the progression of detected conditions.

\subsubsection{Techniques for situation prediction}
Similar to situation detection, the most common technique used in situation prediction is rules, which is used in nine of the 12 systems.
Of these, seven specify Semantic Web rules in particular \cite{alti_agent-based_2021,chiang_context-aware_2015,hristoskova_ontology-driven_2014,reda_heterogeneous_2022,rhayem_semantic-enabled_2021,yu_semantic_2017,zafeiropoulos_evaluating_2024}.
As discussed in Section~\ref{situation-detection}, both Chiang and Liang \cite{chiang_context-aware_2015} and Fenza et al. \cite{bobillo_fuzzy_2016} use fuzzy rules to enhance crisp rules.
ML is used for situation prediction in 4 of the systems \cite{ali_smart_2020,peral_ontology-oriented_2018,zafeiropoulos_evaluating_2024,zhou_design_2022}.
Ali et al. \cite{ali_smart_2020} use an ensemble deep learning classifier, which consists of a five-layer feed-forward network that incorporates a boosting algorithm, to predict future heart attacks, while
Peral et al. \cite{peral_ontology-oriented_2018} use support vector machine and logistic regression models to forecast future blood glucose measurements.
Although De Brouwer et al. \cite{de_brouwer_mbrain_2022} use ML to detect headache triggers, the actual prediction of headaches is knowledge-based using SPARQL queries on stored data.

Besides rules and ML, Bayesian networks are used in two of the systems. These are probabilistic models in the form of directed acyclic graphs that can represent causal relationships among variables in a domain.
Mcheik et al. \cite{mcheick_stroke_2016} use a Bayesian network to calculate the risk of stroke occurring in the next seven days, based on risk factors such as age, presence of diabetes, high blood pressure, and symptom duration. 
This approach is also taken by Kordestani et al. \cite{kordestani_extended_2021} to determine the probability of the occurrence of kidney disease.
The use of fuzzy rules and Bayesian networks is discussed in greater detail in Sections~\ref{explainability} on explainability and ~\ref{uncertainty} on uncertainty handling.

\subsubsection{The role of Semantic Web technologies}
By providing a structured representation of risk factors, temporal concepts, health outcomes, and situations, and supporting reasoning over these concepts, Semantic Web technologies support more effective situation prediction.
Similarly to situation detection, rules remain the most used reasoning approach for situation prediction among the systems, with seven of them using Semantic Web rules specifically.
However, as discussed in Section~\ref{situation-detection}, semantic-based approaches to situation analysis face drawbacks such as scaling difficulties and the inability to handle uncertainty inherently. 
These limitations can be mitigated by combining them with complementary techniques such as ML, fuzzy logic, and Bayesian networks.
In fact, several extensions of Semantic Web standards and languages have been proposed that incorporate these uncertainty handling techniques, and these are discussed in Section~\ref{uncertainty-role}.
However, none of these are used in the reviewed systems.

\subsection{Decision Support} 
\label{decision-support}
Decision support is the natural next step after situation analysis. 
Based on the detected and predicted situations, targeted support can be offered to mitigate adverse situations and promote favourable health outcomes.
45 systems implement some form of decision support.

\subsubsection{Forms of decision support}

Messages and notifications are used in majority of the systems to warn of potentially dangerous situations, issue reminders, and prompt
mitigating action.
These notifications can be sent to monitored individuals, caregivers, clinicians, and emergency services depending on the severity of the situation.
The most common form of decision support is alerting users of adverse situations.
This is done in nearly all of the selected systems, with the exception of 11 which do not mention this crucial functionality \cite{ali_intelligent_2021,ali_smart_2020,alti_agent-based_2021,chatterjee_automatic_2021,el-sappagh_mobile_2019,fenza_hybrid_2012,kilintzis_supporting_2019,kim_ontology_2014,minutolo_hybrid_2016,reda_heterogeneous_2022,yu_semantic_2017}.
A well-documented issue with alerts in the health domain is alert fatigue, a phenomenon in which users become desensitized to alerts due to their frequency \cite{cash_alert_2009}.
Esposito et al. \cite{esposito_smart_2018} mitigate this by differentiating between critical and non-critical abnormal situations, with the latter being sent out in a daily summary report email rather than an instantaneous notification for each case.
Additionally, 10 systems send reminder messages of some kind to users, for instance about medications, exercise, and other interventions \cite{ammar_using_2021,chiang_context-aware_2015,hadjadj_integration_2021,henaien_combined_2020,kordestani_extended_2021,mavropoulos_smart_2021,stavropoulos_detection_2021,titi_ontology-based_2019,vadillo_enhancement_2013,yu_improving_2022}.
For example, the system proposed by Kordestani et al. \cite{kordestani_extended_2021} can remind clinicians to order additional laboratory tests when previously taken tests become out of date.

In addition to alerts and reminders, health monitoring systems can also trigger actions in response to adverse situations.
For example, the system proposed by Alti et al. \cite{alti_agent-based_2021} triggers the injection of insulin in response to a blood glucose level above a certain threshold, 
while the system proposed by Hadjadj and Halimi \cite{hadjadj_integration_2021} can trigger the opening of a vehicle door. 
Such systems must be integrated with an actuation device capable of carrying out the action.
The system proposed by Titi et al. \cite{titi_ontology-based_2019} includes several actuators such as a smoke alarm.
Other systems are integrated with actuators capable of opening doors and moving beds \cite{bampi_ontology-driven_2025}, turning on lights and heaters \cite{chiang_context-aware_2015}, making emergency calls \cite{rhayem_semantic-enabled_2021}, or turning off water or gas if detected \cite{vadillo_enhancement_2013}.

Another form of decision support is the generation of suggestions or recommendations for the mitigation of adverse situations, which is implemented in 29 systems. 
The recommendations include lifestyle modifications, such as diet and activity \cite{ali_smart_2020,ali_type-2_2018,alti_agent-based_2021,ammar_using_2021,chatterjee_automatic_2021,el-sappagh_mobile_2019,garcia-valverde_heart_2014,hadjadj_integration_2021,hussain_big-ecg_2021,kim_ontology_2014,lopes_ontology-driven_2023,rhayem_semantic-enabled_2021,spoladore_ontology-based_2021,villarreal_mobile_2014,yu_improving_2022}, 
medication \cite{alti_agent-based_2021,el-sappagh_mobile_2019,elhadj_do-care_2021,hristoskova_ontology-driven_2014,kordestani_extended_2021,lopes_ontology-driven_2023,peral_ontology-oriented_2018,rhayem_semantic-enabled_2021,titi_ontology-based_2019,xu_design_2017}, 
suitable environmental conditions \cite{chiang_context-aware_2015}, 
hospital visits \cite{zhou_design_2022}, 
and other unspecified treatments and mitigations \cite{ali_intelligent_2021,akhtar_multi-agent_2022,de_brouwer_mbrain_2022,martella_semantically_2025,mavropoulos_smart_2021,zafeiropoulos_evaluating_2024}.
Two important considerations when choosing appropriate medications are their side effects and how they interact with other medications.
Ali et al. \cite{ali_intelligent_2021} use drug review websites to collect data on side effects, while
Elhadj et al. \cite{elhadj_do-care_2021} keep track of medication interactions as well as patient allergies.
Similarly, the system proposed by Ammar et al. \cite{ammar_using_2021} includes an app where monitored individuals can report medication side effects.
This information assists clinicians in prescribing appropriate medications for each patient.
Related to recommendations is the ability for the monitored individual to seek out relevant and trusted medical information.
For example, the system proposed by Rhayem et al. \cite{rhayem_semantic-enabled_2021} includes a notification module that allows patients to contact a clinician and receive recommendations and treatments from them. 

\subsubsection{Techniques for decision support}
Much like situation detection and situation prediction, rule-based reasoning is the most commonly used technique for decision support in the selected systems. 
Rules are used by 37 of the 45 systems that implement decision support, of which 27 systems use Semantic Web rules in particular
\cite{akhtar_multi-agent_2022,ali_smart_2020,ali_type-2_2018,alti_agent-based_2021,bampi_ontology-driven_2025,chatterjee_automatic_2021,chiang_context-aware_2015,el-sappagh_mobile_2019,elhadj_do-care_2021,esposito_smart_2018,garcia-valverde_heart_2014,hadjadj_integration_2021,henaien_combined_2020,hooda_semantic_2020,hristoskova_ontology-driven_2014,kilintzis_supporting_2019,kim_ontology_2014,lopes_ontology-driven_2023,mezghani_semantic_2015,rhayem_semantic-enabled_2021,spoladore_ontology-based_2021,stavropoulos_detection_2021,titi_ontology-based_2019,yu_semantic_2017,zafeiropoulos_evaluating_2024,zeshan_iot-enabled_2023,zhang_knowledge-based_2014}.
They are typically implemented based on detected and predicted situations, such that adverse situations trigger alerts or recommendations.
Zafeiropoulos et al. \cite{zafeiropoulos_evaluating_2024} combine ML and rules, using a graph convolution network to classify alerts into one of two levels (medium or high), after which the alerts are sent based on rules.

Similar to their approach for situation detection, Xu et al. \cite{xu_design_2017} adopt case-based reasoning rather than rules for decision support.
They compare treatments and clinical paths in similar patients, and, if differences are found, the system sends an alert to flag the treatment plan for adjustment.
Among the agent-based systems, the functionality related to decision support is often delegated to agents, such as the notification and alert agents in the systems proposed by Ivașcu et al. \cite{ivascu_multi-agent_2015} and Ivașcu and Negru \cite{ivascu_activity-aware_2021}.
Additionally, interactive agents and chatbots play an important role in decision support.
Ammar et al. \cite{ammar_using_2021} propose digital assistants that provide personalised alerts and suggestions and summarise text.
Similarly, Mavropoulos et al. \cite{mavropoulos_smart_2021} propose a smart virtual agent that clinicians can interact with via voice commands, while
Yu et al. \cite{yu_improving_2022} implement an AI chatbot to answer user questions.

\subsubsection{Quality of decision support}
An important factor in the quality of decision support is user agency.
Rather than simply providing recommendations, decision support tools should allow decision-makers the agency to engage in decision-making by helping them to: 1.) identify and narrow down \textit{options}; 2.) identify \textit{possible outcomes} for each option; 3.) \textit{judge} which outcomes are most likely;  4.) identify the \textit{value} of each option based on their impact on stakeholders; 5.) make \textit{trade-offs} using the aforementioned criteria; and finally, 6.) \textit{understand} how and why the tools work \cite{miller_explainable_2023}. 
This approach to decision support aligns with the human-centred AI paradigm, which advocates for AI systems to augment and enhance human capabilities and performance, rather than automating them away \cite{de_cremer_road_2022}.
Considering the first five of these criteria, none of the selected systems demonstrate this level of decision support.
Among the systems that provide intervention recommendations, none offer more than a single option for a particular situation, nor do they identify the possible outcomes of the recommendation.
Thus, users are likely to either dismiss the recommended interventions or accept them blindly \cite{miller_explainable_2023}, neither of which is optimal.
We discuss the sixth criterion in Section~\ref{explainability} on explainability.

Furthermore, the soundness of recommended interventions can be ensured by incorporating established and clinically validated medical guidelines or vetted information from medical or allied health professional bodies.
This can contribute to the acceptance of health monitoring systems by medical professionals and regulatory bodies.
Despite this, only 15 of the selected systems mention the use of such guidelines or vetted medical information.
They include Lopes de Souza et al. \cite{lopes_ontology-driven_2023}, who incorporate risk level classifications from the American Heart Association, and De Brouwer et al. \cite{de_brouwer_mbrain_2022}, who use the International Classification of Headache Disorders' criteria to issue relevant alerts.
Garcia-Moreno et al. \cite{garcia-moreno_systematic_2023} and Spoladore et al. \cite{spoladore_ontology-based_2021} use the International Classification of Functioning, Disability and Health for health status classification; while Ali et al. \cite{ali_smart_2020} and Hristoskova et al. \cite{hristoskova_ontology-driven_2014} use the Framingham Risk Score to determine congestive heart failure risk.
El-Sappagh et al. \cite{el-sappagh_mobile_2019} extract knowledge from bodies such as the American and Canadian diabetes associations, while Yu et al. \cite{yu_semantic_2017} use guidelines from the United Kingdom's National Health Service.
Finally, the system proposed by Ammar et al. \cite{ammar_using_2021} includes an option for clinicians to refer patients to trusted medical knowledge sources such as the Centers for Disease Control and Prevention (CDC).

\subsubsection{The role of Semantic Web technologies}
The main value that Semantic Web technologies add to decision support in the selected systems is their ability to represent important concepts, particularly situations and domain knowledge, which can then be reasoned over.
This is consistent with our findings on the role of Semantic Web technologies in situation detection and situation prediction. It also aligns with previous research findings on the use of Semantic Web technologies for decision support \cite{jing_ontologies_2023} \cite{blomqvist_use_2014}.
Rules are the most common reasoning tool among the selected systems, with Semantic Web rules used for decision support in 27 of them.
However, rule-based reasoning for decision support faces the same limitations in scalability and lack of inherent uncertainty handling as discussed in the previous sections.

\subsection{Context awareness}
\label{context awareness}
An important aspect of health monitoring is the ability to take context into consideration, which is critical for situation analysis since contextual information enhances sensor data and supports its interpretation. 
This, in turn, impacts decision support.
Consider a case where an individual's heart rate is suddenly elevated. If the individual is engaged in exercise, the increased heart rate is expected. However, if the individual is at rest, this could be a cause for alarm and may necessitate intervention.
Therefore, health monitoring systems must be able to adapt based on the context of the individual being monitored.
The four most common aspects of context are location, time, identity (of a person or agent), and activity (or events) \cite{ye_ontology-based_2007,stevenson_ontonym_2009}. 

\subsubsection{Location}
Ye et al. \cite{ye_ontology-based_2007} highlight three types of locations that can be represented: symbolic locations, coordinate locations, and regions. 
The systems proposed by Akhtar et al. \cite{akhtar_multi-agent_2022}, Bampi et al. \cite{bampi_ontology-driven_2025}, Chiang and Liang \cite{chiang_context-aware_2015}, and Vadillo et al. \cite{vadillo_enhancement_2013} keep track of the different rooms in a house where an individual may be, while those proposed by Khozouie et al. \cite{khozouie_ontological_2018}, Titi et al. \cite{titi_ontology-based_2019}, and Zeshan et al. \cite{zeshan_iot-enabled_2023} indicate more generally the place the monitored individual is (for example, ``home'' or ``hospital'').
These are symbolic locations.
One purpose of such locations is to allow the systems to suggest relevant services based on the type of space currently occupied, as is the case in the system proposed by Chiang and Liang \cite{chiang_context-aware_2015}.
In the system proposed by Hristoskova et al. \cite{hristoskova_ontology-driven_2014}, the clinician's location (i.e. the room they occupy in a hospital) is used to determine which device to send notifications to, optimising for the closest device.
Likewise, Zeshan et al. \cite{zeshan_iot-enabled_2023} determine the closeness between the monitored individual and their caregivers in order to select which caregiver or clinician to notify in case of abnormal sensor observations.
This is similar to the system proposed by Alti et al. \cite{alti_agent-based_2021}, which supports a GPS sensor that captures the current coordinates of the monitored individual. In this system, location is used to select devices closest to the user from which to deploy health services so as to increase efficiency and minimise inter-device communication costs.
Coordinate locations also serve the purpose of alerting caregivers and emergency services of the exact location of a person in the event of a medical emergency, as is suggested by Rhayem et al. \cite{rhayem_semantic-enabled_2021}. 
The system proposed by Hadjadj and Halimi \cite{hadjadj_integration_2021} integrates health monitoring in the public transport system, and therefore includes location sensors in public transportation vehicles.
The final type of location is regions, which are geometrical two- or three-dimensional representations of locations \cite{ye_ontology-based_2007}.
This type of location is used in the system proposed by Kim et al. \cite{kim_ontology_2014} in order to advise users of region-specific situations, such as adverse or dangerous weather. 
Similarly, Ammar et al. \cite{ammar_using_2021} use ZIP codes, which represent a small geographic region, to extract neighbourhood-specific social determinants of health such as walkability.
El-Sappagh et al. \cite{el-sappagh_mobile_2019} use the spatial region class from the Basic Formal Ontology to represent the patient's current location, as well as the placement of the sensors.
Despite the importance of location as an aspect of context, only 21 of the 48 systems \cite{akhtar_multi-agent_2022, alti_agent-based_2021, ammar_using_2021, bampi_ontology-driven_2025, chiang_context-aware_2015, de_brouwer_mbrain_2022, el-sappagh_mobile_2019, elhadj_do-care_2021, garcia-moreno_systematic_2023, hadjadj_integration_2021, henaien_combined_2020, hristoskova_ontology-driven_2014, khozouie_ontological_2018, kim_ontology_2014, martella_semantically_2025, reda_heterogeneous_2022, rhayem_semantic-enabled_2021, titi_ontology-based_2019, vadillo_enhancement_2013, yu_semantic_2017, zeshan_iot-enabled_2023} include it.

\subsubsection{Time} 
In contrast, all of the systems include the concept of time with the exception of five \cite{ali_smart_2020, ali_type-2_2018, fenza_hybrid_2012, hooda_semantic_2020, yu_improving_2022}. 
Observation time is the most common way time is incorporated in the systems, with 25 systems capturing the exact timestamp for each sensor observation \cite{bampi_ontology-driven_2025,garcia-moreno_systematic_2023,kilintzis_supporting_2019,martella_semantically_2025} \cite{alti_agent-based_2021,chatterjee_automatic_2021,de_brouwer_mbrain_2022,el-sappagh_mobile_2019,elhadj_do-care_2021,esposito_smart_2018,hadjadj_integration_2021,hussain_big-ecg_2021,ivascu_multi-agent_2015,khozouie_ontological_2018,lopes_ontology-driven_2023,mavropoulos_smart_2021,minutolo_hybrid_2016,peral_ontology-oriented_2018,reda_heterogeneous_2022,rhayem_semantic-enabled_2021,stavropoulos_detection_2021,titi_ontology-based_2019,vadillo_enhancement_2013,zafeiropoulos_evaluating_2024,zeshan_iot-enabled_2023}.
Besides observation time, the time at which certain events occur can be captured, for example calls to emergency services \cite{alti_agent-based_2021}.
This allows the systems to display or analyse trends over time.
Additionally, Alti et al. \cite{alti_agent-based_2021} capture the time intervals in which reports should be sent.
Rather than a timestamp, Ali et al. \cite{ali_intelligent_2021} record the general time of day during which daily activities occur, i.e. morning, afternoon, or evening.
Similarly, Peral et al. \cite{peral_ontology-oriented_2018} use mealtimes as a point of reference, which is particularly important when taking blood glucose measurements. They distinguish between pre-breakfast, pre-lunch and pre-dinner readings.

Duration and frequency are other important aspects of time.
Duration can be captured for physical activity \cite{chatterjee_automatic_2021,spoladore_ontology-based_2021}, sleep \cite{chatterjee_automatic_2021,stavropoulos_detection_2021,zhou_design_2022}, disease \cite{zafeiropoulos_evaluating_2024}, symptoms \cite{mcheick_stroke_2016}, and treatment \cite{titi_ontology-based_2019,xu_design_2017}.
De Brouwer et al. \cite{de_brouwer_mbrain_2022} capture headache duration as well as the duration of events that influence headaches, such as stress and sleep.
Symptom duration can influence the risk for certain illnesses, while specifying treatment duration ensures medication reminders are sent only during the prescribed period. 
When combined with thresholds, duration can be useful in identifying different situations. 
For example, Stavropoulos et al. \cite{stavropoulos_detection_2021} determine that an individual has a lack of movement if they have fewer than 500 steps and their heart rate has been less than 100 beats per minute for longer than 800 minutes.
Frequency is used by Chiang and Liang \cite{chiang_context-aware_2015}, Spoladore et al. \cite{spoladore_ontology-based_2021}, and Yu et al. \cite{yu_semantic_2017} as a metric for physical activity.
Other examples of frequency in the systems are frequency of sensor observations \cite{mezghani_semantic_2015} and frequency of disease occurrence \cite{villarreal_mobile_2014}.

Notably, valuable features can be extracted from changes in time series sensor data. 
For instance, Hussain and Park \cite{hussain_big-ecg_2021} and Ivașcu and Negru \cite{ivascu_activity-aware_2021} use the time-domain features of the ECG to calculate heart rate and heart rate variability.
Additionally, the multi-agent system proposed by Akhtar et al. \cite{akhtar_multi-agent_2022} incorporates temporal logic, which allows for the formalization of temporal ordering operators such as ``next'', ``always'', ``until'', and ``while'' without referencing actual times \cite{alagar_temporal_2011}.
Another interesting time-related aspect is trajectory, which combines both spatial and temporal properties to represent the mobility of a sensor. 
This is incorporated in the system proposed by Rhayem et al. \cite{rhayem_semantic-enabled_2021} to define a source and destination of a sensor within a particular duration of time.


\subsubsection{Identity}
Identity, which pertains to the actors in a system, is another important aspect of context \cite{ye_ontology-based_2007}. 
This includes the definition of individuals and their properties, such as name, address, gender, and age. 
For health monitoring, this can include additional information such as weight, height, and blood group.
This is the most ubiquitous aspect of context in the systems, with every system including personal information about the monitored individuals.
Besides personal properties, identity also encompasses different user roles within the system.
Nearly all of the systems support different users besides the individual being monitored, typically including clinicians and in some cases, caregivers and family members, with the exception of 11 systems \cite{chiang_context-aware_2015, fenza_hybrid_2012, garcia-valverde_heart_2014, henaien_combined_2020, khozouie_ontological_2018, kim_ontology_2014, martella_semantically_2025, minutolo_hybrid_2016, vadillo_enhancement_2013, yu_semantic_2017, zhang_knowledge-based_2014}.
Identity also includes agents, which are used in the agent-based systems \cite{akhtar_multi-agent_2022,alti_agent-based_2021,ammar_using_2021,ivascu_activity-aware_2021,ivascu_multi-agent_2015,martella_semantically_2025,mavropoulos_smart_2021,vadillo_enhancement_2013}.
Agents\footnote{An agent is a computer system situated in some environment that is capable of acting autonomously in order to achieve some goal(s) \cite{wooldridge_intelligent_2013}} have been applied extensively in the health domain \cite{isern_systematic_2016} as well as in sensor-based systems \cite{savaglio_agent-based_2020}.
The agent-based approach offers several advantages.
For example, agents can be used as personal assistants to support humans in performing tasks and services \cite{montagna_complementing_2020}, such as the interactive virtual agent in the system proposed by Mavropoulos et al. \cite{mavropoulos_smart_2021}. 
Agent-based architectures and their advantages are discussed in greater detail in Section~\ref{architectures}.

\subsubsection{Activity}
The fourth essential aspect of context is activity. 
This can refer to physical activity or the different activities of daily living such as eating and sleeping, both of which are important considerations for situation analysis.
Activity can be derived from sensors such as accelerometers, or can be deduced from location or time (for example, a person in a bedroom in the middle of the night can be assumed to be sleeping).
Physical activity is closely tied to health, and there are many physical activity guidelines issued by governments and global health organisations, including the World Health Organisation \cite{bull_world_2020}. 
Due to this link between physical activity and health, 31 of the systems include physical activity as contextual information.
Such systems monitor physical activity using smartphones, smart watches, or inertial measurement units, which combine accelerometers, gyroscopes, and in some cases, magnetometers \cite{garcia-moreno_systematic_2023} \cite{ali_intelligent_2021,ali_smart_2020,de_brouwer_mbrain_2022,el-sappagh_mobile_2019,esposito_smart_2018,garcia-valverde_heart_2014,ivascu_activity-aware_2021,ivascu_multi-agent_2015,khozouie_ontological_2018,lopes_ontology-driven_2023,mavropoulos_smart_2021,minutolo_hybrid_2016,zafeiropoulos_evaluating_2024}.
Chiang and Liang \cite{chiang_context-aware_2015} monitor body movement using motion sensors placed around the home. This serves two purposes. Firstly, the individual's movement within the home is able to be monitored. This can determine their location at any given time. Secondly, they are able to interact with the system using body movements, such as hand-waving to activate the system.
Ali et al. \cite{ali_type-2_2018} similarly use motion sensors to keep track of body movement. They use range of motion as a metric, which is particularly important for elderly patients who may lose their ability to perform daily activities as their range of motion decreases.
The systems proposed by Ivașcu et al. \cite{ivascu_multi-agent_2015} and Zafeiropoulos et al. \cite{zafeiropoulos_evaluating_2024} perform gait analysis, capturing features such as freezing of gait, postural instability, and rigidity.

Self-reported information can also be used to determine physical activity, but this may not be accurate. To mitigate this, Chatterjee et al. \cite{chatterjee_automatic_2021} use a combination of sensor and questionnaire data. Sensors are used to monitor number of steps and duration of activity, while questionnaires are used to determine the type of activity, for example running or weightlifting.
Beyond tracking physical activity, activity recognition is also important in health monitoring.
It can help in the detection of adverse events like falls, as is done in the systems proposed by Chiang and Liang \cite{chiang_context-aware_2015}, Vadillo et al. \cite{vadillo_enhancement_2013}, and Zafeiropoulos et al. \cite{zafeiropoulos_evaluating_2024}.
Additionally, seven systems are able to recognise daily activities such as sitting, walking, and sleeping \cite{de_brouwer_mbrain_2022,garcia-moreno_systematic_2023,garcia-valverde_heart_2014,ivascu_activity-aware_2021,mavropoulos_smart_2021,rhayem_semantic-enabled_2021,zafeiropoulos_evaluating_2024}.

\subsubsection{Other types of context}

Besides location, time, identity, and activity, other types of contextual information are incorporated in the systems. 
Alti et al. \cite{alti_agent-based_2021} include hardware and network information as part of context, such as available communication protocols, CPU speed, battery power, and memory size. This information is used to ensure the efficient deployment of health services.
Similarly, Zeshan et al. \cite{zeshan_iot-enabled_2023} use battery level and device response time to determine which caregiver or clinician's device to send notifications to.
Hristoskova et al. \cite{hristoskova_ontology-driven_2014} incorporate media devices and their properties in their interpretation of context. For example, the screen size of devices such as mobile phones and tablets is used to determine how to display the health monitoring results. For small screens, the results are summarised. 
An important factor in health monitoring is the state of a person's environment. 15 of the systems use weather data such as temperature and humidity from ambient sensors to provide additional context \cite{akhtar_multi-agent_2022,bampi_ontology-driven_2025,chatterjee_automatic_2021,chiang_context-aware_2015,elhadj_do-care_2021,fenza_hybrid_2012,garcia-moreno_systematic_2023,henaien_combined_2020,khozouie_ontological_2018,kim_ontology_2014,kordestani_extended_2021,martella_semantically_2025,rhayem_semantic-enabled_2021,titi_ontology-based_2019,vadillo_enhancement_2013,zhou_design_2022}.
Weather data sources such as forecasts and indices are used by Kim et al. \cite{kim_ontology_2014} to supplement sensor data, while
the other systems \cite{akhtar_multi-agent_2022,khozouie_ontological_2018,martella_semantically_2025,vadillo_enhancement_2013,zhou_design_2022} include sensors to monitor air quality by checking the levels of different gases and/or inhalable particulate matter in the air.
Contextual information can also include details about an individual's diet, medication, and emotional state.
These details are collected in the system proposed by De Brouwer et al. \cite{de_brouwer_mbrain_2022} through self-reporting via a mobile app.
Finally, Martella et al. \cite{martella_semantically_2025} and Zafeiropoulos et al. \cite{zafeiropoulos_evaluating_2024} capture information about the user's fatigue level.

\subsubsection{The role of Semantic Web technologies}
Ontologies appear to be particularly useful for the representation of contextual information, and majority of the selected systems reuse existing ontologies to do so.
For instance, OWL-Time\footnote{\url{https://www.w3.org/TR/owl-time}}, an ontology that describes temporal properties of real-world objects such as sensors, is reused by a number of systems \cite{de_brouwer_mbrain_2022,mavropoulos_smart_2021,rhayem_semantic-enabled_2021,titi_ontology-based_2019,yu_semantic_2017}. 
Friend of a Friend (FOAF)\footnote{\url{http://xmlns.com/foaf/0.1}}, an ontology that describes people profiles, is also widely reused among the systems \cite{ammar_using_2021,garcia-moreno_systematic_2023} \cite{elhadj_do-care_2021,henaien_combined_2020,mavropoulos_smart_2021,reda_heterogeneous_2022,spoladore_ontology-based_2021,titi_ontology-based_2019,yu_semantic_2017}.
Additionally, sensor ontologies, while not focused solely on contextual information, also include some aspects of context.
For example, SAREF and SSN/SOSA ontologies include timestamps for sensor observations.
Knowledge graphs can also be used to represent contextual information, as is done by Ammar et al. \cite{ammar_using_2021} and Yu et al. \cite{yu_improving_2022}, who use a knowledge graph to capture patient profile data from health records.
Linked data is also useful for context awareness.
For instance, Ammar et al. \cite{ammar_using_2021} use linked open data to access global knowledge from the web, including data relating to social determinants of health.
Additionally, Martella et al. \cite{martella_semantically_2025} mention the possibility of using linked open data as a source of external information, although this is not implemented in the system.

Table~\ref{context-table} summarises the contextual information included in the systems and indicates which types of contextual information are captured using Semantic Web technologies.

\scriptsize
\setlength{\extrarowheight}{2pt}
\begin{longtable}{>{\raggedright\arraybackslash}p{0.02\NetTableWidth}>{\raggedright\arraybackslash}p{0.19\NetTableWidth}>{\raggedright\arraybackslash}p{0.59\NetTableWidth}>{\raggedright\arraybackslash}p{0.2\NetTableWidth}}
\caption{Summary of contextual information captured in the systems and represented using Semantic Web technologies.} \label{context-table}\\
\hline
\textbf{\#} & \textbf{System} & \textbf{Types of contextual information} & \textbf{Represented using Semantic Web technologies}\tabularnewline
\hline
\endfirsthead

\multicolumn{4}{c}{\tablename\ \thetable\ -- continued from previous page} \\
\hline
\textbf{\#} & \textbf{System} & \textbf{Types of contextual information} & \textbf{Represented using Semantic Web technologies}\tabularnewline
\hline
\endhead

\hline
\multicolumn{4}{>{\raggedright\arraybackslash}p{\NetTableWidth}}{\textbf{L} - location; \textbf{T} - time; \textbf{I} - identity; \textbf{A} - activity; \textbf{O} - other \hfill \raggedleft \textbf{Table continued on next page.}}\\
\endfoot

\hline
\multicolumn{4}{>{\raggedright\arraybackslash}p{\NetTableWidth}}{\textbf{L} - location; \textbf{T} - time; \textbf{I} - identity; \textbf{A} - activity; \textbf{O} - other}
\endlastfoot

\hline
\multicolumn{4}{>{\raggedright\arraybackslash}p{\NetTableWidth}}{\textbf{L} - location; \textbf{T} - time; \textbf{I} - identity; \textbf{A} - activity; \textbf{O} - other}
\endlastfoot

1 & Akhtar et al. \cite{akhtar_multi-agent_2022} & \textbf{L} (patient); \textbf{T} (temporal logic); \textbf{I} (profile, user roles); \textbf{O} (air quality, weather) & \textbf{L}; \textbf{I}; \textbf{O}\tabularnewline
\rowcolor{gray!4}
2 & Ali et al. \cite{ali_intelligent_2021} & \textbf{T} (activity); \textbf{I} (profile, user roles); \textbf{A} (step count, intensity level) & None\tabularnewline
3 & Ali et al. \cite{ali_smart_2020} & \textbf{I} (profile, user roles); \textbf{A} (intensity level) & \textbf{I}; \textbf{A}\tabularnewline
\rowcolor{gray!4}
4 & Ali et al. \cite{ali_type-2_2018} & \textbf{I} (profile, user roles); \textbf{A} (range of motion, intensity level) & \textbf{I}; \textbf{A}\tabularnewline
5 & Alti et al. \cite{alti_agent-based_2021} & \textbf{L} (patient, device); \textbf{T} (observation timestamps, report intervals); \textbf{I} (profile, user roles); \textbf{O} (hardware info, network info) & \textbf{L}; \textbf{T}; \textbf{I}; \textbf{O}\tabularnewline
\rowcolor{gray!4}
6 & Ammar et al. \cite{ammar_using_2021} & \textbf{L} (patient); \textbf{I} (profile, user roles); \textbf{A} (exercise); \textbf{O} (social determinants of health) & \textbf{L}; \textbf{I}; \textbf{A}; \textbf{O}\tabularnewline
7 & Bampi et al. \cite{bampi_ontology-driven_2025} & \textbf{L} (patient); \textbf{T} (observation timestamps \& latency); \textbf{I} (profile, user roles); \textbf{O} (weather) & \textbf{L}; \textbf{T}; \textbf{I}; \textbf{O}\tabularnewline
\rowcolor{gray!4}
8 & Chatterjee et al. \cite{chatterjee_automatic_2021} & \textbf{T} (observation timestamps, activity duration, sleep duration); \textbf{I} (profile, user roles); \textbf{A} (step count, intensity level, exercise type); \textbf{O} (weather) & \textbf{T}; \textbf{I}; \textbf{A}; \textbf{O}\tabularnewline
9 & Chiang and Liang \cite{chiang_context-aware_2015} & \textbf{L} (patient); \textbf{T} (activity, exercise frequency); \textbf{I} (profile); \textbf{A} (detection, motion); \textbf{O} (weather, illumination) & \textbf{L}; \textbf{T}; \textbf{I}; \textbf{A}; \textbf{O}\tabularnewline
\rowcolor{gray!4}
10 & De Brouwer et al. \cite{de_brouwer_mbrain_2022} & \textbf{L} (patient, headache location); \textbf{T} (observation timestamps, trigger duration); \textbf{I} (profile, user roles); \textbf{A} (sleep, physical activity, commute); \textbf{O} (diet, medication, emotion) &  \textbf{L}; \textbf{T}; \textbf{I}; \textbf{A}; \textbf{O}\tabularnewline
11 & El-Sappagh et al. \cite{el-sappagh_mobile_2019} & \textbf{L} (patient, sensor); \textbf{T} (observation timestamps); \textbf{I} (profile, user roles); \textbf{A} (intensity level) & \textbf{L}; \textbf{T}; \textbf{I}; \textbf{A}\tabularnewline
\rowcolor{gray!4}
12 & Elhadj et al. \cite{elhadj_do-care_2021} & \textbf{L} (patient); \textbf{T} (observation timestamps); \textbf{I} (profile, user roles); \textbf{O} (weather) & \textbf{L}; \textbf{T}; \textbf{A}; \textbf{O}\tabularnewline
13 & Esposito et al. \cite{esposito_smart_2018} & \textbf{T} (observation timestamps); \textbf{I} (profile, user roles); \textbf{A} (step count, intensity level) & \textbf{T}; \textbf{I}; \textbf{A}\tabularnewline
\rowcolor{gray!4}
14 & Fenza et al. \cite{fenza_hybrid_2012} & \textbf{I} (profile); \textbf{O} (weather) & None\tabularnewline
15 & Garcia-Moreno et al. \cite{garcia-moreno_systematic_2023} & \textbf{L} (activity); \textbf{T} (observation timestamps, duration, sampling frequency, time-domain features); \textbf{I} (profile, user roles); \textbf{A} (recognition) & \textbf{L}; \textbf{T}; \textbf{I}; \textbf{A}\tabularnewline
\rowcolor{gray!4}
16 & Garcia-Valverde et al. \cite{garcia-valverde_heart_2014} & \textbf{T} (situation timestamps); \textbf{I} (profile); \textbf{A} (recognition, intensity level) & \textbf{T}; \textbf{I}; \textbf{A}\tabularnewline
17 & Hadjadj and Halimi \cite{hadjadj_integration_2021} & \textbf{L} (vehicles, bus stop); \textbf{T} (observation timestamps); \textbf{I} (profile, user roles); \textbf{O} (passenger count, vehicle status) & \textbf{L}; \textbf{T}; \textbf{I}; \textbf{O}\tabularnewline
\rowcolor{gray!4}
18 & Henaien et al. \cite{henaien_combined_2020} & \textbf{L} (patient); \textbf{I} (profile); \textbf{A} (motion); \textbf{O} (weather) & \textbf{L}; \textbf{I}; \textbf{A}\tabularnewline
19 & Hooda and Rani \cite{hooda_semantic_2020} & \textbf{I} (profile, user roles) & \textbf{I}\tabularnewline
\rowcolor{gray!4}
20 & Hristoskova et al. \cite{hristoskova_ontology-driven_2014} & \textbf{L} (clinician, device); \textbf{T} (risk horizon); \textbf{I} (profile, user roles); \textbf{O} (device size) & \textbf{L}; \textbf{I}; \textbf{A}; \textbf{O}\tabularnewline
21 & Hussain and Park \cite{hussain_big-ecg_2021} & \textbf{T} (observation timestamps; time-domain features); \textbf{I} (profile, user roles) & None\tabularnewline
\rowcolor{gray!4}
22 & Ivașcu and Negru \cite{ivascu_activity-aware_2021} & \textbf{T} (time-domain features); \textbf{I} (profile, user roles);  \textbf{A} (recognition, intensity level) & \textbf{T}; \textbf{I}; \textbf{A}\tabularnewline
23 & Ivașcu et al. \cite{ivascu_multi-agent_2015} & \textbf{T} (observation timestamps); \textbf{I} (profile, user roles); \textbf{A} (sleep quality, gait analysis) & \textbf{A}\tabularnewline
\rowcolor{gray!4}
24 & Khozouie et al. \cite{khozouie_ontological_2018} & \textbf{L} (patient); \textbf{T} (observation timestamps \& intervals); \textbf{I} (profile); \textbf{A} (type); \textbf{O} (air quality, weather) & \textbf{L}; \textbf{T}; \textbf{I}; \textbf{A}; \textbf{O}\tabularnewline
25 & Kilintzis et al. \cite{kilintzis_supporting_2019} & \textbf{T} (observation timestamps); \textbf{I} (profile, user roles); \textbf{A} (step count, sleep stage ratio) & \textbf{T}; \textbf{I}; \textbf{A}\tabularnewline
\rowcolor{gray!4}
26 & Kim et al. \cite{kim_ontology_2014} & \textbf{L} (patient's region); \textbf{I} (profile); \textbf{O} (weather) & \textbf{L}; \textbf{A}; \textbf{O}\tabularnewline
27 & Kordestani et al. \cite{kordestani_extended_2021} & \textbf{T} (episode timestamps); \textbf{I} (profile, user roles); \textbf{O} (weather) & \textbf{I}; \textbf{O}\tabularnewline
\rowcolor{gray!4}
28 & Lopes de Souza et al. \cite{lopes_ontology-driven_2023} & \textbf{T} (observation timestamps); \textbf{I} (profile, user roles); \textbf{A} (movement) & \textbf{T}; \textbf{I}; \textbf{A} \tabularnewline
29 & Martella et al. \cite{martella_semantically_2025} & \textbf{L} (patient); \textbf{T} (observation timestamps \& duration); \textbf{I} (profile); \textbf{A} (physical effort); \textbf{O} (air quality, weather, ambient light, sound pressure, electromagnetic emissions, fatigue) & \textbf{L}; \textbf{T}; \textbf{I}; \textbf{A}; \textbf{O}\tabularnewline
\rowcolor{gray!4}
30 & Mavropoulos et al. \cite{mavropoulos_smart_2021} &  \textbf{T} (observation timestamps; time-domain features); \textbf{I} (profile, user roles); \textbf{A} (recognition) & \textbf{T}; \textbf{I}; \textbf{A}\tabularnewline
31 & Mcheick et al. \cite{mcheick_stroke_2016} & \textbf{T} (symptom duration); \textbf{I} (profile, user roles) & \textbf{I}\tabularnewline
\rowcolor{gray!4}
32 & Mezghani et al. \cite{mezghani_semantic_2015} & \textbf{T} (observation start/end date, observation frequency, anomaly timestamps); \textbf{I} (profile, user roles) & \textbf{T}; \textbf{I}\tabularnewline
33 & Minutolo et al. \cite{minutolo_hybrid_2016} & \textbf{T} (observation timestamps); \textbf{I} (profile); \textbf{A} (step count) & \textbf{T}; \textbf{I}; \textbf{A}\tabularnewline
\rowcolor{gray!4}
34 & Peral et al. \cite{peral_ontology-oriented_2018} & \textbf{T} (observation timestamps); \textbf{I} (profile, user roles) & \textbf{T}; \textbf{I}\tabularnewline
35 & Reda et al. \cite{reda_heterogeneous_2022} & \textbf{L} (patient); \textbf{T} (observation timeframe); \textbf{I} (profile, user roles); \textbf{A} (step count, type, intensity) & \textbf{L}; \textbf{T}; \textbf{I}; \textbf{A}\tabularnewline
\rowcolor{gray!4}
36 & Rhayem et al. \cite{rhayem_semantic-enabled_2021} & \textbf{L} (patient, device trajectory); \textbf{T} (observation timestamps); \textbf{I} (profile, user roles); \textbf{A} (recognition); \textbf{O} (weather) & \textbf{L}; \textbf{T}; \textbf{I}; \textbf{A}; \textbf{O}\tabularnewline
37 & Spoladore et al. \cite{spoladore_ontology-based_2021} & \textbf{T} (exercise timestamps, duration \& frequency); \textbf{I} (profile, user roles); \textbf{A} (exercise type) & \textbf{T}; \textbf{I}; \textbf{A}\tabularnewline
\rowcolor{gray!4}
38 & Stavropoulos et al. \cite{stavropoulos_detection_2021} & \textbf{T} (observation timestamps, sleep duration, time taken to fall asleep); \textbf{I} (profile, user roles); \textbf{A} (sleep quality, step count, intensity level) & \textbf{T}; \textbf{A}\tabularnewline
39 & Titi et al. \cite{titi_ontology-based_2019} & \textbf{L} (patient); \textbf{T} (observation timestamps, intervals, \& duration);  \textbf{I} (profile, user roles); \textbf{A} (type, intensity level); \textbf{O} (weather) & \textbf{L}; \textbf{T}; \textbf{I}; \textbf{A}; \textbf{O}\tabularnewline
\rowcolor{gray!4}
40 & Vadillo et al. \cite{vadillo_enhancement_2013} & \textbf{L} (patient); \textbf{T} (observation timestamps);  \textbf{I} (profile); \textbf{A} (detection); \textbf{O} (air quality, weather) & \textbf{L}; \textbf{I}\tabularnewline
41 & Villarreal et al. \cite{villarreal_mobile_2014} & \textbf{T} (disease duration \& frequency); \textbf{I} (profile, user roles); \textbf{A} (type) & \textbf{T}; \textbf{I}; \textbf{A}\tabularnewline
\rowcolor{gray!4}
42 & Xu et al. \cite{xu_design_2017} & \textbf{T} (treatment duration); \textbf{I} (profile, user roles) & \textbf{I}\tabularnewline
43 & Yu et al. \cite{yu_improving_2022} & \textbf{I} (profile, user roles); \textbf{A} (exercise); \textbf{O} (diet, medication) &  \textbf{I}; \textbf{A}; \textbf{O}\tabularnewline
\rowcolor{gray!4}
44 & Yu et al. \cite{yu_semantic_2017} & \textbf{L} (patient); \textbf{T} (disease progression, medical event timestamp; exercise frequency); \textbf{I} (profile); \textbf{A} (type) & \textbf{L}; \textbf{T}; \textbf{I}; \textbf{A}\tabularnewline
45 & Zafeiropoulos et al. \cite{zafeiropoulos_evaluating_2024} & \textbf{T} (disease duration, observation timestamp); \textbf{I} (profile, user roles); \textbf{A} (gait analysis, sleep quality); \textbf{O} (fatigue) & \textbf{T}; \textbf{I}; \textbf{A}; \textbf{O}\tabularnewline
\rowcolor{gray!4}
46 & Zeshan et al. \cite{zeshan_iot-enabled_2023} & \textbf{L} (patient, caregiver, clinician); \textbf{T} (observation timestamps); \textbf{I} (profile, user roles); \textbf{O} (battery level; response time) & \textbf{L}; \textbf{T}; \textbf{I}\tabularnewline
47 & Zhang et al. \cite{zhang_knowledge-based_2014} & \textbf{T} (observation timestamps); \textbf{I} (profile) & \textbf{I}\tabularnewline
\rowcolor{gray!4}
48 & Zhou et al. \cite{zhou_design_2022} & \textbf{T} (movement timestamps, sleep duration); \textbf{I} (profile, user roles); \textbf{A} (sleep quality); \textbf{O} (air quality, weather) & None\tabularnewline
\end{longtable}
\normalsize

\subsection{Explainability} 
\label{explainability}
A critical consideration in health monitoring systems, particularly those that incorporate AI, is explainability.
For the purposes of this study, we adopt the perspective that explainability is essentially equivalent to interpretability \cite{miller_explanation_2019}, which in turn can be defined as the degree to which a system's operations can be understood by a human \cite{biran_explanation_2017}.
Explainability contributes significantly to the overall trustworthiness and adoption of systems in the health domain.
Two complementary approaches to explainability in AI systems are prioritising human understanding of generated outputs, and providing explicit explanations for those outputs \cite{miller_explanation_2019}.
The former can be achieved by using techniques that are inherently intuitive and easily comprehensible (intrinsic explainability), or by explaining the workings of a model after it has been trained (post hoc explainability) \cite{molnar_interpretable_2023}.
We examine both intrinsic and post hoc explainability below, followed by a consideration of explicit explanations and their quality.

\subsubsection{Intrinsic explainability}
It can be argued that Semantic Web technologies are inherently intuitive and can contribute to the development of explainable systems; we discuss this in Section~\ref{role-explainability}.
However, the use of these technologies does not guarantee the explainability of the system as a whole. 
Other technologies and techniques implemented within the systems, and the ways in which they are combined, can also play a role in either enhancing or hindering the overall explainability of the system.
For example, rules are inherently easy to understand \cite{hagras_toward_2018}, and nearly all the selected systems implement some form of rule-based reasoning using Semantic Web rule languages.
However, in some cases, rules are implemented for one aspect of the system while less interpretable techniques are used for other components.
An instance of this is the system proposed by Ali et al. \cite{ali_smart_2020}, where a deep learning model is used for disease prediction, while rule-based reasoning is applied for recommendation generation. 
This results in the decision support functionality being highly explainable, while the situation prediction component remains less so.

Bayesian networks can also be considered highly interpretable, as they can perform predictive and diagnostic reasoning \cite{derks_taxonomy_2020} in a way that can be visually interpreted due to their graphical structure \cite{kyrimi_comprehensive_2021}.
Predictive reasoning is done by Mcheick et al. \cite{mcheick_stroke_2016}, who use a Bayesian network to determine whether a person has a high risk of stroke based on risk factors. The reason for whether the risk is high or not can be traced back to the presence of risk factors. 
On the other hand, Kordestani et al. \cite{kordestani_extended_2021} use a Bayesian network for diagnostic reasoning, allowing for the understanding of a kidney disease diagnosis based on its causes.
Fuzzy logic represents another class of interpretable techniques, since it allows variables and their classifications to be presented in a way that is intuitive \cite{hagras_toward_2018}.  
Fuzzy approaches are used in five of the selected systems; we discuss them in more detail in Section~\ref{fuzzy-logic}.

While ML is often criticized for its tendency to produce black box models, certain ML models are intrinsically interpretable, such as logistic or linear regression models and decision trees \cite{petch_opening_2022,molnar_interpretable_2023}.
However, even such models can be rendered uninterpretable as their complexity and scale increase, as is the case with large decision trees \cite{petch_opening_2022}.
In cases where less interpretable ML models are used, they can be combined with Semantic Web technologies to enhance explainability. 
This is closely related to neuro-symbolic AI, in which the strengths of neural networks and symbolic AI are combined to achieve the best of both worlds \cite{sarker_neuro-symbolic_2021}.
Zafeiropoulos et al. \cite{zafeiropoulos_evaluating_2024} and Zhou et al. \cite{zhou_design_2022} explore this hybrid approach, with both using a knowledge graph to provide training data for deep learning models.
Additionally, Ali et al. \cite{ali_intelligent_2021} use an ontology for feature extraction, providing some transparency into the selection of features for their BiLSTM model.

\subsubsection{Post hoc explainability}
The use of inherently interpretable models should be prioritised in high-stakes domains \cite{rudin_stop_2019}.
Nonetheless, explaining the workings of a model after it has been trained is a well-established approach to explainability \cite{molnar_interpretable_2023}.
Two well-known post hoc explainability techniques are Local Interpretable Model-agnostic Explanations (LIME) \cite{ribeiro_why_2016} and SHapley Additive exPlanations (SHAP) \cite{lundberg_unified_2017}.
LIME approximates a black box model with an interpretable surrogate model that is then used to explain the original model's predictions.
SHAP, by contrast, uses Shapley values from coalitional game theory to explain individual predictions by quantifying how much each feature contributes to the outcome.
Beyond these methods, plotting techniques can also be used to visualise feature effects.
They include partial dependence plots, which show the marginal effects of one or two features on predictions; accumulated local effects plots, which provide more accurate feature effects by accounting for feature correlations; and individual conditional expectation plots, which show how an individual prediction changes when a feature changes \cite{molnar_interpretable_2023}.
However, such post hoc explainability methods are not mentioned in any of the selected systems.

\subsubsection{Explanation quality}
Once the reasoning behind the system's outputs has been determined, the next step is to communicate this to the users of the system through explicit explanations.
An explanation can be defined as the answer to a why-question \cite{miller_explanation_2019}, such as \textit{"Why was the situation classified as an emergency?"}, \textit{"Why was an alert sent to the clinician?"}, or \textit{"Why was this medication recommended?"}.
 Good explanations have the following characteristics \cite{molnar_interpretable_2023}:
\begin{enumerate}
    \item They are \textit{contrastive}, showing the contrast between the selected outcome versus other possibilities.
    \item They are \textit{selected}, distilling long lists of causes into one to three main ones.
    \item They \textit{highlight abnormalities}, specifying causes that had a small probability but nevertheless occurred and influenced the outcome.
    \item They are \textit{consistent with existing knowledge or prior beliefs}, as humans tend to devalue or ignore explanations that differ from their beliefs due to confirmation bias. 
    This is especially important for clinicians, since explanations that contradict established medical knowledge can undermine their trust.
    \item They are \textit{general and probable}, identifying causes that can explain many similar outcomes.
\end{enumerate}

The target audience is also an important consideration in the communication of explanations, since different audiences may value distinct aspects based on their role and perspective \cite{barredo_arrieta_explainable_2020}. 
Health monitoring systems should therefore tailor explanations to the needs of various audiences, such as monitored individuals, clinicians, and caregivers.
For certain audiences, like regulatory stakeholders, it may be more appropriate to include explanations as part of the overall system documentation.
Only eight of the selected systems report or indicate that some form of explanation is made available to users \cite{chatterjee_automatic_2021,martella_semantically_2025} \cite{akhtar_multi-agent_2022,chiang_context-aware_2015,elhadj_do-care_2021,mavropoulos_smart_2021,rhayem_semantic-enabled_2021,villarreal_mobile_2014,yu_improving_2022}.
This is typically through a user interface, messaging platform, or, as in the system proposed by Akhtar et al. \cite{akhtar_multi-agent_2022}, as part of system logs. 
Chatbots and virtual agents, as implemented by Mavroppoulos et al. \cite{mavropoulos_smart_2021} and Yu et al. \cite{yu_improving_2022}, can also provide explanations, since they are designed to answer queries from users.

\subsubsection{The role of Semantic Web technologies}
\label{role-explainability}
Domain knowledge can enhance explainability \cite{tocchetti_role_2022}, and Semantic Web technologies excel at structuring such knowledge formally and unambiguously \cite{confalonieri_multiple_2023}.
Ontologies can contribute to the development of explainable systems from three perspectives: by providing sound and explicit knowledge reference models; by supporting common-sense reasoning through the representation of context-aware semantic information; and by facilitating flexible knowledge abstraction and refinement \cite{confalonieri_multiple_2023}.
Since most of the systems use ontologies to represent expert knowledge and contextual information, this can be seen as a first step towards achieving explainability.
Moreover, many ontology reasoners such as HermiT \cite{shearer_hermit_2008} and Pellet \cite{sirin_pellet_2007} provide justifications for their inferences.
These justifications can be generated into human-readable explanations, which in turn can be presented to system users \cite{van_woensel_explanations_2024}.
Although 11 systems \cite{ali_type-2_2018,chatterjee_automatic_2021,el-sappagh_mobile_2019,elhadj_do-care_2021,esposito_smart_2018,hooda_semantic_2020,hristoskova_ontology-driven_2014,khozouie_ontological_2018,titi_ontology-based_2019,vadillo_enhancement_2013,zafeiropoulos_evaluating_2024} mention using ontology reasoners, none of them indicate that the reasoner justifications are presented to users as explanations.
Beyond reasoners, explanations themselves can be represented using ontologies.
For instance, the Explanation Ontology \cite{chari_explanation_2024} is a general-purpose ontology that connects explanations to underlying data and knowledge.
Similarly, the Evidence and Conclusion Ontology \cite{nadendla_eco_2022} captures evidence to support annotations and assertions in the biomedical domain, which can be used to generate explanations.
Yet, none of the reviewed systems formalize explanations in this way.

Knowledge graphs can also contribute to better understanding of system outputs in several ways, including providing a graph-based visualisation of concepts, entity and relation extraction from unstructured data, enrichment of datasets, and inference and reasoning \cite{rajabi_knowledge_2022}.
Finally, although the connection between linked data and explainability is not direct, the open accessibility and interconnection of knowledge shared using a linked data approach can nonetheless contribute to explainability.

\subsection{Uncertainty handling} 
\label{uncertainty}
Given the uncertainty inherent in health decision-making as well as the high likelihood of ambiguity, noise, and missing values in sensor observations, health monitoring systems are greatly enhanced by being able to handle uncertainty. 
Despite this, only 22 of the systems addressed some aspect of this \cite{akhtar_multi-agent_2022,ali_intelligent_2021,ali_smart_2020,ali_type-2_2018,chiang_context-aware_2015,esposito_smart_2018,fenza_hybrid_2012,garcia-moreno_systematic_2023,garcia-valverde_heart_2014,hooda_semantic_2020,hristoskova_ontology-driven_2014,hussain_big-ecg_2021,kilintzis_supporting_2019,kordestani_extended_2021,martella_semantically_2025,mcheick_stroke_2016,mezghani_semantic_2015,minutolo_hybrid_2016,reda_heterogeneous_2022,rhayem_semantic-enabled_2021,titi_ontology-based_2019,zhou_design_2022}.
The approaches used to handle uncertainty are summarised in Table~\ref{uncertainty-table} and discussed in detail in the remainder of this subsection.

\setlength{\extrarowheight}{2pt}
\begin{table*}[ht]
\caption{Approaches to handle uncertainty and the systems that use them.} \label{uncertainty-table}
\begin{tabular}{p{.4\NetTableWidth}p{.6\NetTableWidth}}
\hline
\textbf{Approach} & \textbf{Systems}\\
\hline
Fuzzy logic & Ali et al. \cite{ali_type-2_2018}, Chiang and Liang \cite{chiang_context-aware_2015}, Esposito et al. \cite{esposito_smart_2018}; Fenza et al. \cite{fenza_hybrid_2012}; Minutolo et al. \cite{minutolo_hybrid_2016}\\
\rowcolor{gray!4}Bayesian networks & Kordestani et al. \cite{kordestani_extended_2021}, Mcheick et al. \cite{mcheick_stroke_2016}\\
Answer set programming with probabilistic rules & Kordestani et al. \cite{kordestani_extended_2021}\\
\rowcolor{gray!4}Probabilistic risk classification & Hristoskova et al. \cite{hristoskova_ontology-driven_2014}\\
Defeasible logic & Akhtar et al. \cite{akhtar_multi-agent_2022}\\
\rowcolor{gray!4}Replacing or eliminating missing or invalid sensor data & Ali et al. \cite{ali_intelligent_2021,ali_smart_2020}; Garcia-Moreno et al. \cite{garcia-moreno_systematic_2023}; Hooda and Rani \cite{hooda_semantic_2020}; Hussain and Park \cite{hussain_big-ecg_2021}; Martella et al. \cite{martella_semantically_2025}; Reda at al. \cite{reda_heterogeneous_2022}; Rhayem et al. \cite{rhayem_semantic-enabled_2021}; Titi et al. \cite{titi_ontology-based_2019}\\
Filtering sensor data & Ali et al. \cite{ali_intelligent_2021,ali_smart_2020}; Garcia-Valverde et al. \cite{garcia-valverde_heart_2014}\\
\hline
\end{tabular}
\end{table*}

\subsubsection{Fuzzy logic}
\label{fuzzy-logic}
Fuzzy logic is a widely used technique for representing ambiguity and vagueness in sensor data \cite{khaleghi_multisensor_2011}.
It is used by five of the systems, making it the most commonly implemented uncertainty handling approach among the systems, besides the preprocessing of sensor data.
In fuzzy logic, the truth of a statement is not binary (i.e. either true or false), but can rather be represented in a range from false to true.
Therefore, rather than having crisp thresholds for different categories, fuzzy logic allows for values with different degrees of membership for the different categories. The process of converting crisp inputs into fuzzy sets is called fuzzification.
For example, heart rate is represented in beats per minute, which can be classified into  crisp categories.
Generally, a heart rate greater than 100 beats per minute can be categorised as ``fast'', a heart rate between 60 and 100 beats per minute can be categorised as ``normal'', and a heart rate below 60 beats per minute can be categorised as ``slow'' \cite{bennett_bennetts_2013}.
However, with fuzzy logic, any given heart rate value has a certain degree of membership to any of the categories. 
For instance, a heart rate of 80 beats per minute may have a high degree of membership to the ``normal'' category (for example, 75\%), a lower degree of membership to the ``fast'' category (for example, 20\%), and an even lower degree of membership to the ``slow'' category (for example, 5\%). Fuzzy logic provides a better approach to deal with boundary conditions.
For example, when the heart rate is either 100 or 101, it can be reflected as mostly normal and to a lesser degree fast. Both Ali et al. \cite{ali_type-2_2018} and Chiang and Liang \cite{chiang_context-aware_2015} fuzzify sensor data such as blood pressure and heart rate, as well as attributes such as age and weight.
Similarly, Esposito et al. \cite{esposito_smart_2018} fuzzify the intensity of physical activity, which provides important context for heart rate thresholds.
Fenza et al. \cite{fenza_hybrid_2012} incorporate fuzzy logic with rules to determine the degree of membership to different situation categories based on different combinations of vital signs, while 
Minutolo et al. \cite{minutolo_hybrid_2016} use hybrid rules that incorporate both crisp and fuzzy variables. Fuzzy logic provides a simple but effective mechanism for representing imprecision and vagueness in sensor observations and allows this to be taken into account for more effective situation detection.  

\subsubsection{Bayesian networks}
Bayesian networks are well known for modelling uncertainty and have been widely used in the health domain \cite{kyrimi_comprehensive_2021}.
Kordestani et al. \cite{kordestani_extended_2021} use a Bayesian network for probabilistic diagnosis of acute kidney injury. 
The Bayesian network models immediate (short-term) and background (long-term) causes of acute kidney injury, as well as its symptoms.
They used experts to determine the conditional probabilities of the presence of acute kidney injury given these variables.
Similarly, Mcheick et al. \cite{mcheick_stroke_2016} represent risk factors for stroke using a Bayesian network.

\subsubsection{Nonmonotonic reasoning}
Monotonic reasoning holds that the rejection of an earlier conclusion must only be done if the evidence for the conclusion is also rejected.
Contrastingly, nonmonotonic reasoning holds that an earlier conclusion can be rejected based on new evidence, even when earlier evidence was valid \cite{nute_defeasible_2003}.
This ability to revise conclusions in the face of new evidence is useful in handling uncertainty. 
Defeasible logic is an example of nonmonotonic reasoning in which there are three kinds of rules: strict rules which can never have exceptions, defeasible rules which are typically true but can have exceptions, and undercutting defeaters which are weak possibilities \cite{nute_defeasible_2003}.
Akhtar et al. \cite{akhtar_multi-agent_2022} use defeasible logic to handle inconsistencies in sensor data as well as patient information.
Another type of nonmonotonic reasoning is answer set programming (ASP), which is used by Kordestani et al. \cite{kordestani_extended_2021} to automatically customise treatments for each patient.
They combine ASP with probability to reason with uncertain knowledge regarding treatment.
Using probabilistic ASP rules, their proposed system obtains all possible treatment options for a medical episode and the associated probability of the episode occurring.
If the probability of the episode occurring decreases with a particular treatment, then the treatment's award value is increased.
The treatment with the highest award value is ultimately selected by the system.

\subsubsection{Probabilistic risk classification}
Hristoskova et al. \cite{hristoskova_ontology-driven_2014} account for uncertainty in situation prediction by classifying patients into risk stages based on the four-year probability of congestive heart failure.
They use probabilistic rules defined using SWRL to analyse risk factors and classify individuals based on the Framingham Risk Score.

\subsubsection{Preprocessing sensor data}
Uncertainty can stem from various factors in sensor data, including ambiguous or imprecise readings, noise, or missing values caused by sensor malfunctions or network failures \cite{gravina_multi-sensor_2017,khaleghi_multisensor_2011}.
16 systems have addressed the issue of missing or invalid values in sensor data.
Ali et al. \cite{ali_intelligent_2021,ali_smart_2020} replace them with mean and median values from existing data, while
Hooda and Rani \cite{hooda_semantic_2020} replace them using the preceding value.
Rhayem et al. \cite{rhayem_semantic-enabled_2021} take the approach of removing any missing or unusual values, for example those outside the device measurement ranges.
Similarly, Kilintzis et al. \cite{kilintzis_supporting_2019}, Martella et al. \cite{martella_semantically_2025}, Titi et al. \cite{titi_ontology-based_2019} and Reda et al. \cite{reda_heterogeneous_2022} use rules to check whether sensor data falls within the expected minimum and maximum bounds. 
In their proposed system, Hussain and Park \cite{hussain_big-ecg_2021} use the Pan-Tompkins algorithm to detect the QRS complex in the ECG. This identifies beats without a QRS complex, which may be premature, missing, or ectopic, and are subsequently eliminated.
While Garcia-Moreno et al. \cite{garcia-moreno_systematic_2023} mention missing value imputation as part of their ML pipeline, they do not specify a technique for this.
To deal with noisy data, three systems use filters to improve signal quality.
Ali et al. \cite{ali_intelligent_2021,ali_smart_2020} use a Kalman filter to remove noise, while
Garcia-Valverde et al. \cite{garcia-valverde_heart_2014} use a moving average filter for the same purpose.

\subsubsection{The role of Semantic Web technologies}
\label{uncertainty-role}
As discussed in the previous challenges, Semantic Web technologies provide limited inherent support for uncertainty handling, but this can be mitigated through the techniques discussed in this section.
Several extensions for Semantic Web standards have been proposed in the literature that make use of fuzzy logic and Bayesian inference, such as BayesOWL \cite{ding_bayesowl_2006} and Bayes-SWRL \cite{liu_bayes-swrl_2013}, probabilistic extensions for OWL and SWRL respectively, as well as fuzzyDL \cite{bobillo_fuzzy_2016}, a fuzzy ontology reasoner.
However, none of the selected systems report using any such extensions, opting instead to define custom solutions.
Another way that Semantic Web technologies can support uncertainty handling is through using ontology reasoners for inconsistency detection.
Missing values or otherwise invalid data can be detected through ontology reasoners based on specified and inferred axioms; the use of these tools are discussed in greater detail in Section~\ref{system-quality}.
Additionally, rule-based reasoning can be combined with ontologies to detect invalid data, an approach used by five of the selected systems \cite{rhayem_semantic-enabled_2021,kilintzis_supporting_2019,martella_semantically_2025,titi_ontology-based_2019,reda_heterogeneous_2022}.
Finally, Semantic Web technologies can indirectly support uncertainty handling by representing data quality properties that affect uncertainty. For instance, Garcia-Moreno et al. \cite{garcia-moreno_systematic_2023} model quality properties of sensor data using an ontology, including correctness and completeness.

\subsection{Other challenges}
While we consider the seven challenges discussed above to be particularly salient in sensor-based personal health monitoring systems, we acknowledge that there are other factors that such systems must take into account.
This subsection briefly discusses a few of them.
As an in-depth analysis of these additional challenges is outside the scope of this study, we also include references to relevant articles that interested readers can consult.

\subsubsection{Security and privacy}
As health-related information is highly sensitive, security and privacy are important to consider. Particular aspects of this include security of data storage, network and transmission security, user authentication and access control, consent management, and the use of privacy-preserving techniques such as federated learning. For insights on security and privacy, we direct interested readers to the following articles: Rasool et al. \cite{rasool_security_2022} review security and privacy in the context of the Internet of Medical Things; Thapa and Camtepe \cite{thapa_precision_2021} explore security and privacy challenges and techniques for health data in general; and finally, Kirrane et al. \cite{kirrane_privacy_2018} provide an overview of security and privacy issues that relate to Semantic Web technologies.

\subsubsection{Usability}
Usability is another factor that health monitoring systems should consider, and can broadly be defined as the ease of use of a system \cite{sagar_systematic_2017}. It is a multi-faceted concept with several contributing factors, including understandability (which is closely related to explainability), attractiveness, and overall user satisfaction \cite{sagar_systematic_2017,saeed_exploration_2020}.
Interested readers can refer to the following articles for more information: 
Saeed at el. \cite{saeed_exploration_2020} explore pertinent usability issues in health monitoring systems and identify possible solutions.
With regard to the evaluation of usability, Maramba et al. \cite{maramba_methods_2019} identify current methods used in usability testing in health monitoring applications, while Cho et al. \cite{cho_multi-level_2018} present a usability evaluation framework for mobile health applications.
Finally, considering the usability of the sensors themselves, the reviews by Cusack et al. \cite{cusack_reviewsmart_2024}, Dias and Cunha \cite{dias_wearable_2018}, and Andreu-Perez et al. \cite{andreu-perez_wearable_2015} highlight the types and characteristics of wearable sensors available for health monitoring.
    
\subsubsection{Scalability}
Scalability generally refers to the ability of a system to handle increased workload \cite{weinstock_system_2006}.
In the context of sensor-based health monitoring, this workload could arise from an increased number of sensors, other data sources, users, and services provided by the system.
For further reading on scalability in IoT applications, readers can consult the review by Fortino et al. \cite{fortino_internet_2021}.
The reviews by Albahri et al. \cite{albahri_real-time_2018} and Phillip et al. \cite{philip_internet_2021} discuss scalability in the context of IoT and healthcare.

\subsubsection{Ethics and regulatory compliance}
The high-stakes nature of the health domain necessitates careful consideration of ethical issues.
Although there are many benefits of technology-enabled personal health monitoring, there are also potential harms that it exposes.
Many of these ethical issues overlap with the challenges already discussed, such as situation detection and situation prediction (how accurate and reliable are the detected and predicted situations?), decision support (how appropriate are the suggested recommendations and how much autonomy do system users have?), explainability (to what extent can system outputs be understood?), and security and privacy (how secure is user data and how is consent managed?). 
An additional ethical concern is the cascade of care, a phenomenon in which incidental findings from screenings or monitoring result in further clinical care.
This may cause undue anxiety and psychological harm to the monitored individual, while also potentially leading to costly or invasive diagnostic procedures.
Some of these ethical considerations can be enforced through regulation.
Readers seeking further exploration on this challenge may consult the following articles: 
Morley et al. \cite{morley_ethics_2020} comprehensively map the ethics of AI in healthcare; 
Kwan et al. \cite{kwan_healthcare_2017} highlight the ethical issues associated with health monitoring applications; 
Nittari et al. \cite{nittari_telemedicine_2020} review ethical and legal challenges in telemedicine more broadly; 
Hassanaly and Dufour \cite{hassanaly_analysis_2021} explore the regulation of mobile health applications in the United States, the European Union, and France; and
Thapa and Camtepe \cite{thapa_precision_2021} explore legality and regulatory compliance from the perspective of security and privacy.

\subsection{Challenges assessment}
\label{challenges-assessment}
In this subsection, we conduct an assessment of each system based on the seven key challenges.
Table~\ref{challenges-aspects} summarises the different aspects related to the challenges.
While we consider these aspects to be highly important for achieving effective sensor-based personal health monitoring, it is possible that some of them may exceed the requirements for specific health conditions or applications, and may therefore not be essential in those particular cases.
To assess the degree to which each system tackles the seven challenges, we use a rating scheme based on the identified aspects. 
A score of 1 point is assigned to each system for every aspect that is addressed. All challenges are equally weighted.
We use a four-point rating scale as follows:
\begin{enumerate}
    \item \xmark: None of the aspects are addressed by the system
    \item \textbf{Low}: 40\% or less of the aspects are addressed
    \item \textbf{Medium}: between 41\% and 69\% of the aspects are addressed
    \item \textbf{High}: more than 70\% of the aspects are addressed
\end{enumerate} 

\footnotesize
\begin{longtable}{p{.1\linewidth}p{.83\linewidth}}
\caption{Important aspects related to the seven key challenges.} \label{challenges-aspects}\\
\hline
\textbf{Challenge} & \textbf{Aspects}\\
\hline
\endfirsthead

\multicolumn{2}{c}{\tablename\ \thetable\ -- continued from previous page} \\
\hline
\textbf{Challenge} & \textbf{Aspects}\\
\hline
\endhead

\hline
\multicolumn{2}{r}{\textbf{Table continued on next page.}}\\
\endfoot

\hline
\endlastfoot

Interoperability &
\parbox{\linewidth}{\begin{enumerate}
     \item It is mentioned or illustrated how the system addresses the technical interoperability between the sensors and the rest of the system, e.g. using a gateway device, base unit/station, or established data transmission standards and protocols. 
    \item The system incorporates established standards or ontologies for describing sensor data, such as the SSN and SAREF ontologies.
    \item The system makes use of established health and medical terminologies and nomenclatures such as SNOMED CT, ICD, and ICNP.
    \item The system makes use of existing health data standards such as ISO/IEEE 11073, FHIR, and HL7 V2. 
    \item The system integrates existing health and medical records.
    \item The system integrates other sources of data such as weather forecasts, social networks, and other web data.
\end{enumerate}}
\\
\rowcolor{gray!4}Situation detection & 
\parbox{\linewidth}{\begin{enumerate}
    \item The system can detect deviations or abnormalities in physiological measurements based on historical observations or known thresholds.
    \item The system can classify individuals or situations into predefined categories, levels, or states related to health.
    \item The system can detect medical conditions or diseases that are currently being experienced.
\end{enumerate}}
\\
Situation prediction &
\parbox{\linewidth}{\begin{enumerate}
    \item The system can predict the risk of medical conditions, diseases, or other adverse effects in the future.
    \item The system can predict future physiological measurements based on current or historical sensor observations.
    \item The system can predict the prognosis of detected diseases.
\end{enumerate}}
\\
\rowcolor{gray!4}Decision support & 
\parbox{\linewidth}{\begin{enumerate}
    \item The system sends alerts and notifications for potentially dangerous situations.
    \item The system sends reminders, e.g. for medication and exercise.
    \item The system provides suggestions or recommendations for mitigation or treatment of adverse situations, e.g. medication, diet, or exercise.
    \item The system provides decision support for more than one type of user, e.g. individuals, clinicians, caregivers, etc.
    \item It is mentioned that the system incorporates established clinical practice workflows, medical guidelines, risk scores, scales, or information from medical or allied health professional bodies, the details of which are specified.
    \item Rather than solely making recommendations, the system provides advanced support such as narrowing down options, identifying possible outcomes and their likelihood, and making trade-offs between options.
    \end{enumerate}}
\\
Context awareness &
\parbox{\linewidth}{\begin{enumerate}
    \item The system includes and makes use of the concept of location, e.g. GPS coordinates, symbolic locations (``home'', ``hospital'', ``kitchen''), or geographic regions. 
    \item The system includes and makes use of the concept of time, e.g. observation timestamps, duration, etc.
    \item The system includes different user roles, such as patient, caregiver, and clinician.
    \item The system captures information related to an individual's identity, such as name and address.
    \item The system includes and makes use of the concept of activity, e.g. physical activity monitoring or activity recognition.
    \item The system incorporates ambient sensor data in addition to physiological data from body sensors.
    \item The system includes other types of contextual information, e.g. hardware and networking considerations.
\end{enumerate}}
\\
\rowcolor{gray!4}Explainability & 
\parbox{\linewidth}{\begin{enumerate}
    \item The system uses inherently interpretable techniques as part of the situation detection or situation prediction processes (for example when classifying sensor data), or else applies a post hoc explainability method.
    \item The system uses inherently interpretable techniques as part of the decision support process (for example when making recommendations), or else applies a post hoc explainability method.
    \item It is mentioned or demonstrated that the system presents explicit explanations to the user for generated situations.
    \item It is mentioned or demonstrated that the system presents explicit explanations to the user for generated decisions.
    \item The presented explanations meet two or more of the criteria for good explanations as explained by Molnar \cite{molnar_interpretable_2023}, e.g. contrastiveness, conciseness, generalisability, and a focus on abnormalities.
    \item The presented explanations are tailored to different target audiences (monitored individuals, clinicians, caregivers, regulators, etc), e.g. by providing different levels of detail or highlighting different aspects of the situation or decision.
\end{enumerate}}
\\
Uncertainty handling &
\parbox{\linewidth}{\begin{enumerate}
    \item The system is able to handle uncertainty in the situation detection or situation prediction processes.
    \item The system is able to handle uncertainty in the decision support process. 
    \item The system is able to handle missing, noisy, or otherwise invalid sensor data.
\end{enumerate}}
\\
\end{longtable}
\normalsize

\begin{figure}[ht]
\includegraphics[width=8cm]{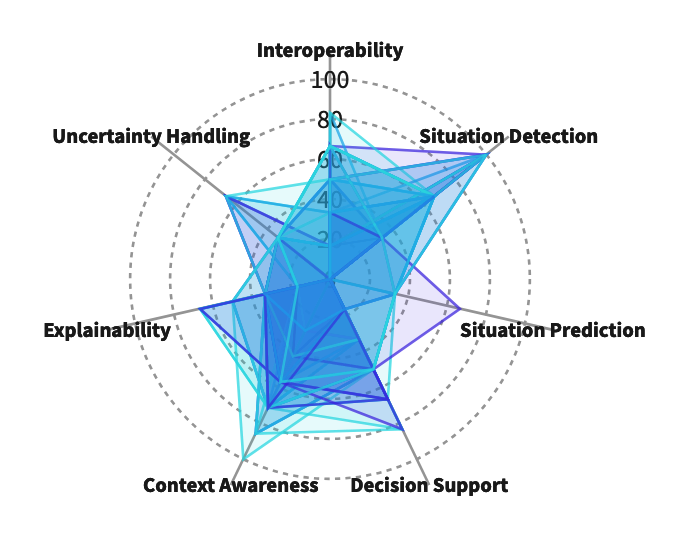}
\caption{Combined radar chart showing the extent to which all the systems address the seven challenges. \label{challenges_radar_chart}}
\end{figure}

Table~\ref{challenges-summary-counts} shows the number of systems with a particular rating for each challenge.
The combined radar chart in Figure~\ref{challenges_radar_chart} provides a visualisation of how well all the systems address the seven challenges.
Separate radar charts for the individual systems are also available\footnote{\url{https://public.flourish.studio/visualisation/17843904}}, and the individual system ratings are shown in Table~\ref{challenges-summary-systems} in the appendix.

\begin{table*}[ht]
\caption{Counts of number of systems with each rating across the seven challenges.} \label{challenges-summary-counts}
\begin{tabular}{l|ccccccc}
\hline
 \diagbox{\textbf{Rating}}{\textbf{Challenge}} & \textbf{Interoperability} & \bfseries{\makecell[c]{Situation \\detection}} & \bfseries{\makecell[c]{Situation \\prediction}} & \bfseries{\makecell[c]{Decision \\support}} & \bfseries{\makecell[c]{Context \\awareness}} & \textbf{Explainability} & \bfseries{\makecell[c]{Uncertainty \\handling}}\tabularnewline
\hline
\xmark & 3 & 0 & 36 & 3 & 0 & 0 & 26 \tabularnewline
\rowcolor{gray!4}\textbf{Low} & 24 & 15 & 11 & 22 & 3 & 40 & 17 \tabularnewline
\textbf{Medium} & 19 & 24 & 1 & 20 & 23 & 8 & 5 \tabularnewline
\rowcolor{gray!4}\textbf{High} & 2 & 9 & 0 & 3 & 22 & 0 & 0 \tabularnewline
\hline
\end{tabular}
\end{table*}

\subsection{Summary}
\label{challenges-summary}
This section has provided an in-depth analysis of seven key challenges that must be addressed in health monitoring systems, as well as a brief discussion of additional challenges that are important in such systems.  
The role played by Semantic Web technologies in overcoming the seven key challenges has been critically examined.
Additionally, non-semantic techniques that are incorporated in the systems have also been discussed.
A full list of reused semantic resources, systems that reuse them, and the challenges they address can be found in Table~\ref{reused-summary} in the appendix.
These resources include ontologies, knowledge graphs, and linked data, as well as vocabularies, taxonomies, and classifications.
To summarise the section, we discuss the results of the assessment, highlighting the challenges that are most neglected among the systems.

Based on the challenges assessment described in Section~\ref{challenges-assessment}, it is evident that more work is needed to address situation prediction, uncertainty handling, explainability, and to a lesser extent, interoperability and decision support.
Situation prediction stands out as the most neglected challenge, with 36 of the 48 systems failing to address it altogether.
This is closely followed by uncertainty handling, which 26 systems do not address at all.
Although all of the selected systems have some degree of explainability through their use of inherently interpretable Semantic Web technologies, 40 of them have a low rating for this challenge, with the remaining eight having a medium rating.
Notably, none of the selected systems achieve a high rating for situation prediction, uncertainty handling, or explainability.
Interoperability is not very well addressed among the systems, with half of them (24) scoring a low rating for this challenge, although two systems achieve a high score.
Similarly, only three systems achieve a high score for decision support. There is a nearly even split between low and medium ratings, with 22 and 20 systems respectively, while three systems do not provide decision support at all.
Context awareness is the best addressed challenge among the systems, with 22 achieving a high score, 23 achieving a medium score, and the remaining three attaining a low score.
Situation detection is also fairly well addressed, with nine high scores, 24 medium scores, and 15 low scores.

\section{System quality}
\label{system-quality}
In this section, we examine the quality of the selected systems as reported in the respective research articles.
We consider four main criteria: the data sources and devices used to collect the data; the development methodologies and tools used; the evaluation approaches and rigour; and finally, the accessibility of research outputs.
These factors all contribute to the credibility, reliability, and reproducibility of the reported systems.
Additionally, this assessment can also be used to inform benchmarking for future research and development of such systems.
We begin with a discussion of the methods of data collection, sources of data, and sensors reported in the systems.
We then review the methodologies and tools used for the development of the different components of the system, including programming languages, libraries, frameworks, and other software.
Next, we examine the evaluation approaches used to evaluate the system components and the systems as a whole. 
The last criteria we discuss is the accessibility of the resources and outputs of each system, including ontologies, data, code, and even user interfaces. 
We conclude by outlining the different aspects related to each criteria and then scoring the selected systems based on these aspects.

\subsection{Data and devices}
\label{data-devices}

The data collection methodology varies among the systems.
14 systems used existing datasets from publicly available repositories such as PhysioNet\footnote{\url{https://physionet.org}} and the University of California, Irvine (UCI) ML repository\footnote{\url{https://archive.ics.uci.edu}}.
These systems and the datasets they use are summarised in Table~\ref{existing-datasets}.

\footnotesize
\begin{longtable}{>{\raggedright}p{.22\NetTableWidth}>{\raggedright}p{.44\NetTableWidth}>{\raggedright}p{.34\NetTableWidth}}
\caption{Existing health datasets used.} \label{existing-datasets}\\
\hline
\textbf{System} & \textbf{Dataset} & \textbf{Source}\tabularnewline
\hline
\endfirsthead

\multicolumn{3}{c}{\tablename\ \thetable\ -- continued from previous page} \\
\hline
\textbf{System} & \textbf{Dataset} & \textbf{Source}\tabularnewline
\hline
\endhead

\hline
\multicolumn{3}{r}{\textbf{Table continued on next page.}}\\
\endfoot

\hline
\endlastfoot

\multirow{3}{*}{Ali et al. \cite{ali_intelligent_2021}} & 1. Pima Indians diabetes dataset & 1. UCI ML Repository \tabularnewline 
& 2. MIMIC-II & 2. PhysioNet \tabularnewline
& 3. Drug review dataset & 3. UCI ML Repository \tabularnewline 
\rowcolor{gray!4}
Ali et al. \cite{ali_smart_2020} & Heart disease dataset (Cleveland, Hungary) & UCI ML Repository\tabularnewline
De Brouwer et al. \cite{de_brouwer_mbrain_2022} & WESAD dataset & Schmidt et al. \cite{schmidt_introducing_2018}\tabularnewline
\rowcolor{gray!4}
Garcia-Valverde et al. \cite{garcia-valverde_heart_2014} & PAMAP2 Physical Activity Monitoring dataset & UCI ML Repository\tabularnewline
Hadjadj and Halimi \cite{hadjadj_integration_2021} & Vital signs of 15 Volunteers & Figshare\tabularnewline 
\rowcolor{gray!4}
Henaien et al. \cite{henaien_combined_2020} & Vital signs dataset & University of Queensland\tabularnewline
\multirow{2}{*}{Hooda and Rani \cite{hooda_semantic_2020}} &
1. Pima Indians diabetes dataset & 1. UCI ML Repository \tabularnewline
& 2. Heart disease dataset (Cleveland) & 2. UCI ML Repository \tabularnewline
\rowcolor{gray!4}
Ivașcu and Negru \cite{ivascu_activity-aware_2021} & Mobile health dataset & UCI ML Repository\tabularnewline
\multirow{2}{*}{Kordestani et al. \cite{kordestani_extended_2021}} & 1. Chronic kidney disease dataset & 1. UCI ML Repository \tabularnewline
& 2. Dermatology dataset & 2. UCI ML Repository \tabularnewline
\rowcolor{gray!4}
& 1. Heterogeneity Human Activity Recognition dataset & 1. UCI ML Repository \tabularnewline 
\rowcolor{gray!4}
\multirow{-2}{*}{Mavropoulos et al. \cite{mavropoulos_smart_2021}} & 2. CoNNL2003 dataset & 2. Sang et al. \cite{sang_introduction_2003} \tabularnewline
\multirow{2}{*}{Peral et al. \cite{peral_ontology-oriented_2018}} & 1. Diabetes dataset & 1. UCI ML Repository \tabularnewline 
& 2. Health Facts database & 2. Strack et al. \cite{strack_impact_2014} \tabularnewline
\rowcolor{gray!4}
Rhayem et al. \cite{rhayem_semantic-enabled_2021} & Various undisclosed datasets & PhysioNet\tabularnewline
Yu et al. \cite{yu_improving_2022} & Undisclosed dataset & \makecell[l]{Children's Hospital, Zhejiang University \\School of Medicine}\tabularnewline
\rowcolor{gray!4}
Zeshan et al. \cite{zeshan_iot-enabled_2023} & Human vital signs dataset & Kaggle\tabularnewline
\end{longtable}
\normalsize

10 systems \cite{ali_type-2_2018,de_brouwer_mbrain_2022,esposito_smart_2018,kilintzis_supporting_2019,garcia-moreno_systematic_2023,hristoskova_ontology-driven_2014,hussain_big-ecg_2021,stavropoulos_detection_2021,vadillo_enhancement_2013,villarreal_mobile_2014} used data collected from participants rather than existing data.
For example, Ali et al. \cite{ali_type-2_2018} collected data from 44 diabetes patients, while Esposito et al. \cite{esposito_smart_2018} collected data from 10 healthy volunteers.
Another approach was to simulate or manually generate the data.
This was done by Alti et al. \cite{alti_agent-based_2021} who simulated temperature and camera data;
Bampi et al. \cite{bampi_ontology-driven_2025} who simulated time-varying sensor data;
Chatterjee et al. \cite{chatterjee_automatic_2021}, who simulated the sensor, interview, and questionnaire data of four dummy participants; 
Martella et al. \cite{martella_semantically_2025}, who used a script to simulate sensor-generated data streams;
and Zafeiropoulos et al. \cite{zafeiropoulos_evaluating_2024}, who simulated the sensor observations and health records of three virtual patients.
Mcheik et al. \cite{mcheick_stroke_2016} similarly generated 513 data records.
Additionally, Stavropoulos et al. \cite{stavropoulos_detection_2021} simulated records in order to test the scalability of their proposed system.
18 systems \cite{akhtar_multi-agent_2022,chiang_context-aware_2015,el-sappagh_mobile_2019,elhadj_do-care_2021,fenza_hybrid_2012,ivascu_multi-agent_2015,khozouie_ontological_2018,lopes_ontology-driven_2023,mezghani_semantic_2015,minutolo_hybrid_2016,reda_heterogeneous_2022,spoladore_ontology-based_2021,titi_ontology-based_2019,vadillo_enhancement_2013,xu_design_2017,yu_semantic_2017,zhang_knowledge-based_2014,zhou_design_2022} indicated the types of data and sensors supported by the systems, but did not mention the source of the data.
It is unclear whether these systems were validated using actual sensor data, beyond a theoretical validation of the system functionality.
Only 19 of the systems gave specific details of the devices used for data collection.
Table~\ref{devices-used} indicates these systems and the types and descriptions of devices mentioned.
While most of these are commercially available devices, the systems proposed by Kim et al. \cite{kim_ontology_2014} and Lopes de Souza et al. \cite{kim_ontology_2014} used custom-made prototypes.

\scriptsize
\begin{longtable}{>{\raggedright}p{.18\NetTableWidth}>{\raggedright}p{.09\NetTableWidth}>{\raggedright}p{.43\NetTableWidth}>{\raggedright}p{.3\NetTableWidth}}
\caption{Devices mentioned for data collection in the selected systems.} \label{devices-used}\\
\hline
\textbf{System} & \textbf{Device type} & \textbf{Name/description} & \textbf{Data type}\tabularnewline
\hline
\endfirsthead

\multicolumn{4}{c}{\tablename\ \thetable\ -- continued from previous page} \\
\hline
\textbf{System} & \textbf{Device type} & \textbf{Name/description} & \textbf{Data type}\tabularnewline
\hline
\endhead

\hline
\multicolumn{3}{>{\raggedright\arraybackslash}p{0.75\NetTableWidth}}{BG - blood glucose; BP - blood pressure; BT - body temperature; BVP - blood volume pulse; CO\textsubscript{2} - carbon dioxide; ECG - electrocardiogram; GSR -  galvanic skin response; HR - heart rate; SpO\textsubscript{2} - blood oxygen saturation; ST - skin temperature; TVOC - total volatile organic compounds} & \multicolumn{1}{>{\raggedleft\arraybackslash}p{0.25\NetTableWidth}}{\textbf{Table continued on next page.}}\\
\endfoot

\hline
\multicolumn{4}{>{\raggedright\arraybackslash}p{\NetTableWidth}}{BG - blood glucose; BP - blood pressure; BT - body temperature; BVP - blood volume pulse; CO\textsubscript{2} - carbon dioxide; ECG - electrocardiogram; GSR -  galvanic skin response; HR - heart rate; SpO\textsubscript{2} - blood oxygen saturation; ST - skin temperature; TVOC - total volatile organic compounds}
\endlastfoot

Bampi et al. \cite{bampi_ontology-driven_2025} & Commercial & Emfit QS device & HR \tabularnewline
\rowcolor{gray!4}
Chiang and Liang \cite{chiang_context-aware_2015} & Commercial & Kinect & Motion \tabularnewline
De Brouwer et al. \cite{de_brouwer_mbrain_2022} & Commercial & Empatica E4 wristband & Acceleration, HR, BVP, GSR, ST \tabularnewline
\rowcolor{gray!4}
& & 1. Amiigo wristband & 1. BT, HR, SpO\textsubscript{2} \tabularnewline
\rowcolor{gray!4}
\multirow{-2}{*}{Esposito et al. \cite{esposito_smart_2018}} & \multirow{-2}{*}{Commercial} & 2. Omron HJ-112 digital pocket pedometer & 2. Acceleration \tabularnewline
\multirow{3}{*}{Garcia-Moreno et al. \cite{garcia-moreno_systematic_2023}} & \multirow{3}{*}{Commercial} & 1. Samsung Gear S3 smartwatch & 1. HR \tabularnewline
& & 2. Empatica E4 wristband & 2. Accelerometer, electrodermal activity, HR, BT \tabularnewline
\rowcolor{gray!4}
& & 1. A\&D UA-767PBT blood pressure monitor & 1. BP, HR \tabularnewline
\rowcolor{gray!4}
& & 2. Nonin Avant 4000 digital pulse oximeter & 2. SpO\textsubscript{2} \tabularnewline
\rowcolor{gray!4}
& & 3. A\&D UC-321PBT weight scale & 3. Body weight \tabularnewline
\rowcolor{gray!4}
\multirow{-4}{*}{Hristoskova et al. \cite{hristoskova_ontology-driven_2014}} & \multirow{-4}{*}{Commercial} & 4. Welch Allyn Cardio Perfect 12 Lead ECG & 4. ECG \tabularnewline
\multirow{2}{*}{Hussain and Park \cite{hussain_big-ecg_2021}} & \multirow{2}{*}{Commercial} & 1. BioNomadix respiration (RSP) with ECG amplifier & \multirow{2}{*}{ECG} \tabularnewline
& & 2. ECG patch from Life Science Technology Inc. &  \tabularnewline
\rowcolor{gray!4}
Ivașcu and Negru \cite{ivascu_activity-aware_2021} & Commercial & Shimmer2 & Acceleration, ECG \tabularnewline
\multirow{4}{*}{Kim et al. \cite{kim_ontology_2014}} & \multirow{4}{*}{Prototype} & Smart wear consisting of upper body clothing made of flexible and stretchy material to which various sensors can be attached. To minimise the use of wires, a coin cell battery and wireless communication is used. & \multirow{4}{*}{BP, other unnamed vital signs} \tabularnewline
\rowcolor{gray!4}
\multirow{5}{*}{Lopes de Souza et al. \cite{lopes_ontology-driven_2023}} & \multirow{5}{*}{Prototype} & Various sensors (i.e. MPU-6050 gyroscope and accelerometer; MKB0805 HR and BP sensor; DS18B20 digital thermometer), are assembled on a T7 V1.3 MINI 32 ESP32 board which is then affixed using 3D printed plastic frame onto a bracelet. & \multirow{5}{*}{Acceleration, orientation, HR, BP, BT} \tabularnewline
\multirow{5}{*}{Martella et al. \cite{martella_semantically_2025}} & \multirow{5}{*}{Commercial} & \multirow{2}{*}{1. Fitbit wristband} & 1. HR, BT \tabularnewline
& & \multirow{3}{*}{2. Aircare device} & 2. TVOC, CO\textsubscript{2}, particulate matter, sound pressure level, temperature, relative humidity, ambient light, atmospheric pressure \tabularnewline
\rowcolor{gray!4}
& & 1. Samsung Galaxy S3 Mini smartphone & \tabularnewline
\rowcolor{gray!4}
\multirow{-2}{*}{Mavropoulos et al. \cite{mavropoulos_smart_2021}} & \multirow{-2}{*}{Commercial} & 2. LG Nexus 4 smartphone & \multirow{-2}{*}{Video} \tabularnewline
\multirow{2}{*}{Reda et al. \cite{reda_heterogeneous_2022}} & \multirow{2}{*}{Commercial} & 1. Fitbit wristband & \tabularnewline
& & 2. Jawbone wristband & \multirow{-2}{*}{Step count, HR, calories burned} \tabularnewline
\rowcolor{gray!4}
 & & 1. Polar H1 chest strap & \tabularnewline
\rowcolor{gray!4}
\multirow{-2}{*}{Spoladore et al. \cite{spoladore_ontology-based_2021}} & \multirow{-2}{*}{Commercial} & 2. COSMED E100 cycle ergometer & \multirow{-2}{*}{ECG} \tabularnewline
Stavropoulos et al. \cite{stavropoulos_detection_2021} & Commercial & Fitbit Charge 3 wristband & Step count, sleep stages, HR\tabularnewline
\rowcolor{gray!4}
& & 1. Arduino e-Health sensor platform & 1. BG, HR, BT, BP, SpO\textsubscript{2} \tabularnewline
\rowcolor{gray!4}
\multirow{-2}{*}{Vadillo et al. \cite{vadillo_enhancement_2013}} & \multirow{-2}{*}{Commercial} & 2. Tunstall Lifeline Connect+ home unit & 2. Motion \tabularnewline
Villarreal et al. \cite{villarreal_mobile_2014} & Commercial & BodyTel Glucotel & BG \tabularnewline
\rowcolor{gray!4}
& & 1. Fitbit wristband & 1. Step count \tabularnewline
\rowcolor{gray!4}
\multirow{-2}{*}{Yu et al. \cite{yu_semantic_2017}} & \multirow{-2}{*}{Commercial} & 2. Withings scale & 2. Body weight \tabularnewline
Zhang et al. \cite{zhang_knowledge-based_2014} & Commercial & Equivital multi-parameter sensor & BP, BT, HR, SpO\textsubscript{2} \tabularnewline
\end{longtable}
\normalsize

\subsection{System and components development}

\subsubsection{Development methodologies}
The use of a development methodology can streamline the process of developing Semantic Web technologies.
In particular, the literature on ontology development methodologies is quite rich, with a large number of established methodologies proposed \cite{fernandez-lopez_overview_2002,iqbal_analysis_2013}.
There have also been several proposed approaches towards developing knowledge graphs \cite{cimiano_knowledge_2017} and ensuring the quality of linked data \cite{kontokostas_test-driven_2014,debattista_luzzumethodology_2016}.
Despite this, only five systems specified an existing methodology for Semantic Web technologies, with all five methodologies being ontology-focused.
Hadjadj and Halimi \cite{hadjadj_integration_2021} used the NeOn framework \cite{carmen_suarez-figueroa_neon_2015}, a scenario-based methodology for building ontologies, while
Zafeiropoulos et al. \cite{zafeiropoulos_evaluating_2024} used the Human-Centered Ontology engineering MEthodology (HCOME) \cite{kotis_human-centered_2006}.
Titi et al. \cite{titi_ontology-based_2019} used an existing case-based ontology engineering methodology \cite{el-sappagh_ontological_2014}.
Similarly, Kilintzis et al. \cite{kilintzis_supporting_2019} followed four of the ontology construction steps recommended by Spear \cite{spear_ontology_2006}.
Although not a development methodology, Peral et al. \cite{peral_ontology-oriented_2018} used the SemanTic Refinement of Ontology MAppings (STROMA) \cite{arnold_enriching_2014} approach for aligning corresponding concepts between different ontologies.  

Additionally, two systems reported the use of existing methodologies for overall system development. 
Garcia-Moreno et al. \cite{garcia-moreno_systematic_2023} used the design science research methodology \cite{hevner_design_2010}, while
Vadillo et al. \cite{vadillo_enhancement_2013} used CommonKADS (Knowledge Acquisition and Development Systems) \cite{kingston_designing_1998}, a knowledge-based system design methodology, and its extension for multi-agent systems, MAS-CommonKADS \cite{iglesias_analysis_1998}.
Overall, only 16 of the systems reported or described the use of a systematic methodology, whether existing or novel, in the development of either the Semantic Web technologies, other system components, or the system as a whole.

\subsubsection{Development tools}
\label{development-tools}
Various languages, frameworks, and libraries were used to develop the systems.
When it comes to rule languages,
SWRL is the most commonly used, with 24 systems either explicitly mentioning it or demonstrating SWRL syntax in code snippets \cite{akhtar_multi-agent_2022,ali_smart_2020,ali_type-2_2018,alti_agent-based_2021,bampi_ontology-driven_2025,chatterjee_automatic_2021,el-sappagh_mobile_2019,elhadj_do-care_2021,esposito_smart_2018,hadjadj_integration_2021,henaien_combined_2020,hooda_semantic_2020,hristoskova_ontology-driven_2014,lopes_ontology-driven_2023,mezghani_semantic_2015,minutolo_hybrid_2016,reda_heterogeneous_2022,rhayem_semantic-enabled_2021,spoladore_ontology-based_2021,titi_ontology-based_2019,yu_semantic_2017,zafeiropoulos_evaluating_2024,zeshan_iot-enabled_2023,zhang_knowledge-based_2014}. 
Apache Jena includes a general purpose rule-based reasoner which is used by three systems \cite{chiang_context-aware_2015,garcia-valverde_heart_2014,kim_ontology_2014}.
Other Semantic Web-based rules languages include SHACL, used by Stavropoulos et al. \cite{stavropoulos_detection_2021} and SPIN, used by Kilintzis et al. \cite{kilintzis_supporting_2019}.
Although Mavropoulos et al. \cite{mavropoulos_smart_2021} mention using OWL to create rules, this approach is inherently limited for situation analysis and decision support because OWL lacks the expressivity for if-then rules.
The authors indicate that they will implement more complex reasoning rules in future work.

Programming languages can also be used to configure rules, as is done by Khozouie et al. \cite{khozouie_ontological_2018} using Java.
Similarly, Fenza et al. \cite{fenza_hybrid_2012} use MATLAB and Fuzzy Control Language to define their fuzzy rules, while Kordestani et al. \cite{kordestani_extended_2021} use Drools\footnote{\url{https://www.drools.org}}, a business rule management system.
Therefore, among the 41 systems that implement rules, 25 systems use Semantic Web-based rule languages, while three use non-Semantic Web languages.
Eight systems do not mention or indicate a specific formal language in which rules are defined \cite{ali_intelligent_2021,ammar_using_2021,hussain_big-ecg_2021,ivascu_activity-aware_2021,ivascu_multi-agent_2015,martella_semantically_2025,peral_ontology-oriented_2018,yu_improving_2022}.
For queries, 26 of the systems use SPARQL, with five also using Apache Jena Fuseki, a SPARQL server, to publish their SPARQL endpoints.

Among the systems that incorporate ontologies, Protégé\footnote{\url{https://protege.stanford.edu}} is most commonly cited as the ontology development platform of choice, used in 27 of the systems \cite{akhtar_multi-agent_2022,ali_intelligent_2021,ali_smart_2020,ali_type-2_2018,alti_agent-based_2021,bampi_ontology-driven_2025,chatterjee_automatic_2021,chiang_context-aware_2015,de_brouwer_mbrain_2022,el-sappagh_mobile_2019,elhadj_do-care_2021,esposito_smart_2018,hadjadj_integration_2021,henaien_combined_2020,hooda_semantic_2020,hussain_big-ecg_2021,ivascu_activity-aware_2021,ivascu_multi-agent_2015,khozouie_ontological_2018,kim_ontology_2014,martella_semantically_2025,spoladore_ontology-based_2021,titi_ontology-based_2019,vadillo_enhancement_2013,zafeiropoulos_evaluating_2024,zeshan_iot-enabled_2023,zhang_knowledge-based_2014}. Protégé is an ontology editor that supports the latest OWL and RDF specifications.
Another commonly used platform is Apache Jena\footnote{\url{https://jena.apache.org}}, a Java framework for building Semantic Web and Linked Data applications, used in 18 of the systems \cite{ali_type-2_2018,alti_agent-based_2021,bampi_ontology-driven_2025,chatterjee_automatic_2021,chiang_context-aware_2015,de_brouwer_mbrain_2022,elhadj_do-care_2021,garcia-valverde_heart_2014,hadjadj_integration_2021,hooda_semantic_2020,ivascu_activity-aware_2021,ivascu_multi-agent_2015,kim_ontology_2014,rhayem_semantic-enabled_2021,titi_ontology-based_2019,vadillo_enhancement_2013,yu_semantic_2017,zeshan_iot-enabled_2023}.
27 systems report using OWL \cite{ali_intelligent_2021,ali_smart_2020,ali_type-2_2018,ammar_using_2021,chatterjee_automatic_2021,el-sappagh_mobile_2019,elhadj_do-care_2021,esposito_smart_2018,fenza_hybrid_2012,hadjadj_integration_2021,hooda_semantic_2020,hristoskova_ontology-driven_2014,khozouie_ontological_2018,kilintzis_supporting_2019,kim_ontology_2014,mavropoulos_smart_2021,minutolo_hybrid_2016,reda_heterogeneous_2022,rhayem_semantic-enabled_2021,spoladore_ontology-based_2021,stavropoulos_detection_2021,titi_ontology-based_2019,vadillo_enhancement_2013,yu_semantic_2017,zafeiropoulos_evaluating_2024,zeshan_iot-enabled_2023,zhang_knowledge-based_2014}, while 16 report using RDF \cite{ammar_using_2021,bampi_ontology-driven_2025,chatterjee_automatic_2021,chiang_context-aware_2015,de_brouwer_mbrain_2022,hadjadj_integration_2021,hooda_semantic_2020,hussain_big-ecg_2021,kilintzis_supporting_2019,lopes_ontology-driven_2023,mezghani_semantic_2015,reda_heterogeneous_2022,titi_ontology-based_2019,yu_semantic_2017,zafeiropoulos_evaluating_2024,zhang_knowledge-based_2014}.
Ammar et al. \cite{ammar_using_2021} use Solid (Social Linked Data)\footnote{\url{https://solidproject.org}}, a platform for developing decentralised linked data applications with the goal of enhanced privacy and data ownership.
They also mention the JavaScript-based LDflex\footnote{\url{https://github.com/LDflex/LDflex}} and the Python-based RDFlib\footnote{\url{https://github.com/RDFLib/rdflib}} for linked data manipulation and querying, the latter of which is also used by Zafeiropoulos et al. \cite{zafeiropoulos_evaluating_2024}.
These platforms and libraries are all free and open source.

For storage, Mavropoulos \cite{mavropoulos_smart_2021} and Stavropoulos et al. \cite{stavropoulos_detection_2021} use GraphDB, while Spoladore et al. \cite{spoladore_ontology-based_2021} use Stardog\footnote{\url{https://www.stardog.com}}.
Both of these are enterprise semantic databases.
Non-semantic database management systems are also used by 11 of the systems, with SQL-based systems being more popular than NoSQL alternatives. 

Four systems use MySQL\footnote{\url{https://www.mysql.com}} \cite{alti_agent-based_2021,chiang_context-aware_2015,titi_ontology-based_2019,villarreal_mobile_2014}, three use SQLite\footnote{\url{https://sqlite.org}} \cite{el-sappagh_mobile_2019,esposito_smart_2018,lopes_ontology-driven_2023}, and one uses PostgreSQL\footnote{\url{https://www.postgresql.org/}} \cite{martella_semantically_2025}, while
three systems use the NoSQL database system MongoDB\footnote{\url{https://www.mongodb.com}} \cite{de_brouwer_mbrain_2022,elhadj_do-care_2021,martella_semantically_2025}.
A summary of all the development tools used in the selected systems is included in Table~\ref{development-evaluation-summary} in the appendix.

\subsection{Rigour of evaluation}
\label{evaluation}
A variety of evaluation approaches are used by the selected systems.
The most common approach is case-based evaluation.
21 systems were evaluated through use case scenarios, which generally describe the sequence of events when a user interacts with the system \cite{ammar_using_2021,bampi_ontology-driven_2025,chiang_context-aware_2015,de_brouwer_mbrain_2022,el-sappagh_mobile_2019,elhadj_do-care_2021,garcia-valverde_heart_2014,hadjadj_integration_2021,ivascu_multi-agent_2015,khozouie_ontological_2018,kordestani_extended_2021,mcheick_stroke_2016,mezghani_semantic_2015,spoladore_ontology-based_2021,stavropoulos_detection_2021,vadillo_enhancement_2013,yu_improving_2022,zafeiropoulos_evaluating_2024,zeshan_iot-enabled_2023,zhang_knowledge-based_2014,zhou_design_2022}.
Nine systems were evaluated using case studies, which are similar to use case scenarios but are more extensive and detailed \cite{akhtar_multi-agent_2022,alti_agent-based_2021,esposito_smart_2018,fenza_hybrid_2012,henaien_combined_2020,martella_semantically_2025,minutolo_hybrid_2016,peral_ontology-oriented_2018,villarreal_mobile_2014,xu_design_2017}.
12 systems were evaluated by running user studies with real users \cite{ali_type-2_2018,de_brouwer_mbrain_2022,esposito_smart_2018,garcia-moreno_systematic_2023,hristoskova_ontology-driven_2014,hussain_big-ecg_2021,kilintzis_supporting_2019,kim_ontology_2014,mavropoulos_smart_2021,stavropoulos_detection_2021,villarreal_mobile_2014,yu_improving_2022}.
Among these systems, three used Likert scales to measure user feedback \cite{kim_ontology_2014,mavropoulos_smart_2021,stavropoulos_detection_2021}.
Additionally, 16 compared their systems with existing ones, showing how their results performed against the state of the art using a set criteria \cite{ali_smart_2020,ali_type-2_2018,el-sappagh_mobile_2019,hussain_big-ecg_2021,ivascu_activity-aware_2021,mavropoulos_smart_2021,rhayem_semantic-enabled_2021,xu_design_2017}.

21 systems were evaluated based on non-functional requirements.
For example, Esposito et al. \cite{esposito_smart_2018} and Vadillo et al. \cite{vadillo_enhancement_2013} used the Architecture-Level Modifiability Analysis (ALMA) method to evaluate the potential costs associated with modifying their systems, such as by adding more sensors.
Similarly, Alti et al. \cite{alti_agent-based_2021} evaluated their system based on execution time, optimality, application’s lifetime and number of discovered services.
Domain-specific quality metrics were also considered.
Yu et al. \cite{yu_improving_2022} evaluated their system using the Chronic Care Model (CCM), an established framework for chronic care management that includes criteria such as system design, self-management support, and decision support.

Additionally, five systems used simulation as a means to investigate the system functionality.
For example, Akhtar et al. \cite{akhtar_multi-agent_2022} used Netlogo, a multi-agent modelling platform, to simulate the use of their system.
Chiang and Liang \cite{chiang_context-aware_2015} used a fuzzy logic simulation tool to validate their fuzzy inference module.
Ivașcu and Negru \cite{ivascu_activity-aware_2021} simulated the system functionality by using each subject in the dataset as the target user, while 
Martella et al. \cite{martella_semantically_2025} used a testbed environment to simulate user scenarios and test how well the system responds to varying workloads. Finally,
Reda et al. \cite{reda_heterogeneous_2022} used a web portal with sample data for testing purposes.
Expert validation was also used to evaluate the systems, with the aim of ensuring maximum similarity between the system output and expert opinion.
This approach was taken by Ali et al. \cite{ali_smart_2020}, El-Sappagh et al. \cite{el-sappagh_mobile_2019}, Hadjadj and Halimi \cite{hadjadj_integration_2021}, Hristoskova et al. \cite{hristoskova_ontology-driven_2014} and Khozouie et al. \cite{khozouie_ontological_2018}.
Additionally, 10 systems used query-based validation, where the system is validated by checking the answers to SPARQL queries \cite{ali_type-2_2018,chatterjee_automatic_2021,de_brouwer_mbrain_2022,el-sappagh_mobile_2019,hadjadj_integration_2021,kilintzis_supporting_2019,kim_ontology_2014,titi_ontology-based_2019,yu_semantic_2017,zafeiropoulos_evaluating_2024}.

In addition to the overall system, the system components were also evaluated.
Inconsistencies in ontologies can be detected using ontology reasoners, which check whether there are contradictions in class hierarchies or class instances \cite{ye_situation_2011}.
Reasoners were used in 11 of the systems to evaluate the structural consistency of ontologies.
Eight systems used Pellet \cite{ali_type-2_2018,elhadj_do-care_2021,esposito_smart_2018,hooda_semantic_2020,hristoskova_ontology-driven_2014,khozouie_ontological_2018,titi_ontology-based_2019,vadillo_enhancement_2013,zafeiropoulos_evaluating_2024}, one system used HermiT \cite{chatterjee_automatic_2021}, and one system reported using both Pellet and HermiT \cite{el-sappagh_mobile_2019}.
Additionally, three systems used ontology evaluation frameworks, namely OntOlogy Pitfall Scanner! (OOPS!) \cite{poveda-villalon_oops_2014}, which was used by El-Sappagh et al. \cite{el-sappagh_mobile_2019} and Zafeiropoulos et al. \cite{zafeiropoulos_evaluating_2024}, and OQuaRE \cite{duque-ramos_oquare_2011}, which was used by Rhayem et al. \cite{rhayem_semantic-enabled_2021}.
Two systems also evaluated the effect of different components within the same system through ablation studies.
Ali et al. \cite{ali_intelligent_2021} tested the performance of their BiLSTM model for classifying healthcare data while using an ontology and without using an ontology. 
The results showed an increase in the accuracy of the model when combined with an ontology.
Similarly, Ali et al. \cite{ali_smart_2020} compared the performance of their proposed ensemble deep learning model with and without feature selection.
Finally, 13 systems, mainly those that implement ML, used well-known metrics such as accuracy, precision, recall, F-score, and mean square error \cite{ali_intelligent_2021,ali_smart_2020,ali_type-2_2018,alti_agent-based_2021,garcia-moreno_systematic_2023,garcia-valverde_heart_2014,hussain_big-ecg_2021,ivascu_activity-aware_2021,kim_ontology_2014,mavropoulos_smart_2021,rhayem_semantic-enabled_2021,yu_improving_2022,zafeiropoulos_evaluating_2024,zeshan_iot-enabled_2023}.

\subsection{Accessibility of research outputs}
The sharing of research outputs, such as code, ontologies, knowledge graphs, and data, is a critical aspect of ensuring research is reproducible and verifiable.
These resources can also be built upon by other researchers, contributing to their reuse for more efficient system development.
This is severely neglected among the selected systems, with only six articles including links to their research outputs.
Among them are Chatterjee et al. \cite{chatterjee_automatic_2021}, who include their OWL ontology, simulated data, propositional variables, rule base, and queries as multimedia appendices.
Using platforms like GitHub rather than static files has the advantage of version control, allowing researchers to manage future updates and revisions.
Three systems take this approach.
The system proposed by Bampi et al. \cite{bampi_ontology-driven_2025} is available on GitHub, including the backend, frontend, ontology, rules, and queries.
Similarly, Zafeiropoulos et al. \cite{zafeiropoulos_evaluating_2024} make their proposed ontology, queries, rules, code, and even research papers available via a GitHub repository.
De Brouwer et al. \cite{de_brouwer_mbrain_2022} include a link to a GitHub repository associated with the Data Analytics for Health and Connected Care ontology, which their proposed ontology extends and was developed by their research group.
However, the new mBrain ontology reported in the article is not made available.
In contrast, the ontology proposed by El-Sappagh et al. \cite{el-sappagh_mobile_2019} has been published on Bioportal\footnote{\url{https://bioportal.bioontology.org}}, a popular repository of biomedical ontologies.
However, no other research outputs, such as rules and queries, are made available.

An important consideration when sharing system resources is ensuring their long-term accessibility.
For instance, although Kilintzis et al. \cite{kilintzis_supporting_2019} and Reda et al. \cite{reda_heterogeneous_2022} provide links to their systems, the web pages were unavailable at the time of writing this article.
Kilintizis et al. \cite{kilintzis_supporting_2019} stated that the link made their ontology available either for reuse or review purposes, while the link from Reda et al. \cite{reda_heterogeneous_2022} was to a portal with a video tutorial and sample datasets for testing purposes. 
As these resources are now inaccessible, we are unable to ascertain if they were ever operational and for how long they may have been active.
Additionally, while GitHub repositories allow for easy accessibility and future updates, they can also be deleted or made private.
A good alternative is Zenodo\footnote{\url{https://zenodo.org}}, a general-purpose research repository designed for long-term preservation by ensuring that published records can be updated but not unpublished. 
Further, Zenodo provides persistent  digital object identifiers (DOIs) for each upload, ensuring easy referencing.

Alti et al. \cite{alti_agent-based_2021} and Garcia-Moreno et al. \cite{garcia-moreno_systematic_2023} include a note that the data associated with their studies are available upon request.
However, this is a suboptimal approach as it is impossible to guarantee the authors' willingness or ability to consistently respond to such requests over time, potentially leading to prolonged delays or even a complete lack of response.
Publishers can mitigate this through well-defined data availability policies.
Finally, we note that researchers may be restricted from sharing participant data due to privacy concerns. 
A potential solution would be to seek participants' consent in sharing their data anonymised and non-identifiable form.
Table~\ref{system-outputs} indicates the links shared by researchers.

\begin{table*}[ht]
\caption{Links to system outputs as shared by researchers.} \label{system-outputs}
\begin{tabular}{lll}
\hline
\textbf{System} & \textbf{Type} & \textbf{Link}\tabularnewline
\hline
Bampi et al. \cite{bampi_ontology-driven_2025} & GitHub repository & \url{https://github.com/mbampi/aal-system} \tabularnewline
\rowcolor{gray!4}Chatterjee et al. \cite{chatterjee_automatic_2021} & Static files & \url{https://www.jmir.org/2021/4/e24656\#app1} \tabularnewline
De Brouwer et al. \cite{de_brouwer_mbrain_2022} & GitHub repository & \url{https://github.com/predict-idlab/DAHCC-Sources}\tabularnewline 
\rowcolor{gray!4}El-Sappagh et al. \cite{el-sappagh_mobile_2019} & Bioportal & \url{https://bioportal.bioontology.org/ontologies/FASTO} \tabularnewline 
Kilintzis et al. \cite{kilintzis_supporting_2019} & Web page - currently inaccessible & \url{http://lomi.med.auth.gr/ontologies/WELCOME_entities} \tabularnewline
\rowcolor{gray!4}Reda et al. \cite{reda_heterogeneous_2022} & Web page - currently inaccessible & \url{http://137.204.74.19:8080/IFOPlatform/welcomePage.jsp}\tabularnewline 
Zafeiropoulos et al. \cite{zafeiropoulos_evaluating_2024} & GitHub repository & \url{https://github.com/KotisK/Wear4PDmove}\tabularnewline 
\hline
\end{tabular}
\end{table*}

\subsection{Quality assessment}
Mirroring our assessment of how well the systems tackle the key challenges in Section~\ref{challenges-assessment}, we have also evaluated the quality of the systems as reported in the corresponding research articles. 
We base our evaluation on the aspects of the quality criteria which are summarised in Table~\ref{quality-critera}, and use the same four-point rating scale (\xmark, Low, Medium, and High) determined by the percentage of aspects that each system has met. 
The quality ratings for each system are shown in Table~\ref{quality_summary} in the appendix.

\footnotesize
\begin{longtable}{>{\raggedright}p{.18\linewidth}p{.75\linewidth}}
\caption{Important aspects related to the quality evaluation criteria.} \label{quality-critera} \\
\hline
\textbf{Criteria} & \textbf{Aspects and scoring guide}\\
\hline
\endfirsthead

\multicolumn{2}{c}{\tablename\ \thetable\ -- continued from previous page} \\
\hline
\textbf{Criteria} & \textbf{Aspects and scoring guide}\\
\hline
\endhead

\hline
\multicolumn{2}{r}{\textbf{Table continued on next page.}}\\
\endfoot

\hline
\endlastfoot

Data and devices &
\parbox{\linewidth}{\begin{enumerate}
     \item Details are given regarding the source of sensor data and other health-related data used in developing or evaluating the systems, e.g.: whether existing or simulated datasets were used or new data collected, the nature of the data, the availability of the data, the number and description of the people the data was collected from. 
     \\0: no details on data sources are given; 1: only partial details are given; 2: comprehensive details are given.
     \item Details of the specific sensor device(s) used to collect data are given. This means that at least one device is explicitly named if it is an existing or commercially available device, or described if it is a novel prototype. A description based on the type of data measured (e.g. temperature sensor or blood pressure monitor) is insufficient.
     \\0: no specific device is mentioned; 1: at least one specific device is mentioned.
\end{enumerate}}
\\
\rowcolor{gray!4}
System and components development &
\parbox{\linewidth}{\begin{enumerate}
    \item A methodology has been followed for the development of Semantic Web technologies, other system components, and/or the system as a whole. The methodology can be existing or novel, but it must outline the steps followed in a systematic manner.
    \\0: no methodology is mentioned; 1: a methodology is mentioned for at least one system component.
    \item Details of the languages, platforms, tools, and other software used for the development of the system and its components are given.
    \\0: no languages, platforms, tools, or other software is mentioned; 1: only partial details are given, e.g. for some system components but not others; 2: comprehensive details are given.
\end{enumerate}}
\\
Rigour of evaluation &
\parbox{\linewidth}{\begin{enumerate}
    \item The individual system components (Semantic Web technologies or other techniques) are evaluated using appropriate methods e.g. the use of reasoners, competency questions, evaluation frameworks, and metrics like precision, recall or F1 score. 
    \\0: no evaluation of components is mentioned. 1: evaluation of one component is done and results are given. 2: more than one component is evaluated, and/or an established evaluation framework is used. 3: in addition to the evaluation of the individual components, the impact of the different components and the way the work together is evaluated, e.g. through ablation studies.
    \item The system is compared with other approaches or systems (i.e. the state of the art) in a systematic way using formally defined criteria, and/or evaluation of the system is done by domain experts.
    \\0: not done; 1: done
    \item The potential real-world functionality of the system is evaluated, either through a use case scenario, case study, simulation, or deployment.
    \\0: potential real world functionality is not considered; 1: a use case scenario/case study/simulation is used; 2: the system is deployed and/or user studies are carried out.
    \item Non-functional requirements have been considered and details of this have been provided (e.g. scalability, adaptability, usability, security). 
    \\0: no NFRs have been considered or insufficient details are provided; 1: at least one NFR has been considered and explained.
\end{enumerate}}
\\
\rowcolor{gray!4}
Accessibility of system outputs &
    \parbox{\linewidth}{\vspace{0.5em}Research outputs, including but not limited to code, ontologies, knowledge graphs, rules, queries, and data, have been made publicly and readily available without the need to contact the authors.
    \newline 0: no resources are available; 1: at least one resource is readily available; 2: more than one resource is readily available. \vspace{0.5em}}
\\
\end{longtable}
\normalsize

Table~\ref{quality-summary-counts} shows the number of systems with a particular rating for each criteria, while
the combined radar chart in Figure~\ref{challenges_radar_chart} provides a visualisation of the overall quality of the systems.
Separate radar charts for the individual systems are also available\footnote{\url{https://public.flourish.studio/visualisation/17845323}}.

\begin{table*}[ht]
\caption{Counts of number of systems with each rating across the four main quality criteria.} \label{quality-summary-counts}
\begin{tabular}{l|cccc}
\hline
 \diagbox{\textbf{Rating}}{\textbf{Criteria}} & \bfseries{\makecell[c]{Data \& \\devices}} & \bfseries{\makecell[c]{System \& components \\development}} & \bfseries{\makecell[c]{Rigour of \\evaluation}} & \bfseries{\makecell[c]{Accessibility of \\research outputs}} \tabularnewline
\hline
\xmark & 11 & 1 & 2 & 42 \tabularnewline
\rowcolor{gray!4}\textbf{Low} & 14 & 10 & 20 & N/A \tabularnewline
\textbf{Medium} & 13 & 25 & 20 & 3 \tabularnewline
\rowcolor{gray!4}\textbf{High} & 10 & 12 & 6 & 3 \tabularnewline
\hline
\end{tabular}
\end{table*}

\begin{figure}[ht]
\includegraphics[width=7cm]{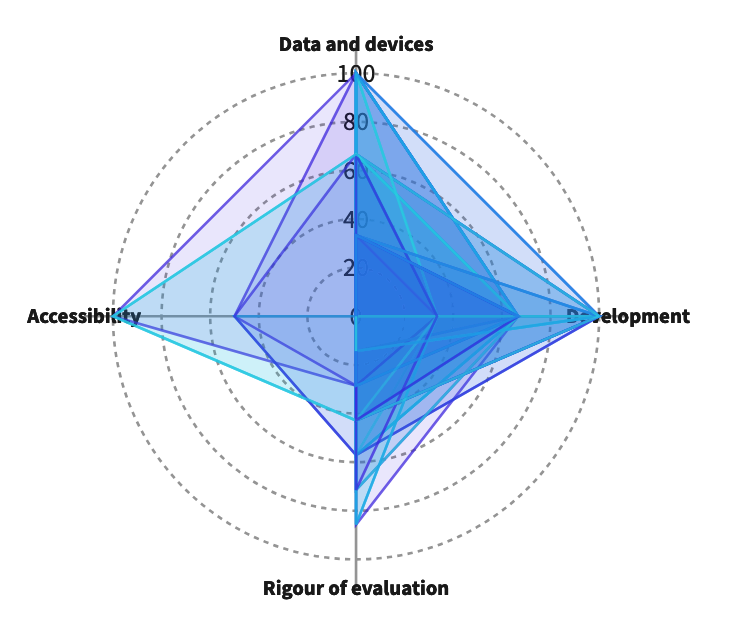}
\caption{Combined radar chart showing the quality of the systems based on the specified criteria. \label{quality_radar_chart}}
\end{figure}

\subsection{Summary}
This section has critically examined the quality of the selected systems, with a focus on four criteria: the data sources and devices used to collect the data; the development methodologies
and tools used; the evaluation approaches and rigour; and the accessibility of research outputs. 
A summary list of the development tools development tools, evaluation approaches, and evaluation metrics used by the systems, can be found in Table~\ref{development-evaluation-summary} in the appendix.
Though this analysis extends beyond Semantic Web technologies, we also consider several factors that are specific to Semantic Web technologies such as methodologies, languages, frameworks, and semantic databases.
To summarise the section, we discuss the results of the quality assessment, highlighting the aspects that are poorly addressed among the systems.

The accessibility of research outputs is by far the most overlooked quality criterion, with only six systems making their resources publicly accessible. 
With regard to the rigour of evaluation, 20 systems achieved a low rating, with another 20 achieving a medium rating. 
This can be primarily attributed to the
fact that numerous researchers only reported the evaluation of one system component, failing to evaluate or account for the impact of other components.
The potential real-world functionality of the system was another poorly addressed aspect of evaluation.
Most systems were not tested with actual users, with only 12 systems being user validated. Case studies and scenarios were the most common approach, used in 29 of the systems.
Additionally, most researchers (32 systems) overlooked the importance of evaluating their systems against external benchmarks, such as drawing comparisons with similar existing systems or seeking evaluations from domain experts.

The description of data collection methods or existing datasets was generally well done among the systems, with 30 of them giving adequate details regarding the number and demographics of participants or dataset records and properly citing reused datasets.
However, only 19 of the systems gave details of the specific devices used to collect the sensor data. 
A likely explanation for this is that many of the proposed systems are conceptual proposals rather than functional implementations, and therefore they were not tested on real sensor data collected from actual devices.
Finally, with regard to system development, 36 papers comprehensively reported on the tools used to develop the different components of their proposed systems.
However, only 16 of the papers reported the use of an existing development methodology, or else adequately described the systematic steps taken to develop each system component.

\section{System architectures}
\label{architectures}
The architecture of a system can be defined as an abstraction of the system in the form of a set of software structures needed to reason about it \cite{bass_software_2021}.
An important concept when discussing system architectures is the architectural style, which defines constraints on the form and structure of an architecture \cite{garlan_introduction_1995}.
This is closely related to the architectural pattern, which is a reusable, well-established architectural solution to a recurring design problem \cite{bass_software_2021}.
As summarised in Table~\ref{selected-systems}, the systems implement a range of architectural styles and patterns.
This section will discuss the architectures of the systems, including how they support the achievement of the seven key challenges discussed in Section~\ref{key-challenges}.

\subsection{Architectural styles and patterns}
The selected systems implement established architectural styles and patterns, including layered, modular, service-oriented, and agent-based architectures \cite{buschmann_pattern-oriented_1996,richards_software_2015}.
We discuss each in turn below.

\subsubsection{Layered architecture}
The most common type of architecture among the systems is the layered architecture, implemented in 28 of the systems \cite{akhtar_multi-agent_2022,ali_intelligent_2021,ali_smart_2020,ali_type-2_2018,alti_agent-based_2021,ammar_using_2021,elhadj_do-care_2021,esposito_smart_2018,fenza_hybrid_2012,garcia-moreno_systematic_2023,hadjadj_integration_2021,henaien_combined_2020,kilintzis_supporting_2019,kim_ontology_2014,kordestani_extended_2021,lopes_ontology-driven_2023,martella_semantically_2025,mavropoulos_smart_2021,mcheick_stroke_2016,mezghani_semantic_2015,reda_heterogeneous_2022,spoladore_ontology-based_2021,titi_ontology-based_2019,vadillo_enhancement_2013,villarreal_mobile_2014,xu_design_2017,yu_semantic_2017,zhang_knowledge-based_2014}.
In this pattern, each layer consists of a group of subtasks, with each group being at a particular level of abstraction \cite{buschmann_pattern-oriented_1996}.
This offers several advantages. It is simple to understand, and the separation of concerns among the different layers makes it easy to test and maintain the systems developed using this architecture \cite{richards_software_2015}.
Among the systems, there are variations in the number of layers and their functionality.
However, the first layer is typically dedicated to data collection from wearable or ambient sensors as well as other data sources. 
It may be named the data collection layer, as in the systems by Ali et al. \cite{ali_intelligent_2021,ali_smart_2020}, the sensing layer, as in the systems by Elhadj et al. \cite{elhadj_do-care_2021} and Esposito et al. \cite{esposito_smart_2018}, or the user layer as in the system by Alti et al. \cite{alti_agent-based_2021}.
Other typical layers include a data storage layer in which data is securely stored; networking layer which manages data communication and transmission in the system; inference and data analysis layer, in which the raw data is processed and analysed to derive important insights; and finally, presentation layer in the form of a user interface where individuals and in some cases, their clinicians and caregivers, can receive visualisations and alerts.  
Other specialised layers may also be included, such as the security layer in the system by Ali et al. \cite{ali_type-2_2018}, or the agents modelling and reasoning layer as proposed by Akhtar et al. \cite{akhtar_multi-agent_2022}.

\subsubsection{Modular architecture}
Similar to the layered architecture is the modular architecture, in which the system is subdivided into modules, blocks, or subsystems.
This is the second most common architectural pattern among the systems, with some kind of modular pattern implemented in 21 of the systems \cite{bampi_ontology-driven_2025,chatterjee_automatic_2021,chiang_context-aware_2015,de_brouwer_mbrain_2022,el-sappagh_mobile_2019,hooda_semantic_2020,hussain_big-ecg_2021,ivascu_activity-aware_2021,ivascu_multi-agent_2015,khozouie_ontological_2018,lopes_ontology-driven_2023,martella_semantically_2025,mavropoulos_smart_2021,minutolo_hybrid_2016,rhayem_semantic-enabled_2021,stavropoulos_detection_2021,yu_improving_2022,zafeiropoulos_evaluating_2024,zeshan_iot-enabled_2023,zhang_knowledge-based_2014,zhou_design_2022}.
Modular and layered architectural patterns can be used concurrently, as is done in four systems \cite{lopes_ontology-driven_2023,martella_semantically_2025,mavropoulos_smart_2021,zhang_knowledge-based_2014}. 
For example, in the system proposed by Zhang et al. \cite{zhang_knowledge-based_2014}, the client management module has a middleware with a layered architecture.
Additionally, because layered architectures tend to be monolithic, making them less agile and difficult to scale and deploy \cite{richards_software_2015}, modularity of layered architectures is advised, in which each layer consists of a modular set of components with a single function or purpose \cite{meyer_modular_2005}.
This is implemented by Mavropoulos et al. \cite{mavropoulos_smart_2021}, whose architecture has 3 levels (layers), with each containing specific modules. For example, the sensors management level contains a data analysis module, while the communication understanding level contains a natural language processing module.
Similarly, Lopes de Souza et al. \cite{lopes_ontology-driven_2023} implement a semantic module within their layered architecture.

\subsubsection{Service-oriented architecture}
Another well-known architectural pattern is the service-oriented architecture, a distributed pattern in which system components provide and consume services \cite{bass_software_2021}.
This is used in eight of the systems \cite{alti_agent-based_2021,fenza_hybrid_2012,garcia-moreno_systematic_2023,hristoskova_ontology-driven_2014,kilintzis_supporting_2019,martella_semantically_2025,mezghani_semantic_2015,xu_design_2017}.
In service-oriented architectures, the different aspects of the challenges can be achieved using specialised services.
For example, Hristoskova et al. \cite{hristoskova_ontology-driven_2014} implement services such as a notification service to generate alerts (decision support) and a user location service to localize specific users (context awareness).
While the service-oriented architectural pattern is powerful and offers a high level of abstraction, it is often overly complex and difficult to understand \cite{richards_software_2015}. 
A way of mitigating these issues is to implement services in a layered architecture, as is done in several other systems \cite{alti_agent-based_2021,fenza_hybrid_2012,mezghani_semantic_2015,xu_design_2017}.
Additionally, agents can be used to effectively manage services, as is the case in the systems proposed by Alti et al. \cite{alti_agent-based_2021} and Fenza et al. \cite{fenza_hybrid_2012}.

\subsubsection{Agent-based architecture}
Among the systems, nine implement an agent-based architecture. Eight of these use a multi-agent architecture \cite{akhtar_multi-agent_2022,alti_agent-based_2021,ammar_using_2021,fenza_hybrid_2012,ivascu_activity-aware_2021,ivascu_multi-agent_2015,martella_semantically_2025,vadillo_enhancement_2013}, while one implements a single-agent architecture \cite{mavropoulos_smart_2021}. 
Multi-agent systems are characterised by the existence of more than one agent acting autonomously within the system.
Typically, each agent manages a particular aspect of the system, which enables decentralisation, efficiency, and scalability.
For example, Alti et al. \cite{alti_agent-based_2021} implement situation detection using a situation reasoning agent and a diseases classifying agent, while Ivașcu and Negru \cite{ivascu_activity-aware_2021} and Ivașcu et al. \cite{ivascu_multi-agent_2015} have notification and alert agents that enhance decision support.
Similarly, the system proposed by Vadillo et al. \cite{vadillo_enhancement_2013} has a sensor validation agent to verify sensor observations thereby managing uncertainty in sensor data, a location agent to mange user locations thereby contributing to context awareness, and a medication agent to oversee the administering of medication, which contributes to decision support.
Among the multi-agent systems that incorporate a service-oriented architecture, agents are instrumental in managing the complexity of the services. Both  Alti et al. \cite{alti_agent-based_2021} and Fenza et al. \cite{fenza_hybrid_2012} use agents to handle service discovery and selection. 
Agents can also enhance decision support by interacting directly with users of the system. 
This is demonstrated by Mavropoulos et al. \cite{mavropoulos_smart_2021}, who use a smart virtual agent capable of dialogue to communicate with clinicians and support their decision-making.

\subsection{Proposed reference architecture}
\begin{figure}[ht]
\includesvg[width=\textwidth]{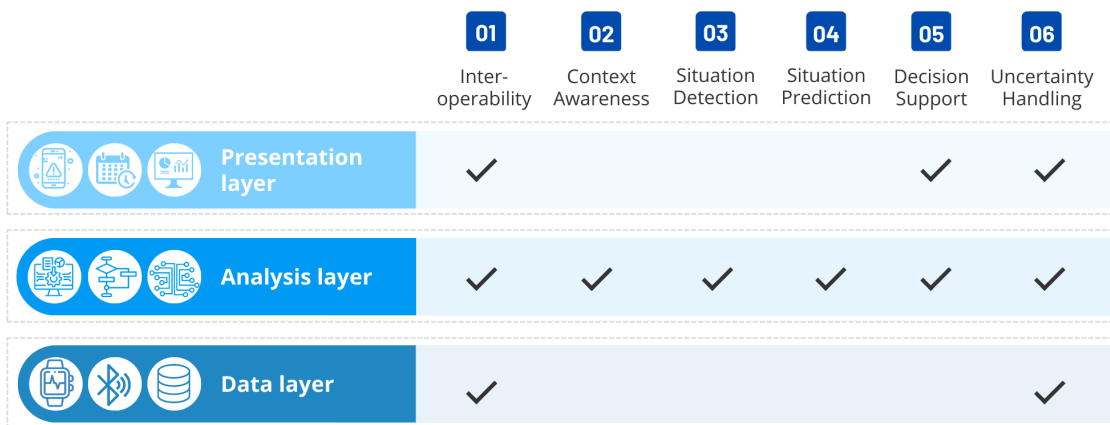}
\caption{Reference architecture for sensor-based personal health monitoring systems.\label{architecture}}
\end{figure}

Based on an analysis of the systems as well as an overview of general sensor-based systems, a reference architecture for personal health monitoring systems is presented in Figure~\ref{architecture}.

The architecture consists of three layers as described below:

\begin{enumerate}
    \item The \textbf{data layer} contains two modules. 
    The \textit{data acquisition module} supports the acquisition of data from body sensors as well as ambient sensors, health records, and user-submitted sources such as questionnaires and social media content. 
    The \textit{data preprocessing module} supports the preprocessing of the acquired data, including data cleaning, normalisation, and feature extraction. 
    
    \item The \textbf{analysis and decision layer} consists of the \textit{situation analysis module}, which provides functionality to derive relevant detected and predicted situations from the data using techniques such as rules, ML, Bayesian networks, and fuzzy logic; and the \textit{decision support module}, which follows up on the detected and predicted situations to recommend interventions that mitigate adverse situations and promote favourable ones.
    Central to both modules is expert health knowledge, including established clinical guidelines.
    
    \item The \textbf{presentation layer} provides functionality through which users can receive communication from and interact with the system.
    The \textit{user communication module} provides support for the system-generated communication of situations and recommended interventions through mediums such as text messages and emails, while 
    the \textit{user interface module} provides web and mobile applications with which users can interact with the system.
\end{enumerate}

The architecture also includes a cross-cutting \textbf{knowledge graph} which represents heterogeneous health data. 
The underlying data schema is defined through an ontology, allowing for the semantic annotation of data and reuse of existing semantic resources such as the SAREF core ontology and its extensions for the health domain.
The knowledge graph can be implemented and stored in graph databases such as GraphDB.
A linked data approach is recommended to ensure related data is interlinked, thereby enabling data integration and reuse.
This architecture is not only consistent with the layered architectures proposed in related reviews \cite{albahri_real-time_2018,bollineni_iot_2025,philip_internet_2021,yin_internet_2016,compton_survey_2009,honti_review_2019}, but also includes modules to separate related but distinct functionalities within each layer, thereby mitigating the monolithicity of the layered approach.
Table~\ref{architecture-table} highlights the key functionalities, recommended tools and techniques, and the key challenges addressed in each layer and module. 

\begin{table*}[ht]
\setcellgapes{5pt}
\makegapedcells
\renewcommand\multirowsetup{\centering} 
\scriptsize
\caption{Summary of the layers and modules within the proposed reference architecture} \label{architecture-table}
\begin{tabular}
{>{\raggedright}p{.03\linewidth}>{\raggedright}p{.09\linewidth}>{\raggedright}p{.28\linewidth}>{\raggedright}p{.32\linewidth}>{\raggedright}p{.14\linewidth}}
\hline
\textbf{Layer} & \textbf{Module} & \textbf{Inputs, processing tasks, and outputs} & \textbf{Suggested development tools} & \textbf{Challenges addressed}\tabularnewline
\hline

\multirow{9}{*}{\rotatebox[origin=c]{90}{Presentation layer}} & \makecell[l]{User \\communication \\module} & \makecell[l]{- Inputs: detected and predicted situations, \\recommended interventions, and explanations \\- Processing: content structuring and adaptation
 \\- Outputs: text messages and emails} & \makecell[l]{- Communication protocols e.g. Internet Message \\Access Protocol (IMAP), Simple Mail Transfer \\Protocol (SMTP), Short Message Service (SMS) \\- Messaging communication software e.g. Twilio, \\Plivo \\- Asynchronous message queuing software e.g. \\RabbitMQ, Apache Kafka \\- Templating engines} & \makecell[l]{- Interoperability \\- Context awareness \\- Explainability} \tabularnewline \greyclineone
 & User interface module & \makecell[l]{- Inputs: software requirements, multimedia \\content, dynamic situations, interventions and \\explanations \\- Processing: development of software applic-\\ations in accordance with software development \\methodologies \\- Outputs: web and mobile interfaces \\and documentation} & \makecell[l]{- Web development languages e.g. HTML, CSS, \\Python, Javascript \\- Web development frameworks e.g. Python's Django \\and Flask and Javascript's Angular and React \\- Mobile development languages e.g. Kotlin, Swift, \\Java} & \makecell[l]{- Interoperability \\- Context awareness \\- Explainability} \tabularnewline \greyline

\multirow{8}{*}{\rotatebox[origin=c]{90}{Analysis and decision layer}} & Situation analysis module & \makecell[l]{- Inputs: preprocessed health data, expert \\knowledge \\- Processing:
execution of situation analysis \\rules, algorithms and models \\- Outputs: detected and predicted situations \\and explanations} & \multirow{-3}{*}{\makecell[l]{- Rule expression languages and extensions e.g. \\SHACL, SPIN, SWRL, Bayes-SWRL \\- Semantic query languages e.g. SPARQL \\- Programming languages for ML model \\development and/or fuzzy logic implementation \\e.g. Python and MATLAB  \\- Semantic Web editors and frameworks e.g. \\Protégé and Apache Jena \\- Reasoners e.g. Pellet and HermiT \\- ML and deep learning libraries and\\ frameworks e.g. Python's scikit-learn, PyTorch, \\TensorFlow \\- Fuzzy logic development libraries e.g. MATLAB's \\Fuzzy Logic Toolbox and Python's Scikit-Fuzzy \\- Bayesian network modelling software e.g. Netica}} &  \makecell[l]{- Interoperability \\- Context awareness \\- Situation detection \\- Situation prediction \\- Explainability \\- Uncertainty handling\\\\} \tabularnewline 
\greyclinetwo \greyclinethree
& Decision support module & \makecell[l]{\\- Inputs: detected and predicted situations, \\expert knowledge \\- Processing:
execution of decision support \\rules, algorithms, and models \\- Outputs: recommended interventions \\and explanations} &  & \makecell[l]{- Interoperability \\- Context awareness \\- Decision support \\- Explainability \\- Uncertainty handling} \tabularnewline \greyline
 
\multirow{9}{*}{\rotatebox[origin=c]{90}{Data layer}} & Data acquisition module & 
\makecell[l]{- Inputs: sensor devices, electronic health \\records, and user-submitted sources e.g. \\questionnaires, social media content \\- Processing: acquisition of data \\- Outputs: acquired heterogeneous health \\data}
& \makecell[l]{- Data serialisation formats e.g. XML, JSON, \\Turtle \\- Database query languages e.g. SQL \\- Data transmission protocols e.g. Bluetooth, \\Bluetooth Low Energy, ANT+ \\- Social media APIs} & \makecell[l]{- Interoperability \\ - Context awareness} \tabularnewline \greyclineone
& \makecell[l]{Data \\preprocessing \\module} & \makecell[l]{- Inputs: raw health data \\- Processing: data cleaning (handling missing, \\noisy, or erroneous data), signal processing, \\feature engineering \\- Outputs: preprocessed health data} & \makecell[l]{- Programming languages for data analysis and \\signal processing e.g. \\Python, MATLAB, R \\- Data analysis libraries e.g. Python's NumPy, \\SciPy, Pandas \\- Data visualisation platforms e.g. Tableau, \\Microsoft Power BI} & Uncertainty handling \tabularnewline
\arrayrulecolor{black}\hline
\end{tabular}
\end{table*}

\section{Discussion}
\label{discussion}

\subsection{Summary of findings}

\begin{figure}[ht]
\includesvg[width=\textwidth]{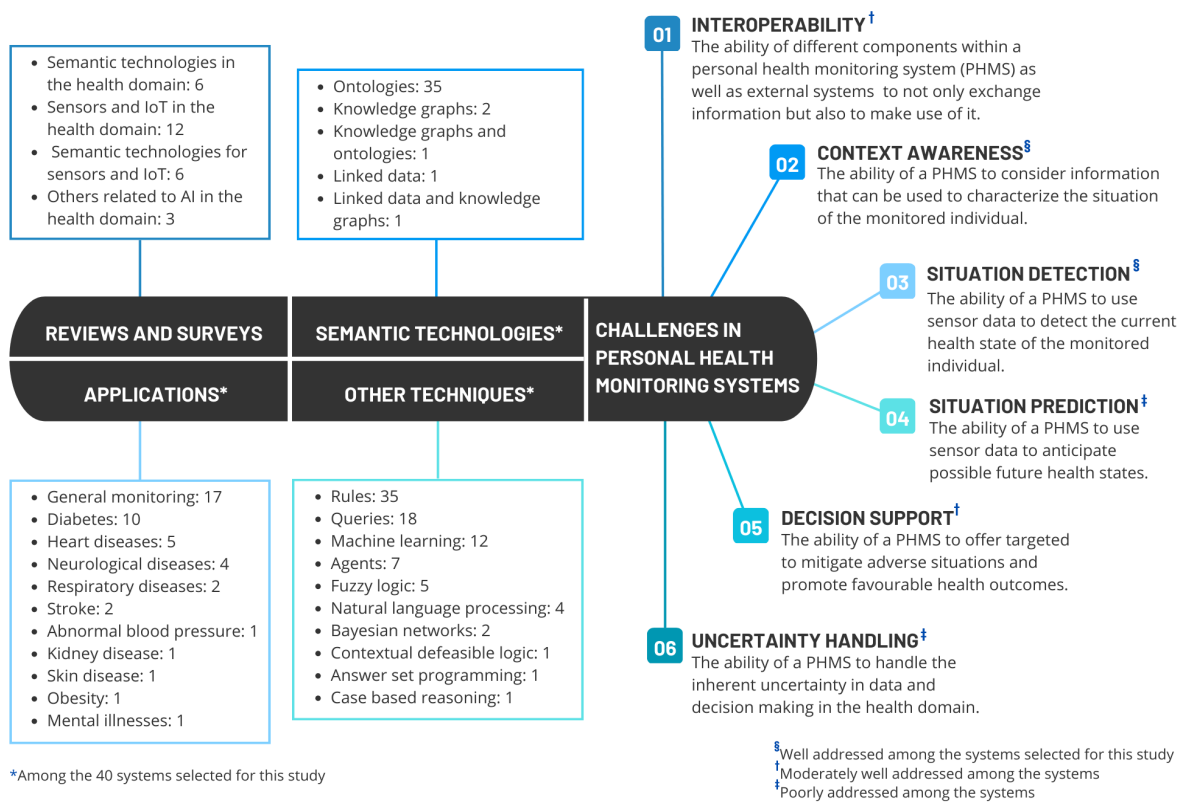}
\caption{Map showing the current state of the field. \label{map}}
\end{figure}

Figure~\ref{map} shows a map outlining the current state of the field.
In the remainder of this section, the inadequacies and limitations in current systems will be highlighted, paving the way for opportunities for future research.

\subsubsection{Summary of the extent to which key challenges are addressed}

Our findings show that three of the seven key challenges are particularly poorly addressed among the systems: situation prediction, explainability, and uncertainty handling. 
Most of the systems included in this study do not adequately address the challenge of situation prediction, with only 12 systems being capable of predicting health risks or giving insight into how detected conditions may progress with time. 
In order to achieve the vision of precision health, it is important for health monitoring systems to go beyond detecting current health states and move towards the anticipation and mitigation of adverse health states.
With regard to explainability, all the reviewed systems use Semantic Web technologies, which are inherently interpretable, and three systems provide chatbots or digital assistants which can respond to user queries. However, only nine systems present explicit explanations for system outputs. 
Additionally, none of the systems implement the criteria for good explanations as outlined by Molnar et al. \cite{molnar_interpretable_2023}, nor do any systems mention tailoring the explanations presented to suit different audiences.

Uncertainty handling is similarly poorly implemented or not addressed at all in majority of the systems. While 16 of the systems consider the impact of sensor limitations such as noise and missing values, only 11 address the inherent uncertainty present in situation analysis and decision support in the health domain. This hinders their ability to perform reliably when faced with ambiguous data or vague or limited knowledge, thus reducing their trustworthiness and dependability. 
Both situation prediction and uncertainty handling can be enhanced by a combination of techniques, as suggested by Behera et al. \cite{behera_emerging_2019}, such as ML and Bayesian networks. Few of the systems take such an approach, with the majority using solely rule-based reasoning.

To a lesser extent, interoperability and decision support are also not fully addressed in the selected systems.
Considering interoperability, we found that only 15 of the systems take advantage of established sensor ontologies such as SAREF.
Neglecting to use such ontologies limits the standardisation and expressiveness of the descriptions of sensors and, importantly, sensor data.
This results in less effective querying of and reasoning on sensor data, which in turn negatively impacts situation analysis.
Further, while 20 systems incorporate established medical terminologies such as SNOMED CT and ICD, all but two systems fail to consider existing health data standards. This significantly limits their ability to effectively integrate electronic health records and other standardized clinical data sources.

There is also significant room for improvement in addressing the challenge of decision support. 
While 37 of the included systems incorporate alerts to warn of hazardous situations, many do not offer recommendations or reminders for medication or lifestyle factors such as diet and exercise. 
Similarly, only 15 of the systems report using established medical guidelines, clinical workflows, or information from vetted health professional bodies, which can help to provide a sound justification for any recommendations made, thereby enhancing the trustworthiness of the systems.
However, the most overlooked aspect of decision support remains the human-centered aspect.
Systems should support users' agency to cognitively engage in decision-making by presenting them with various options and their potential outcomes, and allowing them to be the final decision-makers.
While 29 of the selected systems do suggest recommendations, none offer more than one potential option or present their potential outcomes.

\subsubsection{Summary of the quality assessment}
With regard to the quality assessment, we found that 18 of the systems did not report the data collection methods or sources.
Additionally, only 19 of the systems reported the specific devices used for data collection.
It can be assumed that such systems may not have been properly validated using realistic data, which casts doubts on the claims made regarding the system functionality and performance.
To mitigate this, researchers should clearly indicate which data was used to validate their systems, including how the data was collected, who it was collected from, and the devices that were used.
Concerning the development of the systems, only 16 systems used an existing methodology or else systematically outlined the development steps followed for any of the system components. 
However, nearly all the systems provided details of the languages, platforms, tools, and other software used in the development process.

When it comes to the evaluation of the systems and components, nearly all the systems reported on the methods, metrics, and results of the evaluation process.
However, only 16 included some kind of external evaluation, whether through a systematic comparison with other similar systems or seeking evaluation from domain experts.
Furthermore, as was found in the review by Haque et al. \cite{haque_semantic_2022}, most of the selected systems are yet to be evaluated in real-world settings. 
While this is to be expected in an emerging area, it is imperative that more systems be evaluated in real-world settings going forward, so that practical challenges and user feedback can be identified early on and considered in future system proposals. 
This feedback loop is essential for undertaking further research into personal health monitoring systems that fully harness the potential of Semantic Web technologies. 
Finally, accessibility of resources was poor among the systems, with only six systems providing access to relevant system files.
Wherever possible, researchers should include the research outputs such as ontologies, data, and code as publicly accessible supplementary material in order to enhance reproducibility and verifiability.

\subsection{Future research directions}
This study highlights the fact that many personal health monitoring systems do not fully leverage reusable resources, and instead opt to build resources from scratch.
We also find that an overwhelming majority of the systems are not built with reusability in mind, as evidenced by the limited availability of research outputs from various researchers. 
Although different health conditions may require specific features and functionalities, Semantic Web technologies have the potential to be extendable, allowing for the addition of knowledge as it evolves and making them suitable for reuse across a wide range of health monitoring applications.
We therefore invite researchers to not only reuse existing resources but also to build generalisable semantic and non-semantic system components and make them publicly available.
This would play an important role in accelerating the development of personal health monitoring systems by avoiding redundant efforts.

Another takeaway that has clearly emerged from this study is the advantage of combining Semantic Web technologies with other AI techniques such as ML and Bayesian networks.
Integrating these approaches can significantly improve the tackling of the seven key challenges identified in the study.
ML can also be leveraged to support the development of Semantic Web technologies \cite{damato_machine_2020,kotis_machine_2021}. 
We encourage researchers to explore recent software libraries such as DeepOnto \cite{he_deeponto_2024}, which support ontology engineering tasks using deep learning tools and pre-trained language models.
Indeed, the use of large language models and generative AI is gaining traction in the Semantic Web community for tasks ranging from the generation of competency questions, OWL files, and documentation in ontology engineering \cite{garijo_llms_2024}, to link prediction \cite{shu_knowledge_2024} and graph population \cite{meyer_llm-assisted_2024} in knowledge graph engineering.
Future research should explore how generative AI can accelerate the development of Semantic Web-based personal health monitoring systems.
We also note that most systems do not report the use of recently-proposed Semantic Web standards, with only one system using SHACL, one using SPIN, and none using RDF-star or Notation3.
We encourage researchers to explore these and other state-of-the-art standards.

Additionally, many of the systems do not take into account factors such as diet, exercise, and other determinants of health. 
The next generation of personal health monitoring systems must be more holistic, focusing not only on disease but also on overall wellness.
This includes the monitoring of emotional and mental states, which has been shown to be linked to physical health \cite{uskul_emotions_2015}.
Such information can be represented using Semantic Web technologies, including ontologies \cite{alani_survey_2018} and knowledge graphs \cite{gyrard_interdisciplinary_2022}.
The inadequately addressed challenges, together with the need for more holistic health monitoring, present interesting and important directions for future research in the field.

\subsection{Limitations of this study}
While we believe that this article offers a comprehensive overview of the use of Semantic Web technologies in personal health monitoring systems, it is a very broad area and thus we have necessarily had to delimit the scope of the article.
We focus the depth of coverage on the seven key challenges, the quality analysis of the selected systems, and the proposed reference architecture.
While a few additional challenges are discussed, they are not included in the in-depth analysis, and other potential challenges may not be mentioned in this article.
Furthermore, although we provide an overview of sensors used for health monitoring and highlight some hardware-based interoperability challenges, the practical aspects of the seamless integration of physical sensors and the real-time processing of sensor data are not discussed in depth.

Additionally, we recognise that the challenge assessment framework may not be equally applicable to all health monitoring systems, as some aspects may not be relevant to a particular system's design goals.
Rather than providing a definitive score on system excellence, it should be viewed as a framework for considering different aspects of the seven challenges and identifying existing systems that have addressed them effectively. 
We encourage researchers and developers to adapt the framework to align with their specific system requirements.
Finally, as our inclusion criteria specified peer-reviewed journal articles and conference papers published in English, we acknowledge possible publication and language bias.
Despite these limitations, we believe that the study provides valuable insights and offers a foundation for future research in this rapidly evolving field.

\section{Conclusion}
\label{conclusion}

This systematic mapping study has analysed the landscape of sensor-based personal health monitoring systems that incorporate Semantic Web technologies.
After a careful consideration of the pertinent issues in this application area, we identified seven key challenges that such systems must address.
In a systematic process, we selected 48 systems as representative of the state of the art in the field, and critically analysed them based on their capacity to address the seven challenges.
We also evaluated the quality of the research undertaken to develop them.
Moreover, we discussed the architectures of the selected systems, and proposed a reference architecture to streamline the development of such systems.
Lastly, we discussed the key findings of the study and highlighted opportunities for future research.
It is our hope that this study will serve as a comprehensive overview of the field and spur further high-quality research in effective personal health monitoring systems.

\begin{ack}
This work was financially supported by the Hasso Plattner Institute for Digital Engineering through the HPI Research School at the University of Cape Town.
It was also supported in part by the National Research Foundation of South Africa (grant numbers: 151217, SRUG2204264808).
\end{ack}

\begin{appendix}

\begin{landscape}

\section{Reused semantic resources}
\label{appendix-re-used}

\scriptsize
\setlength{\extrarowheight}{2pt}
\begin{longtable}{>{\raggedright\arraybackslash}m{.29\NetTableWidth}>{\raggedright\arraybackslash}m{.52\NetTableWidth}>{\raggedright\arraybackslash}m{.15\NetTableWidth}>{\raggedright\arraybackslash}m{.39\NetTableWidth}}
\caption{Reused semantic resources, the systems that reuse them, and the challenges they address. Each resource includes either a hyperlink or a reference.}\label{reused-summary}\\
\hline
\textbf{Reused resource} & \textbf{Description} & \textbf{Systems that reuse it} & \textbf{Challenge(s) that it addresses}\tabularnewline
\hline
\endfirsthead

\multicolumn{4}{c}{\tablename\ \thetable\ -- continued from previous page} \\
\hline
\textbf{Reused resource} & \textbf{Description} & \textbf{Systems that reuse it} & \textbf{Challenge(s) that it addresses}\tabularnewline
\hline
\endhead

\hline
\multicolumn{4}{r}{\textbf{Table continued on next page.}}\\
\endfoot

\hline
\endlastfoot

Amigo Device Ontology \cite{vallee_dynamic_2005} & Provides support for the description of devices, user context, quality of service parameters, and communication protocols. & \cite{hristoskova_ontology-driven_2014} & Interoperability; Context awareness \tabularnewline
\rowcolor{gray!4}
\href{http://purl.org/ontology/ao/core}{Association Ontology} & Provides basic concepts and properties for describing specific association statements to something. & \cite{yu_semantic_2017} & Interoperability \tabularnewline
\href{https://basic-formal-ontology.org}{Basic Formal Ontology} & A foundational ontology designed for use in supporting information retrieval, analysis, and integration in scientific and other domains. & \cite{el-sappagh_mobile_2019} & Interoperability \tabularnewline
\rowcolor{gray!4}
\href{https://www.w3.org/2003/01/geo}{Basic Geo (WGS84 - lat/long)} & A basic RDF vocabulary that provides a namespace for representing lat(itude), long(itude), and other information about spatially-located things. & \cite{martella_semantically_2025,reda_heterogeneous_2022} & Interoperability; Context awareness \tabularnewline
\href{https://bioportal.bioontology.org/ontologies/DIAB}{BioMedBridges Diabetes Ontology} & Represents expert knowledge about stages and phenotypes of type 2 diabetes. & \cite{ali_intelligent_2021} & Interoperability; Situation detection; Situation prediction; Explainability \tabularnewline
\rowcolor{gray!4}
Chronic Obstructive Pulmonary Disease Ontology (COPDology) \cite{ajami_ontology-based_2018} & Contains concepts related to the disease, environment, equipment, patient data, and treatment. & \cite{mavropoulos_smart_2021} & Interoperability; Situation detection; Situation prediction; Decision support; Context awareness; Explainability \tabularnewline
Context-Driven Adaptation of Mobile Services (CoDAMoS) Ontology \cite{preuveneers_towards_2004} & An adaptable and extensible context ontology for creating context-aware computing infrastructures by representing users, environments, hardware and software platforms, and services. & \cite{martella_semantically_2025} & Interoperability; Context awareness \tabularnewline
\rowcolor{gray!4}
Core Ontology Model for Crowd-Sensing (COMCS) Ontology \cite{wang_wselector_2015} & A core ontology modelling upper-level concepts for worker selection in different crowd-sensing tasks. & \cite{martella_semantically_2025} & Interoperability; Context awareness \tabularnewline
\href{https://cyc.com/platform}{Cyc Ontology} & A large knowledge base of common sense and background knowledge. & \cite{peral_ontology-oriented_2018} & Interoperability \tabularnewline
\rowcolor{gray!4}
\href{https://dahcc.idlab.ugent.be}{Data Analytics for Health and Connected Care (DAHCC) Ontology} & Captures care, patient, daily life activity recognition and lifestyle domain knowledge. & \cite{de_brouwer_mbrain_2022,zafeiropoulos_evaluating_2024} & Interoperability; Situation detection; Situation prediction; Context awareness; Explainability \tabularnewline
\href{https://www.dbpedia.org}{DBPedia} & A knowledge base of extracted structured information from Wikipedia in the form of an open knowledge graph served as Linked Data. & \cite{reda_heterogeneous_2022,xu_design_2017} & Interoperability \tabularnewline
\rowcolor{gray!4}
Diabetes Mellitus Diagnosis Ontology \cite{el-sappagh_ddo_2016} & A diabetes knowledge base that supports automatic reasoning for solving problems related to diabetes diagnosis. & \cite{el-sappagh_mobile_2019} & Interoperability; Situation detection; Situation prediction; Decision support; Explainability \tabularnewline
\href{https://bioportal.bioontology.org/ontologies/DMTO}{Diabetes Mellitus Treatment Ontology} & Provides knowledge about type 2 diabetes and its patients including complications, symptoms, tests, interactions, and treatment plans. & \cite{ali_intelligent_2021,el-sappagh_mobile_2019} & Interoperability; Situation detection; Situation prediction; Decision support; Explainability \tabularnewline
\rowcolor{gray!4}
Diseasome \cite{goh_human_2007} & A.k.a. the human disease network; a graph that captures all genetic disorders and disease genes and the links between them. & \cite{xu_design_2017} & Interoperability; Situation detection; Situation prediction; Explainability \tabularnewline
\href{https://iot-ontologies.github.io/dogont}{DogOnt Ontology} & Offers a uniform, extensible model for smart IoT environments, including the location, capabilities, and configurations of devices, and the composition of the smart environment. & \cite{bampi_ontology-driven_2025} & Interoperability; Context awareness \tabularnewline
\rowcolor{gray!4}
\href{http://www.ontologydesignpatterns.org/ont/dul/DUL.owl}{DOLCE+DnS Ultra Lite (DUL) Ontology} & A foundational ontology that provides a set of upper level concepts for easier interoperability among middle and lower level ontologies. & \cite{de_brouwer_mbrain_2022,stavropoulos_detection_2021} & Interoperability \tabularnewline
\href{https://go.drugbank.com/}{DrugBank} & A knowledge resource for drug, drug–target and related pharmaceutical information. & \cite{xu_design_2017} & Interoperability; Decision support; Explainability\tabularnewline
\rowcolor{gray!4}
\href{https://bioportal.bioontology.org/ontologies/DTO}{Drug Target Ontology} & Provides formalized and standardized classifications and annotations of druggable protein targets. & \cite{ali_intelligent_2021} & Interoperability; Decision support; Explainability \tabularnewline
\href{https://w3id.org/MON/event.owl}{Event Ontology} & Deals with the notion of reified events, which may have a location, a time, active agents, factors and products. & \cite{yu_semantic_2017} & Interoperability; Context awareness \tabularnewline
\rowcolor{gray!4}
\href{https://w3c.github.io/hcls-fhir-rdf/spec/ontology.html}{Fast Healthcare Interoperability Resources (FHIR) OWL Ontology} & OWL ontology for FHIR Resources represented in RDF. & \cite{el-sappagh_mobile_2019} & Interoperability \tabularnewline
FHIR \& SSN-based T1 Diabetes Ontology (FASTO) \cite{el-sappagh_mobile_2019} & Ontology integrating FHIR standard and SSN ontology with clinical practice guidelines for real-time management of insulin for diabetes patients. & \cite{ali_intelligent_2021} & Interoperability; Situation detection; Situation prediction; Decision support; Explainability \tabularnewline
\rowcolor{gray!4}
\href{https://foodon.org}{Food Ontology (FoodOn)} & A lightweight ontology for describing recipes, ingredients, menus, and diets. & \cite{spoladore_ontology-based_2021} & Interoperability; Decision support; Explainability \tabularnewline
\href{http://xmlns.com/foaf/spec}{Friend of a Friend (FOAF) Ontology} & An ontology describing persons, their activities, and their relations to other people and objects. & \cite{ammar_using_2021,elhadj_do-care_2021,garcia-moreno_systematic_2023,henaien_combined_2020,mavropoulos_smart_2021,reda_heterogeneous_2022,spoladore_ontology-based_2021,titi_ontology-based_2019,yu_semantic_2017} & Interoperability; Context awareness \tabularnewline
\rowcolor{gray!4}
General User Model Ontology (GUMO) \cite{heckmann_gumo_2005} & A user model ontology for use in user-adaptive or ubiquitous computing systems. & \cite{mavropoulos_smart_2021} & Interoperability; Context awareness \tabularnewline
\href{https://www.geonames.org/ontology/documentation.html}{GeoNames Ontology} & Provides elements of description for geographical features. & \cite{rhayem_semantic-enabled_2021} & Interoperability; Context awareness \tabularnewline
\rowcolor{gray!4}
HealthIoT Ontology \cite{rhayem_healthiot_2017} & Provides semantic representation of both the medical connected objects (i.e. sensors) and their data. & \cite{ivascu_activity-aware_2021} & Interoperability; Context awareness \tabularnewline
\href{https://bioportal.bioontology.org/ontologies/HFO}{Heart Failure Ontology} & Defines heart-failure-relevant information including the causes and risk factors, signs and symptoms, diagnostic tests and results, and treatment. & \cite{hristoskova_ontology-driven_2014} & Interoperability; Situation detection; Situation prediction; Decision support; Explainability \tabularnewline
\rowcolor{gray!4}
\href{https://bioportal.bioontology.org/ontologies/DOID}{Human Disease Ontology} & Represents common and rare disease concepts. & \cite{ali_intelligent_2021} & Interoperability \tabularnewline
\href{https://obofoundry.org/ontology/ico.html}{Informed Consent Ontology} & Represents processes and information pertaining to obtaining informed consent in medical investigations. & \cite{yu_semantic_2017} & Interoperability; Context awareness \tabularnewline
\rowcolor{gray!4}
\href{https://www.icn.ch/how-we-do-it/projects/ehealth-icnptm/about-icnp}{International Classification for Nursing Practice (ICNP)} & Provides a standardised terminology that can be used to record the observations and interventions of nurses. & \cite{elhadj_do-care_2021,henaien_combined_2020} & Interoperability \tabularnewline
\href{https://icd.who.int/en/}{International Classification of Diseases (ICD)} & A tool for recording, reporting and grouping conditions and factors that influence health. & \cite{spoladore_ontology-based_2021,titi_ontology-based_2019,yu_improving_2022} & Interoperability; Situation detection; Explainability \tabularnewline
\rowcolor{gray!4}
\href{https://icd.who.int/browse/2025-01/icf/en}{International Classification of Functioning, Disability and Health (ICF)} & A framework for describing and organising information on functioning and disability. & \cite{garcia-moreno_systematic_2023,spoladore_ontology-based_2021} & Interoperability; Situation detection; Explainability \tabularnewline
\href{https://ichd-3.org}{International Classification of Headache Disorders (ICHD-3)} & A detailed hierarchical classification of all headache-related disorders. & \cite{de_brouwer_mbrain_2022} & Interoperability; Situation detection; Explainability \tabularnewline
\rowcolor{gray!4}
\href{https://www.w3.org/submissions/iot-lite}{IoT-lite Ontology} & A lightweight ontology to represent IoT resources, entities, and services. & \cite{garcia-moreno_systematic_2023,rhayem_semantic-enabled_2021} & Interoperability; Context awareness \tabularnewline
IoT-Stream ontology \cite{elsaleh_iot-stream_2020} & A lightweight ontology to semantically annotate stream data from IoT devices. & \cite{garcia-moreno_systematic_2023} & Interoperability; Context awareness \tabularnewline
\rowcolor{gray!4}
\href{https://loinc.org}{Logical Observation Identifiers Names and Codes (LOINC)} & An international standard for identifying health measurements, observations, and documents. & \cite{el-sappagh_mobile_2019} & Interoperability \tabularnewline
\href{https://www.w3.org/TR/ldn/}{Linked Data Notifications (LDN)} & A protocol that describes how servers (receivers) can have messages pushed to them by applications (senders), as well as how other applications (consumers) may retrieve those messages. & \cite{ammar_using_2021} & Interoperability \tabularnewline
\rowcolor{gray!4}
\href{https://www.w3.org/TR/ldp/}{Linked Data Platform (LDP)} & Defines a set of rules for HTTP operations on web resources, some based on RDF, to provide an architecture for read-write Linked Data on the web. & \cite{ammar_using_2021} & Interoperability \tabularnewline
\href{https://www.nlm.nih.gov/mesh/meshhome.html}{Medical Subject Headings (MeSH)} & A vocabulary for indexing, cataloguing, and searching biomedical and health-related information. & \cite{reda_heterogeneous_2022} & Interoperability \tabularnewline
\rowcolor{gray!4}
MIMU-Wear Ontology \cite{villalonga_mimu-wear_2017} & Describes wearable sensor platforms consisting of mainstream magnetic and inertial measurement units. & \cite{ivascu_activity-aware_2021} & Interoperability; Context awareness \tabularnewline
Mobile Crowd Sensing (MCS) Ontology \cite{gasmi_oaisis_2016} & A semantic model designed to structure collected data from crowd sensing systems. & \cite{martella_semantically_2025} & Interoperability; Context awareness \tabularnewline
\rowcolor{gray!4}
Mobile Object Ontology \cite{wannous_modelling_2013} & Models temporal, spatial, and domain related information about mobile objects. & \cite{rhayem_semantic-enabled_2021} & Interoperability; Context awareness \tabularnewline
\href{https://bioportal.bioontology.org/ontologies/MWLA}{Medical Web Lifestyle Aggregator Ontology} & Represents user data from the web. & \cite{yu_semantic_2017} & Interoperability; Context awareness \tabularnewline
\rowcolor{gray!4}
\href{https://bioportal.bioontology.org/ontologies/ONS}{Ontology for Nutritional Studies} & A comprehensive resource for the description of concepts in the broader human nutrition domain. & \cite{ali_intelligent_2021} & Interoperability; Decision support; Explainability \tabularnewline
\href{https://bioportal.bioontology.org/ontologies/OM}{Ontology of units of Measure (OM)} & An ontology that models concepts and relations important to scientific research, with a strong focus on units, quantities, measurements, and dimensions. & \cite{martella_semantically_2025} & Interoperability; Context awareness \tabularnewline
\rowcolor{gray!4}
\href{https://www.openthesaurus.de}{OpenThesaurus} & A multilingual web-based thesaurus. & \cite{peral_ontology-oriented_2018} & Interoperability \tabularnewline
OwlSpeak Ontology \cite{heinroth_owlspeak_2010} & Represents static and dynamic concepts related to spoken dialogue. & \cite{mavropoulos_smart_2021} & Interoperability; Context awareness \tabularnewline
\rowcolor{gray!4}
\href{https://www.w3.org/submissions/OWL-S}{OWL-S Ontology} & Describes Semantic Web services for automatic service discovery, invocation, composition, and interoperation. & \cite{fenza_hybrid_2012,hristoskova_ontology-driven_2014,martella_semantically_2025} & Interoperability; Situation detection; Situation prediction; Decision support; Context awareness \tabularnewline
\href{https://www.w3.org/TR/owl-time/}{OWL-Time Ontology} & Describes the temporal properties of resources. & \cite{de_brouwer_mbrain_2022,martella_semantically_2025,mavropoulos_smart_2021,rhayem_semantic-enabled_2021,titi_ontology-based_2019,yu_semantic_2017} & Interoperability; Context awareness \tabularnewline
\rowcolor{gray!4}
\href{https://bioportal.bioontology.org/ontologies/PMDO}{Parkinson and Movement Disorder Ontology} & Captures neurological findings, treatment plans, and instruments used to evaluate various traits of Parkinson's Disorder. & \cite{zafeiropoulos_evaluating_2024} & Interoperability; Situation detection; Situation prediction; Decision support; Context awareness; Explainability \tabularnewline
Physical Activity Concept Ontology \cite{kim_developing_2019} & Captures various concepts that are required to describe one's physical activity. & \cite{ivascu_activity-aware_2021} & Interoperability; Situation detection; Situation prediction; Decision support; Context awareness; Explainability \tabularnewline
\rowcolor{gray!4}
\href{https://lov.linkeddata.es/dataset/lov/vocabs/place}{Places Ontology} & A lightweight ontology for describing places of geographic interest. & \cite{yu_semantic_2017} & Interoperability; Context awareness \tabularnewline
\href{https://www.w3.org/2005/Incubator/ssn/ssnx/qu/qu-rec20.html}{Quantities and Units (QU) Ontology} & An ontology to model quantities and units of measurement. & \cite{garcia-moreno_systematic_2023} & Interoperability; Context awareness \tabularnewline
\rowcolor{gray!4}
\href{https://www.nlm.nih.gov/research/umls/rxnorm/index.html}{RxNorm} & Provides normalized names for clinical drugs and links to other drug vocabularies. & \cite{el-sappagh_mobile_2019} & Interoperability \tabularnewline
\href{https://saref.etsi.org/saref4ehaw}{SAREF for eHealth and Ageing Well (SAREF4EHAW) Ontology} & An extension of the SAREF ontology for applications related to eHealth and ageing well. & \cite{de_brouwer_mbrain_2022,lopes_ontology-driven_2023} & Interoperability; Context awareness \tabularnewline
\rowcolor{gray!4}
\href{https://saref.etsi.org/saref4wear}{SAREF for Wearables (SAREF4WEAR) Ontology} & An extension of the SAREF ontology for applications related to wearables. & \cite{hadjadj_integration_2021} & Interoperability; Context awareness \tabularnewline
\href{https://www.w3.org/TR/vocab-ssn}{Semantic Sensor Network (SSN)/Sensor, Observation, Sample, and Actuator (SOSA) Ontology} & Describes sensors and their capabilities, measurement processes, observations, and deployments. & \cite{bampi_ontology-driven_2025,chatterjee_automatic_2021,de_brouwer_mbrain_2022,el-sappagh_mobile_2019,elhadj_do-care_2021,garcia-moreno_systematic_2023,henaien_combined_2020,ivascu_activity-aware_2021,martella_semantically_2025,rhayem_semantic-enabled_2021,stavropoulos_detection_2021,titi_ontology-based_2019,zafeiropoulos_evaluating_2024} & Interoperability; Context awareness \tabularnewline
\rowcolor{gray!4}
\href{https://saref.etsi.org/core}{Smart Applications REFerence (SAREF) Ontology} & An ontology to enable semantic interoperability for smart appliances. & \cite{zafeiropoulos_evaluating_2024} & Interoperability; Context awareness \tabularnewline
\href{https://www.etsi.org/technologies/smart-body-area-networks}{Smart Body Area Network (SmartBAN) Ontology} & Describes data related to sensors in BANs. & \cite{el-sappagh_mobile_2019} & Interoperability; Context awareness \tabularnewline
\rowcolor{gray!4}
Stream Annotation Ontology \cite{kolozali_knowledge-based_2014} & An ontology for real-time semantic annotation of streaming IoT data to support dynamic integration into the Web & \cite{garcia-moreno_systematic_2023} & Interoperability; Context awareness \tabularnewline
\href{https://github.com/protegeproject/swrlapi/wiki/SWRLAPITemporal}{SWRL Temporal Ontology} & Defines defines a set of built-ins that can be used to represent temporal information in SWRL rules. & \cite{el-sappagh_mobile_2019} & Interoperability; Context awareness \tabularnewline
\rowcolor{gray!4}
\href{https://bioportal.bioontology.org/ontologies/SNOMEDCT}{Systematized Nomenclature of Medicine Clinical Terms (SNOMED CT)} & A collection of medical terms providing codes, terms, synonyms and definitions used in clinical documentation and reporting. & \cite{bampi_ontology-driven_2025,chatterjee_automatic_2021,el-sappagh_mobile_2019,kilintzis_supporting_2019,kordestani_extended_2021,lopes_ontology-driven_2023,reda_heterogeneous_2022,rhayem_semantic-enabled_2021,titi_ontology-based_2019,zhou_design_2022} & Interoperability \tabularnewline
\href{https://bioportal.bioontology.org/ontologies/TMO}{Translational Medicine Ontology} & & \cite{yu_semantic_2017} & Interoperability; Context awareness \tabularnewline
\rowcolor{gray!4}
\href{https://www.nlm.nih.gov/research/umls/index.html}{Unified Medical Language System (UMLS)} & Provides a mapping structure among different health and biomedical terminologies, classifications, coding standards, and vocabularies. & \cite{peral_ontology-oriented_2018,zhou_design_2022} & Interoperability \tabularnewline
\href{https://bioportal.bioontology.org/ontologies/VSO}{Vital Sign Ontology} & Provides
a controlled structured vocabulary for describing vital sign
measurement data, the processes of measuring vital signs,
and the relevant anatomical entities. & \cite{el-sappagh_mobile_2019,ivascu_activity-aware_2021} & Interoperability \tabularnewline
\rowcolor{gray!4}
\href{https://www.w3.org/wiki/WebAccessControl}{Web Access Control Ontology} & Defines and enforces authorization conditions on Web resources, allowing resource owners to control who can access their data and how. & \cite{ammar_using_2021} & Interoperability \tabularnewline
\href{https://www.w3.org/ns/oa}{Web Annotation Ontology} & Structures annotations (i.e. information about a resource or associations between resources) to enable sharing and reuse across platforms. & \cite{ammar_using_2021} & Interoperability \tabularnewline
\rowcolor{gray!4}
\href{https://wordnet.princeton.edu/}{WordNet} & A lexical English language database of semantic relations between words, linking them into semantic relations. & \cite{peral_ontology-oriented_2018,reda_heterogeneous_2022} & Interoperability \tabularnewline
\end{longtable}
\normalsize

\section{Challenges assessment}
\label{appendix-challenges}

\scriptsize

\begin{longtable}{llllllllll}
\caption{Summary of the extent to which each system addresses the seven key challenges.} \label{challenges-summary-systems}\\
\hline
\textbf{\#} & \textbf{System} & \textbf{Year} & \bfseries\makecell[l]{Interoperability \\(score out of 6)} & \bfseries\makecell[l]{Situation detection \\(score out of 3)} & \bfseries\makecell[l]{Situation prediction \\(score out of 3)} & \bfseries\makecell[l]{Decision support \\(score out of 6)} & \bfseries\makecell[l]{Context awareness \\(score out of 7)} & \bfseries\makecell[l]{Explainability \\(score out of 6)} & \bfseries\makecell[l]{Uncertainty handling \\(score out of 6)} \tabularnewline
\hline
\endfirsthead

\multicolumn{10}{c}{\tablename\ \thetable\ -- continued from previous page} \\
\hline
\textbf{\#} & \textbf{System} & \textbf{Year} & \bfseries\makecell[l]{Interoperability \\(score out of 6)} & \bfseries\makecell[l]{Situation detection \\(score out of 3)} & \bfseries\makecell[l]{Situation prediction \\(score out of 3)} & \bfseries\makecell[l]{Decision support \\(score out of 6)} & \bfseries\makecell[l]{Context awareness \\(score out of 7)} & \bfseries\makecell[l]{Explainability \\(score out of 6)} & \bfseries\makecell[l]{Uncertainty handling \\(score out of 6)} \tabularnewline
\hline
\endhead

\hline
\multicolumn{10}{r}{\textbf{Table continued on next page.}}\\
\endfoot

\hline
\endlastfoot

1 & Akhtar et al. \cite{akhtar_multi-agent_2022} & 2022 & Low (1) & Medium (2) & \xmark (0) & Medium (3) & High (5) & Medium (3) & Low (1) \tabularnewline
\rowcolor{gray!4}2 & Ali et al. \cite{ali_intelligent_2021} & 2021 & Medium (3) & High (3) & \xmark (0) & Low (2) & Medium (4) & Low (2) & Low (1) \tabularnewline
3 & Ali et al. \cite{ali_smart_2020} & 2020 & Low (2) & Medium (2) & Low (1) & Medium (3) & Medium (3) & Low (1) & Medium (2) \tabularnewline
\rowcolor{gray!4}4 & Ali et al. \cite{ali_type-2_2018} & 2018 & Low (1) & High (3) & \xmark (0) & Medium (3) & Medium (3) & Low (2) & Medium (2) \tabularnewline
5 & Alti et al. \cite{alti_agent-based_2021} & 2021 & Low (1) & Medium (2) & Low (1) & Low (1) & High (5) & Low (2) & \xmark (0) \tabularnewline
\rowcolor{gray!4}6 & Ammar et al. \cite{ammar_using_2021} & 2021 & Medium (3) & Low (1) & \xmark (0) & High (5) & High (5) & Low (2) & \xmark (0) \tabularnewline
7 & Bampi et al. \cite{bampi_ontology-driven_2025} & 2025 & Medium (4)& Medium (2)& \xmark (0) & Low (2) & High (5) & Low (2) & \xmark (0) \tabularnewline
\rowcolor{gray!4}8 & Chatterjee et al. \cite{chatterjee_automatic_2021} & 2021 & Medium (4) & Medium (2) & \xmark (0) & Low (2) & High (5) & Medium (3) & \xmark (0) \tabularnewline
9 & Chiang and Liang \cite{chiang_context-aware_2015} & 2015 & Low (1) & Medium (2) & Low (1) & Medium (4) & High (5) & Medium (3) & Low (1) \tabularnewline
\rowcolor{gray!4}10 & De Brouwer et al. \cite{de_brouwer_mbrain_2022} & 2022 & Medium (4) & High (3) & Low (1) & Medium (3) & High (6) & Low (2) & \xmark (0) \tabularnewline
11 & El-Sappagh et al. \cite{el-sappagh_mobile_2019} & 2019 & High (5) & Low (1) & \xmark (0) & Medium (3) & High (5) & Low (2) & \xmark (0) \tabularnewline
\rowcolor{gray!4}12 & Elhadj et al. \cite{elhadj_do-care_2021} & 2021 & High (5) & Medium (2) & \xmark (0) & Low (2) & High (5) & Medium (4) & \xmark (0) \tabularnewline
13 & Esposito et al. \cite{esposito_smart_2018} & 2018 & Low (1) & High (3) & \xmark (0) & Low (2) & Medium (4) & Low (2) & Medium (2) \tabularnewline
\rowcolor{gray!4}14 & Fenza et al. \cite{fenza_hybrid_2012} & 2012 & Low (2) & High (3) & Low (1) & \xmark (0) & Low (2) & Low (1) & Medium (2) \tabularnewline
15 & Garcia-Moreno et al. \cite{garcia-moreno_systematic_2023} & 2023 & Medium (4)& Medium (2)& \xmark (0) & Medium (3) & High (6) & Low (1) & Low (1) \tabularnewline
\rowcolor{gray!4}16 & Garcia-Valverde et al. \cite{garcia-valverde_heart_2014} & 2014 & \xmark (0) & Low (1) & \xmark (0) & Low (2) & Medium (3) & Low (2) & Low (1) \tabularnewline
17 & Hadjadj and Halimi \cite{hadjadj_integration_2021} & 2021 & Medium (3) & Low (1) & \xmark (0) & Medium (3) & High (5) & Low (2) & \xmark (0) \tabularnewline
\rowcolor{gray!4}18 & Henaien et al. \cite{henaien_combined_2020} & 2020 & Medium (4) & Low (1) & \xmark (0) & Medium (3) & Medium (4) & Low (2) & \xmark (0) \tabularnewline
19 & Hooda and Rani \cite{hooda_semantic_2020} & 2020 & Low (1) & Medium (2) & \xmark (0) & Low (2) & Low (2) & Low (2) & Low (1) \tabularnewline
\rowcolor{gray!4}20 & Hristoskova et al. \cite{hristoskova_ontology-driven_2014} & 2014 & Medium (3) & Medium (2) & Low (1) & Medium (3) & High (6) & Low (2) & Low (1) \tabularnewline
21 & Hussain and Park \cite{hussain_big-ecg_2021} & 2021 & Low (2) & Medium (2) & \xmark (0) & Low (2) & Medium (3) & Low (2) & Low (1) \tabularnewline
\rowcolor{gray!4}22 & Ivașcu and Negru \cite{ivascu_activity-aware_2021} & 2021 & Low (1) & Low (1) & \xmark (0) & Low (1) & Medium (4) & Low (2) & \xmark (0) \tabularnewline
23 & Ivașcu et al. \cite{ivascu_multi-agent_2015} & 2015 & Low (1) & Medium (2) & \xmark (0) & Low (1) & High (5) & Low (2) & \xmark (0) \tabularnewline
\rowcolor{gray!4}24 & Khozouie et al. \cite{khozouie_ontological_2018} & 2018 & Low (2) & Low (1) & \xmark (0) & Low (1) & High (5) & Low (2) & \xmark (0) \tabularnewline
25 & Kilintzis et al. \cite{kilintzis_supporting_2019} & 2019 & Medium (3) & Low (1) & \xmark (0) & Low (1) & Medium (4)& Low (1) & Low (1) \tabularnewline
\rowcolor{gray!4}26 & Kim et al. \cite{kim_ontology_2014} & 2014 & Low (1) & High (3) & \xmark (0) & Low (1) & Medium (3) & Low (1) & \xmark (0) \tabularnewline
27 & Kordestani et al. \cite{kordestani_extended_2021} & 2021 & Medium (3) & Medium (2) & \xmark (0) & Medium (4) & Medium (4) & Low (2) & Medium (2) \tabularnewline
\rowcolor{gray!4}28 & Lopes de Souza et al. \cite{lopes_ontology-driven_2023} & 2023 & Medium (3) & High (3) & \xmark (0) & Medium (4) & Medium (4) & Low (2) & \xmark (0) \tabularnewline
29 & Martella et al. \cite{martella_semantically_2025} & 2025 & Medium (3) & Low (1) & \xmark (0) & Low (2) & High (5) & Low (2) & Low (1) \tabularnewline
\rowcolor{gray!4}30 & Mavropoulos et al. \cite{mavropoulos_smart_2021} & 2021 & Low (2) & Medium (2) & \xmark (0) & Medium (4) & High (5) & Medium (4) & \xmark (0) \tabularnewline
31 & Mcheick et al. \cite{mcheick_stroke_2016} & 2016 & \xmark (0) & Low (1) & Low (1) & Low (1) & Medium (3) & Low (2) & Low (1) \tabularnewline
\rowcolor{gray!4}32 & Mezghani et al. \cite{mezghani_semantic_2015} & 2015 & Low (1) & Low (1) & \xmark (0) & Low (1) & Medium (3) & Low (1) & Low (1) \tabularnewline
33 & Minutolo et al. \cite{minutolo_hybrid_2016} & 2016 & Low (1) & High (3) & \xmark (0) & \xmark (0) & Medium (3) & Low (1) & Low (1) \tabularnewline
\rowcolor{gray!4}34 & Peral et al. \cite{peral_ontology-oriented_2018} & 2018 & Medium (4) & Low (1) & Medium (2) & Medium (3) & Medium (3) & Low (1) & \xmark (0) \tabularnewline
35 & Reda et al. \cite{reda_heterogeneous_2022} & 2022 & Low (1) & Medium (2) & Low (1) & \xmark (0) & High (5) & Low (1) & Low (1) \tabularnewline
\rowcolor{gray!4}36 & Rhayem et al. \cite{rhayem_semantic-enabled_2021} & 2021 & Medium (4) & Medium (2) & Low (1) & Medium (3) & High (7) & Medium (3) & Low (1) \tabularnewline
37 & Spoladore et al. \cite{spoladore_ontology-based_2021} & 2021 & Low (2) & Medium (2) & \xmark (0) & Medium (4) & Medium (4) & Low (2) & \xmark (0) \tabularnewline
\rowcolor{gray!4}38 & Stavropoulos et al. \cite{stavropoulos_detection_2021} & 2021 & Low (1) & Medium (2) & \xmark (0) & Low (2) & Medium (4) & Low (2) & \xmark (0) \tabularnewline
39 & Titi et al. \cite{titi_ontology-based_2019} & 2019 & Medium (4) & Medium (2) & \xmark (0) & High (5) & High (6) & Low (2) & Low (1) \tabularnewline
\rowcolor{gray!4}40 & Vadillo et al. \cite{vadillo_enhancement_2013} & 2013 & Low (2) & Low (1) & \xmark (0) & Medium (4) & High (5) & Low (2) & \xmark (0) \tabularnewline
41 & Villarreal et al. \cite{villarreal_mobile_2014} & 2014 & Low (1) & Medium (2) & \xmark (0) & Medium (3) & Medium (4) & Medium (4) & \xmark (0) \tabularnewline
\rowcolor{gray!4}42 & Xu et al. \cite{xu_design_2017} & 2017 & Low (2) & Medium (2) & \xmark (0) & Low (2) & Medium (3) & Low (2) & \xmark (0) \tabularnewline
43 & Yu et al. \cite{yu_improving_2022} & 2022 & Low (2) & Low (1) & \xmark (0) & High (5) & Medium (4) & Medium (4) & \xmark (0) \tabularnewline
\rowcolor{gray!4}44 & Yu et al. \cite{yu_semantic_2017} & 2017 & Medium (3) & Medium (2) & Low (1) & Low (1) & Medium (4) & Low (2) & \xmark (0) \tabularnewline
45 & Zafeiropoulos et al. \cite{zafeiropoulos_evaluating_2024} & 2024 & Medium (3) & High (3) & \xmark (0) & Medium (3) & High (5) & Low (2) & \xmark (0) \tabularnewline
\rowcolor{gray!4}46 & Zeshan et al. \cite{zeshan_iot-enabled_2023} & 2023 & \xmark (0) & Medium (2) & \xmark (0) & Low (1) & High (5) & Low (2) & \xmark (0) \tabularnewline
47 & Zhang et al. \cite{zhang_knowledge-based_2014} & 2014 & Low (2) & Medium (2) & \xmark (0) & Low (1) & Low (2) & Low (2) & \xmark (0) \tabularnewline
\rowcolor{gray!4}48 & Zhou et al. \cite{zhou_design_2022} & 2022 & Medium (4) & Low (1) & Low (1) & Medium (3) & Medium (4) & Low (1) & Low (1) \tabularnewline
\end{longtable}
\normalsize

\section{Development tools, evaluation approaches, and evaluation metrics}

\scriptsize
\begin{longtable}
{>{\raggedright}m{.02\NetTableWidth}>{\raggedright}m{.19\NetTableWidth}>{\raggedright}m{.23\NetTableWidth}>{\raggedright}m{.16\NetTableWidth}>{\raggedright}m{.18\NetTableWidth}>{\raggedright}m{.2\NetTableWidth}>{\raggedright}m{.25\NetTableWidth}}

\caption{Summary of development tools, evaluation approaches, and evaluation metrics used by the systems.} \label{development-evaluation-summary}\\
\hline
\textbf{\#} & \textbf{System} & \textbf{Semantic Web-based languages, standards, reasoners, and development frameworks} & \textbf{Other programming languages, libraries, and frameworks} & \textbf{Database systems, data repositories, and data analysis tools} & \textbf{Evaluation approaches} & \textbf{Evaluation metrics}\tabularnewline
\hline
\endfirsthead

\multicolumn{7}{c}{\tablename\ \thetable\ -- continued from previous page} \\
\hline
\textbf{\#} & \textbf{System} & \textbf{Semantic Web-based languages, standards, reasoners, and development frameworks} & \textbf{Other programming languages, libraries, and frameworks} & \textbf{Database systems, data repositories, and data analysis tools} & \textbf{Evaluation approaches} & \textbf{Evaluation metrics}\tabularnewline
\hline
\endhead

\hline
\multicolumn{6}{>{\raggedright\arraybackslash}p{\NetTableWidth}}{ALMA - Architecture-Level Modifiability Analysis; AUC - Area Under the Curve;  BN - Bayesian Network; CCM - Chronic Care Model; JADE - Java Agent DEvelopment Framework; MAE - Mean Absolute Error; ML - Machine Learning; NFR - non-functional requirements; OWL - Web Ontology Language; RDF - Resource Description Framework; RML - RDF Mapping Language; (R)MSE - (Root) Mean Square Error; SHACL - Shapes Constraint Language; Solid - Social Linked Data; SOTA - State of the Art; SPARQL -  SPARQL Protocol and RDF Query Language; SWRL - Semantic Web Rule Language; WEKA - Waikato Environment for Knowledge Analysis} & \multicolumn{1}{>{\raggedleft\arraybackslash}p{0.25\NetTableWidth}}{\textbf{Table continued on next page.}}\\
\endfoot

\hline
\multicolumn{7}{>{\raggedright\arraybackslash}p{\NetTableWidth}}{ALMA - Architecture-Level Modifiability Analysis; AUC - Area Under the Curve;  BN - Bayesian Network; CCM - Chronic Care Model; JADE - Java Agent DEvelopment Framework; MAE - Mean Absolute Error; ML - Machine Learning; NFR - non-functional requirements; OWL - Web Ontology Language; RDF - Resource Description Framework; RML - RDF Mapping Language; (R)MSE - (Root) Mean Square Error; SHACL - Shapes Constraint Language; Solid - Social Linked Data; SOTA - State of the Art; SPARQL -  SPARQL Protocol and RDF Query Language; SWRL - Semantic Web Rule Language; WEKA - Waikato Environment for Knowledge Analysis}
\endlastfoot

1 & Akhtar et al. \cite{akhtar_multi-agent_2022} & Protégé; SWRL & NetLogo & None mentioned & Case study; System simulation & None mentioned\tabularnewline 
\rowcolor{gray!4}
2 & Ali et al. \cite{ali_intelligent_2021} & OWL; Protégé & Java; WEKA & Amazon S3; Apache Pig; Apache Hadoop & ML model performance; Comparison with SOTA; Ablation study & Accuracy; Precision; Recall; RMSE; MAE\tabularnewline
3 & Ali et al. \cite{ali_smart_2020} & OWL; Protégé; SWRL & Java; WEKA & None mentioned & ML model performance; Comparison with SOTA; Ablation study & Accuracy; Precision; Recall; F-score; RMSE; MAE\tabularnewline
\rowcolor{gray!4}
4 & Ali et al. \cite{ali_type-2_2018} & Apache Jena; OWL; Pellet; Protégé; SPARQL; SWRL & Java & None mentioned & User evaluation with real users; Comparison with SOTA; Expert evaluation of system; Ontology validation; System queries & Accuracy; Precision; Recall; F-score \tabularnewline
5 & Alti et al. \cite{alti_agent-based_2021} & Apache Jena; Protégé; SWRL & JADE; Java; Java Expert System Shell & MySQL & Case study; Comparison with SOTA; NFR
validation & Execution time; Optimality; Application lifetime; No. of discovered services\tabularnewline
\rowcolor{gray!4}
6 & Ammar et al. \cite{ammar_using_2021} & LDflex; OWL; RDF; RDFlib; Solid; SPARQL & JavaScript; Python & None mentioned & Use case scenarios; Comparison with SOTA & Utilization of public health library; incorporation of observations of daily living; incorporation of social determinants of health\tabularnewline
7 & Bampi et al. \cite{bampi_ontology-driven_2025} & Apache Jena; Protégé; RDF; SPARQL; SWRL & None mentioned & None mentioned & Use case scenario; System simulation; NFR
validation & System latency; resource usage (memory \& CPU); Comparison with SOTA\tabularnewline
\rowcolor{gray!4}
8 & Chatterjee et al. \cite{chatterjee_automatic_2021} & Apache Jena; Apache Jena Fuseki; HermiT; OWL; Protégé; RDF; SPARQL; SWRL & None mentioned & None mentioned & Ontology validation; System queries & Ontology reasoning time\tabularnewline
9 & Chiang and Liang \cite{chiang_context-aware_2015} & Apache Jena; Protégé; RDF & C\#; Java; MATLAB & MySQL & Use case scenarios; System simulation & None mentioned\tabularnewline
\rowcolor{gray!4}
10 & De Brouwer et al. \cite{de_brouwer_mbrain_2022} & Apache Jena; Protégé; RDF; SPARQL & None mentioned & MongoDB & System queries; Use case scenarios with real users & None mentioned\tabularnewline
11 & El-Sappagh et al. \cite{el-sappagh_mobile_2019} & HermiT; OWL; Pellet; Protégé; SPARQL; SWRL & SQL & SQLite & Ontology validation; Comparison with SOTA; Comparison with expert opinion; Use case scenarios; NFR validation; System queries & Correctness; Completeness; Extensibility; Conciseness; Organisational fitness\tabularnewline
\rowcolor{gray!4}
12 & Elhadj et al. \cite{elhadj_do-care_2021} & Apache Jena; Apache Jena Fuseki; OWL; Pellet; Protégé; SPARQL; SWRL & Java & MongoDB & Use case scenarios & None mentioned\tabularnewline
13 & Esposito et al. \cite{esposito_smart_2018} & OWL; Pellet; Protégé; SWRL & Java & SQLite & Case study with real users; ALMA method; Ontology validation; NFR validation & Modifiability\tabularnewline
\rowcolor{gray!4}
14 & Fenza et al. \cite{fenza_hybrid_2012} & OWL; SPARQL & JADE; MATLAB; Fuzzy Control Language & None mentioned & Case study; Comparison with logic-based matching & Precision; Recall\tabularnewline
15 & Garcia-Moreno et al. \cite{garcia-moreno_systematic_2023} & None mentioned & None mentioned & None mentioned & ML model performance; NFR validation; User evaluation with real users & Accuracy; F-score; Sensitivity; Specificity; Reusability; Extensibility; Adaptability\tabularnewline
\rowcolor{gray!4}
16 & Garcia-Valverde et al. \cite{garcia-valverde_heart_2014} & Apache Jena & None mentioned & None mentioned & Use case scenario & Accuracy; Precision; F-score\tabularnewline
17 & Hadjadj and Halimi \cite{hadjadj_integration_2021} & Apache Jena; OWL; Protégé; RDF; SPARQL; SWRL & None mentioned & None mentioned & Use case scenario; Comparison with expert opinion; System queries & Similarity between system and expert opinion\tabularnewline
\rowcolor{gray!4}
18 & Henaien et al. \cite{henaien_combined_2020} & Protégé; SWRL & WEKA & None mentioned & None mentioned & None mentioned \tabularnewline
19 & Hooda and Rani \cite{hooda_semantic_2020} & Apache Jena; OWL; Pellet; Protégé; RDF; SPARQL; SWRL & None mentioned & None mentioned & Ontology validation & None mentioned\tabularnewline
\rowcolor{gray!4}
20 & Hristoskova et al. \cite{hristoskova_ontology-driven_2014} & OWL; Pellet; SWRL & None mentioned & None mentioned & Expert evaluation; User evaluation with real users; Ontology validation; NFR validation & Performance; Scalability\tabularnewline
21 & Hussain and Park \cite{hussain_big-ecg_2021} & Protégé; RDF & None mentioned & Apache ActiveMQ; Apache Hadoop; Apache Spark; Elasticsearch; MariaDB & Comparison with SOTA; User evaluation with real users & AUC; Accuracy; Precision; Recall; Neg. predictive value\tabularnewline
\rowcolor{gray!4}
22 & Ivașcu and Negru \cite{ivascu_activity-aware_2021} & Apache Jena; Apache Jena Fuseki; Protégé; SPARQL & JADE; WEKA & None mentioned & Comparison with SOTA; System simulation & Accuracy; Precision; Recall; F-score \tabularnewline
23 & Ivașcu et al. \cite{ivascu_multi-agent_2015} & Apache Jena; Protégé & None mentioned & None mentioned & Use case scenarios & None mentioned\tabularnewline
\rowcolor{gray!4}
24 & Khozouie et al. \cite{khozouie_ontological_2018} & OWL; Pellet; Protégé & Java & None mentioned & Use case scenario; Ontology validation; Expert evaluation & None mentioned\tabularnewline
25 & Kilintzis et al. \cite{kilintzis_supporting_2019} & Apache Jena Fuseki; OWL; RDF; SPARQL; SPIN & None mentioned & Virtuoso & User evaluation with real users; NFR validation; System queries & Robustness; Data integrity\tabularnewline
\rowcolor{gray!4}
26 & Kim et al. \cite{kim_ontology_2014} & Apache Jena; OWL; Protégé & SQL & None mentioned & Ontology validation; User evaluation; System queries with real users & Precision; Recall; F-score; Likert scale\tabularnewline
27 & Kordestani et al. \cite{kordestani_extended_2021} & None mentioned & Drools & None mentioned & Comparison between BN and ML diagnosis; Use case scenarios & F-score\tabularnewline
\rowcolor{gray!4}
28 & Lopes de Souza et al. \cite{lopes_ontology-driven_2023} & RDF; SWRL; SPARQL & GraphQL; Java; Javascript; Node.js; React & SQLite & Ontology evaluation & None mentioned\tabularnewline
29 & Martella et al. \cite{martella_semantically_2025} & Protégé; SPARQL & Java; Python & Apache Spark; MongoDB; PostgreSQL & Case study with simulated users; NFR validation; System simulation & Response time; Throughput; Delay threshold\tabularnewline
\rowcolor{gray!4}
30 & Mavropoulos et al. \cite{mavropoulos_smart_2021} & OWL; SPARQL & None mentioned & GraphDB & Comparison with SOTA; User evaluation with real users & Accuracy; Precision; Recall; F-score; Likert scale\tabularnewline
31 & Mcheick et al. \cite{mcheick_stroke_2016} & None mentioned & Netica & None mentioned & Use case scenarios; NFR validation & Adaptability\tabularnewline
\rowcolor{gray!4}
32 & Mezghani et al. \cite{mezghani_semantic_2015} & Apache Jena Fuseki; RDF; SPARQL; SWRL & SQL & None mentioned & Use case scenario & None mentioned\tabularnewline
33 & Minutolo et al. \cite{minutolo_hybrid_2016} & OWL; SPARQL; SWRL & None mentioned & None mentioned & Case study & None mentioned\tabularnewline
\rowcolor{gray!4}
34 & Peral et al. \cite{peral_ontology-oriented_2018} & None mentioned & None mentioned & None mentioned & Case study & Similarity between actual and predicted values\tabularnewline
35 & Reda et al. \cite{reda_heterogeneous_2022} & OWL; RDF; RML; SPARQL; SWRL & None mentioned & None mentioned & System simulation & Not mentioned\tabularnewline
\rowcolor{gray!4}
36 & Rhayem et al. \cite{rhayem_semantic-enabled_2021} & Apache Jena; OWL; SPARQL; SWRL & Drools & None mentioned & Comparison with SOTA; Ontology validation & F-score; Precision; Recall; Response time; Ontology coverage\tabularnewline
37 & Spoladore et al. \cite{spoladore_ontology-based_2021} & OWL; Protégé; SPARQL; SWRL & None mentioned & Stardog & Use case scenarios & None mentioned\tabularnewline
\rowcolor{gray!4}
38 & Stavropoulos et al. \cite{stavropoulos_detection_2021} & OWL; SHACL; SPARQL & Python; Django & GraphDB & Use case scenarioswith real users; NFR validation; Focus group with clinicians & Scalability; Likert scale\tabularnewline
39 & Titi et al. \cite{titi_ontology-based_2019} & Apache Jena; Pellet; Protégé; RDF; SPARQL; SWRL & Java; Java Server Faces; PrimeFaces & MySQL & System queries & None mentioned\tabularnewline
\rowcolor{gray!4}
40 & Vadillo et al. \cite{vadillo_enhancement_2013} & Apache Jena; OWL; Pellet; Protégé & JADE & None mentioned & Use case scenarios; Ontology validation & Processing time\tabularnewline
41 & Villarreal et al. \cite{villarreal_mobile_2014} & None mentioned & Java & MySQL & Case study; ALMA method; User evaluation with real users; NFR validation & Response time; Usability; Recommendation suitability\tabularnewline
\rowcolor{gray!4}
42 & Xu et al. \cite{xu_design_2017} & SPARQL & None mentioned & None mentioned & Case study; Comparison with SOTA & None mentioned\tabularnewline
43 & Yu et al. \cite{yu_improving_2022} & None mentioned & Python; spaCy; NetworkX & None mentioned & ML model performance; Use case scenarios with real users & AUC; Chronic Care Model criteria\tabularnewline
\rowcolor{gray!4}
44 & Yu et al. \cite{yu_semantic_2017} & Apache Jena; OWL; RDF; SPARQL; SWRL & None mentioned & Virtuoso & System queries & None mentioned\tabularnewline
45 & Zafeiropoulos et al. \cite{zafeiropoulos_evaluating_2024} & OWL; Pellet; Protégé; RDF; RDFlib; SPARQL; SWRL; Owlready2 & Python; PyTorch & None mentioned & Use case scenario; Ontology evaluation; System queries; ML model performance & MAE; MSE; R-Squared \tabularnewline
\rowcolor{gray!4}
46 & Zeshan et al. \cite{zeshan_iot-enabled_2023} & Apache Jena; OWL; Protégé; SWRL & None mentioned & None mentioned & Use case scenario & Precision; Recall; Response time \tabularnewline
47 & Zhang et al. \cite{zhang_knowledge-based_2014} & OWL; Protégé; RDF; SPARQL; SWRL & None mentioned & None mentioned & Use case scenario & None mentioned\tabularnewline
\rowcolor{gray!4}
48 & Zhou et al. \cite{zhou_design_2022} & None mentioned & None mentioned & None mentioned & Use case scenario & None mentioned\tabularnewline
\hline
\end{longtable}
\normalsize

\section{Quality assessment}
\label{appendix-quality}

\scriptsize
\begin{longtable}{lllllll}
\caption{Summary of the quality evaluation of each system.} \label{quality_summary}\\
\hline
\textbf{\#} & \textbf{System} & \textbf{Year} & \bfseries\makecell[l]{Data \& devices (score out of 3)} & \bfseries\makecell[l]{System \& components development \\(score out of 3)} & \bfseries\makecell[l]{Rigour of evaluation (score out of 7)} & \bfseries\makecell[l]{Accessibility of research outputs \\(score out of 2)}\tabularnewline
\hline
\endfirsthead

\multicolumn{7}{c}{\tablename\ \thetable\ -- continued from previous page} \\
\hline
\textbf{\#} & \textbf{System} & \textbf{Year} & \bfseries\makecell[l]{Data \& devices (score out of 3)} & \bfseries\makecell[l]{System \& components development \\(score out of 3)} & \bfseries\makecell[l]{Rigour of evaluation (score out of 7)} & \bfseries\makecell[l]{Accessibility of research outputs \\(score out of 2)}\tabularnewline
\hline
\endhead

\hline
\multicolumn{7}{r}{\textbf{Table continued on next page.}}\\
\endfoot

\hline
\endlastfoot

1 & Akhtar et al. \cite{akhtar_multi-agent_2022} & 2022 & \xmark (0) & Medium (2) & Low (1) & \xmark (0) \tabularnewline
\rowcolor{gray!4}2 & Ali et al. \cite{ali_intelligent_2021} & 2021 & Medium (2) & Medium (2) & Medium (4) & \xmark (0) \tabularnewline
3 & Ali et al. \cite{ali_smart_2020} & 2020 & Medium (2) & Medium (2) & Medium (4) & \xmark (0) \tabularnewline
\rowcolor{gray!4}4 & Ali et al. \cite{ali_type-2_2018} & 2018 & Low (1) & Medium (2) & High (6) & \xmark (0) \tabularnewline
5 & Alti et al. \cite{alti_agent-based_2021} & 2021 & Low (1) & Medium (2) & Low (2) & \xmark (0) \tabularnewline
\rowcolor{gray!4}6 & Ammar et al. \cite{ammar_using_2021} & 2021 & Low (1) & Medium (2)& Medium (3) & \xmark (0) \tabularnewline
7 & Bampi et al. \cite{bampi_ontology-driven_2025} & 2025 & High (3) & Medium (2)& Low (2) & High (2) \tabularnewline
\rowcolor{gray!4}8 & Chatterjee et al. \cite{chatterjee_automatic_2021} & 2021 & Medium (2) & High (3) & Medium (3) & High (2) \tabularnewline
9 & Chiang and Liang \cite{chiang_context-aware_2015} & 2015 & Low (1) & High (3) & Low (1) & \xmark (0) \tabularnewline
\rowcolor{gray!4}10 & De Brouwer et al. \cite{de_brouwer_mbrain_2022} & 2022 & High (3) & Medium (2) & Low (2) & Medium (1) \tabularnewline
11 & El-Sappagh et al. \cite{el-sappagh_mobile_2019} & 2019 & \xmark (0) & High (3) & Medium (4) & Medium (1) \tabularnewline
\rowcolor{gray!4}12 & Elhadj et al. \cite{elhadj_do-care_2021} & 2021 & \xmark (0) & High (3) & Medium (3) & \xmark (0) \tabularnewline
13 & Esposito et al. \cite{esposito_smart_2018} & 2018 & High (3) & Medium (2) & High (5) & \xmark (0) \tabularnewline
\rowcolor{gray!4}14 & Fenza et al. \cite{fenza_hybrid_2012} & 2012 & \xmark (0) & Medium (2) & Low (2) & \xmark (0) \tabularnewline
15 & Garcia-Moreno et al. \cite{garcia-moreno_systematic_2023} & 2023 & High (3) & Low (1) & Medium (4)& \xmark (0) \tabularnewline
\rowcolor{gray!4}16 & Garcia-Valverde et al. \cite{garcia-valverde_heart_2014} & 2014 & Medium (2) & Low (1) & Low (2) & \xmark (0) \tabularnewline
17 & Hadjadj and Halimi \cite{hadjadj_integration_2021} & 2021 & Medium (2) & High (3) & Medium (3) & \xmark (0) \tabularnewline
\rowcolor{gray!4}18 & Henaien et al. \cite{henaien_combined_2020} & 2020 & Low (1) & Medium (2) & \xmark (0) & \xmark (0) \tabularnewline
19 & Hooda and Rani \cite{hooda_semantic_2020} & 2020 & Low (1) & Medium (2) & Low (1) & \xmark (0) \tabularnewline
\rowcolor{gray!4}20 & Hristoskova et al. \cite{hristoskova_ontology-driven_2014} & 2014 & High (3) & Medium (2) & Medium (3) & \xmark (0) \tabularnewline
21 & Hussain and Park \cite{hussain_big-ecg_2021} & 2021 & High (3) & Medium (2) & Medium (4) & \xmark (0) \tabularnewline
\rowcolor{gray!4}22 & Ivașcu and Negru \cite{ivascu_activity-aware_2021} & 2021 & High (3) & High (3) & Medium (3) & \xmark (0) \tabularnewline
23 & Ivașcu et al. \cite{ivascu_multi-agent_2015} & 2015 & \xmark (0) & Medium (2) & \xmark (0) & \xmark (0) \tabularnewline
\rowcolor{gray!4}24 & Khozouie et al. \cite{khozouie_ontological_2018} & 2018 & \xmark (0) & Medium (2) & Low (2) & \xmark (0) \tabularnewline
25 & Kilintzis et al. \cite{kilintzis_supporting_2019} & 2019 & Medium (2)& High (3) & Medium (4)& Medium (1) \tabularnewline
\rowcolor{gray!4}26 & Kim et al. \cite{kim_ontology_2014} & 2014 & Medium (2) & Medium (2) & Medium (4) & \xmark (0) \tabularnewline
27 & Kordestani et al. \cite{kordestani_extended_2021} & 2021 & Medium (2) & Low (1) & Low (2) & \xmark (0) \tabularnewline
\rowcolor{gray!4}28 & Lopes de Souza et al. \cite{lopes_ontology-driven_2023} & 2023 & Low (1) & Medium (2) & Low (1) & \xmark (0) \tabularnewline
29 & Martella et al. \cite{martella_semantically_2025} & 2025 & High (3) & High (3) & Medium (3) & \xmark (0) \tabularnewline
\rowcolor{gray!4}30 & Mavropoulos et al. \cite{mavropoulos_smart_2021} & 2021 & High (3) & Low (1) & High (6) & \xmark (0) \tabularnewline
31 & Mcheick et al. \cite{mcheick_stroke_2016} & 2016 & Low (1) & Low (1) & Low (2) & \xmark (0) \tabularnewline
\rowcolor{gray!4}32 & Mezghani et al. \cite{mezghani_semantic_2015} & 2015 & \xmark (0) & Low (1) & Low (2) & \xmark (0) \tabularnewline
33 & Minutolo et al. \cite{minutolo_hybrid_2016} & 2016 & \xmark (0) & Medium (2) & Medium (3) & \xmark (0) \tabularnewline
\rowcolor{gray!4}34 & Peral et al. \cite{peral_ontology-oriented_2018} & 2018 & Medium (2) & Low (1) & Low (2) & \xmark (0) \tabularnewline
35 & Reda et al. \cite{reda_heterogeneous_2022} & 2022 & Low (1) & Medium (2) & Low (1) & \xmark (0) \tabularnewline
\rowcolor{gray!4}36 & Rhayem et al. \cite{rhayem_semantic-enabled_2021} & 2021 & Medium (2) & Medium (2) & High (5) & \xmark (0) \tabularnewline
37 & Spoladore et al. \cite{spoladore_ontology-based_2021} & 2021 & Low (1) & Medium (2) & Low (2) & \xmark (0) \tabularnewline
\rowcolor{gray!4}38 & Stavropoulos et al. \cite{stavropoulos_detection_2021} & 2021 & High (3) & Medium (2) & Medium (3) & \xmark (0) \tabularnewline
39 & Titi et al. \cite{titi_ontology-based_2019} & 2019 & \xmark (0) & High (3) & Low (1) & \xmark (0) \tabularnewline
\rowcolor{gray!4}40 & Vadillo et al. \cite{vadillo_enhancement_2013} & 2013 & Low (1) & High (3) & Medium (3) & \xmark (0) \tabularnewline
41 & Villarreal et al. \cite{villarreal_mobile_2014} & 2014 & Medium (2) & Low (1) & High (6) & \xmark (0) \tabularnewline
\rowcolor{gray!4}42 & Xu et al. \cite{xu_design_2017} & 2017 & \xmark (0) & Low (1) & Medium (3) & \xmark (0) \tabularnewline
43 & Yu et al. \cite{yu_improving_2022} & 2022 & Medium (2) & Low (1) & High (5) & \xmark (0) \tabularnewline
\rowcolor{gray!4}44 & Yu et al. \cite{yu_semantic_2017} & 2017 & Low (1) & High (3) & Low (1) & \xmark (0) \tabularnewline
45 & Zafeiropoulos et al. \cite{zafeiropoulos_evaluating_2024} & 2024 & Medium (2) & High (3) & Medium (3) & High (2) \tabularnewline
\rowcolor{gray!4}46 & Zeshan et al. \cite{zeshan_iot-enabled_2023} & 2023 & Low (1) & Medium (2) & Medium (3) & \xmark (0) \tabularnewline
47 & Zhang et al. \cite{zhang_knowledge-based_2014} & 2014 & Low (1) & Medium (2) & Low (2) & \xmark (0) \tabularnewline
\rowcolor{gray!4}48 & Zhou et al. \cite{zhou_design_2022} & 2022 & \xmark (0) & \xmark (0) & Low (1) & \xmark (0) \tabularnewline
\end{longtable}
\normalsize

\end{landscape}

\end{appendix}

\bibliographystyle{ios1}           
\bibliography{bibliography}        

\begin{thebibliography}{244}
\ifx \bisbn   \undefined \def \bisbn  #1{ISBN #1}\fi
\ifx \binits  \undefined \def \binits#1{#1} \fi
\ifx \bauthor  \undefined \def \bauthor#1{#1} \fi
\ifx \bjtitle  \undefined \def \bjtitle#1{\textit{#1}}\fi
\ifx \batitle  \undefined \def \batitle#1{#1} \fi
\ifx \bctitle  \undefined \def \bctitle#1{#1} \fi
\ifx \bvolume  \undefined \def \bvolume#1{\textbf{#1}}\fi
\ifx \byear  \undefined \def \byear#1{#1} \fi
\ifx \bissue  \undefined \def \bissue#1{#1} \fi
\ifx \bfpage  \undefined \def \bfpage#1{#1} \fi
\ifx \blpage  \undefined \def \blpage #1{#1} \fi
\ifx \burl  \undefined \def \burl#1{#1} \fi
\ifx \doiurl  \undefined \def \doiurl#1{#1} \fi
\ifx \betal  \undefined \def \betal{et al.} \fi
\ifx \binstitute  \undefined \def \binstitute#1{#1} \fi
\ifx \beditor  \undefined \def \beditor#1{#1} \fi
\ifx \bpublisher  \undefined \def \bpublisher#1{#1} \fi
\ifx \bbtitle  \undefined \def \bbtitle#1{\textit{#1}} \fi
\ifx \bedition  \undefined \def \bedition#1{#1} \fi
\ifx \bseriesno  \undefined \def \bseriesno#1{#1} \fi
\ifx \blocation  \undefined \def \blocation#1{#1} \fi
\ifx \bsertitle  \undefined \def \bsertitle#1{#1} \fi
\ifx \bsnm \undefined \def \bsnm#1{#1} \fi
\ifx \bsuffix \undefined \def \bsuffix#1{#1} \fi
\ifx \bparticle \undefined \def \bparticle#1{#1} \fi
\ifx \barticle \undefined \def \barticle#1{#1} \fi
\ifx \botherref \undefined \def \botherref #1{#1} \fi
\ifx \url \undefined \def \url#1{#1} \fi
\ifx \bchapter \undefined \def \bchapter#1{#1} \fi
\ifx \bbook \undefined \def \bbook#1{#1} \fi
\ifx \bcomment \undefined \def \bcomment#1{#1} \fi
\ifx \oauthor \undefined \def \oauthor#1{#1} \fi
\ifx \citeauthoryear \undefined \def \citeauthoryear#1{#1} \fi
\ifx \texttildelow  \undefined \def \texttildelow{\symbol{126}} \fi
\def \endbibitem {}
\ifx \bconflocation  \undefined \def \bconflocation#1{#1} \fi

\bibitem{murphy_household_2020}
\begin{barticle}
\bauthor{\binits{A.}~\bsnm{Murphy}},
\bauthor{\binits{B.}~\bsnm{Palafox}},
\bauthor{\binits{M.}~\bsnm{Walli-Attaei}},
\bauthor{\binits{T.}~\bsnm{Powell-Jackson}},
\bauthor{\binits{S.}~\bsnm{Rangarajan}},
\bauthor{\binits{K.F.}~\bsnm{Alhabib}},
\bauthor{\binits{A.J.}~\bsnm{Avezum}},
\bauthor{\binits{K.B.T.}~\bsnm{Calik}},
\bauthor{\binits{J.}~\bsnm{Chifamba}},
\bauthor{\binits{T.}~\bsnm{Choudhury}},
\bauthor{\binits{G.}~\bsnm{Dagenais}},
\bauthor{\binits{A.L.}~\bsnm{Dans}},
\bauthor{\binits{R.}~\bsnm{Gupta}},
\bauthor{\binits{R.}~\bsnm{Iqbal}},
\bauthor{\binits{M.}~\bsnm{Kaur}},
\bauthor{\binits{R.}~\bsnm{Kelishadi}},
\bauthor{\binits{R.}~\bsnm{Khatib}},
\bauthor{\binits{I.M.}~\bsnm{Kruger}},
\bauthor{\binits{V.R.}~\bsnm{Kutty}},
\bauthor{\binits{S.A.}~\bsnm{Lear}},
\bauthor{\binits{W.}~\bsnm{Li}},
\bauthor{\binits{P.}~\bsnm{Lopez-Jaramillo}},
\bauthor{\binits{V.}~\bsnm{Mohan}},
\bauthor{\binits{P.K.}~\bsnm{Mony}},
\bauthor{\binits{A.}~\bsnm{Orlandini}},
\bauthor{\binits{A.}~\bsnm{Rosengren}},
\bauthor{\binits{I.}~\bsnm{Rosnah}},
\bauthor{\binits{P.}~\bsnm{Seron}},
\bauthor{\binits{K.}~\bsnm{Teo}},
\bauthor{\binits{L.A.}~\bsnm{Tse}},
\bauthor{\binits{L.}~\bsnm{Tsolekile}},
\bauthor{\binits{Y.}~\bsnm{Wang}},
\bauthor{\binits{A.}~\bsnm{Wielgosz}},
\bauthor{\binits{R.}~\bsnm{Yan}},
\bauthor{\binits{K.E.}~\bsnm{Yeates}},
\bauthor{\binits{K.}~\bsnm{Yusoff}},
\bauthor{\binits{K.}~\bsnm{Zatonska}},
\bauthor{\binits{K.}~\bsnm{Hanson}},
\bauthor{\binits{S.}~\bsnm{Yusuf}} and
\bauthor{\binits{M.}~\bsnm{McKee}},
\batitle{The household economic burden of non-communicable diseases in 18 countries},
\bjtitle{BMJ Global Health}
\bvolume{5}(\bissue{2})
(\byear{2020}),
\bfpage{e002040}.
doi:\doiurl{10.1136/BMJGH-2019-002040}.
\end{barticle}
\endbibitem

\bibitem{gambhir_toward_2018}
\begin{botherref}
\oauthor{\binits{S.S.}~\bsnm{Gambhir}},
\oauthor{\binits{T.J.}~\bsnm{Ge}},
\oauthor{\binits{O.}~\bsnm{Vermesh}} and
\oauthor{\binits{R.}~\bsnm{Spitler}},
Toward achieving precision health,
\textit{Science Translational Medicine}
\textbf{10}(430)
(2018).
doi:\doiurl{10.1126/scitranslmed.aao3612}.
\end{botherref}
\endbibitem

\bibitem{de_brouwer_context-aware_2023}
\begin{barticle}
\bauthor{\binits{M.}~\bsnm{De~Brouwer}},
\bauthor{\binits{B.}~\bsnm{Steenwinckel}},
\bauthor{\binits{Z.}~\bsnm{Fang}},
\bauthor{\binits{M.}~\bsnm{Stojchevska}},
\bauthor{\binits{P.}~\bsnm{Bonte}},
\bauthor{\binits{F.}~\bsnm{De~Turck}},
\bauthor{\binits{S.}~\bsnm{Van~Hoecke}} and
\bauthor{\binits{F.}~\bsnm{Ongenae}},
\batitle{Context-aware query derivation for {IoT} data streams with {DIVIDE} enabling privacy by design},
\bjtitle{Semantic Web}
\bvolume{14}(\bissue{5})
(\byear{2023}),
\bfpage{893}--\blpage{941}.
doi:\doiurl{10.3233/SW-223281}.
\end{barticle}
\endbibitem

\bibitem{ye_situation_2011}
\begin{barticle}
\bauthor{\binits{J.}~\bsnm{Ye}},
\bauthor{\binits{S.}~\bsnm{Dobson}} and
\bauthor{\binits{S.}~\bsnm{McKeever}},
\batitle{Situation identification techniques in pervasive computing: {A} review},
\bjtitle{Pervasive and Mobile Computing}
\bvolume{8}(\bissue{1})
(\byear{2011}),
\bfpage{36}--\blpage{66}.
doi:\doiurl{10.1016/j.pmcj.2011.01.004}.
\end{barticle}
\endbibitem

\bibitem{adeleke_integrating_2017}
\begin{barticle}
\bauthor{\binits{J.A.}~\bsnm{Adeleke}},
\bauthor{\binits{D.}~\bsnm{Moodley}},
\bauthor{\binits{G.}~\bsnm{Rens}} and
\bauthor{\binits{A.O.}~\bsnm{Adewumi}},
\batitle{Integrating statistical machine learning in a semantic sensor web for proactive monitoring and control},
\bjtitle{Sensors}
\bvolume{17}(\bissue{4})
(\byear{2017}),
\bfpage{807}.
doi:\doiurl{10.3390/s17040807}.
\end{barticle}
\endbibitem

\bibitem{sutton_overview_2020}
\begin{barticle}
\bauthor{\binits{R.T.}~\bsnm{Sutton}},
\bauthor{\binits{D.}~\bsnm{Pincock}},
\bauthor{\binits{D.C.}~\bsnm{Baumgart}},
\bauthor{\binits{D.C.}~\bsnm{Sadowski}},
\bauthor{\binits{R.N.}~\bsnm{Fedorak}} and
\bauthor{\binits{K.I.}~\bsnm{Kroeker}},
\batitle{An overview of clinical decision support systems: benefits, risks, and strategies for success},
\bjtitle{npj Digital Medicine}
\bvolume{3}(\bissue{1})
(\byear{2020}),
\bfpage{1}--\blpage{10}.
doi:\doiurl{10.1038/s41746-020-0221-y}.
\end{barticle}
\endbibitem

\bibitem{dullabh_challenges_2022}
\begin{barticle}
\bauthor{\binits{P.}~\bsnm{Dullabh}},
\bauthor{\binits{S.F.}~\bsnm{Sandberg}},
\bauthor{\binits{K.}~\bsnm{Heaney-Huls}},
\bauthor{\binits{L.S.}~\bsnm{Hovey}},
\bauthor{\binits{D.F.}~\bsnm{Lobach}},
\bauthor{\binits{A.}~\bsnm{Boxwala}},
\bauthor{\binits{P.J.}~\bsnm{Desai}},
\bauthor{\binits{E.}~\bsnm{Berliner}},
\bauthor{\binits{C.}~\bsnm{Dymek}},
\bauthor{\binits{M.I.}~\bsnm{Harrison}},
\bauthor{\binits{J.}~\bsnm{Swiger}} and
\bauthor{\binits{D.F.}~\bsnm{Sittig}},
\batitle{Challenges and opportunities for advancing patient-centered clinical decision support: findings from a horizon scan},
\bjtitle{Journal of the American Medical Informatics Association}
\bvolume{29}(\bissue{7})
(\byear{2022}),
\bfpage{1233}--\blpage{1243}.
doi:\doiurl{10.1093/jamia/ocac059}.
\end{barticle}
\endbibitem

\bibitem{dey_towards_2000}
\begin{bchapter}
\bauthor{\binits{A.K.}~\bsnm{Dey}} and
\bauthor{\binits{G.D.}~\bsnm{Abowd}},
\bctitle{Towards a {Better} {Understanding} of {Context} and {Context}-{Awareness}},
in: \bbtitle{{CHI} {Workshop} on the {What}, {Who}, {Where}, {When}, {Why} and {How} of {Context}-{Awareness}},
\byear{2000}.
\end{bchapter}
\endbibitem

\bibitem{haque_semantic_2022}
\begin{barticle}
\bauthor{\binits{A.K.M.B.}~\bsnm{Haque}},
\bauthor{\binits{B.M.}~\bsnm{Arifuzzaman}},
\bauthor{\binits{S.A.N.}~\bsnm{Siddik}},
\bauthor{\binits{A.}~\bsnm{Kalam}},
\bauthor{\binits{T.S.}~\bsnm{Shahjahan}},
\bauthor{\binits{T.S.}~\bsnm{Saleena}},
\bauthor{\binits{M.}~\bsnm{Alam}},
\bauthor{\binits{M.R.}~\bsnm{Islam}},
\bauthor{\binits{F.}~\bsnm{Ahmmed}} and
\bauthor{\binits{M.J.}~\bsnm{Hossain}},
\batitle{Semantic {Web} in {Healthcare}: {A} {Systematic} {Literature} {Review} of {Application}, {Research} {Gap}, and {Future} {Research} {Avenues}},
\bjtitle{International Journal of Clinical Practice}
\bvolume{2022}
(\byear{2022}),
\bfpage{e6807484}.
doi:\doiurl{10.1155/2022/6807484}.
\end{barticle}
\endbibitem

\bibitem{kaplan_decision_2005}
\begin{barticle}
\bauthor{\binits{R.M.}~\bsnm{Kaplan}} and
\bauthor{\binits{D.L.}~\bsnm{Frosch}},
\batitle{Decision {Making} in {Medicine} and {Health} {Care}},
\bjtitle{Annual Review of Clinical Psychology}
\bvolume{1}
(\byear{2005}),
\bfpage{525}--\blpage{556}.
doi:\doiurl{10.1146/annurev.clinpsy.1.102803.144118}.
\end{barticle}
\endbibitem

\bibitem{peng_literature_2020}
\begin{barticle}
\bauthor{\binits{C.}~\bsnm{Peng}},
\bauthor{\binits{P.}~\bsnm{Goswami}} and
\bauthor{\binits{G.}~\bsnm{Bai}},
\batitle{A literature review of current technologies on health data integration for patient-centered health management},
\bjtitle{Health Informatics Journal}
\bvolume{26}(\bissue{3})
(\byear{2020}),
\bfpage{1926}--\blpage{1951}.
doi:\doiurl{10.1177/1460458219892387}.
\end{barticle}
\endbibitem

\bibitem{hammad_semantic-based_2020}
\begin{barticle}
\bauthor{\binits{R.}~\bsnm{Hammad}},
\bauthor{\binits{M.}~\bsnm{Barhoush}} and
\bauthor{\binits{B.H.}~\bsnm{Abed-alguni}},
\batitle{A {Semantic}-{Based} {Approach} for {Managing} {Healthcare} {Big} {Data}: {A} {Survey}},
\bjtitle{Journal of Healthcare Engineering}
\bvolume{2020}
(\byear{2020}),
\bfpage{e8865808}.
doi:\doiurl{10.1155/2020/8865808}.
\end{barticle}
\endbibitem

\bibitem{jing_ontologies_2023}
\begin{barticle}
\bauthor{\binits{X.}~\bsnm{Jing}},
\bauthor{\binits{H.}~\bsnm{Min}},
\bauthor{\binits{Y.}~\bsnm{Gong}},
\bauthor{\binits{P.}~\bsnm{Biondich}},
\bauthor{\binits{D.}~\bsnm{Robinson}},
\bauthor{\binits{T.}~\bsnm{Law}},
\bauthor{\binits{C.}~\bsnm{Nohr}},
\bauthor{\binits{A.}~\bsnm{Faxvaag}},
\bauthor{\binits{L.}~\bsnm{Rennert}},
\bauthor{\binits{N.}~\bsnm{Hubig}} and
\bauthor{\binits{R.}~\bsnm{Gimbel}},
\batitle{Ontologies {Applied} in {Clinical} {Decision} {Support} {System} {Rules}: {Systematic} {Review}},
\bjtitle{JMIR Medical Informatics}
\bvolume{11}
(\byear{2023}),
\bfpage{e43053}.
doi:\doiurl{10.2196/43053}.
\end{barticle}
\endbibitem

\bibitem{cui_review_2025}
\begin{barticle}
\bauthor{\binits{H.}~\bsnm{Cui}},
\bauthor{\binits{J.}~\bsnm{Lu}},
\bauthor{\binits{R.}~\bsnm{Xu}},
\bauthor{\binits{S.}~\bsnm{Wang}},
\bauthor{\binits{W.}~\bsnm{Ma}},
\bauthor{\binits{Y.}~\bsnm{Yu}},
\bauthor{\binits{S.}~\bsnm{Yu}},
\bauthor{\binits{X.}~\bsnm{Kan}},
\bauthor{\binits{C.}~\bsnm{Ling}},
\bauthor{\binits{L.}~\bsnm{Zhao}},
\bauthor{\binits{Z.S.}~\bsnm{Qin}},
\bauthor{\binits{J.C.}~\bsnm{Ho}},
\bauthor{\binits{T.}~\bsnm{Fu}},
\bauthor{\binits{J.}~\bsnm{Ma}},
\bauthor{\binits{M.}~\bsnm{Huai}},
\bauthor{\binits{F.}~\bsnm{Wang}} and
\bauthor{\binits{C.}~\bsnm{Yang}},
\batitle{A review on knowledge graphs for healthcare: {Resources}, applications, and promises},
\bjtitle{Journal of Biomedical Informatics}
\bvolume{169}
(\byear{2025}),
\bfpage{104861}.
doi:\doiurl{10.1016/j.jbi.2025.104861}.
\end{barticle}
\endbibitem

\bibitem{khan_landscaping_2019}
\begin{barticle}
\bauthor{\binits{M.U.}~\bsnm{Khan}},
\bauthor{\binits{S.}~\bsnm{Sherin}},
\bauthor{\binits{M.Z.}~\bsnm{Iqbal}} and
\bauthor{\binits{R.}~\bsnm{Zahid}},
\batitle{Landscaping systematic mapping studies in software engineering: {A} tertiary study},
\bjtitle{Journal of Systems and Software}
\bvolume{149}
(\byear{2019}),
\bfpage{396}--\blpage{436}.
doi:\doiurl{10.1016/j.jss.2018.12.018}.
\end{barticle}
\endbibitem

\bibitem{petersen_guidelines_2015}
\begin{barticle}
\bauthor{\binits{K.}~\bsnm{Petersen}},
\bauthor{\binits{S.}~\bsnm{Vakkalanka}} and
\bauthor{\binits{L.}~\bsnm{Kuzniarz}},
\batitle{Guidelines for conducting systematic mapping studies in software engineering: {An} update},
\bjtitle{Information and Software Technology}
\bvolume{64}
(\byear{2015}),
\bfpage{1}--\blpage{18}.
doi:\doiurl{10.1016/j.infsof.2015.03.007}.
\end{barticle}
\endbibitem

\bibitem{budgen_using_2008}
\begin{bchapter}
\bauthor{\binits{D.}~\bsnm{Budgen}},
\bauthor{\binits{M.}~\bsnm{Turner}},
\bauthor{\binits{P.}~\bsnm{Brereton}} and
\bauthor{\binits{B.}~\bsnm{Kitchenham}},
\bctitle{Using {Mapping} {Studies} in {Software} {Engineering}},
in: \bbtitle{Proceedings of the 20th {Psychology} of {Programming} {Interest} {Group} ({PPIG}) workshop},
Vol.~\bseriesno{8},
\byear{2008},
pp.~\bfpage{195}--\blpage{204}.
\end{bchapter}
\endbibitem

\bibitem{gravina_wearable_2021}
\begin{barticle}
\bauthor{\binits{R.}~\bsnm{Gravina}} and
\bauthor{\binits{G.}~\bsnm{Fortino}},
\batitle{Wearable {Body} {Sensor} {Networks}: {State}-of-the-{Art} and {Research} {Directions}},
\bjtitle{IEEE Sensors Journal}
\bvolume{21}(\bissue{11})
(\byear{2021}),
\bfpage{12511}--\blpage{12522}.
doi:\doiurl{10.1109/JSEN.2020.3044447}.
\end{barticle}
\endbibitem

\bibitem{dias_wearable_2018}
\begin{barticle}
\bauthor{\binits{D.}~\bsnm{Dias}} and
\bauthor{\binits{J.P.S.}~\bsnm{Cunha}},
\batitle{Wearable {Health} {Devices}—{Vital} {Sign} {Monitoring}, {Systems} and {Technologies}},
\bjtitle{Sensors}
\bvolume{18}(\bissue{8})
(\byear{2018}),
\bfpage{2414}.
doi:\doiurl{10.3390/s18082414}.
\end{barticle}
\endbibitem

\bibitem{escabi_biosignal_2012}
\begin{bchapter}
\bauthor{\binits{M.}~\bsnm{Escabí}},
\bctitle{Biosignal {Processing}},
in: \bbtitle{Introduction to {Biomedical} {Engineering}},
\bpublisher{Elsevier Inc.},
\byear{2012},
pp.~\bfpage{667}--\blpage{746}.
ISBN \bisbn{978-0-12-374979-6}.
doi:\doiurl{10.1016/B978-0-12-374979-6.00011-3}.
\end{bchapter}
\endbibitem

\bibitem{ferlini_in-ear_2022}
\begin{barticle}
\bauthor{\binits{A.}~\bsnm{Ferlini}},
\bauthor{\binits{A.}~\bsnm{Montanari}},
\bauthor{\binits{C.}~\bsnm{Min}},
\bauthor{\binits{H.}~\bsnm{Li}},
\bauthor{\binits{U.}~\bsnm{Sassi}} and
\bauthor{\binits{F.}~\bsnm{Kawsar}},
\batitle{In-{Ear} {PPG} for {Vital} {Signs}},
\bjtitle{IEEE Pervasive Computing}
\bvolume{21}(\bissue{1})
(\byear{2022}),
\bfpage{65}--\blpage{74}.
doi:\doiurl{10.1109/MPRV.2021.3121171}.
\end{barticle}
\endbibitem

\bibitem{choudhury_earable_2021}
\begin{bchapter}
\bauthor{\binits{R.R.}~\bsnm{Choudhury}},
\bctitle{Earable {Computing}: {A} {New} {Area} to {Think} {About}},
in: \bbtitle{Proceedings of the 22nd {International} {Workshop} on {Mobile} {Computing} {Systems} and {Applications}},
\bpublisher{ACM},
\blocation{Virtual United Kingdom},
\byear{2021},
pp.~\bfpage{147}--\blpage{153}.
ISBN \bisbn{978-1-4503-8323-3}.
doi:\doiurl{10.1145/3446382.3450216}.
\end{bchapter}
\endbibitem

\bibitem{andreu-perez_wearable_2015}
\begin{barticle}
\bauthor{\binits{J.}~\bsnm{Andreu-Perez}},
\bauthor{\binits{D.R.}~\bsnm{Leff}},
\bauthor{\binits{H.M.D.}~\bsnm{Ip}} and
\bauthor{\binits{G.Z.}~\bsnm{Yang}},
\batitle{From {Wearable} {Sensors} to {Smart} {Implants}-{Toward} {Pervasive} and {Personalized} {Healthcare}},
\bjtitle{IEEE Transactions on Biomedical Engineering}
\bvolume{62}(\bissue{12})
(\byear{2015}),
\bfpage{2750}--\blpage{2762}.
doi:\doiurl{10.1109/TBME.2015.2422751}.
\end{barticle}
\endbibitem

\bibitem{straczkiewicz_systematic_2021}
\begin{barticle}
\bauthor{\binits{M.}~\bsnm{Straczkiewicz}},
\bauthor{\binits{P.}~\bsnm{James}} and
\bauthor{\binits{J.-P.}~\bsnm{Onnela}},
\batitle{A systematic review of smartphone-based human activity recognition methods for health research},
\bjtitle{npj Digital Medicine}
\bvolume{4}(\bissue{1})
(\byear{2021}),
\bfpage{1}--\blpage{15}.
doi:\doiurl{10.1038/s41746-021-00514-4}.
\end{barticle}
\endbibitem

\bibitem{xiao_survey_2022}
\begin{barticle}
\bauthor{\binits{J.}~\bsnm{Xiao}},
\bauthor{\binits{H.}~\bsnm{Li}},
\bauthor{\binits{M.}~\bsnm{Wu}},
\bauthor{\binits{H.}~\bsnm{Jin}},
\bauthor{\binits{M.J.}~\bsnm{Deen}} and
\bauthor{\binits{J.}~\bsnm{Cao}},
\batitle{A {Survey} on {Wireless} {Device}-free {Human} {Sensing}: {Application} {Scenarios}, {Current} {Solutions}, and {Open} {Issues}},
\bjtitle{ACM Comput. Surv.}
\bvolume{55}(\bissue{5})
(\byear{2022}),
\bfpage{88:1}--\blpage{88:35}.
doi:\doiurl{10.1145/3530682}.
\end{barticle}
\endbibitem

\bibitem{kumar_cnn-based_2022}
\begin{barticle}
\bauthor{\binits{A.}~\bsnm{Kumar}},
\bauthor{\binits{S.}~\bsnm{Singh}},
\bauthor{\binits{V.}~\bsnm{Rawal}},
\bauthor{\binits{S.}~\bsnm{Garg}},
\bauthor{\binits{A.}~\bsnm{Agrawal}} and
\bauthor{\binits{S.}~\bsnm{Yadav}},
\batitle{{CNN}-based device-free health monitoring and prediction system using {WiFi} signals},
\bjtitle{International Journal of Information Technology}
\bvolume{14}(\bissue{7})
(\byear{2022}),
\bfpage{3725}--\blpage{3737}.
doi:\doiurl{10.1007/s41870-022-01023-7}.
\end{barticle}
\endbibitem

\bibitem{tian_rf-based_2018}
\begin{barticle}
\bauthor{\binits{Y.}~\bsnm{Tian}},
\bauthor{\binits{G.-H.}~\bsnm{Lee}},
\bauthor{\binits{H.}~\bsnm{He}},
\bauthor{\binits{C.-Y.}~\bsnm{Hsu}} and
\bauthor{\binits{D.}~\bsnm{Katabi}},
\batitle{{RF}-{Based} {Fall} {Monitoring} {Using} {Convolutional} {Neural} {Networks}},
\bjtitle{Proc. ACM Interact. Mob. Wearable Ubiquitous Technol.}
\bvolume{2}(\bissue{3})
(\byear{2018}),
\bfpage{137:1}--\blpage{137:24}.
doi:\doiurl{10.1145/3264947}.
\end{barticle}
\endbibitem

\bibitem{cusack_reviewsmart_2024}
\begin{barticle}
\bauthor{\binits{N.M.}~\bsnm{Cusack}},
\bauthor{\binits{P.D.}~\bsnm{Venkatraman}},
\bauthor{\binits{U.}~\bsnm{Raza}} and
\bauthor{\binits{A.}~\bsnm{Faisal}},
\batitle{Review—{Smart} {Wearable} {Sensors} for {Health} and {Lifestyle} {Monitoring}: {Commercial} and {Emerging} {Solutions}},
\bjtitle{ECS Sensors Plus}
\bvolume{3}(\bissue{1})
(\byear{2024}),
\bfpage{017001}.
doi:\doiurl{10.1149/2754-2726/ad3561}.
\end{barticle}
\endbibitem

\bibitem{compton_ssn_2012}
\begin{barticle}
\bauthor{\binits{M.}~\bsnm{Compton}},
\bauthor{\binits{P.}~\bsnm{Barnaghi}},
\bauthor{\binits{L.}~\bsnm{Bermudez}},
\bauthor{\binits{R.}~\bsnm{García-Castro}},
\bauthor{\binits{O.}~\bsnm{Corcho}},
\bauthor{\binits{S.}~\bsnm{Cox}},
\bauthor{\binits{J.}~\bsnm{Graybeal}},
\bauthor{\binits{M.}~\bsnm{Hauswirth}},
\bauthor{\binits{C.}~\bsnm{Henson}},
\bauthor{\binits{A.}~\bsnm{Herzog}},
\bauthor{\binits{V.}~\bsnm{Huang}},
\bauthor{\binits{K.}~\bsnm{Janowicz}},
\bauthor{\binits{W.D.}~\bsnm{Kelsey}},
\bauthor{\binits{D.}~\bsnm{Le~Phuoc}},
\bauthor{\binits{L.}~\bsnm{Lefort}},
\bauthor{\binits{M.}~\bsnm{Leggieri}},
\bauthor{\binits{H.}~\bsnm{Neuhaus}},
\bauthor{\binits{A.}~\bsnm{Nikolov}},
\bauthor{\binits{K.}~\bsnm{Page}},
\bauthor{\binits{A.}~\bsnm{Passant}},
\bauthor{\binits{A.}~\bsnm{Sheth}} and
\bauthor{\binits{K.}~\bsnm{Taylor}},
\batitle{The {SSN} ontology of the {W3C} semantic sensor network incubator group},
\bjtitle{Journal of Web Semantics}
\bvolume{17}
(\byear{2012}),
\bfpage{25}--\blpage{32}.
doi:\doiurl{10.1016/J.WEBSEM.2012.05.003}.
\end{barticle}
\endbibitem

\bibitem{khaleghi_multisensor_2011}
\begin{barticle}
\bauthor{\binits{B.}~\bsnm{Khaleghi}},
\bauthor{\binits{A.}~\bsnm{Khamis}} and
\bauthor{\binits{F.O.}~\bsnm{Karray}},
\batitle{Multisensor data fusion: {A} review of the state-of-the-art},
\bjtitle{Information Fusion}
\bvolume{14}(\bissue{1})
(\byear{2011}),
\bfpage{28}--\blpage{44}.
doi:\doiurl{10.1016/j.inffus.2011.08.001}.
\end{barticle}
\endbibitem

\bibitem{gravina_multi-sensor_2017}
\begin{barticle}
\bauthor{\binits{R.}~\bsnm{Gravina}},
\bauthor{\binits{P.}~\bsnm{Alinia}},
\bauthor{\binits{H.}~\bsnm{Ghasemzadeh}} and
\bauthor{\binits{G.}~\bsnm{Fortino}},
\batitle{Multi-sensor fusion in body sensor networks: {State}-of-the-art and research challenges},
\bjtitle{Information Fusion}
\bvolume{35}
(\byear{2017}),
\bfpage{68}--\blpage{80}.
doi:\doiurl{10.1016/j.inffus.2016.09.005}.
\end{barticle}
\endbibitem

\bibitem{hitzler_semantic_2021}
\begin{barticle}
\bauthor{\binits{P.}~\bsnm{Hitzler}},
\batitle{Semantic {Web}: {A} {Review} {Of} {The} {Field}},
\bjtitle{Communications of the ACM}
\bvolume{64}(\bissue{2})
(\byear{2021}),
\bfpage{76}--\blpage{83}.
doi:\doiurl{10.1145/3397512}.
\end{barticle}
\endbibitem

\bibitem{schreiber_rdf_2014}
\begin{botherref}
\oauthor{\binits{G.}~\bsnm{Schreiber}} and
\oauthor{\binits{Y.}~\bsnm{Raimond}},
{RDF} 1.1 {Primer},
W3C,
2014.
\url{https://www.w3.org/TR/rdf-primer/}.
\end{botherref}
\endbibitem

\bibitem{garcia_etsi_2023}
\begin{bchapter}
\bauthor{\binits{R.}~\bsnm{García‐Castro}},
\bauthor{\binits{M.}~\bsnm{Lefrançois}},
\bauthor{\binits{M.}~\bsnm{Poveda‐Villalón}} and
\bauthor{\binits{L.}~\bsnm{Daniele}},
\bctitle{The {ETSI} {SAREF} {Ontology} for {Smart} {Applications}: {A} {Long} {Path} of {Development} and {Evolution}},
in: \bbtitle{Energy {Smart} {Appliances}},
\beditor{\binits{A.}~\bsnm{Moreno‐Munoz}} and
\beditor{\binits{N.}~\bsnm{Giacomini}}, eds,
\bpublisher{Wiley},
\byear{2023},
pp.~\bfpage{183}--\blpage{215}.
ISBN \bisbn{978-1-119-89942-6 978-1-119-89945-7}.
doi:\doiurl{10.1002/9781119899457.ch7}.
\end{bchapter}
\endbibitem

\bibitem{janowicz_sosa_2019}
\begin{barticle}
\bauthor{\binits{K.}~\bsnm{Janowicz}},
\bauthor{\binits{A.}~\bsnm{Haller}},
\bauthor{\binits{S.J.D.}~\bsnm{Cox}},
\bauthor{\binits{D.}~\bsnm{Le~Phuoc}} and
\bauthor{\binits{M.}~\bsnm{Lefrançois}},
\batitle{{SOSA}: {A} lightweight ontology for sensors, observations, samples, and actuators},
\bjtitle{Journal of Web Semantics}
\bvolume{56}
(\byear{2019}),
\bfpage{1}--\blpage{10}.
doi:\doiurl{10.1016/j.websem.2018.06.003}.
\end{barticle}
\endbibitem

\bibitem{moreira_saref4health_2020}
\begin{barticle}
\bauthor{\binits{J.}~\bsnm{Moreira}},
\bauthor{\binits{L.F.}~\bsnm{Pires}},
\bauthor{\binits{M.}~\bsnm{van Sinderen}},
\bauthor{\binits{L.}~\bsnm{Daniele}} and
\bauthor{\binits{M.}~\bsnm{Girod-Genet}},
\batitle{{SAREF4health}: {Towards} {IoT} standard-based ontology-driven cardiac e-health systems},
\bjtitle{Applied Ontology}
\bvolume{15}(\bissue{3})
(\byear{2020}),
\bfpage{385}--\blpage{410}.
doi:\doiurl{10.3233/AO-200232}.
\end{barticle}
\endbibitem

\bibitem{poveda-villalon_ontological_2020}
\begin{bchapter}
\bauthor{\binits{M.}~\bsnm{Poveda-Villalon}},
\bauthor{\binits{Q.-D.}~\bsnm{Nguyen}} and
\bauthor{\binits{C.}~\bsnm{Roussey}},
\bctitle{Ontological requirement specification for smart irrigation systems: a {SOSA}/{SSN} and {SAREF} comparison},
in: \bbtitle{Proceedings of the 9th {International} {Semantic} {Sensor} {Networks} {Workshop} ({SSN} 2018)},
\byear{2020}.
\end{bchapter}
\endbibitem

\bibitem{hogan_knowledge_2022}
\begin{barticle}
\bauthor{\binits{A.}~\bsnm{Hogan}},
\bauthor{\binits{E.}~\bsnm{Blomqvist}},
\bauthor{\binits{M.}~\bsnm{Cochez}},
\bauthor{\binits{C.}~\bsnm{D’amato}},
\bauthor{\binits{G.D.}~\bsnm{Melo}},
\bauthor{\binits{C.}~\bsnm{Gutierrez}},
\bauthor{\binits{S.}~\bsnm{Kirrane}},
\bauthor{\binits{J.E.L.}~\bsnm{Gayo}},
\bauthor{\binits{R.}~\bsnm{Navigli}},
\bauthor{\binits{S.}~\bsnm{Neumaier}},
\bauthor{\binits{A.-C.N.}~\bsnm{Ngomo}},
\bauthor{\binits{A.}~\bsnm{Polleres}},
\bauthor{\binits{S.M.}~\bsnm{Rashid}},
\bauthor{\binits{A.}~\bsnm{Rula}},
\bauthor{\binits{L.}~\bsnm{Schmelzeisen}},
\bauthor{\binits{J.}~\bsnm{Sequeda}},
\bauthor{\binits{S.}~\bsnm{Staab}} and
\bauthor{\binits{A.}~\bsnm{Zimmermann}},
\batitle{Knowledge {Graphs}},
\bjtitle{ACM Computing Surveys}
\bvolume{54}(\bissue{4})
(\byear{2022}),
\bfpage{1}--\blpage{37}.
doi:\doiurl{10.1145/3447772}.
\end{barticle}
\endbibitem

\bibitem{chandak_building_2023}
\begin{barticle}
\bauthor{\binits{P.}~\bsnm{Chandak}},
\bauthor{\binits{K.}~\bsnm{Huang}} and
\bauthor{\binits{M.}~\bsnm{Zitnik}},
\batitle{Building a knowledge graph to enable precision medicine},
\bjtitle{Scientific Data}
\bvolume{10}(\bissue{1})
(\byear{2023}),
\bfpage{67},
\bcomment{Publisher: Nature Publishing Group}.
doi:\doiurl{10.1038/s41597-023-01960-3}.
\end{barticle}
\endbibitem

\bibitem{rotmensch_learning_2017}
\begin{barticle}
\bauthor{\binits{M.}~\bsnm{Rotmensch}},
\bauthor{\binits{Y.}~\bsnm{Halpern}},
\bauthor{\binits{A.}~\bsnm{Tlimat}},
\bauthor{\binits{S.}~\bsnm{Horng}} and
\bauthor{\binits{D.}~\bsnm{Sontag}},
\batitle{Learning a {Health} {Knowledge} {Graph} from {Electronic} {Medical} {Records}},
\bjtitle{Scientific Reports}
\bvolume{7}(\bissue{1})
(\byear{2017}),
\bfpage{5994}.
doi:\doiurl{10.1038/s41598-017-05778-z}.
\end{barticle}
\endbibitem

\bibitem{chen_robustly_2019}
\begin{bchapter}
\bauthor{\binits{I.Y.}~\bsnm{Chen}},
\bauthor{\binits{M.}~\bsnm{Agrawal}},
\bauthor{\binits{S.}~\bsnm{Horng}} and
\bauthor{\binits{D.}~\bsnm{Sontag}},
\bctitle{Robustly {Extracting} {Medical} {Knowledge} from {EHRs}: {A} {Case} {Study} of {Learning} a {Health} {Knowledge} {Graph}},
in: \bbtitle{Proceedings of the Pacific {Symposium} on {Biocomputing} 2020},
\bconflocation{Kohala Coast, Hawaii, USA}, \byear{2019},
pp.~\bfpage{19}--\blpage{30}.
ISBN \bisbn{9789811215629 9789811215636}.
doi:\doiurl{10.1142/9789811215636\_0003}.
\end{bchapter}
\endbibitem

\bibitem{gyrard_interdisciplinary_2022}
\begin{barticle}
\bauthor{\binits{A.}~\bsnm{Gyrard}} and
\bauthor{\binits{K.}~\bsnm{Boudaoud}},
\batitle{Interdisciplinary {IoT} and {Emotion} {Knowledge} {Graph}-{Based} {Recommendation} {System} to {Boost} {Mental} {Health}},
\bjtitle{Applied Sciences}
\bvolume{12}(\bissue{19})
(\byear{2022}),
\bfpage{9712}.
doi:\doiurl{10.3390/app12199712}.
\end{barticle}
\endbibitem

\bibitem{jiang_graphcare_2024}
\begin{bchapter}
\bauthor{\binits{P.}~\bsnm{Jiang}},
\bauthor{\binits{C.D.}~\bsnm{Xiao}},
\bauthor{\binits{A.}~\bsnm{Cross}} and
\bauthor{\binits{J.}~\bsnm{Sun}},
\bctitle{{GraphCare}: {Enhancing} {Healthcare} {Predictions} with {Personalized} {Knowledge} {Graphs}},
in: \bbtitle{{ICLR} {Proceedings}},
Vol.~\bseriesno{2024},
\byear{2024},
pp.~\bfpage{16839}--\blpage{16866}.
\url{https://proceedings.iclr.cc/paper_files/paper/2024/hash/484d254ff80e99d543159440a06db0de-Abstract-Conference.html}.
\end{bchapter}
\endbibitem

\bibitem{tao_mining_2020}
\begin{barticle}
\bauthor{\binits{X.}~\bsnm{Tao}},
\bauthor{\binits{T.}~\bsnm{Pham}},
\bauthor{\binits{J.}~\bsnm{Zhang}},
\bauthor{\binits{J.}~\bsnm{Yong}},
\bauthor{\binits{W.P.}~\bsnm{Goh}},
\bauthor{\binits{W.}~\bsnm{Zhang}} and
\bauthor{\binits{Y.}~\bsnm{Cai}},
\batitle{Mining health knowledge graph for health risk prediction},
\bjtitle{World Wide Web}
\bvolume{23}(\bissue{4})
(\byear{2020}),
\bfpage{2341}--\blpage{2362}.
doi:\doiurl{10.1007/s11280-020-00810-1}.
\end{barticle}
\endbibitem

\bibitem{zeng_toward_2022}
\begin{barticle}
\bauthor{\binits{X.}~\bsnm{Zeng}},
\bauthor{\binits{X.}~\bsnm{Tu}},
\bauthor{\binits{Y.}~\bsnm{Liu}},
\bauthor{\binits{X.}~\bsnm{Fu}} and
\bauthor{\binits{Y.}~\bsnm{Su}},
\batitle{Toward better drug discovery with knowledge graph},
\bjtitle{Current Opinion in Structural Biology}
\bvolume{72}
(\byear{2022}),
\bfpage{114}--\blpage{126}.
doi:\doiurl{10.1016/j.sbi.2021.09.003}.
\end{barticle}
\endbibitem

\bibitem{rajabi_knowledge_2022}
\begin{barticle}
\bauthor{\binits{E.}~\bsnm{Rajabi}} and
\bauthor{\binits{S.}~\bsnm{Kafaie}},
\batitle{Knowledge {Graphs} and {Explainable} {AI} in {Healthcare}},
\bjtitle{Information}
\bvolume{13}(\bissue{10})
(\byear{2022}),
\bfpage{459}.
doi:\doiurl{10.3390/info13100459}.
\end{barticle}
\endbibitem

\bibitem{lecue_role_2020}
\begin{barticle}
\bauthor{\binits{F.}~\bsnm{Lecue}},
\batitle{On the role of knowledge graphs in explainable {AI}},
\bjtitle{Semantic Web}
\bvolume{11}(\bissue{1})
(\byear{2020}),
\bfpage{41}--\blpage{51}.
doi:\doiurl{10.3233/SW-190374}.
\end{barticle}
\endbibitem

\bibitem{le-phuoc_graph_2016}
\begin{barticle}
\bauthor{\binits{D.}~\bsnm{Le-Phuoc}},
\bauthor{\binits{H.}~\bsnm{Nguyen Mau~Quoc}},
\bauthor{\binits{H.}~\bsnm{Ngo~Quoc}},
\bauthor{\binits{T.}~\bsnm{Tran~Nhat}} and
\bauthor{\binits{M.}~\bsnm{Hauswirth}},
\batitle{The {Graph} of {Things}: {A} step towards the {Live} {Knowledge} {Graph} of connected things},
\bjtitle{Journal of Web Semantics}
\bvolume{37-38}
(\byear{2016}),
\bfpage{25}--\blpage{35}.
doi:\doiurl{10.1016/j.websem.2016.02.003}.
\end{barticle}
\endbibitem

\bibitem{bizer_linked_2011}
\begin{bchapter}
\bauthor{\binits{C.}~\bsnm{Bizer}},
\bauthor{\binits{T.}~\bsnm{Heath}} and
\bauthor{\binits{T.}~\bsnm{Berners-Lee}},
\bctitle{Linked {Data}: {The} {Story} {So} {Far}},
in: \bbtitle{Semantic {Services}, {Interoperability} and {Web} {Applications}: {Emerging} {Concepts}},
\beditor{\binits{A.}~\bsnm{Sheth}}, ed.,
\bpublisher{IGI Global},
\byear{2011},
pp.~\bfpage{205}--\blpage{227}.
doi:\doiurl{10.4018/978-1-60960-593-3}.
\end{bchapter}
\endbibitem

\bibitem{yu_using_2015}
\begin{barticle}
\bauthor{\binits{L.}~\bsnm{Yu}} and
\bauthor{\binits{Y.}~\bsnm{Liu}},
\batitle{Using {Linked} {Data} in a heterogeneous {Sensor} {Web}: challenges, experiments and lessons learned},
\bjtitle{International Journal of Digital Earth}
\bvolume{8}(\bissue{1})
(\byear{2015}),
\bfpage{17}--\blpage{37}.
doi:\doiurl{10.1080/17538947.2013.839007}.
\end{barticle}
\endbibitem

\bibitem{gray_applying_2014}
\begin{barticle}
\bauthor{\binits{A.J.G.}~\bsnm{Gray}},
\bauthor{\binits{P.}~\bsnm{Groth}},
\bauthor{\binits{A.}~\bsnm{Loizou}},
\bauthor{\binits{S.}~\bsnm{Askjaer}},
\bauthor{\binits{C.}~\bsnm{Brenninkmeijer}},
\bauthor{\binits{K.}~\bsnm{Burger}},
\bauthor{\binits{C.}~\bsnm{Chichester}},
\bauthor{\binits{C.T.}~\bsnm{Evelo}},
\bauthor{\binits{C.}~\bsnm{Goble}},
\bauthor{\binits{L.}~\bsnm{Harland}},
\bauthor{\binits{S.}~\bsnm{Pettifer}},
\bauthor{\binits{M.}~\bsnm{Thompson}},
\bauthor{\binits{A.}~\bsnm{Waagmeester}} and
\bauthor{\binits{A.J.}~\bsnm{Williams}},
\batitle{Applying linked data approaches to pharmacology: {Architectural} decisions and implementation},
\bjtitle{Semantic Web}
\bvolume{5}(\bissue{2})
(\byear{2014}),
\bfpage{101}--\blpage{113}.
doi:\doiurl{10.3233/SW-2012-0088}.
\end{barticle}
\endbibitem

\bibitem{pathak_using_2013}
\begin{barticle}
\bauthor{\binits{J.}~\bsnm{Pathak}},
\bauthor{\binits{R.C.}~\bsnm{Kiefer}} and
\bauthor{\binits{C.G.}~\bsnm{Chute}},
\batitle{Using {Linked} {Data} for {Mining} {Drug}-{Drug} {Interactions} in {Electronic} {Health} {Records}},
\bjtitle{Studies in health technology and informatics}
\bvolume{192}
(\byear{2013}),
\bfpage{682}--\blpage{686}.
\end{barticle}
\endbibitem

\bibitem{zenuni_state_2015}
\begin{barticle}
\bauthor{\binits{X.}~\bsnm{Zenuni}},
\bauthor{\binits{B.}~\bsnm{Raufi}},
\bauthor{\binits{F.}~\bsnm{Ismaili}} and
\bauthor{\binits{J.}~\bsnm{Ajdari}},
\batitle{State of the {Art} of {Semantic} {Web} for {Healthcare}},
\bjtitle{Procedia - Social and Behavioral Sciences}
\bvolume{195}
(\byear{2015}),
\bfpage{1990}--\blpage{1998}.
doi:\doiurl{10.1016/j.sbspro.2015.06.213}.
\end{barticle}
\endbibitem

\bibitem{dimitrieski_survey_2016}
\begin{bchapter}
\bauthor{\binits{V.}~\bsnm{Dimitrieski}},
\bauthor{\binits{G.}~\bsnm{Petrović}},
\bauthor{\binits{A.}~\bsnm{Kovačević}},
\bauthor{\binits{I.}~\bsnm{Luković}} and
\bauthor{\binits{H.}~\bsnm{Fujita}},
\bctitle{A {Survey} on {Ontologies} and {Ontology} {Alignment} {Approaches} in {Healthcare}},
in: \bbtitle{Proceedings of the {International} {Conference} on {Industrial}, {Engineering} and {Other} {Applications} of {Applied} {Intelligent} {Systems}},
\bsertitle{Lecture Notes in Computer Science},
Vol.~\bseriesno{9799},
\bpublisher{Springer},
\byear{2016},
pp.~\bfpage{373}--\blpage{385}.
ISBN \bisbn{978-3-319-42006-6 978-3-319-42007-3}.
doi:\doiurl{10.1007/978-3-319-42007-3\_32}.
\end{bchapter}
\endbibitem

\bibitem{amar_electronic_2024}
\begin{barticle}
\bauthor{\binits{F.}~\bsnm{Amar}},
\bauthor{\binits{A.}~\bsnm{April}} and
\bauthor{\binits{A.}~\bsnm{Abran}},
\batitle{Electronic {Health} {Record} and {Semantic} {Issues} {Using} {Fast} {Healthcare} {Interoperability} {Resources}: {Systematic} {Mapping} {Review}},
\bjtitle{Journal of Medical Internet Research}
\bvolume{26}(\bissue{1})
(\byear{2024}),
\bfpage{e45209}.
doi:\doiurl{10.2196/45209}.
\end{barticle}
\endbibitem

\bibitem{miranda_semantic_2024}
\begin{bchapter}
\bauthor{\binits{N.}~\bsnm{Miranda}},
\bauthor{\binits{M.M.}~\bsnm{Machado}} and
\bauthor{\binits{D.A.}~\bsnm{Moreira}},
\bctitle{Semantic {Web} {Technologies} in {Healthcare}: {A} {Scoping} {Review}},
in: \bbtitle{Simpósio {Brasileiro} de {Sistemas} {Multimídia} e {Web} ({WebMedia})},
\byear{2024},
pp.~\bfpage{171}--\blpage{184}.
doi:\doiurl{10.5753/webmedia\_estendido.2024.244455}.
\end{bchapter}
\endbibitem

\bibitem{wu_semantics-driven_2025}
\begin{barticle}
\bauthor{\binits{Y.}~\bsnm{Wu}},
\bauthor{\binits{M.}~\bsnm{Ren}},
\bauthor{\binits{N.}~\bsnm{Chen}} and
\bauthor{\binits{L.}~\bsnm{Yang}},
\batitle{Semantics-driven improvements in electronic health records data quality: a systematic review},
\bjtitle{BMC Medical Informatics and Decision Making}
\bvolume{25}(\bissue{1})
(\byear{2025}),
\bfpage{298}.
doi:\doiurl{10.1186/s12911-025-03146-w}.
\end{barticle}
\endbibitem

\bibitem{islam_internet_2015}
\begin{barticle}
\bauthor{\binits{S.M.R.}~\bsnm{Islam}},
\bauthor{\bsnm{{Daehan Kwak}}},
\bauthor{\binits{M.}~\bsnm{Humaun~Kabir}},
\bauthor{\binits{M.}~\bsnm{Hossain}} and
\bauthor{\bsnm{{Kyung-Sup Kwak}}},
\batitle{The {Internet} of {Things} for {Health} {Care}: {A} {Comprehensive} {Survey}},
\bjtitle{IEEE Access}
\bvolume{3}
(\byear{2015}),
\bfpage{678}--\blpage{708}.
doi:\doiurl{10.1109/ACCESS.2015.2437951}.
\end{barticle}
\endbibitem

\bibitem{yin_internet_2016}
\begin{barticle}
\bauthor{\binits{Y.}~\bsnm{Yin}},
\bauthor{\binits{Y.}~\bsnm{Zeng}},
\bauthor{\binits{X.}~\bsnm{Chen}} and
\bauthor{\binits{Y.}~\bsnm{Fan}},
\batitle{The internet of things in healthcare: {An} overview},
\bjtitle{Journal of Industrial Information Integration}
\bvolume{1}
(\byear{2016}),
\bfpage{3}--\blpage{13}.
doi:\doiurl{10.1016/j.jii.2016.03.004}.
\end{barticle}
\endbibitem

\bibitem{qi_advanced_2017}
\begin{barticle}
\bauthor{\binits{J.}~\bsnm{Qi}},
\bauthor{\binits{P.}~\bsnm{Yang}},
\bauthor{\binits{G.}~\bsnm{Min}},
\bauthor{\binits{O.}~\bsnm{Amft}},
\bauthor{\binits{F.}~\bsnm{Dong}} and
\bauthor{\binits{L.}~\bsnm{Xu}},
\batitle{Advanced internet of things for personalised healthcare systems: {A} survey},
\bjtitle{Pervasive and Mobile Computing}
\bvolume{41}
(\byear{2017}),
\bfpage{132}--\blpage{149}.
doi:\doiurl{10.1016/j.pmcj.2017.06.018}.
\end{barticle}
\endbibitem

\bibitem{philip_internet_2021}
\begin{barticle}
\bauthor{\binits{N.Y.}~\bsnm{Philip}},
\bauthor{\binits{J.J.P.C.}~\bsnm{Rodrigues}},
\bauthor{\binits{H.}~\bsnm{Wang}},
\bauthor{\binits{S.J.}~\bsnm{Fong}} and
\bauthor{\binits{J.}~\bsnm{Chen}},
\batitle{Internet of {Things} for {In}-{Home} {Health} {Monitoring} {Systems}: {Current} {Advances}, {Challenges} and {Future} {Directions}},
\bjtitle{IEEE Journal on Selected Areas in Communications}
\bvolume{39}(\bissue{2})
(\byear{2021}),
\bfpage{300}--\blpage{310}.
doi:\doiurl{10.1109/JSAC.2020.3042421}.
\end{barticle}
\endbibitem

\bibitem{albahri_real-time_2018}
\begin{barticle}
\bauthor{\binits{O.S.}~\bsnm{Albahri}},
\bauthor{\binits{A.A.}~\bsnm{Zaidan}},
\bauthor{\binits{B.B.}~\bsnm{Zaidan}},
\bauthor{\binits{M.}~\bsnm{Hashim}},
\bauthor{\binits{A.S.}~\bsnm{Albahri}} and
\bauthor{\binits{M.A.}~\bsnm{Alsalem}},
\batitle{Real-{Time} {Remote} {Health}-{Monitoring} {Systems} in a {Medical} {Centre}: {A} {Review} of the {Provision} of {Healthcare} {Services}-{Based} {Body} {Sensor} {Information}, {Open} {Challenges} and {Methodological} {Aspects}},
\bjtitle{Journal of Medical Systems}
\bvolume{42}(\bissue{9})
(\byear{2018}),
\bfpage{164}.
doi:\doiurl{10.1007/s10916-018-1006-6}.
\end{barticle}
\endbibitem

\bibitem{babu_wearable_2024}
\begin{barticle}
\bauthor{\binits{M.}~\bsnm{Babu}},
\bauthor{\binits{Z.}~\bsnm{Lautman}},
\bauthor{\binits{X.}~\bsnm{Lin}},
\bauthor{\binits{M.H.B.}~\bsnm{Sobota}} and
\bauthor{\binits{M.P.}~\bsnm{Snyder}},
\batitle{Wearable {Devices}: {Implications} for {Precision} {Medicine} and the {Future} of {Health} {Care}},
\bjtitle{Annual Review of Medicine}
\bvolume{75}(\bissue{1})
(\byear{2024}),
\bfpage{401}--\blpage{415}.
doi:\doiurl{10.1146/annurev-med-052422-020437}.
\end{barticle}
\endbibitem

\bibitem{majumder_wearable_2017}
\begin{barticle}
\bauthor{\binits{S.}~\bsnm{Majumder}},
\bauthor{\binits{T.}~\bsnm{Mondal}} and
\bauthor{\binits{M.J.}~\bsnm{Deen}},
\batitle{Wearable {Sensors} for {Remote} {Health} {Monitoring}},
\bjtitle{Sensors}
\bvolume{17}(\bissue{1})
(\byear{2017}),
\bfpage{130}.
doi:\doiurl{10.3390/s17010130}.
\end{barticle}
\endbibitem

\bibitem{kim_wearable_2019}
\begin{barticle}
\bauthor{\binits{J.}~\bsnm{Kim}},
\bauthor{\binits{A.S.}~\bsnm{Campbell}},
\bauthor{\binits{B.E.-F.}~\bsnm{de~Ávila}} and
\bauthor{\binits{J.}~\bsnm{Wang}},
\batitle{Wearable biosensors for healthcare monitoring},
\bjtitle{Nature Biotechnology}
\bvolume{37}(\bissue{4})
(\byear{2019}),
\bfpage{389}--\blpage{406}.
doi:\doiurl{10.1038/s41587-019-0045-y}.
\end{barticle}
\endbibitem

\bibitem{baig_systematic_2017}
\begin{botherref}
\oauthor{\binits{M.M.}~\bsnm{Baig}},
\oauthor{\binits{H.}~\bsnm{GholamHosseini}},
\oauthor{\binits{A.A.}~\bsnm{Moqeem}},
\oauthor{\binits{F.}~\bsnm{Mirza}} and
\oauthor{\binits{M.}~\bsnm{Lindén}},
A {Systematic} {Review} of {Wearable} {Patient} {Monitoring} {Systems} – {Current} {Challenges} and {Opportunities} for {Clinical} {Adoption},
\textit{Journal of Medical Systems}
\textbf{41}(7)
(2017).
doi:\doiurl{10.1007/s10916-017-0760-1}.
\end{botherref}
\endbibitem

\bibitem{punj_technological_2019}
\begin{barticle}
\bauthor{\binits{R.}~\bsnm{Punj}} and
\bauthor{\binits{R.}~\bsnm{Kumar}},
\batitle{Technological aspects of {WBANs} for health monitoring: a comprehensive review},
\bjtitle{Wireless Networks}
\bvolume{25}(\bissue{3})
(\byear{2019}),
\bfpage{1125}--\blpage{1157}.
doi:\doiurl{10.1007/s11276-018-1694-3}.
\end{barticle}
\endbibitem

\bibitem{banaee_data_2013}
\begin{barticle}
\bauthor{\binits{H.}~\bsnm{Banaee}},
\bauthor{\binits{M.U.}~\bsnm{Ahmed}} and
\bauthor{\binits{A.}~\bsnm{Loutfi}},
\batitle{Data {Mining} for {Wearable} {Sensors} in {Health} {Monitoring} {Systems}: {A} {Review} of {Recent} {Trends} and {Challenges}},
\bjtitle{Sensors}
\bvolume{13}
(\byear{2013}),
\bfpage{17472}--\blpage{17500}.
doi:\doiurl{10.3390/s131217472}.
\end{barticle}
\endbibitem

\bibitem{dang_human-centred_2023}
\begin{barticle}
\bauthor{\binits{T.}~\bsnm{Dang}},
\bauthor{\binits{D.}~\bsnm{Spathis}},
\bauthor{\binits{A.}~\bsnm{Ghosh}} and
\bauthor{\binits{C.}~\bsnm{Mascolo}},
\batitle{Human-centred artificial intelligence for mobile health sensing: challenges and opportunities},
\bjtitle{Royal Society Open Science}
\bvolume{10}(\bissue{11})
(\byear{2023}),
\bfpage{230806}.
doi:\doiurl{10.1098/rsos.230806}.
\end{barticle}
\endbibitem

\bibitem{bollineni_iot_2025}
\begin{barticle}
\bauthor{\binits{C.}~\bsnm{Bollineni}},
\bauthor{\binits{M.}~\bsnm{Sharma}},
\bauthor{\binits{A.}~\bsnm{Hazra}},
\bauthor{\binits{P.}~\bsnm{Kumari}},
\bauthor{\binits{S.}~\bsnm{Manipriya}} and
\bauthor{\binits{A.}~\bsnm{Tomar}},
\batitle{{IoT} for {Next}-{Generation} {Smart} {Healthcare}: {A} {Comprehensive} {Survey}},
\bjtitle{IEEE Internet of Things Journal}
\bvolume{12}(\bissue{16})
(\byear{2025}),
\bfpage{32616}--\blpage{32639}.
doi:\doiurl{10.1109/JIOT.2025.3570188}.
\end{barticle}
\endbibitem

\bibitem{honti_review_2019}
\begin{barticle}
\bauthor{\binits{G.M.}~\bsnm{Honti}} and
\bauthor{\binits{J.}~\bsnm{Abonyi}},
\batitle{A {Review} of {Semantic} {Sensor} {Technologies} in {Internet} of {Things} {Architectures}},
\bjtitle{Complexity}
\bvolume{2019}
(\byear{2019}),
\bfpage{6473160}.
doi:\doiurl{10.1155/2019/6473160}.
\end{barticle}
\endbibitem

\bibitem{rhayem_semantic_2020}
\begin{barticle}
\bauthor{\binits{A.}~\bsnm{Rhayem}},
\bauthor{\binits{M.B.A.}~\bsnm{Mhiri}} and
\bauthor{\binits{F.}~\bsnm{Gargouri}},
\batitle{Semantic {Web} {Technologies} for the {Internet} of {Things}: {Systematic} {Literature} {Review}},
\bjtitle{Internet of Things}
\bvolume{11}
(\byear{2020}),
\bfpage{100206}.
doi:\doiurl{10.1016/j.iot.2020.100206}.
\end{barticle}
\endbibitem

\bibitem{bajaj_study_2017}
\begin{botherref}
\oauthor{\binits{G.}~\bsnm{Bajaj}},
\oauthor{\binits{R.}~\bsnm{Agarwal}},
\oauthor{\binits{P.}~\bsnm{Singh}},
\oauthor{\binits{N.}~\bsnm{Georgantas}} and
\oauthor{\binits{V.}~\bsnm{Issarny}},
A study of existing {Ontologies} in the {IoT}-domain,
\textit{arXiv preprint arXiv:1707.00112}
(2017).
\end{botherref}
\endbibitem

\bibitem{compton_survey_2009}
\begin{bchapter}
\bauthor{\binits{M.}~\bsnm{Compton}},
\bauthor{\binits{C.}~\bsnm{Henson}},
\bauthor{\binits{L.}~\bsnm{Lefort}},
\bauthor{\binits{H.}~\bsnm{Neuhaus}} and
\bauthor{\binits{A.}~\bsnm{Sheth}},
\bctitle{A survey of the semantic specification of sensors},
in: \bbtitle{Proceedings of the 2nd {International} {Conference} on {Semantic} {Sensor} {Networks}},
Vol.~\bseriesno{522},
\byear{2009},
pp.~\bfpage{17}--\blpage{32}.
\end{bchapter}
\endbibitem

\bibitem{harlamova_survey_2017}
\begin{barticle}
\bauthor{\binits{M.}~\bsnm{Harlamova}},
\bauthor{\binits{M.}~\bsnm{Kirikova}} and
\bauthor{\binits{K.}~\bsnm{Sandkuhl}},
\batitle{A {Survey} on {Challenges} of {Semantics} {Application} in the {Internet} of {Things} {Domain}},
\bjtitle{Applied Computer Systems}
\bvolume{21}(\bissue{1})
(\byear{2017}),
\bfpage{13}--\blpage{21}.
doi:\doiurl{10.1515/acss-2017-0002}.
\end{barticle}
\endbibitem

\bibitem{ye_semantic_2015}
\begin{barticle}
\bauthor{\binits{J.}~\bsnm{Ye}},
\bauthor{\binits{S.}~\bsnm{Dasiopoulou}},
\bauthor{\binits{G.}~\bsnm{Stevenson}},
\bauthor{\binits{G.}~\bsnm{Meditskos}},
\bauthor{\binits{E.}~\bsnm{Kontopoulos}},
\bauthor{\binits{I.}~\bsnm{Kompatsiaris}} and
\bauthor{\binits{S.}~\bsnm{Dobson}},
\batitle{Semantic web technologies in pervasive computing: {A} survey and research roadmap},
\bjtitle{Pervasive and Mobile Computing}
\bvolume{23}
(\byear{2015}),
\bfpage{1}--\blpage{25}.
doi:\doiurl{10.1016/j.pmcj.2014.12.009}.
\end{barticle}
\endbibitem

\bibitem{tortorella_healthcare_2020}
\begin{barticle}
\bauthor{\binits{G.L.}~\bsnm{Tortorella}},
\bauthor{\binits{F.S.}~\bsnm{Fogliatto}},
\bauthor{\binits{A.}~\bsnm{Mac Cawley~Vergara}},
\bauthor{\binits{R.}~\bsnm{Vassolo}} and
\bauthor{\binits{R.}~\bsnm{Sawhney}},
\batitle{Healthcare 4.0: trends, challenges and research directions},
\bjtitle{Production Planning \& Control}
\bvolume{31}(\bissue{15})
(\byear{2020}),
\bfpage{1245}--\blpage{1260}.
doi:\doiurl{10.1080/09537287.2019.1702226}.
\end{barticle}
\endbibitem

\bibitem{jayaraman_healthcare_2020}
\begin{barticle}
\bauthor{\binits{P.P.}~\bsnm{Jayaraman}},
\bauthor{\binits{A.R.M.}~\bsnm{Forkan}},
\bauthor{\binits{A.}~\bsnm{Morshed}},
\bauthor{\binits{P.D.}~\bsnm{Haghighi}} and
\bauthor{\binits{Y.-B.}~\bsnm{Kang}},
\batitle{Healthcare 4.0: {A} review of frontiers in digital health},
\bjtitle{WIREs Data Mining and Knowledge Discovery}
\bvolume{10}(\bissue{2})
(\byear{2020}),
\bfpage{e1350}.
doi:\doiurl{10.1002/widm.1350}.
\end{barticle}
\endbibitem

\bibitem{rahman_ai_2025}
\begin{barticle}
\bauthor{\binits{A.}~\bsnm{Rahman}},
\bauthor{\binits{D.}~\bsnm{Kundu}},
\bauthor{\binits{T.}~\bsnm{Debnath}},
\bauthor{\binits{M.}~\bsnm{Rahman}},
\bauthor{\binits{U.K.}~\bsnm{Das}},
\bauthor{\binits{A.S.M.}~\bsnm{Miah}} and
\bauthor{\binits{G.}~\bsnm{Muhammad}},
\batitle{From {AI} to the {Era} of {Explainable} {AI} in {Healthcare} 5.0: {Current} {State} and {Future} {Outlook}},
\bjtitle{Expert Systems}
\bvolume{42}(\bissue{6})
(\byear{2025}),
\bfpage{e70060}.
doi:\doiurl{10.1111/exsy.70060}.
\end{barticle}
\endbibitem

\bibitem{rashid_human-centered_2024}
\begin{barticle}
\bauthor{\binits{S.}~\bsnm{Rashid}} and
\bauthor{\binits{A.}~\bsnm{Nemati}},
\batitle{Human-centered {IoT}-based health monitoring in the {Healthcare} 5.0 era: literature descriptive analysis and future research guidelines},
\bjtitle{Discover Internet of Things}
\bvolume{4}(\bissue{1})
(\byear{2024}),
\bfpage{26}.
doi:\doiurl{10.1007/s43926-024-00082-5}.
\end{barticle}
\endbibitem

\bibitem{kumar_artificial_2023}
\begin{barticle}
\bauthor{\binits{P.}~\bsnm{Kumar}},
\bauthor{\binits{S.}~\bsnm{Chauhan}} and
\bauthor{\binits{L.K.}~\bsnm{Awasthi}},
\batitle{Artificial {Intelligence} in {Healthcare}: {Review}, {Ethics}, {Trust} {Challenges} \& {Future} {Research} {Directions}},
\bjtitle{Engineering Applications of Artificial Intelligence}
\bvolume{120}
(\byear{2023}),
\bfpage{105894}.
doi:\doiurl{10.1016/j.engappai.2023.105894}.
\end{barticle}
\endbibitem

\bibitem{behera_emerging_2019}
\begin{barticle}
\bauthor{\binits{R.K.}~\bsnm{Behera}},
\bauthor{\binits{P.K.}~\bsnm{Bala}} and
\bauthor{\binits{A.}~\bsnm{Dhir}},
\batitle{The emerging role of cognitive computing in healthcare: {A} systematic literature review},
\bjtitle{International Journal of Medical Informatics}
\bvolume{129}
(\byear{2019}),
\bfpage{154}--\blpage{166}.
doi:\doiurl{10.1016/j.ijmedinf.2019.04.024}.
\end{barticle}
\endbibitem

\bibitem{rhayem_semantic-enabled_2021}
\begin{barticle}
\bauthor{\binits{A.}~\bsnm{Rhayem}},
\bauthor{\binits{M.B.A.}~\bsnm{Mhiri}},
\bauthor{\binits{K.}~\bsnm{Drira}},
\bauthor{\binits{S.}~\bsnm{Tazi}} and
\bauthor{\binits{F.}~\bsnm{Gargouri}},
\batitle{A semantic-enabled and context-aware monitoring system for the internet of medical things},
\bjtitle{Expert Systems}
\bvolume{38}(\bissue{2})
(\byear{2021}),
\bfpage{e12629}.
doi:\doiurl{10.1111/exsy.12629}.
\end{barticle}
\endbibitem

\bibitem{moher_preferred_2009}
\begin{barticle}
\bauthor{\binits{D.}~\bsnm{Moher}},
\bauthor{\binits{A.}~\bsnm{Liberati}},
\bauthor{\binits{J.}~\bsnm{Tetzlaff}},
\bauthor{\binits{D.G.}~\bsnm{Altman}} and
\bauthor{\binits{T.P.}~\bsnm{Group}},
\batitle{Preferred {Reporting} {Items} for {Systematic} {Reviews} and {Meta}-{Analyses}: {The} {PRISMA} {Statement}},
\bjtitle{PLOS Medicine}
\bvolume{6}(\bissue{7})
(\byear{2009}),
\bfpage{e1000097}.
doi:\doiurl{10.1371/journal.pmed.1000097}.
\end{barticle}
\endbibitem

\bibitem{kitchenham_systematic_2010}
\begin{barticle}
\bauthor{\binits{B.}~\bsnm{Kitchenham}},
\bauthor{\binits{R.}~\bsnm{Pretorius}},
\bauthor{\binits{D.}~\bsnm{Budgen}},
\bauthor{\binits{O.}~\bsnm{Pearl~Brereton}},
\bauthor{\binits{M.}~\bsnm{Turner}},
\bauthor{\binits{M.}~\bsnm{Niazi}} and
\bauthor{\binits{S.}~\bsnm{Linkman}},
\batitle{Systematic literature reviews in software engineering – {A} tertiary study},
\bjtitle{Information and Software Technology}
\bvolume{52}(\bissue{8})
(\byear{2010}),
\bfpage{792}--\blpage{805}.
doi:\doiurl{10.1016/j.infsof.2010.03.006}.
\end{barticle}
\endbibitem

\bibitem{ouzzani_rayyanweb_2016}
\begin{barticle}
\bauthor{\binits{M.}~\bsnm{Ouzzani}},
\bauthor{\binits{H.}~\bsnm{Hammady}},
\bauthor{\binits{Z.}~\bsnm{Fedorowicz}} and
\bauthor{\binits{A.}~\bsnm{Elmagarmid}},
\batitle{Rayyan—a web and mobile app for systematic reviews},
\bjtitle{Systematic Reviews}
\bvolume{5}(\bissue{1})
(\byear{2016}),
\bfpage{210}.
doi:\doiurl{10.1186/s13643-016-0384-4}.
\end{barticle}
\endbibitem

\bibitem{bampi_ontology-driven_2025}
\begin{barticle}
\bauthor{\binits{M.D.}~\bsnm{Bampi}},
\bauthor{\binits{W.O.d.}~\bsnm{Morais}},
\bauthor{\binits{J.I.}~\bsnm{Olszewska}} and
\bauthor{\binits{E.P.D.}~\bsnm{Freitas}},
\batitle{Ontology-driven monitoring system for ambient assisted living},
\bjtitle{The Knowledge Engineering Review}
\bvolume{40}
(\byear{2025}),
\bfpage{e2}.
doi:\doiurl{10.1017/S0269888925000037}.
\end{barticle}
\endbibitem

\bibitem{chiang_context-aware_2015}
\begin{barticle}
\bauthor{\binits{T.-C.}~\bsnm{Chiang}} and
\bauthor{\binits{W.-H.}~\bsnm{Liang}},
\batitle{A {Context}-{Aware} {Interactive} {Health} {Care} {System} {Based} on {Ontology} and {Fuzzy} {Inference}},
\bjtitle{Journal of Medical Systems}
\bvolume{39}(\bissue{9})
(\byear{2015}),
\bfpage{105}.
doi:\doiurl{10.1007/s10916-015-0287-2}.
\end{barticle}
\endbibitem

\bibitem{elhadj_do-care_2021}
\begin{barticle}
\bauthor{\binits{H.B.}~\bsnm{Elhadj}},
\bauthor{\binits{F.}~\bsnm{Sallabi}},
\bauthor{\binits{A.}~\bsnm{Henaien}},
\bauthor{\binits{L.}~\bsnm{Chaari}},
\bauthor{\binits{K.}~\bsnm{Shuaib}} and
\bauthor{\binits{M.}~\bsnm{Al~Thawadi}},
\batitle{Do-{Care}: {A} dynamic ontology reasoning based healthcare monitoring system},
\bjtitle{Future Generation Computer Systems}
\bvolume{118}
(\byear{2021}),
\bfpage{417}--\blpage{431}.
doi:\doiurl{10.1016/j.future.2021.01.001}.
\end{barticle}
\endbibitem

\bibitem{esposito_smart_2018}
\begin{barticle}
\bauthor{\binits{M.}~\bsnm{Esposito}},
\bauthor{\binits{A.}~\bsnm{Minutolo}},
\bauthor{\binits{R.}~\bsnm{Megna}},
\bauthor{\binits{M.}~\bsnm{Forastiere}},
\bauthor{\binits{M.}~\bsnm{Magliulo}} and
\bauthor{\binits{G.}~\bsnm{De~Pietro}},
\batitle{A smart mobile, self-configuring, context-aware architecture for personal health monitoring},
\bjtitle{Engineering Applications of Artificial Intelligence}
\bvolume{67}
(\byear{2018}),
\bfpage{136}--\blpage{156}.
doi:\doiurl{10.1016/j.engappai.2017.09.019}.
\end{barticle}
\endbibitem

\bibitem{garcia-moreno_systematic_2023}
\begin{barticle}
\bauthor{\binits{F.M.}~\bsnm{Garcia-Moreno}},
\bauthor{\binits{M.}~\bsnm{Bermudez-Edo}},
\bauthor{\binits{J.M.}~\bsnm{Pérez-Mármol}},
\bauthor{\binits{J.L.}~\bsnm{Garrido}} and
\bauthor{\binits{M.J.}~\bsnm{Rodríguez-Fórtiz}},
\batitle{Systematic design of health monitoring systems centered on older adults and {ADLs}},
\bjtitle{BMC Medical Informatics and Decision Making}
\bvolume{23}(\bissue{3})
(\byear{2023}),
\bfpage{300}.
doi:\doiurl{10.1186/s12911-024-02432-3}.
\end{barticle}
\endbibitem

\bibitem{garcia-valverde_heart_2014}
\begin{barticle}
\bauthor{\binits{T.}~\bsnm{Garcia-Valverde}},
\bauthor{\binits{A.}~\bsnm{Muñoz}},
\bauthor{\binits{F.}~\bsnm{Arcas}},
\bauthor{\binits{A.}~\bsnm{Bueno-Crespo}} and
\bauthor{\binits{A.}~\bsnm{Caballero}},
\batitle{Heart {Health} {Risk} {Assessment} {System}: {A} {Nonintrusive} {Proposal} {Using} {Ontologies} and {Expert} {Rules}},
\bjtitle{BioMed Research International}
\bvolume{2014}
(\byear{2014}),
\bfpage{e959645}.
doi:\doiurl{10.1155/2014/959645}.
\end{barticle}
\endbibitem

\bibitem{henaien_combined_2020}
\begin{bchapter}
\bauthor{\binits{A.}~\bsnm{Henaien}},
\bauthor{\binits{H.}~\bsnm{Ben~Elhadj}} and
\bauthor{\binits{L.}~\bsnm{Chaari~Fourati}},
\bctitle{Combined {Machine} {Learning} and {Semantic} {Modelling} for {Situation} {Awareness} and {Healthcare} {Decision} {Support}},
in: \bbtitle{The {Impact} of {Digital} {Technologies} on {Public} {Health} in {Developed} and {Developing} {Countries}},
\beditor{\binits{M.}~\bsnm{Jmaiel}},
\beditor{\binits{M.}~\bsnm{Mokhtari}},
\beditor{\binits{B.}~\bsnm{Abdulrazak}},
\beditor{\binits{H.}~\bsnm{Aloulou}} and
\beditor{\binits{S.}~\bsnm{Kallel}}, eds,
\bsertitle{Lecture {Notes} in {Computer} {Science}},
\bpublisher{Springer International Publishing},
\byear{2020},
pp.~\bfpage{197}--\blpage{209}.
doi:\doiurl{10.1007/978-3-030-51517-1\_16}.
\end{bchapter}
\endbibitem

\bibitem{hooda_semantic_2020}
\begin{bchapter}
\bauthor{\binits{D.}~\bsnm{Hooda}} and
\bauthor{\binits{R.}~\bsnm{Rani}},
\bctitle{Semantic {Driven} {Healthcare} {Monitoring} and {Disease} {Detection} {Framework} from {Heterogeneous} {Sensor} {Data}},
in: \bbtitle{2020 {Sixth} {International} {Conference} on {Parallel}, {Distributed} and {Grid} {Computing} ({PDGC})},
\byear{2020},
pp.~\bfpage{415}--\blpage{420}.
doi:\doiurl{10.1109/PDGC50313.2020.9315793}.
\end{bchapter}
\endbibitem

\bibitem{ivascu_activity-aware_2021}
\begin{barticle}
\bauthor{\binits{T.}~\bsnm{Ivașcu}} and
\bauthor{\binits{V.}~\bsnm{Negru}},
\batitle{Activity-{Aware} {Vital} {Sign} {Monitoring} {Based} on a {Multi}-{Agent} {Architecture}},
\bjtitle{Sensors}
\bvolume{21}(\bissue{12})
(\byear{2021}),
\bfpage{4181}.
doi:\doiurl{10.3390/S21124181}.
\end{barticle}
\endbibitem

\bibitem{ivascu_multi-agent_2015}
\begin{bchapter}
\bauthor{\binits{T.}~\bsnm{Ivascu}},
\bauthor{\binits{B.}~\bsnm{Manate}} and
\bauthor{\binits{V.}~\bsnm{Negru}},
\bctitle{A {Multi}-agent {Architecture} for {Ontology}-{Based} {Diagnosis} of {Mental} {Disorders}},
in: \bbtitle{2015 17th {International} {Symposium} on {Symbolic} and {Numeric} {Algorithms} for {Scientific} {Computing} ({SYNASC})},
\byear{2015},
pp.~\bfpage{423}--\blpage{430}.
doi:\doiurl{10.1109/SYNASC.2015.69}.
\end{bchapter}
\endbibitem

\bibitem{kilintzis_supporting_2019}
\begin{barticle}
\bauthor{\binits{V.}~\bsnm{Kilintzis}},
\bauthor{\binits{I.}~\bsnm{Chouvarda}},
\bauthor{\binits{N.}~\bsnm{Beredimas}},
\bauthor{\binits{P.}~\bsnm{Natsiavas}} and
\bauthor{\binits{N.}~\bsnm{Maglaveras}},
\batitle{Supporting integrated care with a flexible data management framework built upon {Linked} {Data}, {HL7} {FHIR} and ontologies},
\bjtitle{Journal of Biomedical Informatics}
\bvolume{94}
(\byear{2019}),
\bfpage{103179}.
doi:\doiurl{10.1016/j.jbi.2019.103179}.
\end{barticle}
\endbibitem

\bibitem{spoladore_ontology-based_2021}
\begin{barticle}
\bauthor{\binits{D.}~\bsnm{Spoladore}},
\bauthor{\binits{V.}~\bsnm{Colombo}},
\bauthor{\binits{S.}~\bsnm{Arlati}},
\bauthor{\binits{A.}~\bsnm{Mahroo}},
\bauthor{\binits{A.}~\bsnm{Trombetta}} and
\bauthor{\binits{M.}~\bsnm{Sacco}},
\batitle{An {Ontology}-{Based} {Framework} for a {Telehealthcare} {System} to {Foster} {Healthy} {Nutrition} and {Active} {Lifestyle} in {Older} {Adults}},
\bjtitle{Electronics}
\bvolume{10}(\bissue{17})
(\byear{2021}),
\bfpage{2129}.
doi:\doiurl{10.3390/electronics10172129}.
\end{barticle}
\endbibitem

\bibitem{stavropoulos_detection_2021}
\begin{barticle}
\bauthor{\binits{T.G.}~\bsnm{Stavropoulos}},
\bauthor{\binits{G.}~\bsnm{Meditskos}},
\bauthor{\binits{I.}~\bsnm{Lazarou}},
\bauthor{\binits{L.}~\bsnm{Mpaltadoros}},
\bauthor{\binits{S.}~\bsnm{Papagiannopoulos}},
\bauthor{\binits{M.}~\bsnm{Tsolaki}} and
\bauthor{\binits{I.}~\bsnm{Kompatsiaris}},
\batitle{Detection of {Health}-{Related} {Events} and {Behaviours} from {Wearable} {Sensor} {Lifestyle} {Data} {Using} {Symbolic} {Intelligence}: {A} {Proof}-of-{Concept} {Application} in the {Care} of {Multiple} {Sclerosis}},
\bjtitle{Sensors}
\bvolume{21}(\bissue{18})
(\byear{2021}),
\bfpage{6230}.
doi:\doiurl{10.3390/S21186230}.
\end{barticle}
\endbibitem

\bibitem{titi_ontology-based_2019}
\begin{bchapter}
\bauthor{\binits{S.}~\bsnm{Titi}},
\bauthor{\binits{H.B.}~\bsnm{Elhadj}} and
\bauthor{\binits{L.}~\bsnm{Chaari}},
\bctitle{An ontology-based healthcare monitoring system in the internet of things},
in: \bbtitle{Proceedings of the 15th {International} {Wireless} {Communications} and {Mobile} {Computing} {Conference} ({IWCMC} 2019)},
\byear{2019},
pp.~\bfpage{319}--\blpage{324}.
doi:\doiurl{10.1109/IWCMC.2019.8766510}.
\end{bchapter}
\endbibitem

\bibitem{vadillo_enhancement_2013}
\begin{bchapter}
\bauthor{\binits{L.}~\bsnm{Vadillo}},
\bauthor{\binits{M.A.}~\bsnm{Valero}} and
\bauthor{\binits{G.}~\bsnm{Gil}},
\bctitle{Enhancement of a body area network to support smart health monitoring at the digital home},
in: \bbtitle{Proceedings of the 8th {International} {Conference} on {Body} {Area} {Networks}},
\byear{2013},
pp.~\bfpage{213}--\blpage{216}.
ISBN \bisbn{978-1-936968-89-3}.
doi:\doiurl{10.4108/icst.bodynets.2013.253578}.
\end{bchapter}
\endbibitem

\bibitem{villarreal_mobile_2014}
\begin{barticle}
\bauthor{\binits{V.}~\bsnm{Villarreal}},
\bauthor{\binits{J.}~\bsnm{Fontecha}},
\bauthor{\binits{R.}~\bsnm{Hervas}} and
\bauthor{\binits{J.}~\bsnm{Bravo}},
\batitle{Mobile and ubiquitous architecture for the medical control of chronic diseases through the use of intelligent devices: {Using} the architecture for patients with diabetes},
\bjtitle{Future Generation Computer Systems}
\bvolume{34}
(\byear{2014}),
\bfpage{161}--\blpage{175}.
doi:\doiurl{10.1016/j.future.2013.12.013}.
\end{barticle}
\endbibitem

\bibitem{zhou_design_2022}
\begin{bchapter}
\bauthor{\binits{F.}~\bsnm{Zhou}},
\bauthor{\binits{X.}~\bsnm{Wan}},
\bauthor{\binits{X.}~\bsnm{Du}},
\bauthor{\binits{Z.}~\bsnm{Lu}} and
\bauthor{\binits{J.}~\bsnm{Wu}},
\bctitle{Design and {Implementation} of {An} {Intelligent} {Health} {Management} {System} for {Nursing} {Homes}},
in: \bbtitle{Proceedings of the 2022 {IEEE} 7th {International} {Conference} on {Smart} {Cloud}},
\byear{2022},
pp.~\bfpage{145}--\blpage{150}.
doi:\doiurl{10.1109/SmartCloud55982.2022.00029}.
\end{bchapter}
\endbibitem

\bibitem{yu_improving_2022}
\begin{barticle}
\bauthor{\binits{G.}~\bsnm{Yu}},
\bauthor{\binits{M.}~\bsnm{Tabatabaei}},
\bauthor{\binits{J.}~\bsnm{Mezei}},
\bauthor{\binits{Q.}~\bsnm{Zhong}},
\bauthor{\binits{S.}~\bsnm{Chen}},
\bauthor{\binits{Z.}~\bsnm{Li}},
\bauthor{\binits{J.}~\bsnm{Li}},
\bauthor{\binits{L.}~\bsnm{Shu}} and
\bauthor{\binits{Q.}~\bsnm{Shu}},
\batitle{Improving chronic disease management for children with knowledge graphs and artificial intelligence},
\bjtitle{Expert Systems with Applications}
\bvolume{201}
(\byear{2022}),
\bfpage{117026}.
doi:\doiurl{10.1016/j.eswa.2022.117026}.
\end{barticle}
\endbibitem

\bibitem{xu_design_2017}
\begin{barticle}
\bauthor{\binits{B.}~\bsnm{Xu}},
\bauthor{\binits{L.}~\bsnm{Xu}},
\bauthor{\binits{H.}~\bsnm{Cai}},
\bauthor{\binits{L.}~\bsnm{Jiang}},
\bauthor{\binits{Y.}~\bsnm{Luo}} and
\bauthor{\binits{Y.}~\bsnm{Gu}},
\batitle{The design of an m-{Health} monitoring system based on a cloud computing platform},
\bjtitle{Enterprise Information Systems}
\bvolume{11}(\bissue{1})
(\byear{2017}),
\bfpage{17}--\blpage{36}.
doi:\doiurl{10.1080/17517575.2015.1053416}.
\end{barticle}
\endbibitem

\bibitem{reda_heterogeneous_2022}
\begin{barticle}
\bauthor{\binits{R.}~\bsnm{Reda}},
\bauthor{\binits{F.}~\bsnm{Piccinini}},
\bauthor{\binits{G.}~\bsnm{Martinelli}} and
\bauthor{\binits{A.}~\bsnm{Carbonaro}},
\batitle{Heterogeneous self-tracked health and fitness data integration and sharing according to a linked open data approach},
\bjtitle{Computing}
\bvolume{104}(\bissue{4})
(\byear{2022}),
\bfpage{835}--\blpage{857}.
doi:\doiurl{10.1007/s00607-021-00988-w}.
\end{barticle}
\endbibitem

\bibitem{zafeiropoulos_evaluating_2024}
\begin{barticle}
\bauthor{\binits{N.}~\bsnm{Zafeiropoulos}},
\bauthor{\binits{P.}~\bsnm{Bitilis}},
\bauthor{\binits{G.E.}~\bsnm{Tsekouras}} and
\bauthor{\binits{K.}~\bsnm{Kotis}},
\batitle{Evaluating {Ontology}-{Based} {PD} {Monitoring} and {Alerting} in {Personal} {Health} {Knowledge} {Graphs} and {Graph} {Neural} {Networks}},
\bjtitle{Information}
\bvolume{15}(\bissue{2})
(\byear{2024}),
\bfpage{100}.
doi:\doiurl{10.3390/info15020100}.
\end{barticle}
\endbibitem

\bibitem{ammar_using_2021}
\begin{barticle}
\bauthor{\binits{N.}~\bsnm{Ammar}},
\bauthor{\binits{J.E.}~\bsnm{Bailey}},
\bauthor{\binits{R.L.}~\bsnm{Davis}} and
\bauthor{\binits{A.}~\bsnm{Shaban-Nejad}},
\batitle{Using a {Personal} {Health} {Library}–{Enabled} {mHealth} {Recommender} {System} for {Self}-{Management} of {Diabetes} {Among} {Underserved} {Populations}: {Use} {Case} for {Knowledge} {Graphs} and {Linked} {Data}},
\bjtitle{JMIR Formative Research}
\bvolume{5}(\bissue{3})
(\byear{2021}),
\bfpage{e24738}.
doi:\doiurl{10.2196/24738}.
\end{barticle}
\endbibitem

\bibitem{akhtar_multi-agent_2022}
\begin{botherref}
\oauthor{\binits{S.M.}~\bsnm{Akhtar}},
\oauthor{\binits{M.}~\bsnm{Nazir}},
\oauthor{\binits{K.}~\bsnm{Saleem}},
\oauthor{\binits{R.Z.}~\bsnm{Ahmad}},
\oauthor{\binits{A.R.}~\bsnm{Javed}},
\oauthor{\binits{S.}~\bsnm{S.~Band}} and
\oauthor{\binits{A.}~\bsnm{Mosavi}},
A {Multi}-{Agent} {Formalism} {Based} on {Contextual} {Defeasible} {Logic} for {Healthcare} {Systems},
\textit{Frontiers in Public Health}
\textbf{10}
(2022).
\end{botherref}
\endbibitem

\bibitem{ali_intelligent_2021}
\begin{barticle}
\bauthor{\binits{F.}~\bsnm{Ali}},
\bauthor{\binits{S.}~\bsnm{El-Sappagh}},
\bauthor{\binits{S.M.R.}~\bsnm{Islam}},
\bauthor{\binits{A.}~\bsnm{Ali}},
\bauthor{\binits{M.}~\bsnm{Attique}},
\bauthor{\binits{M.}~\bsnm{Imran}} and
\bauthor{\binits{K.-S.}~\bsnm{Kwak}},
\batitle{An intelligent healthcare monitoring framework using wearable sensors and social networking data},
\bjtitle{Future Generation Computer Systems}
\bvolume{114}
(\byear{2021}),
\bfpage{23}--\blpage{43}.
doi:\doiurl{10.1016/j.future.2020.07.047}.
\end{barticle}
\endbibitem

\bibitem{ali_smart_2020}
\begin{barticle}
\bauthor{\binits{F.}~\bsnm{Ali}},
\bauthor{\binits{S.}~\bsnm{El-Sappagh}},
\bauthor{\binits{S.M.R.}~\bsnm{Islam}},
\bauthor{\binits{D.}~\bsnm{Kwak}},
\bauthor{\binits{A.}~\bsnm{Ali}},
\bauthor{\binits{M.}~\bsnm{Imran}} and
\bauthor{\binits{K.-S.}~\bsnm{Kwak}},
\batitle{A smart healthcare monitoring system for heart disease prediction based on ensemble deep learning and feature fusion},
\bjtitle{Information Fusion}
\bvolume{63}
(\byear{2020}),
\bfpage{208}--\blpage{222}.
doi:\doiurl{10.1016/j.inffus.2020.06.008}.
\end{barticle}
\endbibitem

\bibitem{ali_type-2_2018}
\begin{barticle}
\bauthor{\binits{F.}~\bsnm{Ali}},
\bauthor{\binits{S.M.R.}~\bsnm{Islam}},
\bauthor{\binits{D.}~\bsnm{Kwak}},
\bauthor{\binits{P.}~\bsnm{Khan}},
\bauthor{\binits{N.}~\bsnm{Ullah}},
\bauthor{\binits{S.-j.}~\bsnm{Yoo}} and
\bauthor{\binits{K.S.}~\bsnm{Kwak}},
\batitle{Type-2 fuzzy ontology–aided recommendation systems for {IoT}–based healthcare},
\bjtitle{Computer Communications}
\bvolume{119}
(\byear{2018}),
\bfpage{138}--\blpage{155}.
doi:\doiurl{10.1016/j.comcom.2017.10.005}.
\end{barticle}
\endbibitem

\bibitem{alti_agent-based_2021}
\begin{barticle}
\bauthor{\binits{A.}~\bsnm{Alti}} and
\bauthor{\binits{L.}~\bsnm{Laouamer}},
\batitle{Agent-{Based} {Autonomic} {Semantic} {Context}-{Aware} {Platform} for {Smart} {Health} {Monitoring} and {Disease} {Detection}},
\bjtitle{The Computer Journal}
\bvolume{65}(\bissue{3})
(\byear{2021}),
\bfpage{736}--\blpage{755}.
doi:\doiurl{10.1093/comjnl/bxab075}.
\end{barticle}
\endbibitem

\bibitem{chatterjee_automatic_2021}
\begin{barticle}
\bauthor{\binits{A.}~\bsnm{Chatterjee}},
\bauthor{\binits{A.}~\bsnm{Prinz}},
\bauthor{\binits{M.}~\bsnm{Gerdes}} and
\bauthor{\binits{S.}~\bsnm{Martinez}},
\batitle{An {Automatic} {Ontology}-{Based} {Approach} to {Support} {Logical} {Representation} of {Observable} and {Measurable} {Data} for {Healthy} {Lifestyle} {Management}: {Proof}-of-{Concept} {Study}},
\bjtitle{Journal of Medical Internet Research}
\bvolume{23}(\bissue{4})
(\byear{2021}),
\bfpage{e24656}.
doi:\doiurl{10.2196/24656}.
\end{barticle}
\endbibitem

\bibitem{de_brouwer_mbrain_2022}
\begin{barticle}
\bauthor{\binits{M.}~\bsnm{De~Brouwer}},
\bauthor{\binits{N.}~\bsnm{Vandenbussche}},
\bauthor{\binits{B.}~\bsnm{Steenwinckel}},
\bauthor{\binits{M.}~\bsnm{Stojchevska}},
\bauthor{\binits{J.}~\bsnm{Van Der~Donckt}},
\bauthor{\binits{V.}~\bsnm{Degraeve}},
\bauthor{\binits{J.}~\bsnm{Vaneessen}},
\bauthor{\binits{F.}~\bsnm{De~Turck}},
\bauthor{\binits{B.}~\bsnm{Volckaert}},
\bauthor{\binits{P.}~\bsnm{Boon}},
\bauthor{\binits{K.}~\bsnm{Paemeleire}},
\bauthor{\binits{S.}~\bsnm{Van~Hoecke}} and
\bauthor{\binits{F.}~\bsnm{Ongenae}},
\batitle{{mBrain}: towards the continuous follow-up and headache classification of primary headache disorder patients},
\bjtitle{BMC Medical Informatics and Decision Making}
\bvolume{22}(\bissue{1})
(\byear{2022}),
\bfpage{87}.
doi:\doiurl{10.1186/s12911-022-01813-w}.
\end{barticle}
\endbibitem

\bibitem{el-sappagh_mobile_2019}
\begin{barticle}
\bauthor{\binits{S.}~\bsnm{El-Sappagh}},
\bauthor{\binits{F.}~\bsnm{Ali}},
\bauthor{\binits{A.}~\bsnm{Hendawi}},
\bauthor{\binits{J.H.}~\bsnm{Jang}} and
\bauthor{\binits{K.S.}~\bsnm{Kwak}},
\batitle{A mobile health monitoring-and-treatment system based on integration of the {SSN} sensor ontology and the {HL7} {FHIR} standard},
\bjtitle{BMC Medical Informatics and Decision Making}
\bvolume{19}(\bissue{1})
(\byear{2019}),
\bfpage{1}--\blpage{36}.
doi:\doiurl{10.1186/s12911-019-0806-z}.
\end{barticle}
\endbibitem

\bibitem{fenza_hybrid_2012}
\begin{barticle}
\bauthor{\binits{G.}~\bsnm{Fenza}},
\bauthor{\binits{D.}~\bsnm{Furno}} and
\bauthor{\binits{V.}~\bsnm{Loia}},
\batitle{Hybrid approach for context-aware service discovery in healthcare domain},
\bjtitle{Journal of Computer and System Sciences}
\bvolume{78}(\bissue{4})
(\byear{2012}),
\bfpage{1232}--\blpage{1247}.
doi:\doiurl{10.1016/j.jcss.2011.10.011}.
\end{barticle}
\endbibitem

\bibitem{hadjadj_integration_2021}
\begin{barticle}
\bauthor{\binits{A.}~\bsnm{Hadjadj}} and
\bauthor{\binits{K.}~\bsnm{Halimi}},
\batitle{An {Integration} of {Health} {Monitoring} {System} in {Public} {Transport} {Using} the {Semantic} {Web} of {Things}},
\bjtitle{Journal of Universal Computer Science}
\bvolume{27}(\bissue{12})
(\byear{2021}),
\bfpage{1325}--\blpage{1346}.
doi:\doiurl{10.3897/JUCS.76983}.
\end{barticle}
\endbibitem

\bibitem{hristoskova_ontology-driven_2014}
\begin{barticle}
\bauthor{\binits{A.}~\bsnm{Hristoskova}},
\bauthor{\binits{V.}~\bsnm{Sakkalis}},
\bauthor{\binits{G.}~\bsnm{Zacharioudakis}},
\bauthor{\binits{M.}~\bsnm{Tsiknakis}} and
\bauthor{\binits{F.}~\bsnm{De~Turck}},
\batitle{Ontology-{Driven} {Monitoring} of {Patient}’s {Vital} {Signs} {Enabling} {Personalized} {Medical} {Detection} and {Alert}},
\bjtitle{Sensors}
\bvolume{14}(\bissue{1})
(\byear{2014}),
\bfpage{1598}--\blpage{1628}.
doi:\doiurl{10.3390/s140101598}.
\end{barticle}
\endbibitem

\bibitem{hussain_big-ecg_2021}
\begin{barticle}
\bauthor{\binits{I.}~\bsnm{Hussain}} and
\bauthor{\binits{S.J.}~\bsnm{Park}},
\batitle{Big-{ECG}: {Cardiographic} {Predictive} {Cyber}-{Physical} {System} for {Stroke} {Management}},
\bjtitle{IEEE Access}
\bvolume{9}
(\byear{2021}),
\bfpage{123146}--\blpage{123164}.
doi:\doiurl{10.1109/ACCESS.2021.3109806}.
\end{barticle}
\endbibitem

\bibitem{khozouie_ontological_2018}
\begin{barticle}
\bauthor{\binits{N.}~\bsnm{Khozouie}},
\bauthor{\binits{F.}~\bsnm{Fotouhi-Ghazvini}} and
\bauthor{\binits{B.}~\bsnm{Minaei-Bidgoli}},
\batitle{Ontological mobihealth system},
\bjtitle{Indonesian Journal of Electrical Engineering and Computer Science}
\bvolume{10}(\bissue{1})
(\byear{2018}),
\bfpage{309}--\blpage{319}.
doi:\doiurl{10.11591/ijeecs.v10.i1.pp309-319}.
\end{barticle}
\endbibitem

\bibitem{kim_ontology_2014}
\begin{barticle}
\bauthor{\binits{J.}~\bsnm{Kim}},
\bauthor{\binits{J.}~\bsnm{Kim}},
\bauthor{\binits{D.}~\bsnm{Lee}} and
\bauthor{\binits{K.-Y.}~\bsnm{Chung}},
\batitle{Ontology driven interactive healthcare with wearable sensors},
\bjtitle{Multimedia Tools and Applications}
\bvolume{71}(\bissue{2})
(\byear{2014}),
\bfpage{827}--\blpage{841}.
doi:\doiurl{10.1007/s11042-012-1195-9}.
\end{barticle}
\endbibitem

\bibitem{kordestani_extended_2021}
\begin{barticle}
\bauthor{\binits{H.}~\bsnm{Kordestani}},
\bauthor{\binits{R.}~\bsnm{Mojarad}},
\bauthor{\binits{A.}~\bsnm{Chibani}},
\bauthor{\binits{K.}~\bsnm{Barkaoui}},
\bauthor{\binits{Y.}~\bsnm{Amirat}} and
\bauthor{\binits{W.}~\bsnm{Zahran}},
\batitle{Extended {Hapicare}: {A} telecare system with probabilistic diagnosis and self-adaptive treatment},
\bjtitle{Expert Systems with Applications}
\bvolume{186}
(\byear{2021}),
\bfpage{115749}.
doi:\doiurl{10.1016/j.eswa.2021.115749}.
\end{barticle}
\endbibitem

\bibitem{lopes_ontology-driven_2023}
\begin{bchapter}
\bauthor{\binits{P.}~\bsnm{Lopes~de Souza}},
\bauthor{\binits{W.}~\bsnm{Lopes~de Souza}},
\bauthor{\binits{L.}~\bsnm{Ferreira~Pires}},
\bauthor{\binits{J.}~\bsnm{Moreira}},
\bauthor{\binits{R.}~\bsnm{Rodrigues}} and
\bauthor{\binits{R.}~\bsnm{Ciferri}},
\bctitle{Ontology-{Driven} {IoT} {System} for {Monitoring} {Hypertension}},
in: \bbtitle{Proceedings of the 25th {International} {Conference} on {Enterprise} {Information} {Systems}},
\bpublisher{SCITEPRESS - Science and Technology Publications},
\byear{2023},
pp.~\bfpage{757}--\blpage{767}.
ISBN \bisbn{978-989-758-648-4}.
doi:\doiurl{10.5220/0011989100003467}.
\end{bchapter}
\endbibitem

\bibitem{martella_semantically_2025}
\begin{barticle}
\bauthor{\binits{A.}~\bsnm{Martella}},
\bauthor{\binits{A.}~\bsnm{Longo}},
\bauthor{\binits{M.}~\bsnm{Zappatore}},
\bauthor{\binits{B.D.}~\bsnm{Martino}} and
\bauthor{\binits{A.}~\bsnm{Esposito}},
\batitle{A semantically enabled architecture for interoperable edge-cloud continuum applied to the e-health scenario},
\bjtitle{Software: Practice and Experience}
\bvolume{55}(\bissue{3})
(\byear{2025}),
\bfpage{409}--\blpage{447}.
doi:\doiurl{10.1002/spe.3375}.
\end{barticle}
\endbibitem

\bibitem{mavropoulos_smart_2021}
\begin{barticle}
\bauthor{\binits{T.}~\bsnm{Mavropoulos}},
\bauthor{\binits{S.}~\bsnm{Symeonidis}},
\bauthor{\binits{A.}~\bsnm{Tsanousa}},
\bauthor{\binits{P.}~\bsnm{Giannakeris}},
\bauthor{\binits{M.}~\bsnm{Rousi}},
\bauthor{\binits{E.}~\bsnm{Kamateri}},
\bauthor{\binits{G.}~\bsnm{Meditskos}},
\bauthor{\binits{K.}~\bsnm{Ioannidis}},
\bauthor{\binits{S.}~\bsnm{Vrochidis}} and
\bauthor{\binits{I.}~\bsnm{Kompatsiaris}},
\batitle{Smart integration of sensors, computer vision and knowledge representation for intelligent monitoring and verbal human-computer interaction},
\bjtitle{Journal of Intelligent Information Systems}
\bvolume{57}(\bissue{2})
(\byear{2021}),
\bfpage{321}--\blpage{345}.
doi:\doiurl{10.1007/s10844-021-00648-7}.
\end{barticle}
\endbibitem

\bibitem{mcheick_stroke_2016}
\begin{bchapter}
\bauthor{\binits{H.}~\bsnm{Mcheick}},
\bauthor{\binits{H.}~\bsnm{Nasser}},
\bauthor{\binits{M.}~\bsnm{Dbouk}} and
\bauthor{\binits{A.}~\bsnm{Nasser}},
\bctitle{Stroke {Prediction} {Context}-{Aware} {Health} {Care} {System}},
in: \bbtitle{2016 {IEEE} {International} {Conference} on {Connected} {Health}: {Applications}, {Systems} and {Engineering} {Technologies} ({CHASE})},
\byear{2016},
pp.~\bfpage{30}--\blpage{35}.
doi:\doiurl{10.1109/CHASE.2016.49}.
\end{bchapter}
\endbibitem

\bibitem{mezghani_semantic_2015}
\begin{barticle}
\bauthor{\binits{E.}~\bsnm{Mezghani}},
\bauthor{\binits{E.}~\bsnm{Exposito}},
\bauthor{\binits{K.}~\bsnm{Drira}},
\bauthor{\binits{M.}~\bsnm{Da~Silveira}} and
\bauthor{\binits{C.}~\bsnm{Pruski}},
\batitle{A {Semantic} {Big} {Data} {Platform} for {Integrating} {Heterogeneous} {Wearable} {Data} in {Healthcare}},
\bjtitle{Journal of Medical Systems}
\bvolume{39}(\bissue{12})
(\byear{2015}),
\bfpage{185}.
doi:\doiurl{10.1007/s10916-015-0344-x}.
\end{barticle}
\endbibitem

\bibitem{minutolo_hybrid_2016}
\begin{bchapter}
\bauthor{\binits{A.}~\bsnm{Minutolo}},
\bauthor{\binits{M.}~\bsnm{Esposito}} and
\bauthor{\binits{G.}~\bsnm{De~Pietro}},
\bctitle{A hybrid reasoning system for mobile and intelligent health services},
in: \bbtitle{2016 {IEEE} {International} {Conference} on {Systems}, {Man}, and {Cybernetics} ({SMC})},
\byear{2016},
pp.~\bfpage{003399}--\blpage{003404}.
doi:\doiurl{10.1109/SMC.2016.7844759}.
\end{bchapter}
\endbibitem

\bibitem{peral_ontology-oriented_2018}
\begin{barticle}
\bauthor{\binits{J.}~\bsnm{Peral}},
\bauthor{\binits{A.}~\bsnm{Ferrandez}},
\bauthor{\binits{D.}~\bsnm{Gil}},
\bauthor{\binits{R.}~\bsnm{Munoz-Terol}} and
\bauthor{\binits{H.}~\bsnm{Mora}},
\batitle{An {Ontology}-{Oriented} {Architecture} for {Dealing} {With} {Heterogeneous} {Data} {Applied} to {Telemedicine} {Systems}},
\bjtitle{IEEE Access}
\bvolume{6}
(\byear{2018}),
\bfpage{41118}--\blpage{41138}.
doi:\doiurl{10.1109/ACCESS.2018.2857499}.
\end{barticle}
\endbibitem

\bibitem{yu_semantic_2017}
\begin{bchapter}
\bauthor{\binits{H.Q.}~\bsnm{Yu}},
\bauthor{\binits{X.}~\bsnm{Zhao}},
\bauthor{\binits{Z.}~\bsnm{Deng}} and
\bauthor{\binits{F.}~\bsnm{Dong}},
\bctitle{Semantic {Lifting} and {Reasoning} on the {Personalised} {Activity} {Big} {Data} {Repository} for {Healthcare} {Research}},
in: \bbtitle{1st {International} {Workshop} on {Internet} of {Things} and {Big} {Data} for {Healthcare} ({IoTBDH} 2017)},
\byear{2017},
pp.~\bfpage{818}--\blpage{823}.
doi:\doiurl{10.1109/iThings-GreenCom-CPSCom-SmartData.2017.125}.
\end{bchapter}
\endbibitem

\bibitem{zeshan_iot-enabled_2023}
\begin{barticle}
\bauthor{\binits{F.}~\bsnm{Zeshan}},
\bauthor{\binits{A.}~\bsnm{Ahmad}},
\bauthor{\binits{M.I.}~\bsnm{Babar}},
\bauthor{\binits{M.}~\bsnm{Hamid}},
\bauthor{\binits{F.}~\bsnm{Hajjej}} and
\bauthor{\binits{M.}~\bsnm{Ashraf}},
\batitle{An {IoT}-{Enabled} {Ontology}-{Based} {Intelligent} {Healthcare} {Framework} for {Remote} {Patient} {Monitoring}},
\bjtitle{IEEE Access}
\bvolume{11}
(\byear{2023}),
\bfpage{133947}--\blpage{133966}.
doi:\doiurl{10.1109/ACCESS.2023.3332708}.
\end{barticle}
\endbibitem

\bibitem{zhang_knowledge-based_2014}
\begin{barticle}
\bauthor{\binits{W.}~\bsnm{Zhang}},
\bauthor{\binits{K.}~\bsnm{Thurow}} and
\bauthor{\binits{R.}~\bsnm{Stoll}},
\batitle{A {Knowledge}-based {Telemonitoring} {Platform} for {Application} in {Remote} {Healthcare}},
\bjtitle{INTERNATIONAL JOURNAL OF COMPUTERS COMMUNICATIONS \& CONTROL}
\bvolume{9}(\bissue{5})
(\byear{2014}),
\bfpage{644}--\blpage{654}.
\end{barticle}
\endbibitem

\bibitem{benson_principles_2021}
\begin{bbook}
\bauthor{\binits{T.}~\bsnm{Benson}} and
\bauthor{\binits{G.}~\bsnm{Grieve}},
\bbtitle{Principles of {Health} {Interoperability}: {FHIR}, {HL7} and {SNOMED} {CT}},
\bedition{4}th edn,
\bsertitle{Health {Information} {Technology} {Standards}},
\bpublisher{Springer},
\byear{2021}.
ISBN \bisbn{978-3-030-56882-5 978-3-030-56883-2}.
doi:\doiurl{10.1007/978-3-030-56883-2}.
\end{bbook}
\endbibitem

\bibitem{gibbons_coming_2007}
\begin{botherref}
\oauthor{\binits{P.}~\bsnm{Gibbons}},
\oauthor{\binits{N.}~\bsnm{Arzt}},
\oauthor{\binits{S.}~\bsnm{Burke-Beebe}},
\oauthor{\binits{C.}~\bsnm{Chute}},
\oauthor{\binits{G.}~\bsnm{Dickinson}},
\oauthor{\binits{T.}~\bsnm{Flewelling}},
\oauthor{\binits{T.}~\bsnm{Jepsen}},
\oauthor{\binits{D.}~\bsnm{Kamens}},
\oauthor{\binits{J.}~\bsnm{Larson}},
\oauthor{\binits{J.}~\bsnm{Ritter}},
\oauthor{\binits{M.}~\bsnm{Rozen}},
\oauthor{\binits{S.}~\bsnm{Selover}} and
\oauthor{\binits{J.}~\bsnm{Stanford}},
Coming to {Terms}: {Scoping} {Interoperability} for {Health} {Care},
Technical Report,
Health Level Seven EHR Interoperability Work Group,
2007.
\end{botherref}
\endbibitem

\bibitem{hosseini_chapter_2016}
\begin{bchapter}
\bauthor{\binits{M.}~\bsnm{Hosseini}} and
\bauthor{\binits{B.E.}~\bsnm{Dixon}},
\bctitle{Chapter 8 - {Syntactic} {Interoperability} and the {Role} of {Standards}},
in: \bbtitle{Health {Information} {Exchange}},
\bedition{1}st edn,
\beditor{\binits{B.E.}~\bsnm{Dixon}}, ed.,
\bpublisher{Academic Press},
\byear{2016},
pp.~\bfpage{123}--\blpage{136}.
ISBN \bisbn{978-0-12-803135-3}.
doi:\doiurl{10.1016/B978-0-12-803135-3.00008-6}.
\end{bchapter}
\endbibitem

\bibitem{kuziemsky_framework_2016}
\begin{barticle}
\bauthor{\binits{C.E.}~\bsnm{Kuziemsky}} and
\bauthor{\binits{L.}~\bsnm{Peyton}},
\batitle{A framework for understanding process interoperability and health information technology},
\bjtitle{Health Policy and Technology}
\bvolume{5}(\bissue{2})
(\byear{2016}),
\bfpage{196}--\blpage{203}.
doi:\doiurl{10.1016/j.hlpt.2016.02.007}.
\end{barticle}
\endbibitem

\bibitem{rahmani_exploiting_2018}
\begin{barticle}
\bauthor{\binits{A.M.}~\bsnm{Rahmani}},
\bauthor{\binits{T.N.}~\bsnm{Gia}},
\bauthor{\binits{B.}~\bsnm{Negash}},
\bauthor{\binits{A.}~\bsnm{Anzanpour}},
\bauthor{\binits{I.}~\bsnm{Azimi}},
\bauthor{\binits{M.}~\bsnm{Jiang}} and
\bauthor{\binits{P.}~\bsnm{Liljeberg}},
\batitle{Exploiting smart e-{Health} gateways at the edge of healthcare {Internet}-of-{Things}: {A} fog computing approach},
\bjtitle{Future Generation Computer Systems}
\bvolume{78}
(\byear{2018}),
\bfpage{641}--\blpage{658}.
doi:\doiurl{10.1016/j.future.2017.02.014}.
\end{barticle}
\endbibitem

\bibitem{veer_achieving_2008}
\begin{botherref}
\oauthor{\binits{H.v.d.}~\bsnm{Veer}} and
\oauthor{\binits{A.}~\bsnm{Wiles}},
Achieving {Technical} {Interoperability} - the {ETSI} {Approach},
White {Paper},
European Telecommunications Standards Institute (ETSI),
2008.
\end{botherref}
\endbibitem

\bibitem{sheth_semantic_2008}
\begin{barticle}
\bauthor{\binits{A.}~\bsnm{Sheth}},
\bauthor{\binits{C.}~\bsnm{Henson}} and
\bauthor{\binits{S.S.}~\bsnm{Sahoo}},
\batitle{Semantic sensor web},
\bjtitle{IEEE Internet Computing}
\bvolume{12}(\bissue{4})
(\byear{2008}),
\bfpage{78}--\blpage{83}.
doi:\doiurl{10.1109/MIC.2008.87}.
\end{barticle}
\endbibitem

\bibitem{bodenreider_unified_2014}
\begin{barticle}
\bauthor{\binits{O.}~\bsnm{Bodenreider}},
\batitle{The {Unified} {Medical} {Language} {System} ({UMLS}): integrating biomedical terminology},
\bjtitle{Nucleic Acids Research}
\bvolume{32}(\bissue{Database issue})
(\byear{2014}),
\bfpage{D267}--\blpage{D270}.
doi:\doiurl{10.1093/nar/gkh061}.
\end{barticle}
\endbibitem

\bibitem{amaral_foundational_2021}
\begin{barticle}
\bauthor{\binits{G.}~\bsnm{Amaral}},
\bauthor{\binits{F.}~\bsnm{Baião}} and
\bauthor{\binits{G.}~\bsnm{Guizzardi}},
\batitle{Foundational ontologies, ontology-driven conceptual modeling, and their multiple benefits to data mining},
\bjtitle{WIREs Data Mining and Knowledge Discovery}
\bvolume{11}(\bissue{4})
(\byear{2021}),
\bfpage{e1408}.
doi:\doiurl{10.1002/widm.1408}.
\end{barticle}
\endbibitem

\bibitem{daniele_created_2015}
\begin{barticle}
\bauthor{\binits{L.}~\bsnm{Daniele}},
\bauthor{\binits{F.}~\bsnm{den Hartog}} and
\bauthor{\binits{J.}~\bsnm{Roes}},
\batitle{Created in {Close} {Interaction} with the {Industry}: {The} {Smart} {Appliances} {REFerence} ({SAREF}) {Ontology}},
\bjtitle{Lecture Notes in Business Information Processing}
\bvolume{225}
(\byear{2015}),
\bfpage{100}--\blpage{112}.
doi:\doiurl{10.1007/978-3-319-21545-7\_9}.
\end{barticle}
\endbibitem

\bibitem{sharma_implementing_2022}
\begin{bchapter}
\bauthor{\binits{D.K.}~\bsnm{Sharma}},
\bauthor{\binits{E.}~\bsnm{Prud'Hommeaux}},
\bauthor{\binits{D.}~\bsnm{Booth}},
\bauthor{\binits{K.J.}~\bsnm{Peterson}},
\bauthor{\binits{D.J.}~\bsnm{Stone}},
\bauthor{\binits{H.}~\bsnm{Solbrig}},
\bauthor{\binits{G.}~\bsnm{Xiao}},
\bauthor{\binits{E.}~\bsnm{Pfaff}} and
\bauthor{\binits{G.}~\bsnm{Jiang}},
\bctitle{Implementing a {New} {FHIR} {RDF} {Specification} for {Semantic} {Clinical} {Data} {Using} a {JSONLD}- based {Approach}},
in: \bbtitle{{CEUR} {Workshop} {Proceedings}},
Vol.~\bseriesno{3127},
\bpublisher{CEUR-WS},
\byear{2022},
pp.~\bfpage{82}--\blpage{86},
\bcomment{11th International Workshop on Enterprise Modeling and Information Systems Architectures (EMISA 2021)}.
\url{https://pure.johnshopkins.edu/en/publications/implementing-a-new-fhir-rdf-specification-for-semantic-clinical-d}.
\end{bchapter}
\endbibitem

\bibitem{hitzler_neural-symbolic_2020}
\begin{barticle}
\bauthor{\binits{P.}~\bsnm{Hitzler}},
\bauthor{\binits{F.}~\bsnm{Bianchi}},
\bauthor{\binits{M.}~\bsnm{Ebrahimi}} and
\bauthor{\binits{M.K.}~\bsnm{Sarker}},
\batitle{Neural-symbolic integration and the {Semantic} {Web}},
\bjtitle{Semantic Web}
\bvolume{11}(\bissue{1})
(\byear{2020}),
\bfpage{3}--\blpage{11}.
doi:\doiurl{10.3233/SW-190368}.
\end{barticle}
\endbibitem

\bibitem{ravi_deep_2017}
\begin{barticle}
\bauthor{\binits{D.}~\bsnm{Ravì}},
\bauthor{\binits{C.}~\bsnm{Wong}},
\bauthor{\binits{F.}~\bsnm{Deligianni}},
\bauthor{\binits{M.}~\bsnm{Berthelot}},
\bauthor{\binits{J.}~\bsnm{Andreu-Perez}},
\bauthor{\binits{B.}~\bsnm{Lo}} and
\bauthor{\binits{G.-Z.}~\bsnm{Yang}},
\batitle{Deep {Learning} for {Health} {Informatics}},
\bjtitle{IEEE Journal of Biomedical and Health Informatics}
\bvolume{21}(\bissue{1})
(\byear{2017}),
\bfpage{4}--\blpage{21}.
doi:\doiurl{10.1109/JBHI.2016.2636665}.
\end{barticle}
\endbibitem

\bibitem{li_comprehensive_2021}
\begin{barticle}
\bauthor{\binits{W.}~\bsnm{Li}},
\bauthor{\binits{Y.}~\bsnm{Chai}},
\bauthor{\binits{F.}~\bsnm{Khan}},
\bauthor{\binits{S.R.U.}~\bsnm{Jan}},
\bauthor{\binits{S.}~\bsnm{Verma}},
\bauthor{\binits{V.G.}~\bsnm{Menon}},
\bauthor{\bsnm{{Kavita}}} and
\bauthor{\binits{X.}~\bsnm{Li}},
\batitle{A {Comprehensive} {Survey} on {Machine} {Learning}-{Based} {Big} {Data} {Analytics} for {IoT}-{Enabled} {Smart} {Healthcare} {System}},
\bjtitle{Mobile Networks and Applications}
\bvolume{26}(\bissue{1})
(\byear{2021}),
\bfpage{234}--\blpage{252}.
doi:\doiurl{10.1007/s11036-020-01700-6}.
\end{barticle}
\endbibitem

\bibitem{kokar_ontology-based_2009}
\begin{barticle}
\bauthor{\binits{M.M.}~\bsnm{Kokar}},
\bauthor{\binits{C.J.}~\bsnm{Matheus}} and
\bauthor{\binits{K.}~\bsnm{Baclawski}},
\batitle{Ontology-based situation awareness},
\bjtitle{Information Fusion}
\bvolume{10}(\bissue{1})
(\byear{2009}),
\bfpage{83}--\blpage{98}.
doi:\doiurl{10.1016/j.inffus.2007.01.004}.
\end{barticle}
\endbibitem

\bibitem{almeida_towards_2018}
\begin{bchapter}
\bauthor{\binits{J.P.A.}~\bsnm{Almeida}},
\bauthor{\binits{P.D.}~\bsnm{Costa}} and
\bauthor{\binits{G.}~\bsnm{Guizzardi}},
\bctitle{Towards an {Ontology} of {Scenes} and {Situations}},
in: \bbtitle{2018 {IEEE} {Conference} on {Cognitive} and {Computational} {Aspects} of {Situation} {Management} ({CogSIMA})},
\byear{2018},
pp.~\bfpage{29}--\blpage{35}.
doi:\doiurl{10.1109/COGSIMA.2018.8423994}.
\end{bchapter}
\endbibitem

\bibitem{bobillo_fuzzy_2016}
\begin{barticle}
\bauthor{\binits{F.}~\bsnm{Bobillo}} and
\bauthor{\binits{U.}~\bsnm{Straccia}},
\batitle{The fuzzy ontology reasoner {fuzzyDL}},
\bjtitle{Knowledge-Based Systems}
\bvolume{95}
(\byear{2016}),
\bfpage{12}--\blpage{34}.
doi:\doiurl{10.1016/j.knosys.2015.11.017}.
\end{barticle}
\endbibitem

\bibitem{cash_alert_2009}
\begin{barticle}
\bauthor{\binits{J.J.}~\bsnm{Cash}},
\batitle{Alert fatigue},
\bjtitle{American Journal of Health-System Pharmacy}
\bvolume{66}(\bissue{23})
(\byear{2009}),
\bfpage{2098}--\blpage{2101}.
doi:\doiurl{10.2146/ajhp090181}.
\end{barticle}
\endbibitem

\bibitem{miller_explainable_2023}
\begin{bchapter}
\bauthor{\binits{T.}~\bsnm{Miller}},
\bctitle{Explainable {AI} is {Dead}, {Long} {Live} {Explainable} {AI}! {Hypothesis}-driven {Decision} {Support} using {Evaluative} {AI}},
in: \bbtitle{Proceedings of the 2023 {ACM} {Conference} on {Fairness}, {Accountability}, and {Transparency}},
\bpublisher{Association for Computing Machinery},
\byear{2023},
pp.~\bfpage{333}--\blpage{342}.
ISBN \bisbn{9798400701924}.
doi:\doiurl{10.1145/3593013.3594001}.
\end{bchapter}
\endbibitem

\bibitem{de_cremer_road_2022}
\begin{barticle}
\bauthor{\binits{D.}~\bsnm{De~Cremer}},
\bauthor{\binits{D.}~\bsnm{Narayanan}},
\bauthor{\binits{A.}~\bsnm{Deppeler}},
\bauthor{\binits{M.}~\bsnm{Nagpal}} and
\bauthor{\binits{J.}~\bsnm{McGuire}},
\batitle{The road to a human-centred digital society: opportunities, challenges and responsibilities for humans in the age of machines},
\bjtitle{AI and Ethics}
\bvolume{2}(\bissue{4})
(\byear{2022}),
\bfpage{579}--\blpage{583}.
doi:\doiurl{10.1007/s43681-021-00116-6}.
\end{barticle}
\endbibitem

\bibitem{blomqvist_use_2014}
\begin{barticle}
\bauthor{\binits{E.}~\bsnm{Blomqvist}},
\batitle{The use of {Semantic} {Web} technologies for decision support – a survey},
\bjtitle{Semantic Web}
\bvolume{5}(\bissue{3})
(\byear{2014}),
\bfpage{177}--\blpage{201}.
doi:\doiurl{10.3233/SW-2012-0084}.
\end{barticle}
\endbibitem

\bibitem{ye_ontology-based_2007}
\begin{barticle}
\bauthor{\binits{J.}~\bsnm{Ye}},
\bauthor{\binits{L.}~\bsnm{Coyle}},
\bauthor{\binits{S.}~\bsnm{Dobson}} and
\bauthor{\binits{P.}~\bsnm{Nixon}},
\batitle{Ontology-based models in pervasive computing systems},
\bjtitle{The Knowledge Engineering Review}
\bvolume{22}(\bissue{4})
(\byear{2007}),
\bfpage{315}--\blpage{347}.
doi:\doiurl{10.1017/S0269888907001208}.
\end{barticle}
\endbibitem

\bibitem{stevenson_ontonym_2009}
\begin{bchapter}
\bauthor{\binits{G.}~\bsnm{Stevenson}},
\bauthor{\binits{S.}~\bsnm{Knox}},
\bauthor{\binits{S.}~\bsnm{Dobson}} and
\bauthor{\binits{P.}~\bsnm{Nixon}},
\bctitle{Ontonym: a collection of upper ontologies for developing pervasive systems},
in: \bbtitle{Proceedings of the 1st {Workshop} on {Context}, {Information} and {Ontologies}},
\byear{2009},
p.~\bfpage{8}.
\end{bchapter}
\endbibitem

\bibitem{alagar_temporal_2011}
\begin{bchapter}
\bauthor{\binits{V.S.}~\bsnm{Alagar}} and
\bauthor{\binits{K.}~\bsnm{Periyasamy}},
\bctitle{Temporal {Logic}},
in: \bbtitle{Specification of {Software} {Systems}},
\bsertitle{Texts in Computer Science},
\bpublisher{Springer London},
\blocation{London},
\byear{2011},
pp.~\bfpage{177}--\blpage{229}.
ISBN \bisbn{978-0-85729-276-6 978-0-85729-277-3}.
doi:\doiurl{10.1007/978-0-85729-277-3\_11}.
\end{bchapter}
\endbibitem

\bibitem{wooldridge_intelligent_2013}
\begin{bchapter}
\bauthor{\binits{M.}~\bsnm{Wooldridge}},
\bctitle{Intelligent {Agents}},
in: \bbtitle{Multiagent {Systems}},
\bedition{2}nd edn,
\beditor{\binits{G.}~\bsnm{Weiss}}, ed.,
\bpublisher{MIT Press},
\byear{2013},
pp.~\bfpage{3}--\blpage{50}.
\end{bchapter}
\endbibitem

\bibitem{isern_systematic_2016}
\begin{barticle}
\bauthor{\binits{D.}~\bsnm{Isern}} and
\bauthor{\binits{A.}~\bsnm{Moreno}},
\batitle{A {Systematic} {Literature} {Review} of {Agents} {Applied} in {Healthcare}},
\bjtitle{Journal of Medical Systems}
\bvolume{40}(\bissue{2})
(\byear{2016}),
\bfpage{1}--\blpage{14}.
doi:\doiurl{10.1007/s10916-015-0376-2}.
\end{barticle}
\endbibitem

\bibitem{savaglio_agent-based_2020}
\begin{barticle}
\bauthor{\binits{C.}~\bsnm{Savaglio}},
\bauthor{\binits{M.}~\bsnm{Ganzha}},
\bauthor{\binits{M.}~\bsnm{Paprzycki}},
\bauthor{\binits{C.}~\bsnm{Bădică}},
\bauthor{\binits{M.}~\bsnm{Ivanović}} and
\bauthor{\binits{G.}~\bsnm{Fortino}},
\batitle{Agent-based {Internet} of {Things}: {State}-of-the-art and research challenges},
\bjtitle{Future Generation Computer Systems}
\bvolume{102}
(\byear{2020}),
\bfpage{1038}--\blpage{1053}.
doi:\doiurl{10.1016/j.future.2019.09.016}.
\end{barticle}
\endbibitem

\bibitem{montagna_complementing_2020}
\begin{barticle}
\bauthor{\binits{S.}~\bsnm{Montagna}},
\bauthor{\binits{S.}~\bsnm{Mariani}},
\bauthor{\binits{E.}~\bsnm{Gamberini}},
\bauthor{\binits{A.}~\bsnm{Ricci}} and
\bauthor{\binits{F.}~\bsnm{Zambonelli}},
\batitle{Complementing {Agents} with {Cognitive} {Services}: {A} {Case} {Study} in {Healthcare}},
\bjtitle{Journal of Medical Systems}
\bvolume{44}(\bissue{188})
(\byear{2020}),
\bfpage{1}--\blpage{10}.
doi:\doiurl{10.1007/s10916-020-01621-7}.
\end{barticle}
\endbibitem

\bibitem{bull_world_2020}
\begin{barticle}
\bauthor{\binits{F.C.}~\bsnm{Bull}},
\bauthor{\binits{S.S.}~\bsnm{Al-Ansari}},
\bauthor{\binits{S.}~\bsnm{Biddle}},
\bauthor{\binits{K.}~\bsnm{Borodulin}},
\bauthor{\binits{M.P.}~\bsnm{Buman}},
\bauthor{\binits{G.}~\bsnm{Cardon}},
\bauthor{\binits{C.}~\bsnm{Carty}},
\bauthor{\binits{J.-P.}~\bsnm{Chaput}},
\bauthor{\binits{S.}~\bsnm{Chastin}},
\bauthor{\binits{R.}~\bsnm{Chou}},
\bauthor{\binits{P.C.}~\bsnm{Dempsey}},
\bauthor{\binits{L.}~\bsnm{DiPietro}},
\bauthor{\binits{U.}~\bsnm{Ekelund}},
\bauthor{\binits{J.}~\bsnm{Firth}},
\bauthor{\binits{C.M.}~\bsnm{Friedenreich}},
\bauthor{\binits{L.}~\bsnm{Garcia}},
\bauthor{\binits{M.}~\bsnm{Gichu}},
\bauthor{\binits{R.}~\bsnm{Jago}},
\bauthor{\binits{P.T.}~\bsnm{Katzmarzyk}},
\bauthor{\binits{E.}~\bsnm{Lambert}},
\bauthor{\binits{M.}~\bsnm{Leitzmann}},
\bauthor{\binits{K.}~\bsnm{Milton}},
\bauthor{\binits{F.B.}~\bsnm{Ortega}},
\bauthor{\binits{C.}~\bsnm{Ranasinghe}},
\bauthor{\binits{E.}~\bsnm{Stamatakis}},
\bauthor{\binits{A.}~\bsnm{Tiedemann}},
\bauthor{\binits{R.P.}~\bsnm{Troiano}},
\bauthor{\binits{H.P.}~\bsnm{van~der Ploeg}},
\bauthor{\binits{V.}~\bsnm{Wari}} and
\bauthor{\binits{J.F.}~\bsnm{Willumsen}},
\batitle{World {Health} {Organization} 2020 guidelines on physical activity and sedentary behaviour},
\bjtitle{British Journal of Sports Medicine}
\bvolume{54}(\bissue{24})
(\byear{2020}),
\bfpage{1451}--\blpage{1462}.
doi:\doiurl{10.1136/bjsports-2020-102955}.
\end{barticle}
\endbibitem

\bibitem{miller_explanation_2019}
\begin{barticle}
\bauthor{\binits{T.}~\bsnm{Miller}},
\batitle{Explanation in artificial intelligence: {Insights} from the social sciences},
\bjtitle{Artificial Intelligence}
\bvolume{267}
(\byear{2019}),
\bfpage{1}--\blpage{38}.
doi:\doiurl{10.1016/j.artint.2018.07.007}.
\end{barticle}
\endbibitem

\bibitem{biran_explanation_2017}
\begin{bchapter}
\bauthor{\binits{O.}~\bsnm{Biran}} and
\bauthor{\binits{C.}~\bsnm{Cotton}},
\bctitle{Explanation and {Justification} in {Machine} {Learning}: {A} {Survey}},
in: \bbtitle{{IJCAI}-17 {Workshop} on {Explainable} {AI} ({XAI}) {Proceedings}},
\bconflocation{Melbourne, Australia}, \byear{2017},
pp.~\bfpage{8}--\blpage{13}.
\end{bchapter}
\endbibitem

\bibitem{molnar_interpretable_2023}
\begin{bbook}
\bauthor{\binits{C.}~\bsnm{Molnar}},
\bbtitle{Interpretable Machine Learning: A Guide for Making Black Box Models Explainable},
\bedition{2}nd edn,
\byear{2023},
\bcomment{Online; accessed 2024-04-03}.
\url{https://christophm.github.io/interpretable-ml-book/}.
\end{bbook}
\endbibitem

\bibitem{hagras_toward_2018}
\begin{barticle}
\bauthor{\binits{H.}~\bsnm{Hagras}},
\batitle{Toward {Human}-{Understandable}, {Explainable} {AI}},
\bjtitle{Computer}
\bvolume{51}(\bissue{9})
(\byear{2018}),
\bfpage{28}--\blpage{36}.
doi:\doiurl{10.1109/MC.2018.3620965}.
\url{https://ieeexplore.ieee.org/document/8481251/}.
\end{barticle}
\endbibitem

\bibitem{derks_taxonomy_2020}
\begin{barticle}
\bauthor{\binits{I.P.}~\bsnm{Derks}} and
\bauthor{\binits{A.}~\bsnm{de~Waal}},
\batitle{A {Taxonomy} of {Explainable} {Bayesian} {Networks}},
\bjtitle{Artificial Intelligence Research. SACAIR 2021.}
\bvolume{1342}
(\byear{2020}),
\bfpage{220}--\blpage{235}.
doi:\doiurl{10.1007/978-3-030-66151-9\_14}.
\end{barticle}
\endbibitem

\bibitem{kyrimi_comprehensive_2021}
\begin{barticle}
\bauthor{\binits{E.}~\bsnm{Kyrimi}},
\bauthor{\binits{S.}~\bsnm{McLachlan}},
\bauthor{\binits{K.}~\bsnm{Dube}},
\bauthor{\binits{M.R.}~\bsnm{Neves}},
\bauthor{\binits{A.}~\bsnm{Fahmi}} and
\bauthor{\binits{N.}~\bsnm{Fenton}},
\batitle{A comprehensive scoping review of {Bayesian} networks in healthcare: {Past}, present and future},
\bjtitle{Artificial Intelligence in Medicine}
\bvolume{117}
(\byear{2021}),
\bfpage{102108}.
doi:\doiurl{10.1016/j.artmed.2021.102108}.
\end{barticle}
\endbibitem

\bibitem{petch_opening_2022}
\begin{barticle}
\bauthor{\binits{J.}~\bsnm{Petch}},
\bauthor{\binits{S.}~\bsnm{Di}} and
\bauthor{\binits{W.}~\bsnm{Nelson}},
\batitle{Opening the {Black} {Box}: {The} {Promise} and {Limitations} of {Explainable} {Machine} {Learning} in {Cardiology}},
\bjtitle{Canadian Journal of Cardiology}
\bvolume{38}(\bissue{2})
(\byear{2022}),
\bfpage{204}--\blpage{213}.
doi:\doiurl{10.1016/j.cjca.2021.09.004}.
\end{barticle}
\endbibitem

\bibitem{sarker_neuro-symbolic_2021}
\begin{barticle}
\bauthor{\binits{M.K.}~\bsnm{Sarker}},
\bauthor{\binits{L.}~\bsnm{Zhou}},
\bauthor{\binits{A.}~\bsnm{Eberhart}} and
\bauthor{\binits{P.}~\bsnm{Hitzler}},
\batitle{Neuro-symbolic artificial intelligence},
\bjtitle{AI Communications}
\bvolume{34}(\bissue{3})
(\byear{2021}),
\bfpage{197}--\blpage{209}.
doi:\doiurl{10.3233/AIC-210084}.
\end{barticle}
\endbibitem

\bibitem{rudin_stop_2019}
\begin{botherref}
\oauthor{\binits{C.}~\bsnm{Rudin}},
Stop explaining black box machine learning models for high stakes decisions and use interpretable models instead,
\textit{Nature Machine Intelligence}
\textbf{1}(5)
(2019).
doi:\doiurl{10.1038/s42256-019-0048-x}.
\end{botherref}
\endbibitem

\bibitem{ribeiro_why_2016}
\begin{bchapter}
\bauthor{\binits{M.T.}~\bsnm{Ribeiro}},
\bauthor{\binits{S.}~\bsnm{Singh}} and
\bauthor{\binits{C.}~\bsnm{Guestrin}},
\bctitle{"{Why} {Should} {I} {Trust} {You}?": {Explaining} the {Predictions} of {Any} {Classifier}},
in: \bbtitle{Proceedings of the 22nd {ACM} {SIGKDD} {International} {Conference} on {Knowledge} {Discovery} and {Data} {Mining}},
\bsertitle{{KDD} '16},
\bpublisher{Association for Computing Machinery},
\blocation{New York, NY, USA},
\byear{2016},
pp.~\bfpage{1135}--\blpage{1144}.
ISBN \bisbn{978-1-4503-4232-2}.
doi:\doiurl{10.1145/2939672.2939778}.
\end{bchapter}
\endbibitem

\bibitem{lundberg_unified_2017}
\begin{bchapter}
\bauthor{\binits{S.M.}~\bsnm{Lundberg}} and
\bauthor{\binits{S.-I.}~\bsnm{Lee}},
\bctitle{A {Unified} {Approach} to {Interpreting} {Model} {Predictions}},
in: \bbtitle{Advances in {Neural} {Information} {Processing} {Systems}},
Vol.~\bseriesno{30},
\byear{2017}.
\url{https://proceedings.neurips.cc/paper/2017/hash/8a20a8621978632d76c43dfd28b67767-Abstract.html}.
\end{bchapter}
\endbibitem

\bibitem{barredo_arrieta_explainable_2020}
\begin{barticle}
\bauthor{\binits{A.}~\bsnm{Barredo~Arrieta}},
\bauthor{\binits{N.}~\bsnm{Díaz-Rodríguez}},
\bauthor{\binits{J.}~\bsnm{Del~Ser}},
\bauthor{\binits{A.}~\bsnm{Bennetot}},
\bauthor{\binits{S.}~\bsnm{Tabik}},
\bauthor{\binits{A.}~\bsnm{Barbado}},
\bauthor{\binits{S.}~\bsnm{Garcia}},
\bauthor{\binits{S.}~\bsnm{Gil-Lopez}},
\bauthor{\binits{D.}~\bsnm{Molina}},
\bauthor{\binits{R.}~\bsnm{Benjamins}},
\bauthor{\binits{R.}~\bsnm{Chatila}} and
\bauthor{\binits{F.}~\bsnm{Herrera}},
\batitle{Explainable {Artificial} {Intelligence} ({XAI}): {Concepts}, taxonomies, opportunities and challenges toward responsible {AI}},
\bjtitle{Information Fusion}
\bvolume{58}
(\byear{2020}),
\bfpage{82}--\blpage{115}.
doi:\doiurl{10.1016/j.inffus.2019.12.012}.
\end{barticle}
\endbibitem

\bibitem{tocchetti_role_2022}
\begin{barticle}
\bauthor{\binits{A.}~\bsnm{Tocchetti}} and
\bauthor{\binits{M.}~\bsnm{Brambilla}},
\batitle{The {Role} of {Human} {Knowledge} in {Explainable} {AI}},
\bjtitle{Data}
\bvolume{7}(\bissue{7})
(\byear{2022}),
\bfpage{93}.
doi:\doiurl{10.3390/data7070093}.
\end{barticle}
\endbibitem

\bibitem{confalonieri_multiple_2023}
\begin{botherref}
\oauthor{\binits{R.}~\bsnm{Confalonieri}} and
\oauthor{\binits{G.}~\bsnm{Guizzardi}},
On the {Multiple} {Roles} of {Ontologies} in {Explainable} {AI},
\textit{arXiv preprint arXiv:2311.04778}
(2023).
\end{botherref}
\endbibitem

\bibitem{shearer_hermit_2008}
\begin{bchapter}
\bauthor{\binits{R.}~\bsnm{Shearer}},
\bauthor{\binits{B.}~\bsnm{Motik}} and
\bauthor{\binits{I.}~\bsnm{Horrocks}},
\bctitle{{HermiT}: {A} {Highly}-{Eﬃcient} {OWL} {Reasoner}},
in: \bbtitle{Proceedings of the {Fifth} {OWLED} {Workshop} on {OWL}: {Experiences} and {Directions}, collocated with the 7th {International} {Semantic} {Web} {Conference}},
\byear{2008},
p.~\bfpage{10}.
\end{bchapter}
\endbibitem

\bibitem{sirin_pellet_2007}
\begin{barticle}
\bauthor{\binits{E.}~\bsnm{Sirin}},
\bauthor{\binits{B.}~\bsnm{Parsia}},
\bauthor{\binits{B.C.}~\bsnm{Grau}},
\bauthor{\binits{A.}~\bsnm{Kalyanpur}} and
\bauthor{\binits{Y.}~\bsnm{Katz}},
\batitle{Pellet: {A} practical {OWL}-{DL} reasoner},
\bjtitle{Journal of Web Semantics}
\bvolume{5}(\bissue{2})
(\byear{2007}),
\bfpage{51}--\blpage{53}.
doi:\doiurl{10.1016/j.websem.2007.03.004}.
\end{barticle}
\endbibitem

\bibitem{van_woensel_explanations_2024}
\begin{bchapter}
\bauthor{\binits{W.}~\bsnm{Van~Woensel}},
\bauthor{\binits{F.}~\bsnm{Scioscia}},
\bauthor{\binits{G.}~\bsnm{Loseto}},
\bauthor{\binits{O.}~\bsnm{Seneviratne}},
\bauthor{\binits{E.}~\bsnm{Patton}} and
\bauthor{\binits{S.}~\bsnm{Abidi}},
\bctitle{Explanations of {Symbolic} {Reasoning} to {Effect} {Patient} {Persuasion} and {Education}},
in: \bbtitle{Explainable {Artificial} {Intelligence} and {Process} {Mining} {Applications} for {Healthcare}},
\beditor{\binits{J.M.}~\bsnm{Juarez}},
\beditor{\binits{C.}~\bsnm{Fernandez-Llatas}},
\beditor{\binits{C.}~\bsnm{Bielza}},
\beditor{\binits{O.}~\bsnm{Johnson}},
\beditor{\binits{P.}~\bsnm{Kocbek}},
\beditor{\binits{P.}~\bsnm{Larrañaga}},
\beditor{\binits{N.}~\bsnm{Martin}},
\beditor{\binits{J.}~\bsnm{Munoz-Gama}},
\beditor{\binits{G.}~\bsnm{Štiglic}},
\beditor{\binits{M.}~\bsnm{Sepulveda}} and
\beditor{\binits{A.}~\bsnm{Vellido}}, eds,
\bpublisher{Springer Nature Switzerland},
\blocation{Cham},
\byear{2024},
pp.~\bfpage{62}--\blpage{71}.
ISBN \bisbn{978-3-031-54303-6}.
doi:\doiurl{10.1007/978-3-031-54303-6\_7}.
\end{bchapter}
\endbibitem

\bibitem{chari_explanation_2024}
\begin{barticle}
\bauthor{\binits{S.}~\bsnm{Chari}},
\bauthor{\binits{O.}~\bsnm{Seneviratne}},
\bauthor{\binits{M.}~\bsnm{Ghalwash}},
\bauthor{\binits{S.}~\bsnm{Shirai}},
\bauthor{\binits{D.M.}~\bsnm{Gruen}},
\bauthor{\binits{P.}~\bsnm{Meyer}},
\bauthor{\binits{P.}~\bsnm{Chakraborty}} and
\bauthor{\binits{D.L.}~\bsnm{McGuinness}},
\batitle{Explanation {Ontology}: {A} general-purpose, semantic representation for supporting user-centered explanations},
\bjtitle{Semantic Web}
\bvolume{15}(\bissue{4})
(\byear{2024}),
\bfpage{959}--\blpage{989},
\bcomment{Publisher: IOS Press}.
doi:\doiurl{10.3233/SW-233282}.
\end{barticle}
\endbibitem

\bibitem{nadendla_eco_2022}
\begin{barticle}
\bauthor{\binits{S.}~\bsnm{Nadendla}},
\bauthor{\binits{R.}~\bsnm{Jackson}},
\bauthor{\binits{J.}~\bsnm{Munro}},
\bauthor{\binits{F.}~\bsnm{Quaglia}},
\bauthor{\binits{B.}~\bsnm{Mészáros}},
\bauthor{\binits{D.}~\bsnm{Olley}},
\bauthor{\binits{E.T.}~\bsnm{Hobbs}},
\bauthor{\binits{S.M.}~\bsnm{Goralski}},
\bauthor{\binits{M.}~\bsnm{Chibucos}},
\bauthor{\binits{C.J.}~\bsnm{Mungall}},
\bauthor{\binits{S.C.E.}~\bsnm{Tosatto}},
\bauthor{\binits{I.}~\bsnm{Erill}} and
\bauthor{\binits{M.G.}~\bsnm{Giglio}},
\batitle{{ECO}: the {Evidence} and {Conclusion} {Ontology}, an update for 2022},
\bjtitle{Nucleic Acids Research}
\bvolume{50}(\bissue{D1})
(\byear{2022}),
\bfpage{D1515}--\blpage{D1521}.
doi:\doiurl{10.1093/nar/gkab1025}.
\end{barticle}
\endbibitem

\bibitem{bennett_bennetts_2013}
\begin{bbook}
\bauthor{\binits{D.H.}~\bsnm{Bennett}},
\bbtitle{Bennett's {Cardiac} {Arrhythmias}: {Practical} {Notes} on {Interpretation} and {Treatment}},
\bpublisher{John Wiley \& Sons, Ltd.},
\byear{2013},
p.~\bfpage{4}.
ISBN \bisbn{978-0-470-67493-2}.
\end{bbook}
\endbibitem

\bibitem{nute_defeasible_2003}
\begin{bchapter}
\bauthor{\binits{D.}~\bsnm{Nute}},
\bctitle{Defeasible {Logic}},
in: \bbtitle{Web {Knowledge} {Management} and {Decision} {Support}},
\beditor{\binits{G.}~\bsnm{Goos}},
\beditor{\binits{J.}~\bsnm{Hartmanis}},
\beditor{\binits{J.}~\bsnm{van Leeuwen}},
\beditor{\binits{O.}~\bsnm{Bartenstein}},
\beditor{\binits{U.}~\bsnm{Geske}},
\beditor{\binits{M.}~\bsnm{Hannebauer}} and
\beditor{\binits{O.}~\bsnm{Yoshie}}, eds,
\bsertitle{Lecture Notes in Computer Science},
Vol.~\bseriesno{2543},
\bpublisher{Springer Berlin Heidelberg},
\blocation{Berlin, Heidelberg},
\byear{2003},
pp.~\bfpage{151}--\blpage{169}.
ISBN \bisbn{978-3-540-00680-0 978-3-540-36524-2}.
doi:\doiurl{10.1007/3-540-36524-9\_13}.
\end{bchapter}
\endbibitem

\bibitem{ding_bayesowl_2006}
\begin{bchapter}
\bauthor{\binits{Z.}~\bsnm{Ding}},
\bauthor{\binits{Y.}~\bsnm{Peng}} and
\bauthor{\binits{R.}~\bsnm{Pan}},
\bctitle{{BayesOWL}: {Uncertainty} {Modeling} in {Semantic} {Web} {Ontologies}},
in: \bbtitle{Soft {Computing} in {Ontologies} and {Semantic} {Web}},
\beditor{\binits{Z.}~\bsnm{Ma}}, ed.,
\bpublisher{Springer},
\blocation{Berlin, Heidelberg},
\byear{2006},
pp.~\bfpage{3}--\blpage{29}.
ISBN \bisbn{978-3-540-33473-6}.
doi:\doiurl{10.1007/978-3-540-33473-6\_1}.
\end{bchapter}
\endbibitem

\bibitem{liu_bayes-swrl_2013}
\begin{bchapter}
\bauthor{\binits{Y.}~\bsnm{Liu}},
\bauthor{\binits{S.}~\bsnm{Chen}},
\bauthor{\binits{S.}~\bsnm{Li}} and
\bauthor{\binits{Y.}~\bsnm{Wang}},
\bctitle{Bayes-{SWRL}: {A} {Probabilistic} {Extension} of {SWRL}},
in: \bbtitle{2013 {Ninth} {International} {Conference} on {Computational} {Intelligence} and {Security}},
\byear{2013},
pp.~\bfpage{702}--\blpage{706}.
doi:\doiurl{10.1109/CIS.2013.153}.
\end{bchapter}
\endbibitem

\bibitem{rasool_security_2022}
\begin{barticle}
\bauthor{\binits{R.U.}~\bsnm{Rasool}},
\bauthor{\binits{H.F.}~\bsnm{Ahmad}},
\bauthor{\binits{W.}~\bsnm{Rafique}},
\bauthor{\binits{A.}~\bsnm{Qayyum}} and
\bauthor{\binits{J.}~\bsnm{Qadir}},
\batitle{Security and privacy of internet of medical things: {A} contemporary review in the age of surveillance, botnets, and adversarial {ML}},
\bjtitle{Journal of Network and Computer Applications}
\bvolume{201}
(\byear{2022}),
\bfpage{103332}.
doi:\doiurl{10.1016/j.jnca.2022.103332}.
\end{barticle}
\endbibitem

\bibitem{thapa_precision_2021}
\begin{barticle}
\bauthor{\binits{C.}~\bsnm{Thapa}} and
\bauthor{\binits{S.}~\bsnm{Camtepe}},
\batitle{Precision health data: {Requirements}, challenges and existing techniques for data security and privacy},
\bjtitle{Computers in Biology and Medicine}
\bvolume{129}
(\byear{2021}),
\bfpage{104130}.
doi:\doiurl{10.1016/j.compbiomed.2020.104130}.
\end{barticle}
\endbibitem

\bibitem{kirrane_privacy_2018}
\begin{barticle}
\bauthor{\binits{S.}~\bsnm{Kirrane}},
\bauthor{\binits{S.}~\bsnm{Villata}} and
\bauthor{\binits{M.}~\bsnm{d’Aquin}},
\batitle{Privacy, security and policies: {A} review of problems and solutions with semantic web technologies},
\bjtitle{Semantic Web}
\bvolume{9}(\bissue{2})
(\byear{2018}),
\bfpage{153}--\blpage{161}.
doi:\doiurl{10.3233/SW-180289}.
\end{barticle}
\endbibitem

\bibitem{sagar_systematic_2017}
\begin{barticle}
\bauthor{\binits{K.}~\bsnm{Sagar}} and
\bauthor{\binits{A.}~\bsnm{Saha}},
\batitle{A systematic review of software usability studies},
\bjtitle{International Journal of Information Technology}
(\byear{2017}).
doi:\doiurl{10.1007/s41870-017-0048-1}.
\end{barticle}
\endbibitem

\bibitem{saeed_exploration_2020}
\begin{barticle}
\bauthor{\binits{N.}~\bsnm{Saeed}},
\bauthor{\binits{M.}~\bsnm{Manzoor}} and
\bauthor{\binits{P.}~\bsnm{Khosravi}},
\batitle{An exploration of usability issues in telecare monitoring systems and possible solutions: a systematic literature review},
\bjtitle{Disability and Rehabilitation: Assistive Technology}
\bvolume{15}(\bissue{3})
(\byear{2020}),
\bfpage{271}--\blpage{281}.
doi:\doiurl{10.1080/17483107.2019.1578998}.
\end{barticle}
\endbibitem

\bibitem{maramba_methods_2019}
\begin{barticle}
\bauthor{\binits{I.}~\bsnm{Maramba}},
\bauthor{\binits{A.}~\bsnm{Chatterjee}} and
\bauthor{\binits{C.}~\bsnm{Newman}},
\batitle{Methods of usability testing in the development of {eHealth} applications: {A} scoping review},
\bjtitle{International Journal of Medical Informatics}
\bvolume{126}
(\byear{2019}),
\bfpage{95}--\blpage{104}.
doi:\doiurl{10.1016/j.ijmedinf.2019.03.018}.
\end{barticle}
\endbibitem

\bibitem{cho_multi-level_2018}
\begin{barticle}
\bauthor{\binits{H.}~\bsnm{Cho}},
\bauthor{\binits{P.-Y.}~\bsnm{Yen}},
\bauthor{\binits{D.}~\bsnm{Dowding}},
\bauthor{\binits{J.A.}~\bsnm{Merrill}} and
\bauthor{\binits{R.}~\bsnm{Schnall}},
\batitle{A multi-level usability evaluation of mobile health applications: {A} case study},
\bjtitle{Journal of Biomedical Informatics}
\bvolume{86}
(\byear{2018}),
\bfpage{79}--\blpage{89}.
doi:\doiurl{10.1016/j.jbi.2018.08.012}.
\end{barticle}
\endbibitem

\bibitem{weinstock_system_2006}
\begin{botherref}
\oauthor{\binits{C.}~\bsnm{Weinstock}} and
\oauthor{\binits{J.}~\bsnm{Goodenough}},
On {System} {Scalability},
Technical note, CMU/SEI-2006-TN-012,
Carnegie Mellon University,
2006.
\end{botherref}
\endbibitem

\bibitem{fortino_internet_2021}
\begin{barticle}
\bauthor{\binits{G.}~\bsnm{Fortino}},
\bauthor{\binits{C.}~\bsnm{Savaglio}},
\bauthor{\binits{G.}~\bsnm{Spezzano}} and
\bauthor{\binits{M.}~\bsnm{Zhou}},
\batitle{Internet of {Things} as {System} of {Systems}: {A} {Review} of {Methodologies}, {Frameworks}, {Platforms}, and {Tools}},
\bjtitle{IEEE Transactions on Systems, Man, and Cybernetics: Systems}
\bvolume{51}(\bissue{1})
(\byear{2021}),
\bfpage{223}--\blpage{236}.
doi:\doiurl{10.1109/TSMC.2020.3042898}.
\end{barticle}
\endbibitem

\bibitem{morley_ethics_2020}
\begin{barticle}
\bauthor{\binits{J.}~\bsnm{Morley}},
\bauthor{\binits{C.C.V.}~\bsnm{Machado}},
\bauthor{\binits{C.}~\bsnm{Burr}},
\bauthor{\binits{J.}~\bsnm{Cowls}},
\bauthor{\binits{I.}~\bsnm{Joshi}},
\bauthor{\binits{M.}~\bsnm{Taddeo}} and
\bauthor{\binits{L.}~\bsnm{Floridi}},
\batitle{The ethics of {AI} in health care: {A} mapping review},
\bjtitle{Social Science \& Medicine}
\bvolume{260}
(\byear{2020}),
\bfpage{113172}.
doi:\doiurl{10.1016/j.socscimed.2020.113172}.
\url{https://www.sciencedirect.com/science/article/pii/S0277953620303919}.
\end{barticle}
\endbibitem

\bibitem{kwan_healthcare_2017}
\begin{barticle}
\bauthor{\binits{V.}~\bsnm{Kwan}},
\bauthor{\binits{G.}~\bsnm{Hagen}},
\bauthor{\binits{M.}~\bsnm{Noel}},
\bauthor{\binits{K.}~\bsnm{Dobson}} and
\bauthor{\binits{K.}~\bsnm{Yeates}},
\batitle{Healthcare at {Your} {Fingertips}: {The} {Professional} {Ethics} of {Smartphone} {Health}-{Monitoring} {Applications}},
\bjtitle{Ethics \& Behavior}
\bvolume{27}(\bissue{8})
(\byear{2017}),
\bfpage{615}--\blpage{631}.
doi:\doiurl{10.1080/10508422.2017.1285237}.
\end{barticle}
\endbibitem

\bibitem{nittari_telemedicine_2020}
\begin{barticle}
\bauthor{\binits{G.}~\bsnm{Nittari}},
\bauthor{\binits{R.}~\bsnm{Khuman}},
\bauthor{\binits{S.}~\bsnm{Baldoni}},
\bauthor{\binits{G.}~\bsnm{Pallotta}},
\bauthor{\binits{G.}~\bsnm{Battineni}},
\bauthor{\binits{A.}~\bsnm{Sirignano}},
\bauthor{\binits{F.}~\bsnm{Amenta}} and
\bauthor{\binits{G.}~\bsnm{Ricci}},
\batitle{Telemedicine {Practice}: {Review} of the {Current} {Ethical} and {Legal} {Challenges}},
\bjtitle{Telemedicine and e-Health}
\bvolume{26}(\bissue{12})
(\byear{2020}),
\bfpage{1427}--\blpage{1437}.
doi:\doiurl{10.1089/tmj.2019.0158}.
\end{barticle}
\endbibitem

\bibitem{hassanaly_analysis_2021}
\begin{barticle}
\bauthor{\binits{P.}~\bsnm{Hassanaly}} and
\bauthor{\binits{J.C.}~\bsnm{Dufour}},
\batitle{Analysis of the {Regulatory}, {Legal}, and {Medical} {Conditions} for the {Prescription} of {Mobile} {Health} {Applications} in the {United} {States}, {The} {European} {Union}, and {France}},
\bjtitle{Medical Devices: Evidence and Research}
\bvolume{14}
(\byear{2021}),
\bfpage{389}--\blpage{409}.
doi:\doiurl{10.2147/MDER.S328996}.
\end{barticle}
\endbibitem

\bibitem{schmidt_introducing_2018}
\begin{bchapter}
\bauthor{\binits{P.}~\bsnm{Schmidt}},
\bauthor{\binits{A.}~\bsnm{Reiss}},
\bauthor{\binits{R.}~\bsnm{Duerichen}},
\bauthor{\binits{C.}~\bsnm{Marberger}} and
\bauthor{\binits{K.}~\bsnm{Van~Laerhoven}},
\bctitle{Introducing {WESAD}, a {Multimodal} {Dataset} for {Wearable} {Stress} and {Affect} {Detection}},
in: \bbtitle{Proceedings of the 20th {ACM} {International} {Conference} on {Multimodal} {Interaction}},
\bsertitle{{ICMI} '18},
\bpublisher{Association for Computing Machinery},
\blocation{New York, NY, USA},
\byear{2018},
pp.~\bfpage{400}--\blpage{408}.
ISBN \bisbn{978-1-4503-5692-3}.
doi:\doiurl{10.1145/3242969.3242985}.
\end{bchapter}
\endbibitem

\bibitem{sang_introduction_2003}
\begin{botherref}
\oauthor{\binits{E.F.T.K.}~\bsnm{Sang}} and
\oauthor{\binits{F.}~\bsnm{De~Meulder}},
Introduction to the CoNLL-2003 Shared Task: Language-Independent Named Entity Recognition,
arXiv,
2003.
doi:\doiurl{10.48550/ARXIV.CS/0306050}.
\end{botherref}
\endbibitem

\bibitem{strack_impact_2014}
\begin{barticle}
\bauthor{\binits{B.}~\bsnm{Strack}},
\bauthor{\binits{J.P.}~\bsnm{DeShazo}},
\bauthor{\binits{C.}~\bsnm{Gennings}},
\bauthor{\binits{J.L.}~\bsnm{Olmo}},
\bauthor{\binits{S.}~\bsnm{Ventura}},
\bauthor{\binits{K.J.}~\bsnm{Cios}} and
\bauthor{\binits{J.N.}~\bsnm{Clore}},
\batitle{Impact of {HbA1c} {Measurement} on {Hospital} {Readmission} {Rates}: {Analysis} of 70,000 {Clinical} {Database} {Patient} {Records}},
\bjtitle{BioMed Research International}
\bvolume{2014}
(\byear{2014}),
\bfpage{e781670},
\bcomment{Publisher: Hindawi}.
doi:\doiurl{10.1155/2014/781670}.
\end{barticle}
\endbibitem

\bibitem{fernandez-lopez_overview_2002}
\begin{barticle}
\bauthor{\binits{M.}~\bsnm{Fernández-López}} and
\bauthor{\binits{A.}~\bsnm{Gómez-Pérez}},
\batitle{Overview and analysis of methodologies for building ontologies},
\bjtitle{The Knowledge Engineering Review}
\bvolume{17}(\bissue{2})
(\byear{2002}),
\bfpage{129}--\blpage{156}.
doi:\doiurl{10.1017/S0269888902000462}.
\end{barticle}
\endbibitem

\bibitem{iqbal_analysis_2013}
\begin{barticle}
\bauthor{\binits{R.}~\bsnm{Iqbal}},
\bauthor{\binits{M.}~\bsnm{Azrifah}},
\bauthor{\binits{A.}~\bsnm{Murad}},
\bauthor{\binits{A.}~\bsnm{Mustapha}} and
\bauthor{\binits{N.M.}~\bsnm{Sharef}},
\batitle{An analysis of ontology engineering methodologies: {A} literature review},
\bjtitle{Research Journal of Applied Sciences, Engineering and Technology}
\bvolume{6}(\bissue{16})
(\byear{2013}),
\bfpage{2993}--\blpage{3000}.
\end{barticle}
\endbibitem

\bibitem{cimiano_knowledge_2017}
\begin{barticle}
\bauthor{\binits{P.}~\bsnm{Cimiano}} and
\bauthor{\binits{H.}~\bsnm{Paulheim}},
\batitle{Knowledge graph refinement: {A} survey of approaches and evaluation methods},
\bjtitle{Semantic Web}
\bvolume{8}(\bissue{3})
(\byear{2017}),
\bfpage{489}--\blpage{508}.
doi:\doiurl{10.3233/SW-160218}.
\end{barticle}
\endbibitem

\bibitem{kontokostas_test-driven_2014}
\begin{bchapter}
\bauthor{\binits{D.}~\bsnm{Kontokostas}},
\bauthor{\binits{P.}~\bsnm{Westphal}},
\bauthor{\binits{S.}~\bsnm{Auer}},
\bauthor{\binits{S.}~\bsnm{Hellmann}},
\bauthor{\binits{J.}~\bsnm{Lehmann}},
\bauthor{\binits{R.}~\bsnm{Cornelissen}} and
\bauthor{\binits{A.}~\bsnm{Zaveri}},
\bctitle{Test-driven evaluation of linked data quality},
in: \bbtitle{Proceedings of the 23rd international conference on {World} wide web},
\bpublisher{ACM},
\blocation{Seoul Korea},
\byear{2014},
pp.~\bfpage{747}--\blpage{758}.
ISBN \bisbn{978-1-4503-2744-2}.
doi:\doiurl{10.1145/2566486.2568002}.
\end{bchapter}
\endbibitem

\bibitem{debattista_luzzumethodology_2016}
\begin{barticle}
\bauthor{\binits{J.}~\bsnm{Debattista}},
\bauthor{\binits{S.}~\bsnm{Auer}} and
\bauthor{\binits{C.}~\bsnm{Lange}},
\batitle{Luzzu—{A} {Methodology} and {Framework} for {Linked} {Data} {Quality} {Assessment}},
\bjtitle{Journal of Data and Information Quality}
\bvolume{8}(\bissue{1})
(\byear{2016}),
\bfpage{1}--\blpage{32}.
doi:\doiurl{10.1145/2992786}.
\end{barticle}
\endbibitem

\bibitem{carmen_suarez-figueroa_neon_2015}
\begin{barticle}
\bauthor{\binits{M.}~\bsnm{Carmen~Suárez-Figueroa}},
\bauthor{\binits{A.}~\bsnm{Gómez-Pérez}} and
\bauthor{\binits{M.}~\bsnm{Fernández-López}},
\batitle{The {NeOn} {Methodology} framework: {A} scenario-based methodology for ontology development},
\bjtitle{Applied Ontology}
\bvolume{10}
(\byear{2015}),
\bfpage{107}--\blpage{145}.
doi:\doiurl{10.3233/AO-150145}.
\end{barticle}
\endbibitem

\bibitem{kotis_human-centered_2006}
\begin{barticle}
\bauthor{\binits{K.}~\bsnm{Kotis}} and
\bauthor{\binits{G.A.}~\bsnm{Vouros}},
\batitle{Human-centered ontology engineering: {The} {HCOME} methodology},
\bjtitle{Knowledge and Information Systems}
\bvolume{10}(\bissue{1})
(\byear{2006}),
\bfpage{109}--\blpage{131}.
doi:\doiurl{10.1007/s10115-005-0227-4}.
\end{barticle}
\endbibitem

\bibitem{el-sappagh_ontological_2014}
\begin{barticle}
\bauthor{\binits{S.H.}~\bsnm{El-Sappagh}},
\bauthor{\binits{S.}~\bsnm{El-Masri}},
\bauthor{\binits{M.}~\bsnm{Elmogy}},
\bauthor{\binits{A.M.}~\bsnm{Riad}} and
\bauthor{\binits{B.}~\bsnm{Saddik}},
\batitle{An {Ontological} {Case} {Base} {Engineering} {Methodology} for {Diabetes} {Management}},
\bjtitle{Journal of Medical Systems}
\bvolume{38}(\bissue{8})
(\byear{2014}),
\bfpage{67}.
doi:\doiurl{10.1007/s10916-014-0067-4}.
\end{barticle}
\endbibitem

\bibitem{spear_ontology_2006}
\begin{botherref}
\oauthor{\binits{A.D.}~\bsnm{Spear}},
Ontology for the {Twenty} {First} {Century}: {An} {Introduction} with {Recommendations},
Saarbrücken, Germany, 2006.
\end{botherref}
\endbibitem

\bibitem{arnold_enriching_2014}
\begin{barticle}
\bauthor{\binits{P.}~\bsnm{Arnold}} and
\bauthor{\binits{E.}~\bsnm{Rahm}},
\batitle{Enriching ontology mappings with semantic relations},
\bjtitle{Data \& Knowledge Engineering}
\bvolume{93}
(\byear{2014}),
\bfpage{1}--\blpage{18}.
doi:\doiurl{10.1016/j.datak.2014.07.001}.
\end{barticle}
\endbibitem

\bibitem{hevner_design_2010}
\begin{bchapter}
\bauthor{\binits{A.}~\bsnm{Hevner}} and
\bauthor{\binits{S.}~\bsnm{Chatterjee}},
\bctitle{Design {Science} {Research} in {Information} {Systems}},
in: \bbtitle{Design {Research} in {Information} {Systems}: {Theory} and {Practice}},
\bpublisher{Springer US},
\blocation{Boston, MA},
\byear{2010},
pp.~\bfpage{9}--\blpage{22}.
ISBN \bisbn{978-1-4419-5653-8}.
doi:\doiurl{10.1007/978-1-4419-5653-8\_2}.
\end{bchapter}
\endbibitem

\bibitem{kingston_designing_1998}
\begin{barticle}
\bauthor{\binits{J.K.C.}~\bsnm{Kingston}},
\batitle{Designing knowledge based systems: the {CommonKADS} design model},
\bjtitle{Knowledge-Based Systems}
\bvolume{11}(\bissue{5})
(\byear{1998}),
\bfpage{311}--\blpage{319}.
doi:\doiurl{10.1016/S0950-7051(98)00071-9}.
\end{barticle}
\endbibitem

\bibitem{iglesias_analysis_1998}
\begin{bchapter}
\bauthor{\binits{C.A.}~\bsnm{Iglesias}},
\bauthor{\binits{M.}~\bsnm{Garijo}},
\bauthor{\binits{J.C.}~\bsnm{González}} and
\bauthor{\binits{J.R.}~\bsnm{Velasco}},
\bctitle{Analysis and design of multiagent systems using {MAS}-{CommonKADS}},
in: \bbtitle{Intelligent {Agents} {IV} {Agent} {Theories}, {Architectures}, and {Languages}},
\beditor{\binits{M.P.}~\bsnm{Singh}},
\beditor{\binits{A.}~\bsnm{Rao}} and
\beditor{\binits{M.J.}~\bsnm{Wooldridge}}, eds,
\bpublisher{Springer},
\blocation{Berlin, Heidelberg},
\byear{1998},
pp.~\bfpage{313}--\blpage{327}.
ISBN \bisbn{978-3-540-69696-4}.
doi:\doiurl{10.1007/BFb0026768}.
\end{bchapter}
\endbibitem

\bibitem{poveda-villalon_oops_2014}
\begin{barticle}
\bauthor{\binits{M.}~\bsnm{Poveda-Villalón}},
\bauthor{\binits{A.}~\bsnm{Gómez-Pérez}} and
\bauthor{\binits{M.C.}~\bsnm{Suárez-Figueroa}},
\batitle{{OOPS}! ({OntOlogy} {Pitfall} {Scanner}!): {An} {On}-line {Tool} for {Ontology} {Evaluation}},
\bjtitle{International Journal on Semantic Web and Information Systems}
\bvolume{10}(\bissue{2})
(\byear{2014}),
\bfpage{7}--\blpage{34}.
doi:\doiurl{10.4018/ijswis.2014040102}.
\end{barticle}
\endbibitem

\bibitem{duque-ramos_oquare_2011}
\begin{barticle}
\bauthor{\binits{A.}~\bsnm{Duque-Ramos}},
\bauthor{\binits{J.T.}~\bsnm{Fernández-Breis}},
\bauthor{\binits{R.}~\bsnm{Stevens}} and
\bauthor{\binits{N.}~\bsnm{Aussenac-Gilles}},
\batitle{{OQuaRE}: {A} {SQuaRE}-based {Approach} for {Evaluating} the {Quality} of {Ontologies}},
\bjtitle{Journal of Research and Practice in Information Technology}
\bvolume{43}(\bissue{2})
(\byear{2011}),
\bfpage{18}.
\end{barticle}
\endbibitem

\bibitem{bass_software_2021}
\begin{bbook}
\bauthor{\binits{L.}~\bsnm{Bass}},
\bauthor{\binits{P.}~\bsnm{Clements}} and
\bauthor{\binits{R.}~\bsnm{Kazman}},
\bbtitle{Software {Architecture} in {Practice}},
\bedition{4}th edn,
\bpublisher{Addison-Wesley Professional},
\byear{2021}.
ISBN \bisbn{978-0-13-688567-2}.
\end{bbook}
\endbibitem

\bibitem{garlan_introduction_1995}
\begin{barticle}
\bauthor{\binits{D.}~\bsnm{Garlan}} and
\bauthor{\binits{D.}~\bsnm{Perry}},
\batitle{Introduction to the {Special} {Issue} on {Software} {Architecture}},
\bjtitle{IEEE Transactions on Software Engineering}
\bvolume{21}(\bissue{4})
(\byear{1995}),
\bfpage{269}--\blpage{274}.
\end{barticle}
\endbibitem

\bibitem{buschmann_pattern-oriented_1996}
\begin{bbook}
\bauthor{\binits{F.}~\bsnm{Buschmann}},
\bauthor{\binits{R.}~\bsnm{Meunier}},
\bauthor{\binits{H.}~\bsnm{Rohnert}},
\bauthor{\binits{P.}~\bsnm{Sommerlad}} and
\bauthor{\binits{M.}~\bsnm{Stal}},
\bbtitle{Pattern-{Oriented} {Sofware} {Architecture}},
\bpublisher{Wiley},
\byear{1996}.
ISBN \bisbn{978-0-471-95869-7}.
\end{bbook}
\endbibitem

\bibitem{richards_software_2015}
\begin{bbook}
\bauthor{\binits{M.}~\bsnm{Richards}},
\bbtitle{Software {Architecture} {Patterns}: {Understanding} {Common} {Architecture} {Patterns} and {When} to {Use} {Them}},
\bedition{1}st edn,
\bpublisher{O’Reilly Media, Inc.},
\byear{2015}.
\end{bbook}
\endbibitem

\bibitem{meyer_modular_2005}
\begin{barticle}
\bauthor{\binits{M.H.}~\bsnm{Meyer}} and
\bauthor{\binits{P.H.}~\bsnm{Webb}},
\batitle{Modular, layered architecture: the necessary foundation for effective mass customisation in software},
\bjtitle{International Journal of Mass Customisation}
\bvolume{1}(\bissue{1})
(\byear{2005}),
\bfpage{14}.
doi:\doiurl{10.1504/IJMASSC.2005.007349}.
\end{barticle}
\endbibitem

\bibitem{damato_machine_2020}
\begin{barticle}
\bauthor{\binits{C.}~\bsnm{d’Amato}},
\batitle{Machine {Learning} for the {Semantic} {Web}: {Lessons} learnt and next research directions},
\bjtitle{Semantic Web}
\bvolume{11}(\bissue{1})
(\byear{2020}),
\bfpage{195}--\blpage{203}.
doi:\doiurl{10.3233/SW-200388}.
\end{barticle}
\endbibitem

\bibitem{kotis_machine_2021}
\begin{barticle}
\bauthor{\binits{K.I.}~\bsnm{Kotis}},
\bauthor{\binits{K.}~\bsnm{Zachila}} and
\bauthor{\binits{E.}~\bsnm{Paparidis}},
\batitle{Machine {Learning} {Meets} the {Semantic} {Web}},
\bjtitle{Artificial Intelligence Advances}
\bvolume{3}(\bissue{1})
(\byear{2021}),
\bfpage{63}--\blpage{70}.
doi:\doiurl{10.30564/aia.v3i1.3178}.
\end{barticle}
\endbibitem

\bibitem{he_deeponto_2024}
\begin{botherref}
\oauthor{\binits{Y.}~\bsnm{He}},
\oauthor{\binits{J.}~\bsnm{Chen}},
\oauthor{\binits{H.}~\bsnm{Dong}},
\oauthor{\binits{I.}~\bsnm{Horrocks}},
\oauthor{\binits{C.}~\bsnm{Allocca}},
\oauthor{\binits{T.}~\bsnm{Kim}} and
\oauthor{\binits{B.}~\bsnm{Sapkota}},
{DeepOnto}: {A} {Python} {Package} for {Ontology} {Engineering} with {Deep} {Learning},
\textit{Semantic Web}
(2024).
\url{https://semantic-web-journal.net/content/deeponto-python-package-ontology-engineering-deep-learning-0}.
\end{botherref}
\endbibitem

\bibitem{garijo_llms_2024}
\begin{botherref}
\oauthor{\binits{D.}~\bsnm{Garijo}},
\oauthor{\binits{M.}~\bsnm{Poveda-Villalón}},
\oauthor{\binits{E.}~\bsnm{Amador-Domínguez}},
\oauthor{\binits{Z.}~\bsnm{Wang}},
\oauthor{\binits{R.}~\bsnm{García-Castro}} and
\oauthor{\binits{O.}~\bsnm{Corcho}},
{LLMs} for {Ontology} {Engineering}: {A} landscape of {Tasks} and {Benchmarking} challenges,
in: \textit{Proceedings of the {Special} {Session} on {Harmonising} {Generative} {AI} and {Semantic} {Web} {Technologies} ({HGAIS} 2024)},
Vol.~3953,
CEUR Workshop Proceedings.
\url{https://ceur-ws.org/Vol-3953/364.pdf}.
\end{botherref}
\endbibitem

\bibitem{shu_knowledge_2024}
\begin{bchapter}
\bauthor{\binits{D.}~\bsnm{Shu}},
\bauthor{\binits{T.}~\bsnm{Chen}},
\bauthor{\binits{M.}~\bsnm{Jin}},
\bauthor{\binits{C.}~\bsnm{Zhang}},
\bauthor{\binits{M.}~\bsnm{Du}} and
\bauthor{\binits{Y.}~\bsnm{Zhang}},
\bctitle{Knowledge {Graph} {Large} {Language} {Model} ({KG}-{LLM}) for {Link} {Prediction}},
in: \bbtitle{Proceedings of the 16th {Asian} {Conference} on {Machine} {Learning}},
Vol.~\bseriesno{260},
\byear{2024},
pp.~\bfpage{143}--\blpage{158}.
doi:\doiurl{10.48550/arXiv.2403.07311}.
\end{bchapter}
\endbibitem

\bibitem{meyer_llm-assisted_2024}
\begin{bchapter}
\bauthor{\binits{L.-P.}~\bsnm{Meyer}},
\bauthor{\binits{C.}~\bsnm{Stadler}},
\bauthor{\binits{J.}~\bsnm{Frey}},
\bauthor{\binits{N.}~\bsnm{Radtke}},
\bauthor{\binits{K.}~\bsnm{Junghanns}},
\bauthor{\binits{R.}~\bsnm{Meissner}},
\bauthor{\binits{G.}~\bsnm{Dziwis}},
\bauthor{\binits{K.}~\bsnm{Bulert}} and
\bauthor{\binits{M.}~\bsnm{Martin}},
\bctitle{{LLM}-assisted {Knowledge} {Graph} {Engineering}: {Experiments} with {ChatGPT}},
in: \bbtitle{1st {Working} {Conference} on {Artificial} {Intelligence} {Development} for a {Resilient} and {Sustainable} {Tomorrow}},
\beditor{\binits{C.}~\bsnm{Zinke-Wehlmann}} and
\beditor{\binits{J.}~\bsnm{Friedrich}}, eds,
\bpublisher{Springer Fachmedien},
\blocation{Wiesbaden},
\byear{2024},
pp.~\bfpage{103}--\blpage{115}.
ISBN \bisbn{978-3-658-43705-3}.
doi:\doiurl{10.1007/978-3-658-43705-3\_8}.
\end{bchapter}
\endbibitem

\bibitem{uskul_emotions_2015}
\begin{bchapter}
\bauthor{\binits{A.K.}~\bsnm{Uskul}} and
\bauthor{\binits{A.B.}~\bsnm{Horn}},
\bctitle{Emotions and {Health}},
in: \bbtitle{International {Encyclopedia} of the {Social} \& {Behavioral} {Sciences}},
Vol.~\bseriesno{7},
\beditor{\binits{J.D.}~\bsnm{Wright}}, ed.,
\bpublisher{Elsevier},
\byear{2015},
pp.~\bfpage{496}--\blpage{501}.
\bcomment{ISSN 0014-9772}.
ISBN \bisbn{978-0-08-097087-5}.
doi:\doiurl{10.1016/B978-0-08-097086-8.25006-X}.
\end{bchapter}
\endbibitem

\bibitem{alani_survey_2018}
\begin{barticle}
\bauthor{\binits{H.}~\bsnm{Alani}},
\bauthor{\binits{R.}~\bsnm{Abaalkhail}},
\bauthor{\binits{B.}~\bsnm{Guthier}},
\bauthor{\binits{R.}~\bsnm{Alharthi}} and
\bauthor{\binits{A.}~\bsnm{El~Saddik}},
\batitle{Survey on ontologies for affective states and their influences},
\bjtitle{Semantic Web}
\bvolume{9}(\bissue{4})
(\byear{2018}),
\bfpage{441}--\blpage{458}.
doi:\doiurl{10.3233/SW-170270}.
\end{barticle}
\endbibitem

\bibitem{vallee_dynamic_2005}
\begin{bchapter}
\bauthor{\binits{M.}~\bsnm{Vallée}},
\bauthor{\binits{F.}~\bsnm{Ramparany}} and
\bauthor{\binits{L.}~\bsnm{Vercouter}},
\bctitle{Dynamic {Service} {Composition} in {Ambient} {Intelligence} {Environments}: a {Multi}-{Agent} {Approach}},
in: \bbtitle{Proceedings of the {First} {European} {Young} {Researchers} {Workshop} on {Service}-{Oriented} {Computing}},
\byear{2005},
p.~\bfpage{6}.
doi:\doiurl{10.1109/SCC.2009.16}.
\end{bchapter}
\endbibitem

\bibitem{ajami_ontology-based_2018}
\begin{barticle}
\bauthor{\binits{H.}~\bsnm{Ajami}} and
\bauthor{\binits{H.}~\bsnm{Mcheick}},
\batitle{Ontology-{Based} {Model} to {Support} {Ubiquitous} {Healthcare} {Systems} for {COPD} {Patients}},
\bjtitle{Electronics}
\bvolume{7}(\bissue{12})
(\byear{2018}),
\bfpage{371}.
doi:\doiurl{10.3390/electronics7120371}.
\end{barticle}
\endbibitem

\bibitem{preuveneers_towards_2004}
\begin{bchapter}
\bauthor{\binits{D.}~\bsnm{Preuveneers}},
\bauthor{\binits{J.}~\bsnm{Van~den Bergh}},
\bauthor{\binits{D.}~\bsnm{Wagelaar}},
\bauthor{\binits{A.}~\bsnm{Georges}},
\bauthor{\binits{P.}~\bsnm{Rigole}},
\bauthor{\binits{T.}~\bsnm{Clerckx}},
\bauthor{\binits{Y.}~\bsnm{Berbers}},
\bauthor{\binits{K.}~\bsnm{Coninx}},
\bauthor{\binits{V.}~\bsnm{Jonckers}} and
\bauthor{\binits{K.}~\bsnm{De~Bosschere}},
\bctitle{Towards an {Extensible} {Context} {Ontology} for {Ambient} {Intelligence}},
in: \bbtitle{Ambient {Intelligence}},
\beditor{\binits{P.}~\bsnm{Markopoulos}},
\beditor{\binits{B.}~\bsnm{Eggen}},
\beditor{\binits{E.}~\bsnm{Aarts}} and
\beditor{\binits{J.L.}~\bsnm{Crowley}}, eds,
\bpublisher{Springer},
\blocation{Berlin, Heidelberg},
\byear{2004},
pp.~\bfpage{148}--\blpage{159}.
ISBN \bisbn{978-3-540-30473-9}.
doi:\doiurl{10.1007/978-3-540-30473-9\_15}.
\end{bchapter}
\endbibitem

\bibitem{wang_wselector_2015}
\begin{bchapter}
\bauthor{\binits{J.}~\bsnm{Wang}},
\bauthor{\binits{S.}~\bsnm{Helal}},
\bauthor{\binits{Y.}~\bsnm{Wang}} and
\bauthor{\binits{D.}~\bsnm{Zhang}},
\bctitle{{WSelector}: {A} {Multi}-scenario and {Multi}-view {Worker} {Selection} {Framework} for {Crowd}-{Sensing}},
in: \bbtitle{2015 {IEEE} 12th {Intl} {Conf} on {Ubiquitous} {Intelligence} and {Computing} and 2015 {IEEE} 12th {Intl} {Conf} on {Autonomic} and {Trusted} {Computing} and 2015 {IEEE} 15th {Intl} {Conf} on {Scalable} {Computing} and {Communications} and {Its} {Associated} {Workshops} ({UIC}-{ATC}-{ScalCom})},
\byear{2015},
pp.~\bfpage{54}--\blpage{61}.
doi:\doiurl{10.1109/UIC-ATC-ScalCom-CBDCom-IoP.2015.32}.
\end{bchapter}
\endbibitem

\bibitem{el-sappagh_ddo_2016}
\begin{barticle}
\bauthor{\binits{S.}~\bsnm{El-Sappagh}} and
\bauthor{\binits{F.}~\bsnm{Ali}},
\batitle{{DDO}: a diabetes mellitus diagnosis ontology},
\bjtitle{Applied Informatics}
\bvolume{3}(\bissue{1})
(\byear{2016}),
\bfpage{5}.
doi:\doiurl{10.1186/s40535-016-0021-2}.
\end{barticle}
\endbibitem

\bibitem{goh_human_2007}
\begin{barticle}
\bauthor{\binits{K.-I.}~\bsnm{Goh}},
\bauthor{\binits{M.E.}~\bsnm{Cusick}},
\bauthor{\binits{D.}~\bsnm{Valle}},
\bauthor{\binits{B.}~\bsnm{Childs}},
\bauthor{\binits{M.}~\bsnm{Vidal}} and
\bauthor{\binits{A.-L.}~\bsnm{Barab{\'a}si}},
\batitle{The human disease network},
\bjtitle{Proceedings of the National Academy of Sciences}
\bvolume{104}(\bissue{21})
(\byear{2007}),
\bfpage{8685}--\blpage{8690}.
\end{barticle}
\endbibitem

\bibitem{heckmann_gumo_2005}
\begin{bchapter}
\bauthor{\binits{D.}~\bsnm{Heckmann}},
\bauthor{\binits{T.}~\bsnm{Schwartz}},
\bauthor{\binits{B.}~\bsnm{Brandherm}},
\bauthor{\binits{M.}~\bsnm{Schmitz}} and
\bauthor{\binits{M.}~\bsnm{von Wilamowitz-Moellendorff}},
\bctitle{Gumo – {The} {General} {User} {Model} {Ontology}},
in: \bbtitle{User {Modeling} 2005},
\beditor{\binits{L.}~\bsnm{Ardissono}},
\beditor{\binits{P.}~\bsnm{Brna}} and
\beditor{\binits{A.}~\bsnm{Mitrovic}}, eds,
\bpublisher{Springer},
\blocation{Berlin, Heidelberg},
\byear{2005},
pp.~\bfpage{428}--\blpage{432}.
ISBN \bisbn{978-3-540-31878-1}.
doi:\doiurl{10.1007/11527886\_58}.
\end{bchapter}
\endbibitem

\bibitem{rhayem_healthiot_2017}
\begin{bchapter}
\bauthor{\binits{A.}~\bsnm{Rhayem}},
\bauthor{\binits{M.B.A.}~\bsnm{Mhiri}} and
\bauthor{\binits{F.}~\bsnm{Gargouri}},
\bctitle{{HealthIoT} {Ontology} for {Data} {Semantic} {Representation} and {Interpretation} {Obtained} from {Medical} {Connected} {Objects}},
in: \bbtitle{2017 {IEEE}/{ACS} 14th {International} {Conference} on {Computer} {Systems} and {Applications} ({AICCSA})},
\byear{2017},
pp.~\bfpage{1470}--\blpage{1477},
\bcomment{ISSN: 2161-5330}.
doi:\doiurl{10.1109/AICCSA.2017.171}.
\end{bchapter}
\endbibitem

\bibitem{elsaleh_iot-stream_2020}
\begin{barticle}
\bauthor{\binits{T.}~\bsnm{Elsaleh}},
\bauthor{\binits{S.}~\bsnm{Enshaeifar}},
\bauthor{\binits{R.}~\bsnm{Rezvani}},
\bauthor{\binits{S.T.}~\bsnm{Acton}},
\bauthor{\binits{V.}~\bsnm{Janeiko}} and
\bauthor{\binits{M.}~\bsnm{Bermudez-Edo}},
\batitle{{IoT}-{Stream}: {A} {Lightweight} {Ontology} for {Internet} of {Things} {Data} {Streams} and {Its} {Use} with {Data} {Analytics} and {Event} {Detection} {Services}},
\bjtitle{Sensors}
\bvolume{20}(\bissue{4})
(\byear{2020}),
\bfpage{953},
\bcomment{Publisher: Multidisciplinary Digital Publishing Institute}.
doi:\doiurl{10.3390/s20040953}.
\end{barticle}
\endbibitem

\bibitem{villalonga_mimu-wear_2017}
\begin{barticle}
\bauthor{\binits{C.}~\bsnm{Villalonga}},
\bauthor{\binits{H.}~\bsnm{Pomares}},
\bauthor{\binits{I.}~\bsnm{Rojas}} and
\bauthor{\binits{O.}~\bsnm{Banos}},
\batitle{{MIMU}-{Wear}: {Ontology}-based sensor selection for real-world wearable activity recognition},
\bjtitle{Neurocomputing}
\bvolume{250}
(\byear{2017}),
\bfpage{76}--\blpage{100}.
doi:\doiurl{10.1016/j.neucom.2016.09.125}.
\end{barticle}
\endbibitem

\bibitem{gasmi_oaisis_2016}
\begin{bchapter}
\bauthor{\binits{A.}~\bsnm{Gasmi}},
\bauthor{\binits{N.}~\bsnm{Tamani}},
\bauthor{\binits{C.}~\bsnm{Faucher}} and
\bauthor{\binits{Y.}~\bsnm{Ghamri-Doudane}},
\bctitle{{OAISIS}: {An} ontological-based approach for interlinking {CrowdSensing} information systems},
in: \bbtitle{2016 {IEEE} {International} {Conference} on {Systems}, {Man}, and {Cybernetics} ({SMC})},
\byear{2016},
pp.~\bfpage{003995}--\blpage{004000}.
doi:\doiurl{10.1109/SMC.2016.7844858}.
\end{bchapter}
\endbibitem

\bibitem{wannous_modelling_2013}
\begin{bchapter}
\bauthor{\binits{R.}~\bsnm{Wannous}},
\bauthor{\binits{J.}~\bsnm{Malki}},
\bauthor{\binits{A.}~\bsnm{Bouju}} and
\bauthor{\binits{C.}~\bsnm{Vincent}},
\bctitle{Modelling {Mobile} {Object} {Activities} {Based} on {Trajectory} {Ontology} {Rules} {Considering} {Spatial} {Relationship} {Rules}},
in: \bbtitle{Modeling {Approaches} and {Algorithms} for {Advanced} {Computer} {Applications}},
\beditor{\binits{A.}~\bsnm{Amine}},
\beditor{\binits{A.M.}~\bsnm{Otmane}} and
\beditor{\binits{L.}~\bsnm{Bellatreche}}, eds,
\bpublisher{Springer International Publishing},
\blocation{Cham},
\byear{2013},
pp.~\bfpage{249}--\blpage{258}.
ISBN \bisbn{978-3-319-00560-7}.
doi:\doiurl{10.1007/978-3-319-00560-7\_29}.
\end{bchapter}
\endbibitem

\bibitem{heinroth_owlspeak_2010}
\begin{bchapter}
\bauthor{\binits{T.}~\bsnm{Heinroth}},
\bauthor{\binits{D.}~\bsnm{Denich}} and
\bauthor{\binits{A.}~\bsnm{Schmitt}},
\bctitle{{OwlSpeak} - adaptive spoken dialogue within {Intelligent} {Environments}},
in: \bbtitle{2010 8th {IEEE} {International} {Conference} on {Pervasive} {Computing} and {Communications} {Workshops} ({PERCOM} {Workshops})},
\byear{2010},
pp.~\bfpage{666}--\blpage{671}.
doi:\doiurl{10.1109/PERCOMW.2010.5470518}.
\end{bchapter}
\endbibitem

\bibitem{kim_developing_2019}
\begin{barticle}
\bauthor{\binits{H.}~\bsnm{Kim}},
\bauthor{\binits{J.}~\bsnm{Mentzer}} and
\bauthor{\binits{R.}~\bsnm{Taira}},
\batitle{Developing a {Physical} {Activity} {Ontology} to {Support} the {Interoperability} of {Physical} {Activity} {Data}},
\bjtitle{Journal of Medical Internet Research}
\bvolume{21}(\bissue{4})
(\byear{2019}),
\bfpage{e12776}.
doi:\doiurl{10.2196/12776}.
\end{barticle}
\endbibitem

\bibitem{kolozali_knowledge-based_2014}
\begin{bchapter}
\bauthor{\binits{S.}~\bsnm{Kolozali}},
\bauthor{\binits{M.}~\bsnm{Bermudez-Edo}},
\bauthor{\binits{D.}~\bsnm{Puschmann}},
\bauthor{\binits{F.}~\bsnm{Ganz}} and
\bauthor{\binits{P.}~\bsnm{Barnaghi}},
\bctitle{A {Knowledge}-{Based} {Approach} for {Real}-{Time} {IoT} {Data} {Stream} {Annotation} and {Processing}},
in: \bbtitle{2014 {IEEE} {International} {Conference} on {Internet} of {Things}({iThings}), and {IEEE} {Green} {Computing} and {Communications} ({GreenCom}) and {IEEE} {Cyber}, {Physical} and {Social} {Computing} ({CPSCom})},
\bpublisher{IEEE},
\blocation{Taipei, Taiwan},
\byear{2014},
pp.~\bfpage{215}--\blpage{222}.
ISBN \bisbn{978-1-4799-5967-9}.
doi:\doiurl{10.1109/iThings.2014.39}.
\end{bchapter}
\endbibitem

\end{thebibliography}

\end{document}